\documentclass{article} %
\usepackage{iclr2025_conference,times}
\usepackage[T1]{fontenc}

\usepackage{amsmath,amsfonts,bm}

\def\eqref#1{equation~\ref{#1}}

\def\1{\bm{1}}

\DeclareMathAlphabet{\mathsfit}{\encodingdefault}{\sfdefault}{m}{sl}
\SetMathAlphabet{\mathsfit}{bold}{\encodingdefault}{\sfdefault}{bx}{n}

\definecolor{coolblack}{rgb}{0.0, 0.18, 0.39}
\definecolor{cornellred}{rgb}{0.7, 0.11, 0.11}

\usepackage[pagebackref,breaklinks,colorlinks,citecolor=coolblack,linkcolor=cornellred]{hyperref}
\usepackage{url}

\usepackage{booktabs} %
\usepackage{tabularx}
\usepackage{amsmath}
\usepackage{amssymb}
\usepackage{mathtools}
\usepackage{amsthm}
\usepackage{amsmath}
\usepackage{lipsum}  
\usepackage{listings}
\usepackage{caption}
\usepackage{subcaption}
\usepackage{CJKutf8}
\usepackage{xspace}
\usepackage{multirow}
\usepackage[framemethod=TikZ]{mdframed}
\usepackage{tabu}
\usepackage{longtable}
\usepackage{alphalph}
\usepackage[subtle]{savetrees}

\DeclareMathSizes{10}{9}{6}{5}

\title{Analyzing The Language of Visual Tokens}

\author{David M. Chan$^1$, Rodolfo Corona$^1$, Joonyong Park$^2$, Cheol Jun Cho$^1$, Yutong Bai$^1$, Trevor Darrell$^1$ \\
$^1$University of California, Berkeley,$\:$ $^2$The University of Tokyo, Tokyo\\
\texttt{\{davidchan,rcorona,cheoljun,yutongbai,trevordarrell\}@berkeley.edu},\\ \texttt{joonyong-park@g.ecc.u-tokyo.ac.jp }
}

\iclrpreprint
\begin{document}

\maketitle

\begin{abstract}
With the introduction of transformer-based models for vision and language tasks, such as LLaVA and Chameleon, there has been renewed interest in the discrete tokenized representation of images. These models often treat image patches as discrete tokens, analogous to words in natural language, learning joint alignments between visual and human languages. However, little is known about the statistical behavior of these visual languages—whether they follow similar frequency distributions, grammatical structures, or topologies as natural languages. In this paper, we take a natural-language-centric approach to analyzing discrete visual languages and uncover striking similarities and fundamental differences. We demonstrate that, although visual languages adhere to Zipfian distributions, higher token innovation drives greater entropy and lower compression, with tokens predominantly representing object parts, indicating intermediate granularity. We also show that visual languages lack cohesive grammatical structures, leading to higher perplexity and weaker hierarchical organization compared to natural languages. Finally, we demonstrate that, while vision models align more closely with natural languages than other models, this alignment remains significantly weaker than the cohesion found within natural languages. Through these experiments, we demonstrate how understanding the statistical properties of discrete visual languages can inform the design of more effective computer vision models.
\end{abstract}
\section{Introduction}

\begin{figure}[t]
    \centering
    \includegraphics[width=\linewidth]{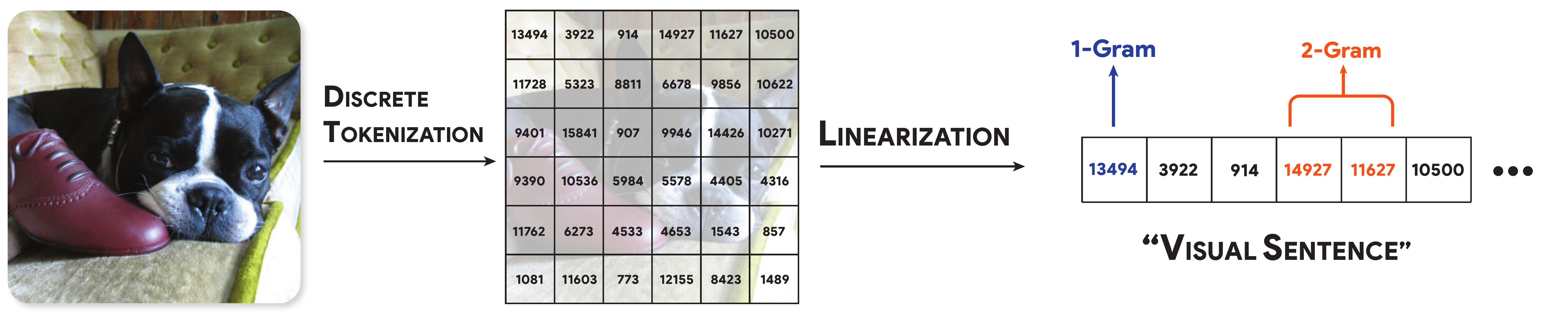}
    \caption{Discrete tokenizers used for visual pre-processing induce ``visual languages'' made up of sentences containing 1-D sequences of discrete tokens extracted from the images in a dataset. In this paper, we explore how the statistics of these ``visual languages'' differ from ``natural languages,'' and understand the implications of such statistical differences.}
    \label{fig:visual_language}
\end{figure}

Transformer-based models have not just advanced, but fundamentally reshaped how we approach both vision and language processing, merging these domains in shared sequential representation spaces. Indeed, most recent multi-modal models including DALL-E \citep{ramesh2022hierarchical},  LLaVA \citep{liu2024visual} and Chameleon \citep{team2024chameleon} operate over joint tokenized representations of images and language, where models decompose images into ``visual languages'': linearized discrete patches or tokens analogous to words in a sentence. This process, shown in \autoref{fig:visual_language}, enables seamless integration of images into transformer architectures and allows models to solve multimodal tasks, ranging from image generation and image captioning to visual question answering and translation.  

Despite the success of such shared-structure models, current research lacks an in-depth understanding of whether the internal structure of visual tokens mirrors the principles governing natural languages. Specifically, the question arises: do languages formed of visual tokens follow the same statistical patterns, such as frequency distributions, grammatical rules, or semantic dependencies, that human languages exhibit? Investigating such statistical behavior of discrete visual tokens extends beyond theoretical curiosity; it has broad implications for practical machine learning applications. While in linguistic theory, phenomena like Zipf’s law and entropy shape natural languages’ structure and shape the design of machine learning algorithms, no such rules exist for visual languages. Such rules, if they exist, have the potential to motivate creating modality-specific models and procedures to capture the unique statistical properties of the underlying visual data. 

In pursuit of such rules, in this paper we inspect the equivalence of visual and natural languages through an empirical analysis of token distributions, segmentation granularity, and syntactic and semantic structures. We start by investigating the frequency statistics of visual words and compare them to natural languages. Our analysis reveals that although visual languages can follow power-law (Zipfian) distributions, they use more tokens more uniformly. This leads to languages with greater per-token entropy and lower compression ratios, and implies that vision models may require more attention heads, larger embeddings, and longer training times with more diverse data compared to natural language models. Noting in these experiments that visual languages have coarser granularity than patches, we demonstrate through correlation analysis that visual tokens operate at an intermediate level of granularity, and typically represent object parts rather than whole objects or sub-parts in images. Correspondingly, we show visual tokens are less effective at representing fine-grained details or whole-object structures. Following this line of reasoning, we explore if tokens have composable structure, and using parse trees generated by Compound Probabilistic Context-Free Grammars (C-PCFG), we show visual languages have grammatical structures that are more fragmented, with grammars trained on them exhibiting higher perplexity compared to natural languages. We confirm these observations by building a co-occurrence based embedding space, and evaluating the topological alignment between natural and visual languages. In this, we find visual languages align more with natural languages than with other visual languages, but less so than natural languages align with each other.

Together, we aim to show through these experiments show that while visual languages have striking similarities to natural languages, there are also notable and fundamental differences, motivating a unique modality-specific approaches to vision-language learning.

\section{Do visual tokens act like words?}

The first, question that we examine is: Do visual tokens themselves (i.e. the patches of an image) act like words? While we often treat these tokens as either a word (or subword), as each token forms a single input sequence element in a transformer, it seems unintuitive that there would be a one to one statistical correlation between the two concepts. In this section, we look at several statistical properties of individual tokens, comparing those observed in natural language to those in visual systems.

\subsection{Preliminaries}
\label{sec:preliminaries}

What, explicitly, is a visual language? In this work, we consider a visual language to be a language induced over ``visual tokens'' by first converting images in a dataset to a discrete set of symbols using a visual tokenizer (often a VQ-VAE), and then linearizing those tokens into one-dimensional sequences (See \autoref{fig:visual_language}). Such a definition parallels efforts in both text-to-image diffusion and large vision and language models which have both explored using discrete visual tokens for vision-language model alignment \citep{team2024chameleon, ramesh2022hierarchical, gu2022vector, razavi2019generating}, as well as in uni-modal models such as LVM \citep{bai2024sequential} and LLamaGen \citep{sun2024autoregressive}. 

We primarily focus on common tokenizers used for recent vision and language models, and our selection of tokenizers is overviewed in \autoref{tab:tokenizers}. These tokenizers are all VQ-VAE-based, trained on varying datasets, and with various methods. While some recent models such as Transfusion \citep{zhou2024transfusion} and LLaVA \cite{liu2024visual} leverage continuous-valued tokens instead of discrete vocabularies, there is still considerable uncertainty about whether discrete or continuous-valued tokens are more effective \citep{mao2021discrete}. While many of our methods in this paper could apply to continuous tokens through a discrete quantization of those tokens, we leave such continuous extensions to future work. For more details on the tokenizers, see \autoref{app:tokenizers}.

We ground our empirical experiments in several common multi-modal datasets, including Conceptual Captions (12M) \citep{sharma2018conceptual}, MS-COCO \citep{lin2014microsoft}, ILSVRC (ImageNet) \citep{russakovsky2015imagenet} and XM-3600 \citep{thapliyal2022crossmodal}. Each of these datasets has a set of images, and (except ILSVRC) paired text in one or more languages. For more information on the datasets, see \autoref{app:datasets}. 

An example visual sentence from MS-COCO (Image ID: 399655) is given in \autoref{fig:visual_language}. In all of the experiments in this paper, we linearize the tokens using a row-wise scan order (for a detailed discussion on scan-order, see \autoref{app:scan-order}). Such linearization is the de facto standard for turning spatial visual tokens into sequences of discrete tokens for use in learning applications.

\subsection{Token Frequency and Zipf's Law}
\label{sec:token_frequency}

\begin{table}
    \scriptsize
    \centering
    \begin{tabularx}{\linewidth}{lXcc}
    \toprule
    Tokenizer & Application & Resolution & Vocab Size \\
    \midrule
    chameleon-512 \citep{team2024chameleon} & Multimodal Foundation Model & $512 \times 512$ & 8192  \\
    compvis-vq-f8-64 \citep{rombach2022high}      & Image Generation & $64 \times 64$ & 16384 \\
    compvis-vq-f8-256 \citep{rombach2022high}  & Image Generation & $256 \times 256$ & 16384 \\
    compvis-vq-imagenet-f16-1024-256 \citep{esser2021taming} & Image Generation & $256 \times 256$ & 1024 \\
    llamagen-vq-ds16-c2i \citep{sun2024autoregressive} & Text $\to$ Image & $256 \times 256$ & 16384 \\
    \bottomrule
    \end{tabularx}
    \caption{Visual tokenizers that we use in this paper. We select several tokenizers across several applications at varying resolutions and vocab sizes.}
    \label{tab:tokenizers}
\end{table}

The statistics of natural language token distributions have long been studied, beginning with \citet{dewey1921relative}, who first plotted the frequency of English words. A key principle that emerged from this research is Zipf’s Law \citep{kingsley1932selected}, which describes a power-law relationship between the frequency of words and their rank in a language where a small number of high-frequency words dominate natural language, while the majority of words occur infrequently. Formally, Zipf's law states that:
\begin{equation}
    f(r) \propto {r^{\alpha + \sigma Z}}
\end{equation}
where $f(r)$ is the frequency of the element with rank $r$ and $\alpha/\sigma$ parameterize a learned Gaussian distribution (close to 1/0 in many natural languages).

Zipf’s law has been observed across many languages \citep{gelbukh2001zipf, yu2018zipf} and non-human communication systems (such as dolphins \citep{mccowan1999quantitative}). As Mandelbrot pointed out, adherence to Zipf-like distributions ensures that communication systems—whether natural or artificial—operate efficiently \citep{mandelbrot1953contribution}. Language models, especially large language models (LLMs), have been shown to follow this same pattern, with token distributions that obey Zipf’s law \citep{patwary2019language}. This statistical regularity in language extends beyond word frequency - Zipf's law has also been observed in images themselves: \citet{ruderman1997origins} showed that the distribution of object sizes and spatial frequencies in natural scenes follows power-law distributions, and \citet{crosier2007zipf} showed that there was Zipfian behavior in image coding schemes such as JPEG.

\begin{figure}[t]
    \centering
    \begin{subfigure}[b]{0.49\textwidth}
        \centering
        \includegraphics[width=\textwidth]{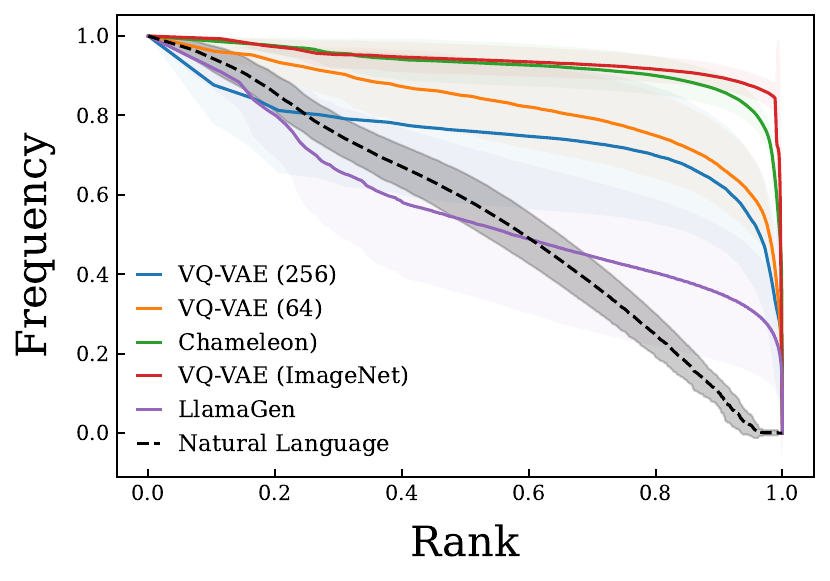}
        \caption{1-grams}
        \label{fig:zipf_ngrams1}
    \end{subfigure}
    \hfill
    \begin{subfigure}[b]{0.49\textwidth}
        \centering
        \includegraphics[width=\textwidth]{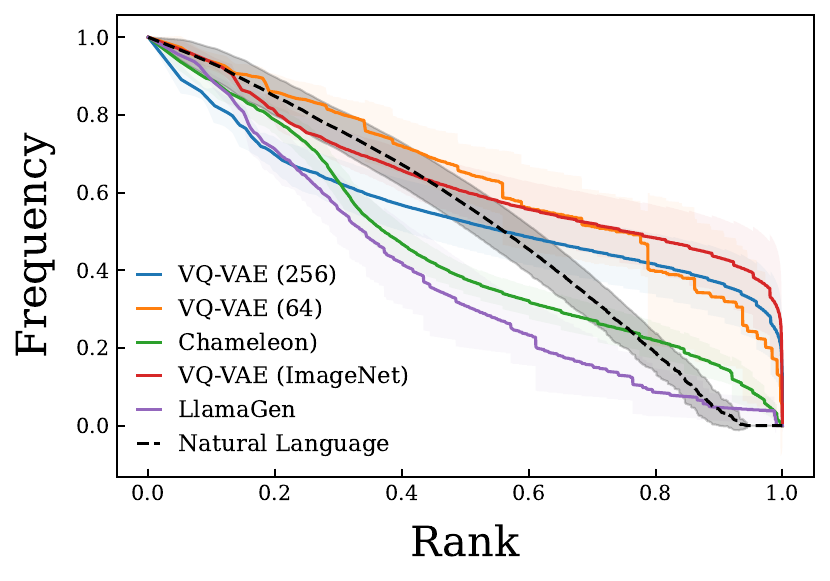}
        \caption{2-grams}
        \label{fig:zipf_ngrams2}
    \end{subfigure}
    \hfill
    \begin{subfigure}[b]{0.49\textwidth}
        \centering
        \includegraphics[width=\textwidth]{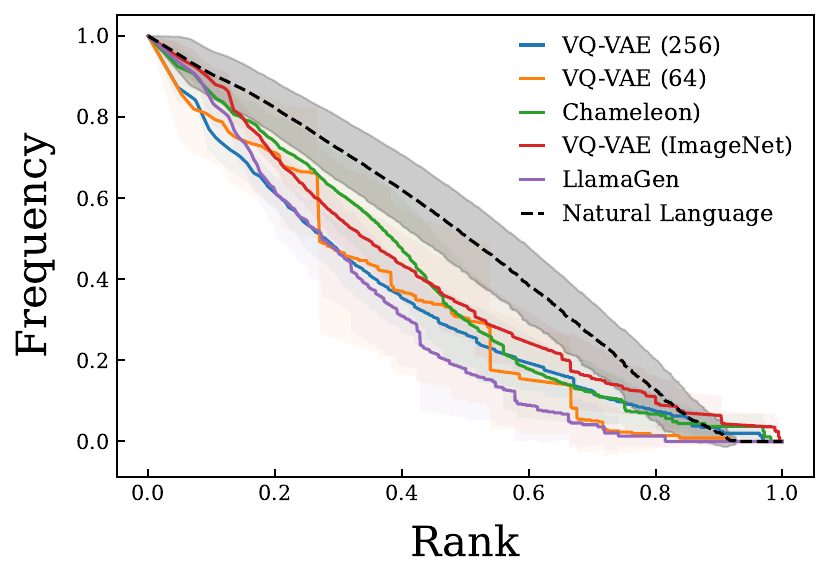}
        \caption{3-grams}
        \label{fig:zipf_ngrams3}
    \end{subfigure}
    \hfill
    \begin{subfigure}[b]{0.49\textwidth}
        \centering
        \includegraphics[width=\textwidth]{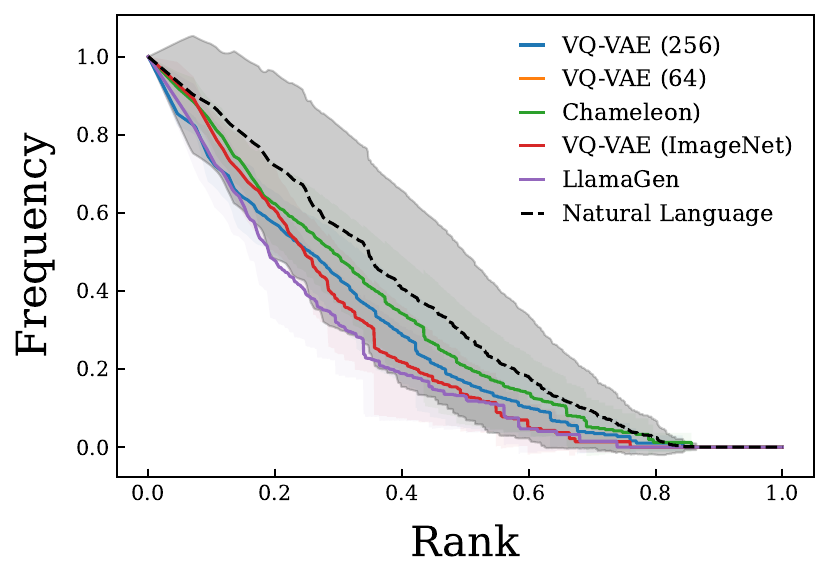}
        \caption{7-grams}
        \label{fig:zipf_ngrams7}
    \end{subfigure}
    \hfill
    \caption{Plot of normalized token log-frequency against normalized Log-Rank for several visual and textual languages for different n-grams, aggregated across datasets. While the tails of visual languages do not conform to Zipf's law well for small values of $N$, for larger values of $N$, the fit becomes more linear.  }
    \label{fig:zipf_n}
\end{figure}

Thus, we first ask the question - \textbf{Do ``visual languages'' follow Zipf's law?} To do this, we tokenize the image datasets according to \autoref{sec:preliminaries} and compute the empirical token-rank frequency distributions on each of the datasets (See \autoref{app:zipf} for details). We show the empirical distributions in \autoref{fig:zipf_n}. If the plots were Zipfian, we would expect them to be linear in the log-log space; while this is the case for natural languages, visual languages do not seem to generally conform to a linear curve, instead, for one and two grams, the plots follow a lognormal distribution, and for higher level N-grams are more convex in nature.

For one/two-grams, this indicates that token utilization is fairly uniform, with most tokens occurring in equal proportion, and the heavier tails of the distribution indicate that ``rare'' are, in practice, not so rare, occurring with much higher frequency than expected under a power-law distribution. Whereas natural languages are often structured with a clear core vocabulary and then more specialized words, it seems like visual features seem to be more evenly distributed, with many features or combinations being equally likely. At higher n-grams, for visual languages there is more convex behavior, suggesting that there are very few common n-grams, instead, n-grams are often unique, and composed in ways that appear very infrequently within the datasets. Such an implication implies that visual languages are highly context-dependent (which is sensible, as visual scenes are quite complex).

To confirm these details, we fit a Zipf's distribution to each of the models, with the results of the fit shown in \autoref{tab:zipf_fit}. Interestingly, the $\alpha$ values have opposite behaviors for visual and natural languages in the light of increasing $N$. In natural languages, the fact that $\alpha$ increases with $N$ means that higher-order N-grams follow steeper power-law distributions, and the distribution of N-gram frequencies becomes more concentrated around a few common combinations, while the frequency of rare combinations decreases rapidly. In visual languages, on the other hand, the decrease in $\alpha$ with increasing $N$ suggests that higher-order combinations of visual features follow flatter distributions: as visual N-grams increase in complexity, there is more diversity in the combinations of features and patterns, leading to richer and more distributed sets of higher-order feature combinations.

These phenomena together suggest that VQ-VAEs are ``spreading'' information between the independent tokens, rather than building compressive and compositional structures, which we explore further in \autoref{sec:token_frequency} (token innovation) and \autoref{sec:entropy} (compression). Indeed, since Zipf’s Law reflects a (theoretically optimal) balance between redundancy and information, it suggests that visual languages are more data-driven, and reflect the underlying complexity and variability of visual scenes, rather than focusing on reducing redundancy for communicative operations. Such a deviation might suggest that models that are more Zipfian, such as chameleon, may be better placed as embedding/alignment models for visual tasks, whereas models such have more convex N-gram distributions are better for high-fidelity generation tasks. 

Beyond model quality/applicability implications, the fact that visual languages don’t follow Zipf’s Law implies that traditional NLP-inspired techniques (e.g., those relying on power-law distributions such as compression algorithms, or memory-based systems based on Zipfian patterns) may not directly apply to visual languages. Beyond this, visual languages likely require different optimization techniques taking into account the non-linear distribution of N-grams -- methods that handle long-tail distributions might be more appropriate than techniques focused on heavy tails. Such differences in distribution could also suggest that higher-order interactions between visual features are more important in vision models than in language models, and model architectures should be designed to capture these higher-order patterns effectively.

\begin{table}[t]
\scriptsize
\centering
\resizebox{\linewidth}{!}{%
\begin{tabular}{lcccccccccc}
\toprule
 & \multicolumn{2}{c}{N=1} & \multicolumn{2}{c}{N=2} & \multicolumn{2}{c}{N=3} & \multicolumn{2}{c}{N=5} & \multicolumn{2}{c}{N=7} \\
\cmidrule(lr){2-3} \cmidrule(lr){4-5} \cmidrule(lr){6-7} \cmidrule(lr){8-9} \cmidrule(lr){10-11}
 & Natural & Visual & Natural & Visual & Natural & Visual & Natural & Visual & Natural & Visual \\
\midrule
$\alpha$ & 1.71$_{ 0.23}$ & 4.37$_{ 1.33}$ & 1.99$_{ 0.25}$ & 4.43$_{ 1.50}$ & 2.28$_{ 0.33}$ & 2.57$_{ 0.82}$ & 2.85$_{ 0.82}$ & 2.35$_{ 0.52}$ & 3.02$_{ 0.73}$ & 2.35$_{ 0.50}$ \\
$\sigma$ & 0.01$_{ 0.02}$ & 0.18$_{ 0.14}$ & 0.01$_{ 0.01}$ & 0.07$_{ 0.16}$ & 0.03$_{ 0.04}$ & 0.09$_{ 0.18}$ & 0.25$_{ 0.54}$ & 0.09$_{ 0.13}$ & 0.28$_{ 0.46}$ & 0.09$_{ 0.14}$ \\
$\overline{\log \mathcal{L}}$ & -4.03$_{ 1.38}$ & -9.72$_{ 3.28}$ & -3.11$_{ 0.41}$ & -4.24$_{ 2.27}$ & -2.92$_{ 0.42}$ & -3.53$_{ 1.68}$ & -2.43$_{ 0.67}$ & -2.99$_{ 1.22}$ & -1.98$_{ 1.02}$ & -2.72$_{ 1.19}$ \\
\bottomrule
\end{tabular}
}
\caption{Comparison of aggregate power law fit metrics ($\alpha,\sigma$, mean log-likelihood) across different N-gram lengths for natural and visual languages. While visual languages do not follow Zipf's law for $N=1$, the fit is significantly better for $N=3$ and above.}
\label{tab:zipf_fit}
\end{table}

\subsection{Token Innovation}
\label{sec:token_innovation}

\begin{figure}[t]
    \centering
    \begin{subfigure}[b]{0.49\textwidth}
        \centering
        \includegraphics[width=\textwidth]{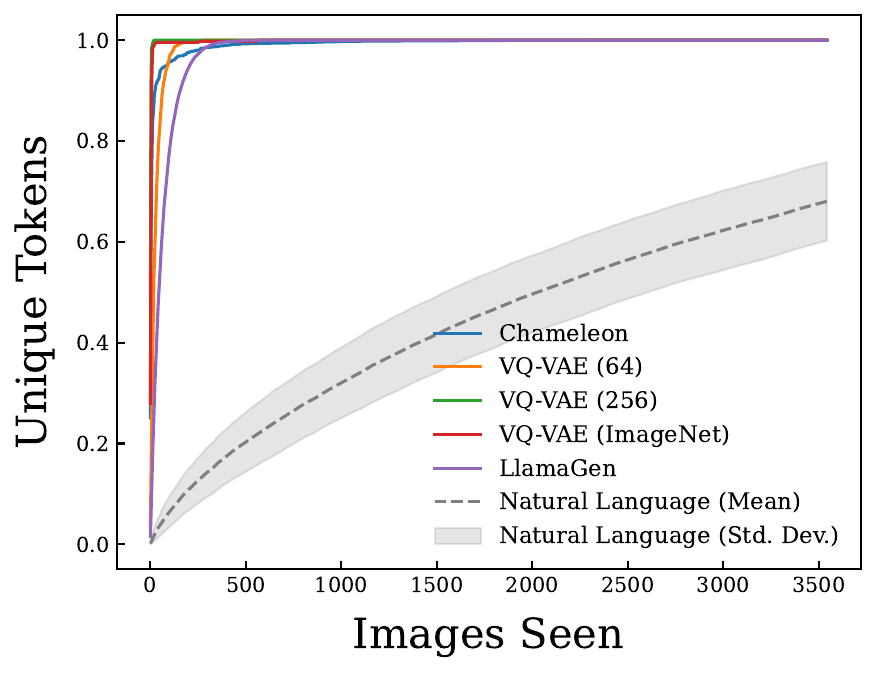}
        \caption{1-grams}
        \label{fig:heaps_ngrams1}
    \end{subfigure}
    \hfill
    \begin{subfigure}[b]{0.49\textwidth}
        \centering
        \includegraphics[width=\textwidth]{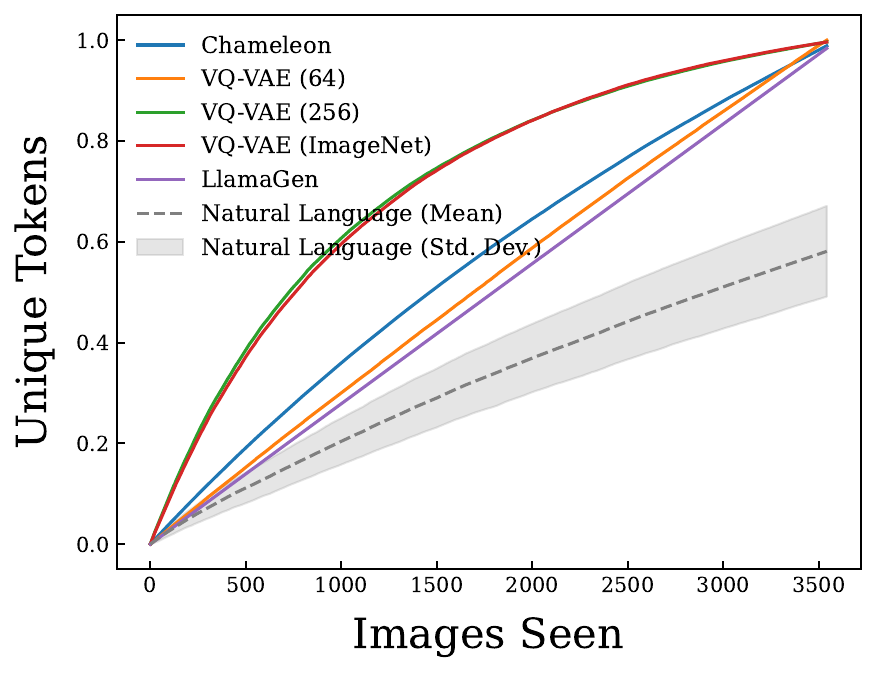}
        \caption{2-grams}
        \label{fig:heaps_ngrams2}
    \end{subfigure}
    \hfill
    \begin{subfigure}[b]{0.49\textwidth}
        \centering
        \includegraphics[width=\textwidth]{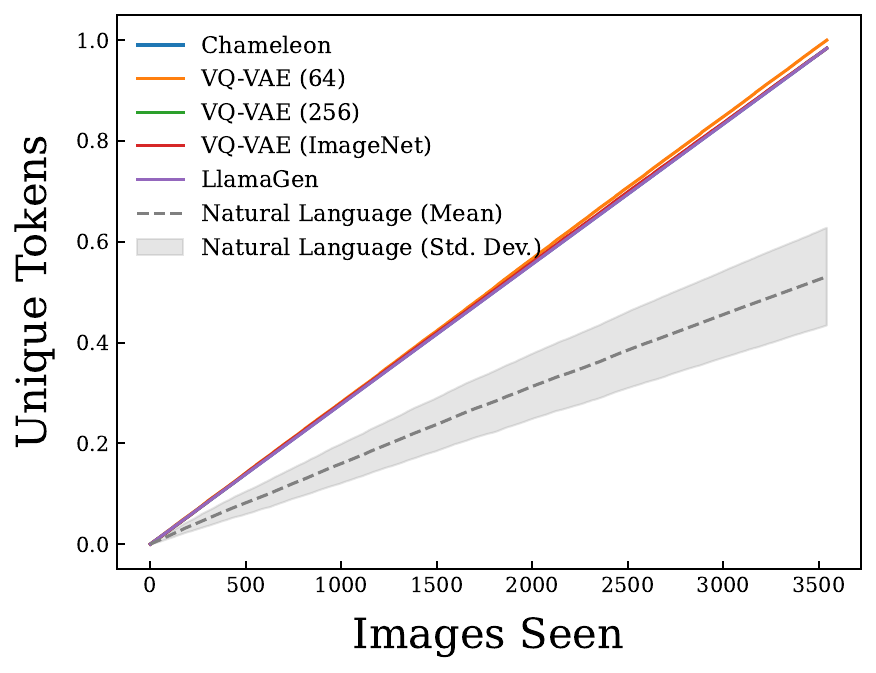}
        \caption{3-grams}
        \label{fig:heaps_ngrams3}
    \end{subfigure}
    \hfill
    \begin{subfigure}[b]{0.49\textwidth}
        \centering
        \includegraphics[width=\textwidth]{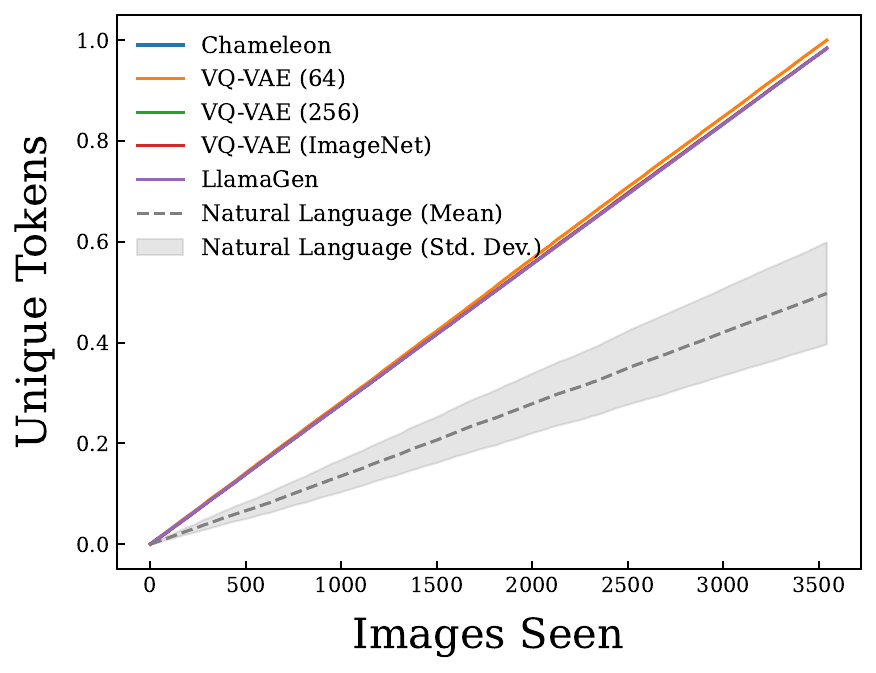}
        \caption{5-grams}
        \label{fig:heaps_ngrams4}
    \end{subfigure}
    
    \caption{Comparison of unique tokens as a function of images seen on the XM-3600 dataset for different N-grams. While higher values of N approach a linear relationship in the visual languages, textual languages are always sub-linear in their growth. Surprisingly, for 3/5-grams, several visual language curves overlap.}
    \label{fig:heaps_xm3600}
\end{figure}

One thing that stands out from the experiments in \autoref{sec:token_frequency} is that single visual tokens appear more uniformly than single words, inspiring the question: do new images consist of mostly new tokens, or do new images re-combine existing tokens in novel ways? In natural language, this has generally been codified by Heaps'/Herdan's law \citep{herdan1964quantitative, heaps1978information}, which says that vocabularies’ sizes are concave increasing power laws of texts’ sizes (See \autoref{app:heaps} for details).

To explore this effect, \autoref{fig:heaps_xm3600} plots the number of unique tokens seen against the number of images seen for the XM-3600 dataset for several visual tokenizers and natural languages. The natural languages follow the expected distribution, with unique tokens increasing sub-linearly with respect to the number of images. The visual tokens, on the other hand, appear much more rapidly. For single tokens, almost all of the tokens in the vocabulary appear within the first 100 images, suggesting that the rate of token innovation is significantly higher than that of natural languages. For 2-grams and 4-grams, the relationship trends linear, but never approaches the sub-linear behavior that is expected of generative systems which follow Heaps' law. Additional experiments on MS-COCO are given in \autoref{app:heaps}. 

\begin{figure}[t]
    \centering
    \begin{subfigure}[b]{0.49\textwidth}
        \centering
        \includegraphics[width=\textwidth]{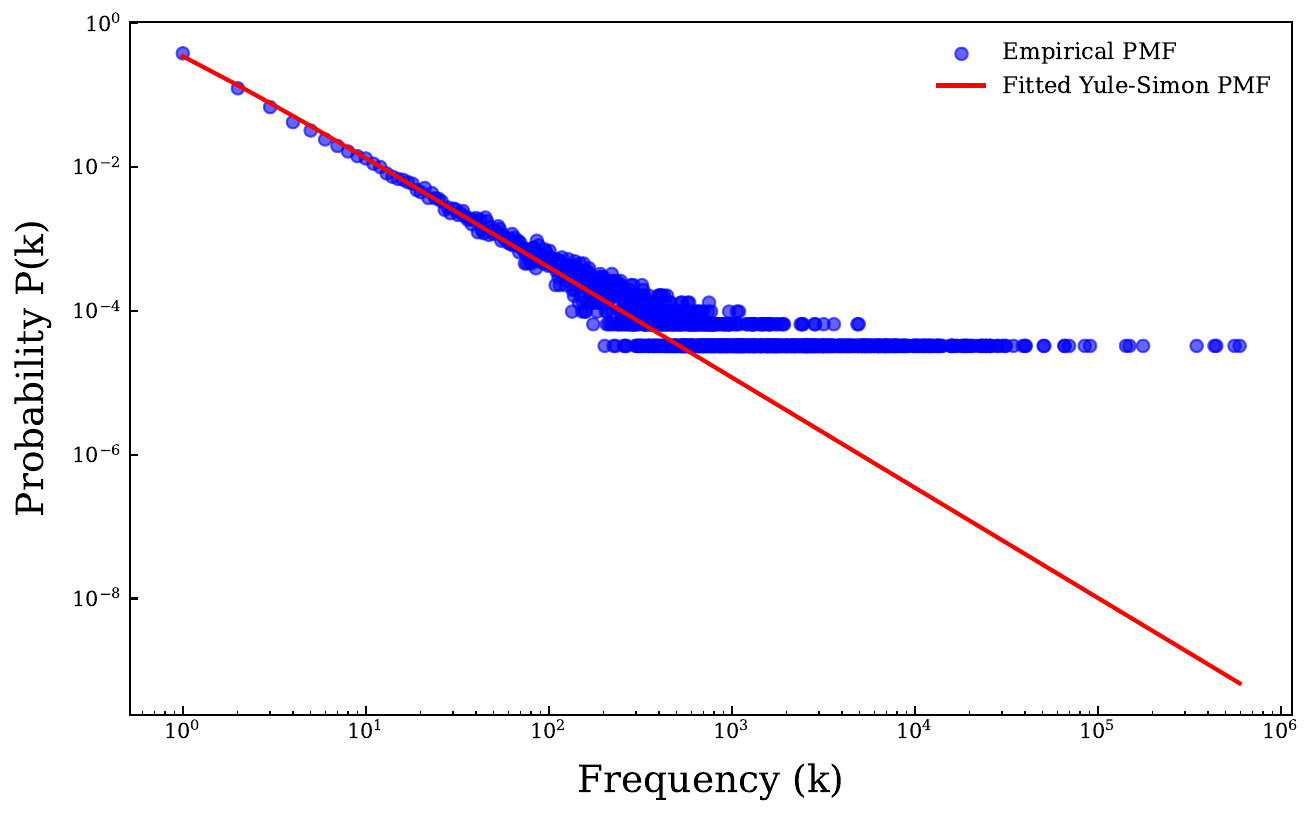}
        \caption{Spanish (1-grams)}
        \label{fig:ngrams1}
    \end{subfigure}
    \hfill
    \begin{subfigure}[b]{0.49\textwidth}
        \centering
        \includegraphics[width=\textwidth]{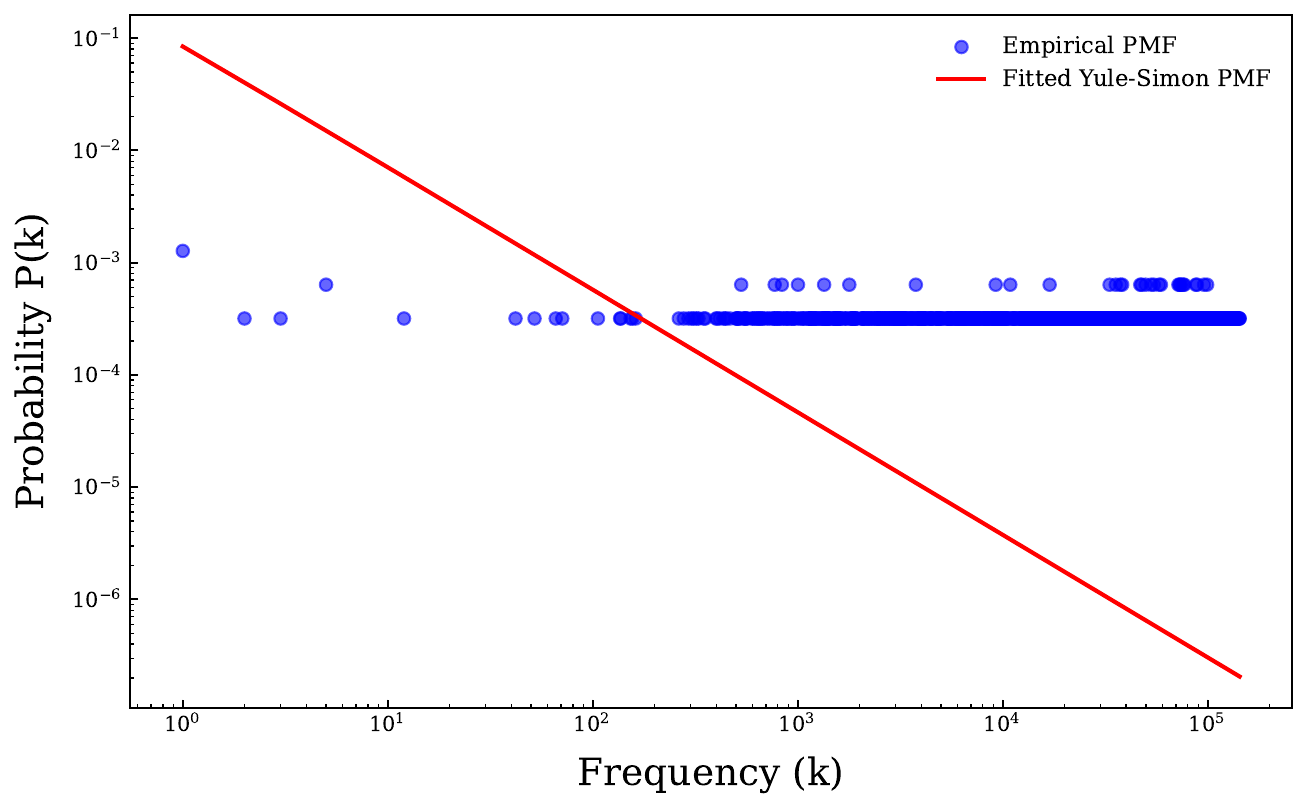}
        \caption{Chameleon (1-grams)}
        \label{fig:ngrams2}
    \end{subfigure}
    \hfill
    \begin{subfigure}[b]{0.49\textwidth}
        \centering
        \includegraphics[width=\textwidth]{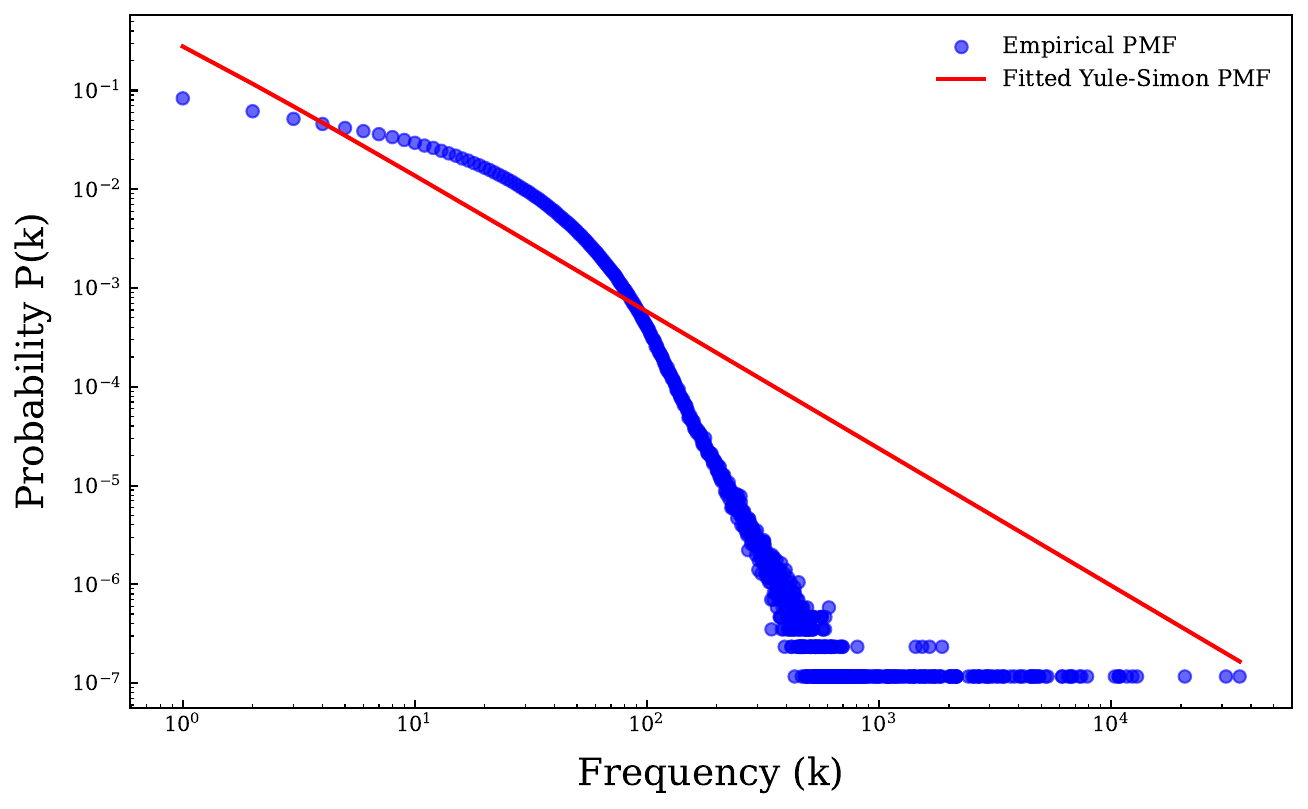}
        \caption{Chameleon (2-grams)}
        \label{fig:ngrams4}
    \end{subfigure}
    \hfill
    \begin{subfigure}[b]{0.49\textwidth}
        \centering
        \includegraphics[width=\textwidth]{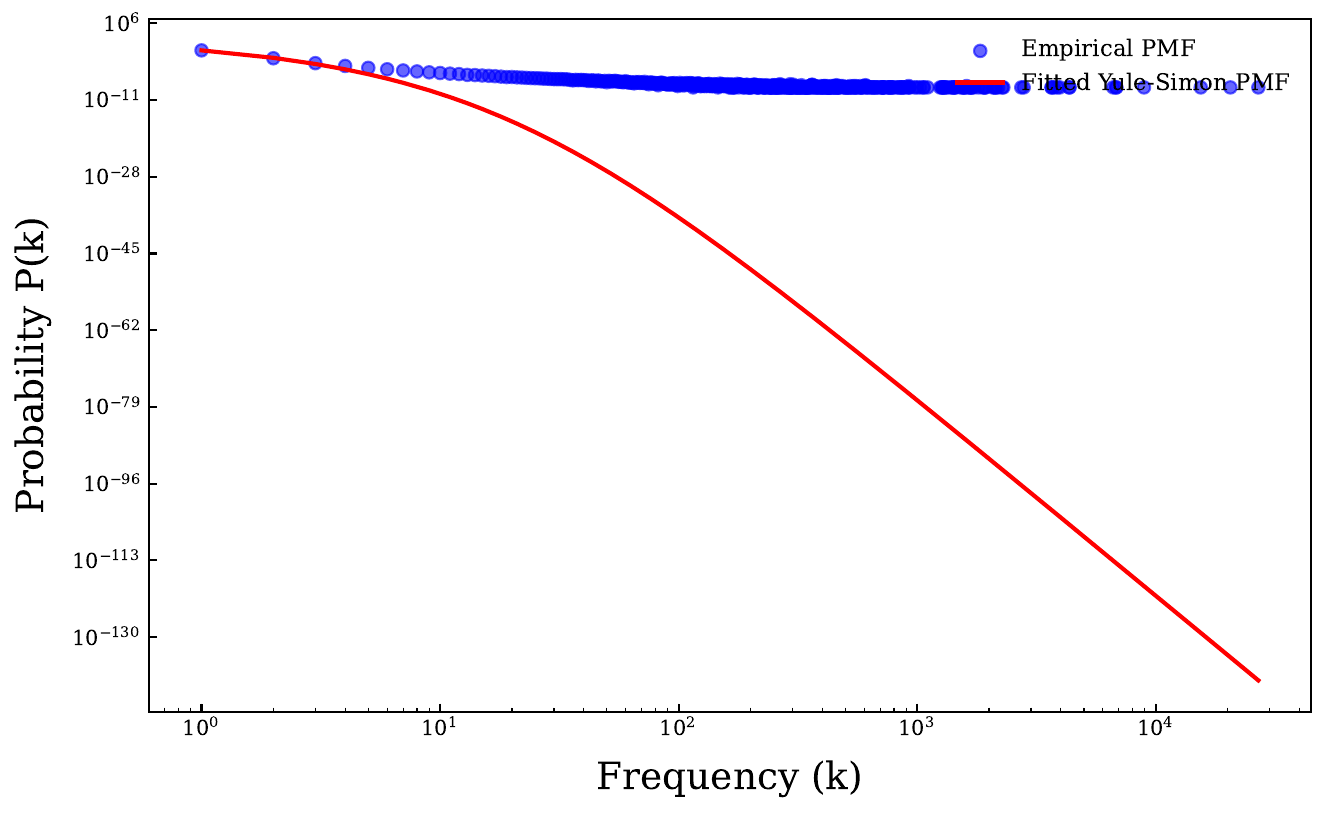}
        \caption{Chameleon (3-grams)}
        \label{fig:ngramsX}
    \end{subfigure}
    
    \caption{Yule-Simon model fit for Chameleon vs. Spanish on the COCO dataset (More models/languages in \autoref{app:yule-simon}). While Spanish (and in general, natural languages) largely fits a Yule-Simon model, Chameleon does not appear to be generated by such a process at any n-gram level. }
    \label{fig:yule-simon}
\end{figure}

We further fit a Yule-Simon distribution \citep{simon1955class} to both the natural and visual languages. The Yule-Simon process is a stochastic model for generating sequences of words or tokens, where the probability of introducing a new token decreases as more tokens are added, leading to a power-law distribution; mathematically, this process is governed by a probability proportional to the current token frequency, combined with a parameter that controls the rate of new token introduction (see \autoref{app:yule-simon} for more details). The results, given in \autoref{fig:yule-simon} and \autoref{app:yule-simon}, demonstrate that the generative process for new tokens largely does not fit with that described by the Yule-Simon process in the visual case, however, fit quite well for many text languages.

The fact that visual tokens have a much higher rate of innovation has several key implications for the design, training, and evaluation of both generative and discriminative models. The high vocabulary diversity of visual tokens means that while generative models will be able to generate higher-fidelity output, discriminative models are at high risk of over-fitting: risking overly specific captions or inconsistencies across similar images (a feature that has already been noted in several works \citep{chan2022s,caglayan2020curious}). Such high vocab diversity also impacts the training efficiency of models: both generative and discriminative models will require longer training times and need more varied datasets to handle expanding token sets than models of natural language (a fact which has been observed explicitly in \citep{touvron2021training}, and more generally with vision transformers). Beyond training, inference and evaluation are also impacted. Decoding approaches that rely on frequency/presence penalties may want to leverage unique/more aggressive penalties for vision compared to language tokens. For evaluation, perhaps already clear from existing work, semantic-based evaluation is likely more effective than token-based evaluation in visual approaches due to the high level of diversity in the local token space \citep{anderson2016spice,hessel2021clipscore}.

\subsection{Naturality}
\label{sec:benfords}

Benford's Law describes the frequency distribution of leading digits in naturally occurring datasets, where smaller digits like 1 and 2 appear disproportionately more often than larger digits like 8 and 9 \citep{benford1938law}. Originally observed in domains such as physics \citep{sambridge2010benford}, economics \citep{todter2009benford}, and demographics \citep{miller2015benford}, recently, there has been growing interest in extending this statistical principle to linguistic data \citep{golbeck2023benford, melian2017zipf,hong2010benford}. One of the primary applications of Benford's law is the detection of anomalies in data: datasets that do not follow Benford's law are likely to be unnatural in nature - here, we ask the question, do visual language token frequencies naturally follow Benford's law? We follow a similar tokenization process to \autoref{sec:token_frequency}, and plot the occurrence of leading digits in the token frequency distribution (See \autoref{app:benfords} for more details). 

\begin{figure}[t]
    \centering
    \begin{subfigure}[b]{0.32\textwidth}
        \centering
        \includegraphics[width=\textwidth]{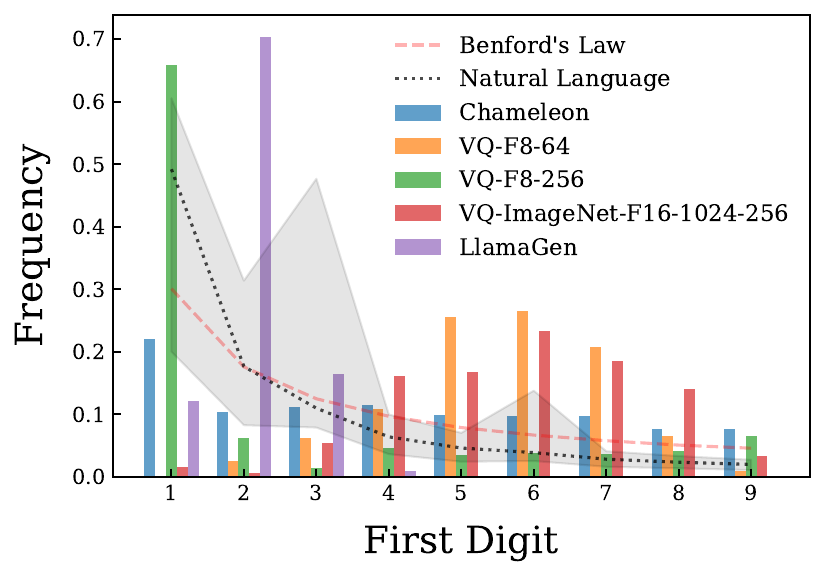}
        \caption{1-grams}
        \label{fig:benfords_ngrams1}
    \end{subfigure}
    \hfill
    \begin{subfigure}[b]{0.32\textwidth}
        \centering
        \includegraphics[width=\textwidth]{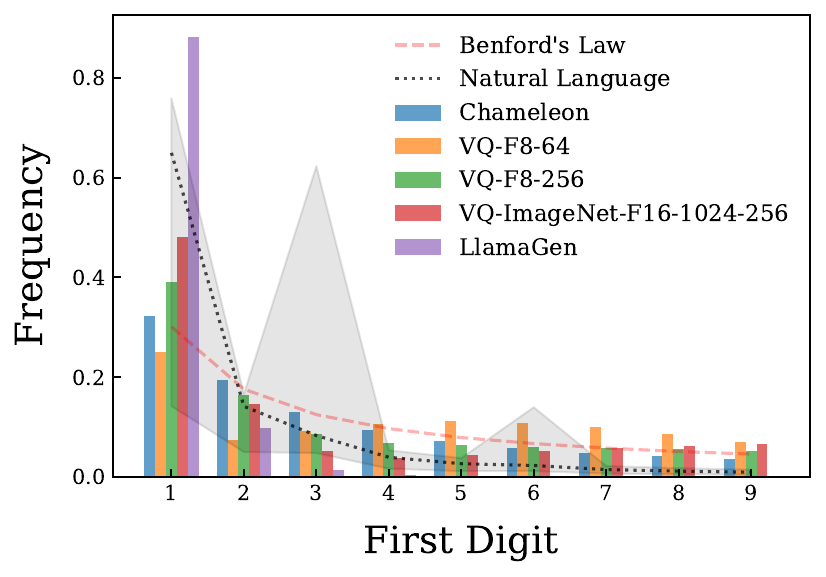}
        \caption{2-grams}
        \label{fig:benfords_ngrams2}
    \end{subfigure}
    \hfill
    \begin{subfigure}[b]{0.32\textwidth}
        \centering
        \includegraphics[width=\textwidth]{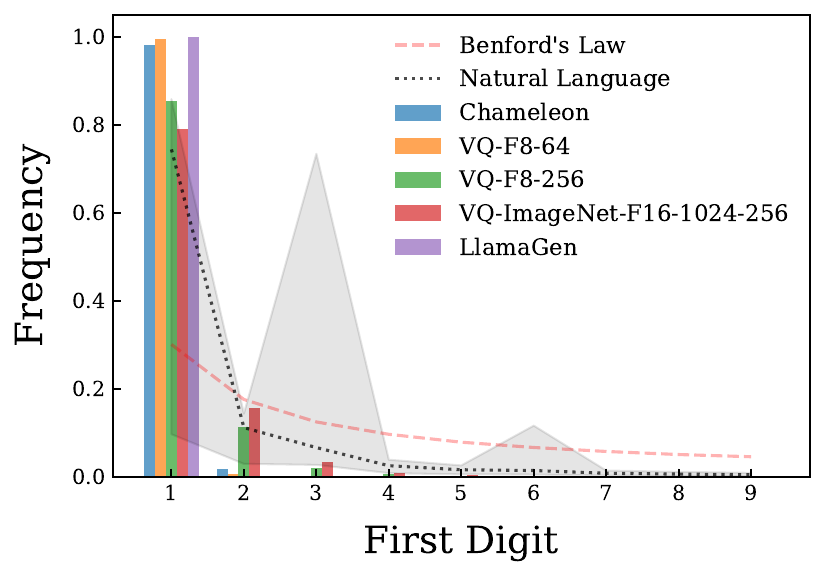}
        \caption{3-grams}
        \label{fig:benfords_ngrams3}
    \end{subfigure}
    
    \caption{Plot of the first digits of the token frequency distribution on the MS-COCO dataset. While 1-grams have a uniquely 1-heavy head, with a Gaussian tail (around 6), 2-grams naturally follow an exponential decay function, and 3-grams are dominated by unique tokens.  The grey area represents the maximum and minimum among the 36 natural languages.}
    \label{fig:benfords_coco}
\end{figure}

Our results are shown in \autoref{fig:benfords_coco} for the MS-COCO dataset, and in \autoref{app:benfords} on other datasets. Interestingly, for single tokens, the distribution is unique-token-heavy, with the remaining tokens having a Gaussian distribution around six. Two-grams are the most natural, with Chameleon following Benford's law almost exactly, with three-grams significantly dominated by low/unique frequency tokens. Interestingly, the highest quality tokenizer, the Chameleon tokenizer, is by far the most natural in \autoref{fig:benfords_ngrams1}, suggesting that tokenizations performing well for vision-text tasks might have more natural distributions. Beyond this effect, \autoref{fig:benfords_ngrams2} shows that distributions of visual-token bi-grams have the most natural distribution curves, implying a potential correspondence in statistics between visual bi-grams and text uni-grams, and suggesting that future work in tokenization could explore vocabularies of token bi-grams or bi-gram compression for vision tokenizers.

\subsection{Entropy and Redundancy}
\label{sec:entropy}

\begin{table}
\centering
\caption{Understanding the entropy and Huffman compression rates of visual and natural languages ($p < 0.01$ across all metrics). While the compression rate improves slightly with two-grams in the visual case, it is reduced significantly in the natural case. Full results in \autoref{tab:huffman_full}.}
\label{tab:huffman}
\resizebox{\linewidth}{!}{
\begin{tabular}{lccccccc}
\toprule
Language & Avg Code Length & Entropy & Fixed Code Length & Orig Bits (M) & Huff Bits (M) & Comp. Rate & \% Reduction \\
\midrule
Visual & 10.7 \(\pm\) 1.9 & 10.7 \(\pm\) 1.9 & 11.0 \(\pm\) 1.8 & 5.35 \(\pm\) 1.3 & 5.20 \(\pm\) 1.3 & 1.03 \(\pm\) 0.02 & 2.9 \(\pm\) 1.9 \\
Visual (N=2) & 18.1 \(\pm\) 0.8 & 18.1 \(\pm\) 0.8 & 18.7 \(\pm\) 0.5 & 9.1 \(\pm\) 1.5 & 8.8 \(\pm\) 1.5 & 1.03 \(\pm\) 0.02 & 3.2 \(\pm\) 2.2 \\
\midrule
Natural  & 9.0  \(\pm\) 0.9 &  8.9 \(\pm\) 0.9 & 13.8 \(\pm\) 0.9 & 4.10 \(\pm\) 3.1 & 2.54 \(\pm\) 1.8 & 1.55 \(\pm\) 0.1  & 34.9 \(\pm\) 6.1 \\
Natural (N=2) & 13.5 \(\pm\) 1.0 & 13.5 \(\pm\) 1.0 & 16.3 \(\pm\) 1.1 & 4.9 \(\pm\) 3.8 & 3.9 \(\pm\) 3.0 & 1.21 \(\pm\) 0.08 & 16.9 \(\pm\) 5.2 \\
\bottomrule
\end{tabular}
}
\end{table}

Building on the foundational work of \citet{shannon1951prediction}, entropy and redundancy have long been understood as key characteristics of natural language, providing insight into its inherent predictability and compressibility. While natural languages, like English, exhibit high redundancy that enables efficient encoding, it is unclear if visual languages might have similar coding behaviors. To evaluate the efficiency of encoding, we use a similar setup to \autoref{sec:token_frequency} and extract token streams for each of the target datasets. We then compute the entropy of the token streams, as well as compute a simple Huffman code/compression \citep{huffman1952method} for each of the resulting streams. Such a hierarchical compression code allows us to estimate the overall ``compressibility'' of the stream (See \autoref{app:huffman} for background/details).

The results are summarized in \autoref{tab:huffman}. We can see that in general, the average code length, entropy, and bits of information/sample are higher for visual languages. This suggests that visual languages have more variability and are inherently more complex to predict and encode than natural language. This is unsurprising, given the complexity and richness of the visual world, compared to the sparsity of natural language, however, it is somewhat surprising that the entropy is not massively different from natural languages, suggesting that visual tokenizers are capable of reducing the richness of natural language to suitably sparse representations for reasoning. Notably different is the ``compressibility'' of the token streams. While natural language tokens are highly compressible using Huffman encoding, visual languages are almost incompressible, suggesting that information is highly distributed amongst the tokens and that there is very little structural reuse between the different images. While we explore grammars further in \autoref{sec:grammars}, this experiment indicates that it is unlikely that models have non-trivial grammars of tokens, instead, these tokens are more local, and particularly high-variance.

These experiments have several potential implications for model design. First, since visual tokens have significantly higher entropy and lower compressibility, it may be necessary to use more attention heads, deeper models, and more dense embeddings, in visual-based models in order to capture a sufficient number of relationships and higher-level representations of visual information. Models like LLaVA \citep{liu2024visual} with simple projection layers between the visual token and text token spaces may not perform as well on downstream visual tasks as models such as mPlug \citep{ye2024mplug} which have more dense transformer-based adapters (a result which is empirically verified by \cite{tong2024cambrian}, who leverage a spatially aware dense connector to achieve significant performance improvements).

It's worth noting that Huffman encoding is independent of the scan order of the images, and instead, focuses only on token frequencies. It would be interesting for future work to explore how scan order impacts compress-ability, and we discuss potential experiments and limitations regarding scan order in \autoref{app:scan-order}.

\subsection{Token Segmentation Granularity}
\label{sec:segmentation}

\begin{table}[t]
    \scriptsize
    \caption{Whole, part, and sub-part purity/part-normalized mutual information on the SPIN dataset. PP: Part Purity (\%), VTP: Visual-Token Purity (\%), PNMI: Part-Normalized Mutual Information.}
    \label{tab:purity}
    \begin{tabularx}{\linewidth}{X ccc ccc ccc}
    \toprule
    \multirow{3}{*}{Tokenizer} & \multicolumn{3}{c}{Wholes} & \multicolumn{3}{c}{Parts} & \multicolumn{3}{c}{Sub-Parts} \\
    \cmidrule(lr){2-4} \cmidrule(lr){5-7} \cmidrule(lr){8-10}
    & PP & VTP & PNMI & PP & VTP & PNMI & PP & VTP & PNMI \\
    \midrule
    chameleon-512 & 2.512 & 0.216 & 1.557 & 4.399 & 0.138 & 0.256 & 1.660& 0.200& 0.898 \\
compvis-vq-f8-64 & 3.061 & 0.526 & 6.148 & 5.653 & 0.308 & 1.760 & 2.611& 0.508& 6.246 \\
compvis-vq-f8-256 & 2.333 & 0.467 & 0.925 & 4.209 & 0.334 & 0.122 & 1.527& 0.426& 0.434 \\
compvis-vq-imagenet & 2.467 & 0.739 & 1.463 & 4.354 & 0.479 & 0.207 & 1.626& 0.623& 0.787 \\
llamagen-vq-ds16-c2i & 4.384 & 0.107 & 13.711 & 6.983 & 0.057 & 4.487 & 3.656& 0.112& 13.273 \\
    \bottomrule
    \end{tabularx}
\end{table}

One common question for many vision researchers is: ``do visual tokens represent objects?'' Indeed, while visual tokens are spatially fixed to patches in the image, because of the VQ-VAE training process, it is unclear if they take on additional non-spatial semantic meaning. Recently, \citet{hsu2021hubert} demonstrated that in audio domains, HuBERT tokens (audio-tokens) have relatively high mutual information with phoneme representations of audio, suggesting that self-supervised models are capable of learning natural structures despite being segmented to fixed-width patches. Can we answer this question for visual languages as well? 

Recently, \citet{myers2024spin} introduced the SPIN dataset, a new labeled dataset of hierarchically segmented objects, where the objects are labeled at the whole, part, and sub-part levels. This gives us per-image annotations of the existence of wholes, parts, and sub-parts. From this, we compute several measures of natural correlation between these part-annotations and the visual token languages, inspired by \citet{hsu2021hubert} (For more details, see \autoref{app:purity}): \textbf{Part Purity}, a metric that measures the average accuracy of assigning a visual-token to its most likely part label, reflecting image-level part consistency within a particular visual token, \textbf{Visual-Token Purity}, a metric that assesses how well images containing the same part label are consistently assigned to the same visual-tokens and \textbf{Part-Normalized Mutual Information}, an information-theoretic metric which measures the percentage of uncertainty about the part-label eliminated after observing a particular visual token.

The results are summarized in \autoref{tab:purity}. In general, tokenizers appear to be most effective at capturing part-level representations, as evidenced by consistently higher Part Purity (PP) values for parts compared to wholes or sub-parts across all models. This suggests that tokenizers are better aligned with mid-level structures (parts), rather than whole objects or fine-grained sub-parts. However, Visual-Token Purity (VTP) remains low across all models and levels of granularity, indicating that images containing the same part-label are not consistently assigned to the same visual tokens, reflecting fragmentation in the clustering. PNMI values are generally higher for sub-parts than for parts or wholes, particularly in models like \texttt{llamagen-vq-ds16-c2i}, which shows the highest PNMI across all levels. This implies that tokenizers can capture more fine-grained information at the sub-part level, though the corresponding decrease in part purity for sub-parts suggests that while they can reduce uncertainty about part labels, their actual clustering of sub-parts is inconsistent. 
\section{Are Visual Languages Structured Like Natural Languages?}
\label{sec:grammars}

In \autoref{sec:entropy} we showed that visual languages are not very compressible using Huffman encodings, suggesting that visual languages may not have hierarchical structures similar to those of natural languages. 
To inquire further into this question, we test whether Context-free Grammars~\citep{chomsky1959algebraic} can approximate the structure of visual languages as well as they can natural languages by fitting grammars to each modality using unsupervised grammar induction techniques. 

Particularly, we use Compound Probabilistic Context-Free Grammars (C-PCFG)~\citep{kim-etal-2019-compound} as the grammar formalism for our experiments. 
C-PCFGs are a type of neural PCFG, where grammar production rules are modeled as compound probability distributions~\citep{robbins1956empirical} -- every production depends on both the set of symbols in the grammar as well as a global latent variable $z$. 
This formulation, trained with variational methods, allows for global sentence information to flow through all parsing decisions in a sentence while remaining compatible with efficient inference methods which standard PCFGs enjoy \citep{baker1979trainable}. 
For more details on C-PCFGs see Appendix \ref{app:cpcfgs}.

\begin{figure}[t]
    \centering
    \begin{subfigure}[c]{0.46\linewidth} %
        \centering
        \scriptsize
        \begin{tabularx}{\linewidth}{Xccccc}
            \toprule
            Dataset & PPL & PPL-R & MBF & FR & CU \\
            \midrule
            COCO-DE & 24.70 & 99.61\% & 3.00 & 2.44 & 1.00 \\
            COCO-EN & 25.18 & 99.62\% & 3.02 & 2.43 & 1.00 \\
            \midrule
            COCO-VQ & 671.93 & 95.80\% & 1.41 & 1.75 & 0.97 \\
            XM3600 & 739.37 & 95.40\% & 6.65 & 6.39 & 0.33 \\
            CC12M & 595.01 & 96.28\% & 2.85 & 2.54 & 1.00 \\
            ILSVRC & 654.25 & 95.92\% & 1.93 & 2.28 & 0.93 \\
            SPIN & 656.61 & 95.89\% & 1.27 & 1.82 & 0.73 \\
            \bottomrule
        \end{tabularx}
        \caption{Generated parse tree statistics}
        \label{tab:grammar_stats}
    \end{subfigure}%
    \hspace{0.05\linewidth} %
    \begin{subfigure}[c]{0.45\linewidth} %
        \centering
        \includegraphics[width=\linewidth]{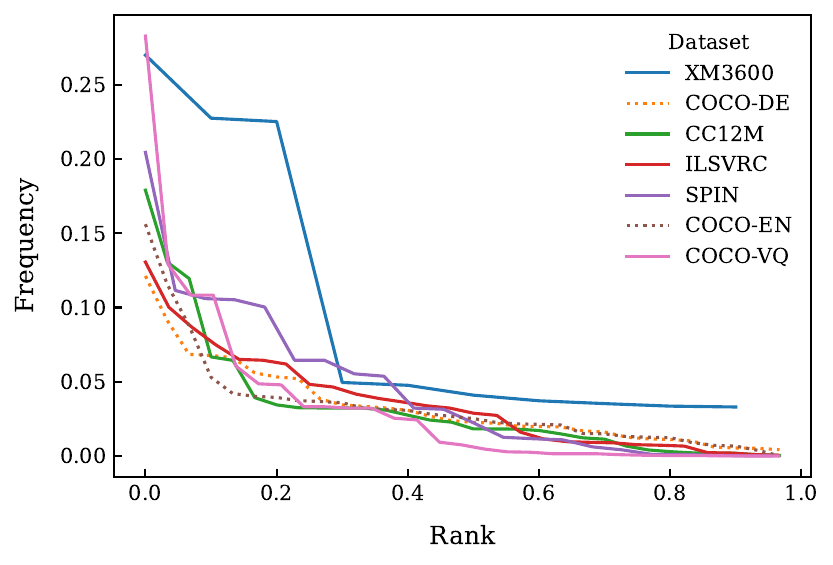} %
        \caption{Parse tree non-terminal node frequencies}
        \label{fig:grammar_freqs}
    \end{subfigure}
    \caption{Comparison between C-PCFG grammars trained on textual and visual languages. Grammars learned over text exhibit greater reduction in perplexities (PPL-R). Both modalities present comparable parse tree heights (FR), right-branching propensity (MBF),  Non-terminal codebook utilization (CU), and non-terminal node label frequencies (b). \vspace{-1em}}
\end{figure}

C-PCFG memory costs are cubic on sentence length, leading us to use the \verb|compvis-vq-f8-64| tokenizer for visual grammars, which provides a tractable 32 tokens per image. 
For each dataset, we train grammars over five seeds for 15 epochs and select the seed with the lowest test set perplexity for analysis. 
We test our pipeline by evaluating parsers learned on English COCO captions (COCO-EN) against silver-label parse trees extracted with Benepar~\citep{kitaev-klein-2018-constituency}, attaining an F1 score of 49 on the best seed, which is comparable to prior work~\citep{zhao-titov-2020-visually}. 

We report test set statistics over learned grammars, such as final parse tree perplexity (PPL) and percentage reduction in perplexity (PPL-R) from random initialization to convergence. 
The mean branching factor~\citep{li-etal-2024-evaluating} (MBF) measures on average whether generated parse trees tend to branch right or left. 
This is achieved by averaging the proportion of leaves between the right and left branches of nodes $n$ across parse trees $t$ in the dataset: 
\begin{equation}
\text{MBF}(t) = \frac{1}{|t|}\sum_{n\in t}\frac{\text{CR}(n)}{\text{CL}(n)}
\end{equation}
Here, CR and CL represent the counts of leaves in the right and left branches of a node, respectively. 
To get a better sense of parse tree topology, we also measure the ratio between tree height (the length of the longest path in the tree) and the minimum possible height for the tree: 
\begin{equation}
\text{FR}(t) = \frac{\text{H}(t)}{\text{log}\text{L}(t)}
\end{equation}
Where H$(t)$ and L$(t)$ are the height and number of tokens in the input sequence, respectively. 
Codebook utilization (CU) measures the percentage of non-terminal labels utilized within generated parse trees. 

We present these statistics in \autoref{tab:grammar_stats}, as well as normalized non-terminal node frequencies for parse trees generated by each grammar in \autoref{fig:grammar_freqs}, with some example parse trees in \autoref{fig:parse_trees}. 
Although both modalities experience a great reduction in perplexity compared to random initialization, textual grammars (COCO-EN and COCO-DE) generally exhibit greater reductions in perplexities than visual grammars, corroborating findings from \autoref{sec:entropy} which suggest that visual tokens are not as compressible as textual tokens.  
Although visual grammars converge to PPL values an order of magnitude greater than the textual grammars, we observed that their PPL values at the start of training are proportionally higher, likely due to the generally longer visual sentence length (32 tokens in these experiments). 
All other measures are generally comparable across modalities -- both modalities show similar proclivities towards right-branching trees (MBF), although visual grammars are somewhat more balanced. 
Both modalities present similar tree heights (FR), with the non-terminal label codebooks being largely utilized. 
The notable exception to these trends is the grammar trained on XM3600 tokens. 
XM3600 contains a significantly lower number of training examples (one order of magnitude less than SPIN, and two orders less than all other datasets), which may have resulted in a degenerate grammar being learned. 

These results suggest that the structure of visual languages may not be as well approximated by context-free grammars as natural languages are. 
This raises the question of whether they may be better fit by other grammatical formalisms, such as mildly context-sensitive grammars~\citep{yang-etal-2023-unsupervised} which allow for dependencies to cross between token spans.

\subsection{Topological Similarity}
\label{sec:topo}

\begin{table}
\scriptsize
\centering
\caption{Summary of Procrustes/Hausdorff alignment distances between vision languages and natural languages on the MS-COCO dataset. While in general, all languages are poorly co-aligned, in general, vision languages align slightly, but significantly, more strongly with natural languages than they do with other vision models.}
\label{tab:topology_summary}
\begin{tabularx}{\linewidth}{lXcclc}
\hline
\toprule
\textbf{Distance} & \textbf{Language} & \begin{tabular}{@{}c@{}} \textbf{Language-to-Vision} \\ \textbf{Distance} \end{tabular} & \begin{tabular}{@{}c@{}} \textbf{Language-to-Natural Language} \\ \textbf{Distance} \end{tabular} & \begin{tabular}{@{}c@{}} \textbf{Closest} \\ \textbf{Language} \end{tabular} & \begin{tabular}{@{}c@{}} \textbf{Closest} \\ \textbf{Distance} \end{tabular} \\
\midrule
\textbf{Procrustes} & Natural (Average) & 0.96689 $\pm$ 0.00425 &  0.96530 $\pm$ 0.00735 & text-hr & 0.96333 \\
 &Chameleon & 0.97699 $\pm$ 0.0158 & 0.96474 $\pm$ 0.00382 & text-no & 0.95580 \\ 
 &VQ-VAE (256) & 0.97886 $\pm$ 0.01427 & 0.96532 $\pm$ 0.00401 & text-ko & 0.95381 \\ 
 &VQ-VAE (64) & 0.97875 $\pm$ 0.01517 & 0.97024 $\pm$ 0.00310 & text-it & 0.96329 \\ 
 &VQ-VAE (ImageNet) & 0.97896 $\pm$ 0.01478 & 0.96731 $\pm$ 0.00386 & text-hu & 0.95709 \\
\midrule
\textbf{Haussdorf} & Natural (Average) & 10.81174 $\pm$ 1.28073 & 9.42697 $\pm$ 1.15902 & text-pl & 7.60177 \\
 &Chameleon & 7.68661 $\pm$ 0.58783 & 6.56173 $\pm$ 0.38642 & text-ko & 5.85738 \\ 
 &VQ-VAE (256) & 7.24126 $\pm$ 0.27003 & 5.95511 $\pm$ 0.34980 & text-zh & 5.36121 \\ 
 &VQ-VAE (64) & 7.97373 $\pm$ 0.29399 & 6.90376 $\pm$ 0.42660 & text-it & 6.02011 \\ 
 &VQ-VAE (ImageNet) & 7.52335 $\pm$ 0.24214 & 5.91748 $\pm$ 0.58758 & text-hr & 4.92164 \\
\bottomrule
\end{tabularx}
\end{table}

To expand our discussion on structural similarity, we further investigate how similar the topological structures of visual and textual tokens are, and whether these similarities can reveal meaningful insights about the underlying representations, i.e. can we observe strong structural alignment points between the natural and visual latent spaces, or are there notable deviations?

We begin by training GloVe embeddings \citep{pennington2014glove} on co-occurrence matrices derived from visual tokens and textual tokens present in the captions (details in \autoref{app:glove}). This gives us a continuous topology of similar dimension within which we can explore potential alignment. We then explore two pairwise distance matrices between the two GloVe vector spaces: Procrustes alignment \citep{gower1975generalized} and directed Haussdorf distance \citep{bowen1979hausdorff}. 

\autoref{fig:procrustes} gives the Procrustes similarity and \autoref{fig:haussdorf} gives the directed Haussdorf distance between the models, with some key aggregates summarized in \autoref{tab:topology_summary}. While there are few clear trends, a key finding is that vision models are largely more aligned with natural language models than they are with each other, with Chameleon being slightly more central than other models (perhaps due to its training process). Overall, the lack of strong alignment trends between different vision models highlights that their latent spaces are more fragmented, suggesting that visual token representations are often model-specific or task-dependent, rather than universally structured. Notably, however, some languages align much better with visual models than others (such as Korean to the Chameleon tokenizers, or Hungarian/Polish in general), suggesting that some tokenizers may be significantly stronger when aligning to specific languages. Another interesting observation is that the directed Hausdorff distance shows that the natural language to vision model alignment is significantly further than the vision model to natural language alignment. This results implies that generation of images from text is much harder than the generation of text from images - something often observed in practice.

Given the overall distances between these structural representations, our experiments suggest that future model architectures should focus on reducing this asymmetry. Specialized models that effectively encode multimodal information - and perhaps aligned tokenization methods (such as CLIP), represent promising future directions for research.

\section{Conclusion}

This paper takes a first look at visual languages from the angle of empirical statistics. While there are similarities between how we currently treat visual and natural languages/sentences - the experiments in this paper show that, at least statistically, visual tokens and natural languages are far from trivially aligned. Such poor statistical alignments motivate both unique model architectures and training procedures for visual transformers - and we hope that this work inspires further research into novel architectures, designs, and hyper-parameters for vision-token based models. Indeed, while some of the hypotheses that we outlined in this paper have already been demonstrated, many of the suggestions (such as increasing frequency penalties when decoding visual languages) remain untested in practice - and it is interesting and necessary future work to close the loop on such potential modifications. We hope, as a whole, that this work inspires additional research into fundamental statistics as a motivation for new architectural decisions and directions.

\bibliography{iclr2025_conference}

\begin{thebibliography}{64}
\providecommand{\natexlab}[1]{#1}
\providecommand{\url}[1]{\texttt{#1}}
\expandafter\ifx\csname urlstyle\endcsname\relax
  \providecommand{\doi}[1]{doi: #1}\else
  \providecommand{\doi}{doi: \begingroup \urlstyle{rm}\Url}\fi

\bibitem[Alstott et~al.(2014)Alstott, Bullmore, and Plenz]{alstott2014powerlaw}
Jeff Alstott, Ed~Bullmore, and Dietmar Plenz.
\newblock powerlaw: a python package for analysis of heavy-tailed distributions.
\newblock \emph{PloS one}, 9\penalty0 (1):\penalty0 e85777, 2014.

\bibitem[Anderson et~al.(2016)Anderson, Fernando, Johnson, and Gould]{anderson2016spice}
Peter Anderson, Basura Fernando, Mark Johnson, and Stephen Gould.
\newblock Spice: Semantic propositional image caption evaluation.
\newblock In \emph{European conference on computer vision}, pp.\  382--398. Springer, 2016.

\bibitem[Bai et~al.(2024)Bai, Geng, Mangalam, Bar, Yuille, Darrell, Malik, and Efros]{bai2024sequential}
Yutong Bai, Xinyang Geng, Karttikeya Mangalam, Amir Bar, Alan~L Yuille, Trevor Darrell, Jitendra Malik, and Alexei~A Efros.
\newblock Sequential modeling enables scalable learning for large vision models.
\newblock In \emph{Proceedings of the IEEE/CVF Conference on Computer Vision and Pattern Recognition}, pp.\  22861--22872, 2024.

\bibitem[Baker(1979)]{baker1979trainable}
James~K Baker.
\newblock Trainable grammars for speech recognition.
\newblock \emph{The Journal of the Acoustical Society of America}, 65\penalty0 (S1):\penalty0 S132--S132, 1979.

\bibitem[Benford(1938)]{benford1938law}
Frank Benford.
\newblock The law of anomalous numbers.
\newblock \emph{Proceedings of the American philosophical society}, pp.\  551--572, 1938.

\bibitem[Bowen(1979)]{bowen1979hausdorff}
Rufus Bowen.
\newblock Hausdorff dimension of quasi-circles.
\newblock \emph{Publications Math{\'e}matiques de l'IH{\'E}S}, 50:\penalty0 11--25, 1979.

\bibitem[Caglayan et~al.(2020)Caglayan, Madhyastha, and Specia]{caglayan2020curious}
Ozan Caglayan, Pranava Madhyastha, and Lucia Specia.
\newblock Curious case of language generation evaluation metrics: A cautionary tale.
\newblock In \emph{Proceedings of the 28th International Conference on Computational Linguistics}, pp.\  2322--2328. International Committee on Computational Linguistics, December 2020.
\newblock \doi{10.18653/v1/2020.coling-main.210}.
\newblock URL \url{https://aclanthology.org/2020.coling-main.210}.

\bibitem[Chan et~al.(2022)Chan, Myers, Vijayanarasimhan, Ross, Seybold, and Canny]{chan2022s}
David~M. Chan, Austin Myers, Sudheendra Vijayanarasimhan, David~A. Ross, Bryan Seybold, and John~F. Canny.
\newblock What's in a caption? dataset-specific linguistic diversity and its effect on visual description models and metrics.
\newblock In \emph{{IEEE/CVF} Conference on Computer Vision and Pattern Recognition Workshops, {CVPR} Workshops 2022, New Orleans, LA, USA, June 19-20, 2022}, pp.\  4739--4748. {IEEE}, 2022.
\newblock \doi{10.1109/CVPRW56347.2022.00520}.

\bibitem[Chomsky \& Sch{\"u}tzenberger(1959)Chomsky and Sch{\"u}tzenberger]{chomsky1959algebraic}
Noam Chomsky and Marcel~P Sch{\"u}tzenberger.
\newblock The algebraic theory of context-free languages.
\newblock In \emph{Studies in Logic and the Foundations of Mathematics}, volume~26, pp.\  118--161. Elsevier, 1959.

\bibitem[Church(2017)]{church2017word2vec}
Kenneth~Ward Church.
\newblock Word2vec.
\newblock \emph{Natural Language Engineering}, 23\penalty0 (1):\penalty0 155--162, 2017.

\bibitem[Crosier \& Griffin(2007)Crosier and Griffin]{crosier2007zipf}
Michael Crosier and Lewis~D Griffin.
\newblock Zipf's law in image coding schemes.
\newblock In \emph{BMVC}, pp.\  1--10. Citeseer, 2007.

\bibitem[Deng et~al.(2009)Deng, Dong, Socher, Li, Li, and Li]{deng2009imagenet}
Jia Deng, Wei Dong, Richard Socher, Li{-}Jia Li, Kai Li, and Fei{-}Fei Li.
\newblock Imagenet: {A} large-scale hierarchical image database.
\newblock In \emph{2009 {IEEE} Computer Society Conference on Computer Vision and Pattern Recognition {(CVPR} 2009), 20-25 June 2009, Miami, Florida, {USA}}, pp.\  248--255. {IEEE} Computer Society, 2009.
\newblock \doi{10.1109/CVPR.2009.5206848}.

\bibitem[Dewey(1921)]{dewey1921relative}
Godfrey Dewey.
\newblock \emph{Relative frequency of English speech sounds}.
\newblock PhD thesis, Harvard Graduate School of Education, 1921.

\bibitem[Esser et~al.(2021)Esser, Rombach, and Ommer]{esser2021taming}
Patrick Esser, Robin Rombach, and Bjorn Ommer.
\newblock Taming transformers for high-resolution image synthesis.
\newblock In \emph{Proceedings of the IEEE/CVF conference on computer vision and pattern recognition}, pp.\  12873--12883, 2021.

\bibitem[Gafni et~al.(2022)Gafni, Polyak, Ashual, Sheynin, Parikh, and Taigman]{gafni2022make}
Oran Gafni, Adam Polyak, Oron Ashual, Shelly Sheynin, Devi Parikh, and Yaniv Taigman.
\newblock Make-a-scene: Scene-based text-to-image generation with human priors.
\newblock In \emph{European Conference on Computer Vision}, pp.\  89--106. Springer, 2022.

\bibitem[Gelbukh \& Sidorov(2001)Gelbukh and Sidorov]{gelbukh2001zipf}
Alexander Gelbukh and Grigori Sidorov.
\newblock Zipf and heaps laws’ coefficients depend on language.
\newblock In \emph{Computational Linguistics and Intelligent Text Processing: Second International Conference, CICLing 2001 Mexico City, Mexico, February 18--24, 2001 Proceedings 2}, pp.\  332--335. Springer, 2001.

\bibitem[Golbeck(2023)]{golbeck2023benford}
Jennifer Golbeck.
\newblock Benford’s law applies to word frequency rank in english, german, french, spanish, and italian.
\newblock \emph{Plos one}, 18\penalty0 (9):\penalty0 e0291337, 2023.

\bibitem[Gower(1975)]{gower1975generalized}
John~C Gower.
\newblock Generalized procrustes analysis.
\newblock \emph{Psychometrika}, 40:\penalty0 33--51, 1975.

\bibitem[Gu et~al.(2022)Gu, Chen, Bao, Wen, Zhang, Chen, Yuan, and Guo]{gu2022vector}
Shuyang Gu, Dong Chen, Jianmin Bao, Fang Wen, Bo~Zhang, Dongdong Chen, Lu~Yuan, and Baining Guo.
\newblock Vector quantized diffusion model for text-to-image synthesis.
\newblock In \emph{Proceedings of the IEEE/CVF conference on computer vision and pattern recognition}, pp.\  10696--10706, 2022.

\bibitem[Heaps(1978)]{heaps1978information}
Harold~Stanley Heaps.
\newblock \emph{Information retrieval: Computational and theoretical aspects}.
\newblock Academic Press, Inc., 1978.

\bibitem[Herdan(1964)]{herdan1964quantitative}
Gustav Herdan.
\newblock Quantitative linguistics.
\newblock \emph{(No Title)}, 1964.

\bibitem[Hessel et~al.(2021)Hessel, Holtzman, Forbes, Le~Bras, and Choi]{hessel2021clipscore}
Jack Hessel, Ari Holtzman, Maxwell Forbes, Ronan Le~Bras, and Yejin Choi.
\newblock {CLIPS}core: A reference-free evaluation metric for image captioning.
\newblock In \emph{Proceedings of the 2021 Conference on Empirical Methods in Natural Language Processing}, pp.\  7514--7528. Association for Computational Linguistics, November 2021.
\newblock \doi{10.18653/v1/2021.emnlp-main.595}.
\newblock URL \url{https://aclanthology.org/2021.emnlp-main.595}.

\bibitem[Hong(2010)]{hong2010benford}
Jung-Ha Hong.
\newblock Benford's law in linguistic texts: Its principle and applications.
\newblock \emph{Language and Information}, 14\penalty0 (1):\penalty0 145--163, 2010.

\bibitem[Hsu et~al.(2021)Hsu, Tsai, Bolte, Salakhutdinov, and Mohamed]{hsu2021hubert}
Wei-Ning Hsu, Yao-Hung~Hubert Tsai, Benjamin Bolte, Ruslan Salakhutdinov, and Abdelrahman Mohamed.
\newblock Hubert: How much can a bad teacher benefit asr pre-training?
\newblock In \emph{ICASSP 2021-2021 IEEE International Conference on Acoustics, Speech and Signal Processing (ICASSP)}, pp.\  6533--6537. IEEE, 2021.

\bibitem[Huffman(1952)]{huffman1952method}
David~A Huffman.
\newblock A method for the construction of minimum-redundancy codes.
\newblock \emph{Proceedings of the IRE}, 40\penalty0 (9):\penalty0 1098--1101, 1952.

\bibitem[Kim et~al.(2019)Kim, Dyer, and Rush]{kim-etal-2019-compound}
Yoon Kim, Chris Dyer, and Alexander Rush.
\newblock Compound probabilistic context-free grammars for grammar induction.
\newblock In Anna Korhonen, David Traum, and Llu{\'\i}s M{\`a}rquez (eds.), \emph{Proceedings of the 57th Annual Meeting of the Association for Computational Linguistics}, pp.\  2369--2385, Florence, Italy, July 2019. Association for Computational Linguistics.
\newblock \doi{10.18653/v1/P19-1228}.
\newblock URL \url{https://aclanthology.org/P19-1228}.

\bibitem[Kingma(2013)]{kingma2013auto}
Diederik~P Kingma.
\newblock Auto-encoding variational bayes.
\newblock \emph{arXiv preprint arXiv:1312.6114}, 2013.

\bibitem[Kingsley~Zipf(1932)]{kingsley1932selected}
George Kingsley~Zipf.
\newblock \emph{Selected studies of the principle of relative frequency in language}.
\newblock Harvard university press, 1932.

\bibitem[Kitaev \& Klein(2018)Kitaev and Klein]{kitaev-klein-2018-constituency}
Nikita Kitaev and Dan Klein.
\newblock Constituency parsing with a self-attentive encoder.
\newblock In Iryna Gurevych and Yusuke Miyao (eds.), \emph{Proceedings of the 56th Annual Meeting of the Association for Computational Linguistics (Volume 1: Long Papers)}, pp.\  2676--2686, Melbourne, Australia, July 2018. Association for Computational Linguistics.
\newblock \doi{10.18653/v1/P18-1249}.
\newblock URL \url{https://aclanthology.org/P18-1249}.

\bibitem[Kuznetsova et~al.(2020)Kuznetsova, Rom, Alldrin, Uijlings, Krasin, Pont-Tuset, Kamali, Popov, Malloci, Kolesnikov, et~al.]{kuznetsova2020open}
Alina Kuznetsova, Hassan Rom, Neil Alldrin, Jasper Uijlings, Ivan Krasin, Jordi Pont-Tuset, Shahab Kamali, Stefan Popov, Matteo Malloci, Alexander Kolesnikov, et~al.
\newblock The open images dataset v4: Unified image classification, object detection, and visual relationship detection at scale.
\newblock \emph{International journal of computer vision}, 128\penalty0 (7):\penalty0 1956--1981, 2020.

\bibitem[Li et~al.(2024)Li, Corona, Mangalam, Chen, Flaherty, Belongie, Weinberger, Malik, Darrell, and Klein]{li-etal-2024-evaluating}
Boyi Li, Rodolfo Corona, Karttikeya Mangalam, Catherine Chen, Daniel Flaherty, Serge Belongie, Kilian Weinberger, Jitendra Malik, Trevor Darrell, and Dan Klein.
\newblock Re-evaluating the need for visual signals in unsupervised grammar induction.
\newblock In Kevin Duh, Helena Gomez, and Steven Bethard (eds.), \emph{Findings of the Association for Computational Linguistics: NAACL 2024}, pp.\  1113--1123, Mexico City, Mexico, June 2024. Association for Computational Linguistics.
\newblock \doi{10.18653/v1/2024.findings-naacl.70}.
\newblock URL \url{https://aclanthology.org/2024.findings-naacl.70}.

\bibitem[Lin et~al.(2014)Lin, Maire, Belongie, Hays, Perona, Ramanan, Doll{\'a}r, and Zitnick]{lin2014microsoft}
Tsung-Yi Lin, Michael Maire, Serge Belongie, James Hays, Pietro Perona, Deva Ramanan, Piotr Doll{\'a}r, and C~Lawrence Zitnick.
\newblock Microsoft coco: Common objects in context.
\newblock In \emph{European conference on computer vision}, pp.\  740--755. Springer, 2014.

\bibitem[Liu et~al.(2024)Liu, Li, Wu, and Lee]{liu2024visual}
Haotian Liu, Chunyuan Li, Qingyang Wu, and Yong~Jae Lee.
\newblock Visual instruction tuning.
\newblock \emph{Advances in neural information processing systems}, 36, 2024.

\bibitem[Mandelbrot(1953)]{mandelbrot1953contribution}
Beno{\^\i}t Mandelbrot.
\newblock Contribution {\`a} la th{\'e}orie math{\'e}matique des jeux de communication.
\newblock In \emph{Annales de l'ISUP}, volume~2, pp.\  3--124, 1953.

\bibitem[Mao et~al.(2021)Mao, Jiang, Dehghani, Vondrick, Sukthankar, and Essa]{mao2021discrete}
Chengzhi Mao, Lu~Jiang, Mostafa Dehghani, Carl Vondrick, Rahul Sukthankar, and Irfan Essa.
\newblock Discrete representations strengthen vision transformer robustness.
\newblock \emph{arXiv preprint arXiv:2111.10493}, 2021.

\bibitem[McCowan et~al.(1999)McCowan, Hanser, and Doyle]{mccowan1999quantitative}
Brenda McCowan, Sean~F Hanser, and Laurance~R Doyle.
\newblock Quantitative tools for comparing animal communication systems: information theory applied to bottlenose dolphin whistle repertoires.
\newblock \emph{Animal behaviour}, 57\penalty0 (2):\penalty0 409--419, 1999.

\bibitem[Meli{\'a}n et~al.(2017)Meli{\'a}n, Conejero, and Ramirez]{melian2017zipf}
Jos{\'e} Alberto~P{\'e}rez Meli{\'a}n, J~Alberto Conejero, and Cesar~Ferri Ramirez.
\newblock Zipf’s and benford’s laws in twitter hashtags.
\newblock In \emph{Proceedings of the Student Research Workshop at the 15th Conference of the European Chapter of the Association for Computational Linguistics}, pp.\  84--93, 2017.

\bibitem[Miller(2015)]{miller2015benford}
Steven~J Miller.
\newblock \emph{Benford's law}.
\newblock Princeton University Press, 2015.

\bibitem[Myers-Dean et~al.(2024)Myers-Dean, Reynolds, Price, Fan, and Gurari]{myers2024spin}
Josh Myers-Dean, Jarek Reynolds, Brian Price, Yifei Fan, and Danna Gurari.
\newblock Spin: Hierarchical segmentation with subpart granularity in natural images.
\newblock \emph{arXiv preprint arXiv:2407.09686}, 2024.

\bibitem[Patwary et~al.(2019)Patwary, Chabbi, Jun, Huang, Diamos, and Church]{patwary2019language}
Mostofa Patwary, Milind Chabbi, Heewoo Jun, Jiaji Huang, Gregory Diamos, and Kenneth Church.
\newblock Language modeling at scale.
\newblock In \emph{2019 IEEE International Parallel and Distributed Processing Symposium (IPDPS)}, pp.\  590--599. IEEE, 2019.

\bibitem[Pennington et~al.(2014)Pennington, Socher, and Manning]{pennington2014glove}
Jeffrey Pennington, Richard Socher, and Christopher Manning.
\newblock {G}lo{V}e: Global vectors for word representation.
\newblock In \emph{Proceedings of the 2014 Conference on Empirical Methods in Natural Language Processing ({EMNLP})}, pp.\  1532--1543. Association for Computational Linguistics, October 2014.
\newblock \doi{10.3115/v1/D14-1162}.
\newblock URL \url{https://aclanthology.org/D14-1162}.

\bibitem[Ramesh et~al.(2022)Ramesh, Dhariwal, Nichol, Chu, and Chen]{ramesh2022hierarchical}
Aditya Ramesh, Prafulla Dhariwal, Alex Nichol, Casey Chu, and Mark Chen.
\newblock Hierarchical text-conditional image generation with clip latents.
\newblock \emph{ArXiv preprint}, abs/2204.06125, 2022.

\bibitem[Razavi et~al.(2019)Razavi, Van~den Oord, and Vinyals]{razavi2019generating}
Ali Razavi, Aaron Van~den Oord, and Oriol Vinyals.
\newblock Generating diverse high-fidelity images with vq-vae-2.
\newblock \emph{Advances in neural information processing systems}, 32, 2019.

\bibitem[Robbins(1956)]{robbins1956empirical}
Herbert Robbins.
\newblock An {E}mpirical {B}ayes {A}pproach to {S}tatistics.
\newblock \emph{Proceedings of the Third Berkeley Symposium on Mathematical Statistics and Probability}, pp.\  157--163, 1956.

\bibitem[Rombach et~al.(2022)Rombach, Blattmann, Lorenz, Esser, and Ommer]{rombach2022high}
Robin Rombach, Andreas Blattmann, Dominik Lorenz, Patrick Esser, and Bj{\"o}rn Ommer.
\newblock High-resolution image synthesis with latent diffusion models.
\newblock In \emph{Proceedings of the IEEE/CVF Conference on Computer Vision and Pattern Recognition}, pp.\  10684--10695, 2022.

\bibitem[Ruderman(1997)]{ruderman1997origins}
Daniel~L Ruderman.
\newblock Origins of scaling in natural images.
\newblock \emph{Vision research}, 37\penalty0 (23):\penalty0 3385--3398, 1997.

\bibitem[Russakovsky et~al.(2015)Russakovsky, Deng, Su, Krause, Satheesh, Ma, Huang, Karpathy, Khosla, Bernstein, et~al.]{russakovsky2015imagenet}
Olga Russakovsky, Jia Deng, Hao Su, Jonathan Krause, Sanjeev Satheesh, Sean Ma, Zhiheng Huang, Andrej Karpathy, Aditya Khosla, Michael Bernstein, et~al.
\newblock Imagenet large scale visual recognition challenge.
\newblock \emph{International journal of computer vision}, 115:\penalty0 211--252, 2015.

\bibitem[Sambridge et~al.(2010)Sambridge, Tkal{\v{c}}i{\'c}, and Jackson]{sambridge2010benford}
Malcolm Sambridge, Hrvoje Tkal{\v{c}}i{\'c}, and A~Jackson.
\newblock Benford's law in the natural sciences.
\newblock \emph{Geophysical research letters}, 37\penalty0 (22), 2010.

\bibitem[Shannon(1951)]{shannon1951prediction}
Claude~E Shannon.
\newblock Prediction and entropy of printed english.
\newblock \emph{Bell system technical journal}, 30\penalty0 (1):\penalty0 50--64, 1951.

\bibitem[Sharma et~al.(2018)Sharma, Ding, Goodman, and Soricut]{sharma2018conceptual}
Piyush Sharma, Nan Ding, Sebastian Goodman, and Radu Soricut.
\newblock Conceptual captions: A cleaned, hypernymed, image alt-text dataset for automatic image captioning.
\newblock In \emph{Proceedings of the 56th Annual Meeting of the Association for Computational Linguistics (Volume 1: Long Papers)}, pp.\  2556--2565. Association for Computational Linguistics, July 2018.
\newblock \doi{10.18653/v1/P18-1238}.
\newblock URL \url{https://aclanthology.org/P18-1238}.

\bibitem[Simon(1955)]{simon1955class}
Herbert~A Simon.
\newblock On a class of skew distribution functions.
\newblock \emph{Biometrika}, 42\penalty0 (3/4):\penalty0 425--440, 1955.

\bibitem[Sun et~al.(2024)Sun, Jiang, Chen, Zhang, Peng, Luo, and Yuan]{sun2024autoregressive}
Peize Sun, Yi~Jiang, Shoufa Chen, Shilong Zhang, Bingyue Peng, Ping Luo, and Zehuan Yuan.
\newblock Autoregressive model beats diffusion: Llama for scalable image generation.
\newblock \emph{arXiv preprint arXiv:2406.06525}, 2024.

\bibitem[Team(2024)]{team2024chameleon}
Chameleon Team.
\newblock Chameleon: Mixed-modal early-fusion foundation models.
\newblock \emph{arXiv preprint arXiv:2405.09818}, 2024.

\bibitem[Thapliyal et~al.(2022)Thapliyal, Tuset, Chen, and Soricut]{thapliyal2022crossmodal}
Ashish~V Thapliyal, Jordi~Pont Tuset, Xi~Chen, and Radu Soricut.
\newblock Crossmodal-3600: A massively multilingual multimodal evaluation dataset.
\newblock In \emph{Proceedings of the 2022 Conference on Empirical Methods in Natural Language Processing}, pp.\  715--729, 2022.

\bibitem[T{\"o}dter(2009)]{todter2009benford}
Karl-Heinz T{\"o}dter.
\newblock Benford’s law as an indicator of fraud in economics.
\newblock \emph{German Economic Review}, 10\penalty0 (3):\penalty0 339--351, 2009.

\bibitem[Tong et~al.(2024)Tong, Brown, Wu, Woo, Middepogu, Akula, Yang, Yang, Iyer, Pan, et~al.]{tong2024cambrian}
Shengbang Tong, Ellis Brown, Penghao Wu, Sanghyun Woo, Manoj Middepogu, Sai~Charitha Akula, Jihan Yang, Shusheng Yang, Adithya Iyer, Xichen Pan, et~al.
\newblock Cambrian-1: A fully open, vision-centric exploration of multimodal llms.
\newblock \emph{arXiv preprint arXiv:2406.16860}, 2024.

\bibitem[Touvron et~al.(2021)Touvron, Cord, Douze, Massa, Sablayrolles, and J{\'e}gou]{touvron2021training}
Hugo Touvron, Matthieu Cord, Matthijs Douze, Francisco Massa, Alexandre Sablayrolles, and Herv{\'e} J{\'e}gou.
\newblock Training data-efficient image transformers \& distillation through attention.
\newblock In \emph{International conference on machine learning}, pp.\  10347--10357. PMLR, 2021.

\bibitem[Van Den~Oord et~al.(2017)Van Den~Oord, Vinyals, et~al.]{van2017neural}
Aaron Van Den~Oord, Oriol Vinyals, et~al.
\newblock Neural discrete representation learning.
\newblock \emph{Advances in neural information processing systems}, 30, 2017.

\bibitem[Willis \& Yule(1922)Willis and Yule]{willis1922some}
John~C Willis and G~Udny Yule.
\newblock Some statistics of evolution and geographical distribution in plants and animals, and their significance.
\newblock \emph{Nature}, 109\penalty0 (2728):\penalty0 177--179, 1922.

\bibitem[Yang et~al.(2023)Yang, Levy, and Kim]{yang-etal-2023-unsupervised}
Songlin Yang, Roger Levy, and Yoon Kim.
\newblock Unsupervised discontinuous constituency parsing with mildly context-sensitive grammars.
\newblock In Anna Rogers, Jordan Boyd-Graber, and Naoaki Okazaki (eds.), \emph{Proceedings of the 61st Annual Meeting of the Association for Computational Linguistics (Volume 1: Long Papers)}, pp.\  5747--5766, Toronto, Canada, July 2023. Association for Computational Linguistics.
\newblock \doi{10.18653/v1/2023.acl-long.316}.
\newblock URL \url{https://aclanthology.org/2023.acl-long.316}.

\bibitem[Ye et~al.(2024)Ye, Xu, Liu, Hu, Yan, Qian, Zhang, Huang, and Zhou]{ye2024mplug}
Jiabo Ye, Haiyang Xu, Haowei Liu, Anwen Hu, Ming Yan, Qi~Qian, Ji~Zhang, Fei Huang, and Jingren Zhou.
\newblock mplug-owl3: Towards long image-sequence understanding in multi-modal large language models.
\newblock \emph{arXiv preprint arXiv:2408.04840}, 2024.

\bibitem[Yu et~al.(2018)Yu, Xu, and Liu]{yu2018zipf}
Shuiyuan Yu, Chunshan Xu, and Haitao Liu.
\newblock Zipf's law in 50 languages: its structural pattern, linguistic interpretation, and cognitive motivation.
\newblock \emph{arXiv preprint arXiv:1807.01855}, 2018.

\bibitem[Zhao \& Titov(2020)Zhao and Titov]{zhao-titov-2020-visually}
Yanpeng Zhao and Ivan Titov.
\newblock Visually grounded compound {PCFG}s.
\newblock In Bonnie Webber, Trevor Cohn, Yulan He, and Yang Liu (eds.), \emph{Proceedings of the 2020 Conference on Empirical Methods in Natural Language Processing (EMNLP)}, pp.\  4369--4379, Online, November 2020. Association for Computational Linguistics.
\newblock \doi{10.18653/v1/2020.emnlp-main.354}.
\newblock URL \url{https://aclanthology.org/2020.emnlp-main.354}.

\bibitem[Zhou et~al.(2024)Zhou, Yu, Babu, Tirumala, Yasunaga, Shamis, Kahn, Ma, Zettlemoyer, and Levy]{zhou2024transfusion}
Chunting Zhou, Lili Yu, Arun Babu, Kushal Tirumala, Michihiro Yasunaga, Leonid Shamis, Jacob Kahn, Xuezhe Ma, Luke Zettlemoyer, and Omer Levy.
\newblock Transfusion: Predict the next token and diffuse images with one multi-modal model.
\newblock \emph{arXiv preprint arXiv:2408.11039}, 2024.

\end{thebibliography}
\bibliographystyle{iclr2025_conference}

\appendix

\renewcommand{\theequation}{\thesection.\arabic{equation}}
\renewcommand{\thefigure}{\thesection.\arabic{figure}}
\renewcommand{\thetable}{\thesection.\arabic{table}}

\makeatletter
\@addtoreset{equation}{section}
\@addtoreset{figure}{section}
\@addtoreset{table}{section}
\makeatother

\clearpage
\section*{Appendix}

The appendix consists of the following further discussion:
\begin{itemize}
    \item \autoref{app:tokenizers} discusses the tokenizers used when constructing the visual languages, with detailed descriptions of Chameleon, Stable Diffusion, and LlamaGen tokenizers.
    \item \autoref{app:datasets} describes the datasets utilized in this work, including Conceptual Captions (CC12M), MS-COCO, ImageNet (ILSVRC), XM-3600, and SPIN.
    \item \autoref{app:limitations} describes potential limitations and opportunities for future work.
    \item \autoref{app:zipf} describes the Zipf experiments in \autoref{sec:token_frequency}, and gives additional experimental details.
    \item \autoref{app:heaps} describes Heaps' law, and gives additional experimental results to complement \autoref{sec:token_innovation}.
    \item \autoref{app:yule-simon} explains the Yule-Simon distribution, the methodology used to fit this distribution to observed token frequencies, and the experimental results from token frequency analysis.
    \item \autoref{app:benfords} discusses the process used for analyzing visual tokens according to Benford's law in \autoref{sec:token_frequency}, including n-gram extraction and first-digit distribution analysis across datasets.
    \item \autoref{app:huffman} explains the Huffman encoding experiments, measuring entropy and compression efficiency of tokenized visual data.
    \item \autoref{app:purity} explores segmentation granularity and how visual tokens correspond to parts and sub-parts in images, using co-occurrence metrics like Part Purity and Visual Token Purity.
    \item \autoref{app:glove} discusses CPFCGs, the process for extracting GloVe embeddings from both vision and language tokenizers, and the topological analysis used in \autoref{sec:topo}.
\end{itemize}

\section{Tokenizers}
\label{app:tokenizers}

In this work, we explore three families of VQ-VAE \citep{van2017neural} based tokenizers for images. While the general details are given in \autoref{tab:tokenizers}, we expand on the details for the tokenizers here.

\textbf{Chameleon \citep{team2024chameleon}:} Chameleon is a family of early-fusion token-based mixed-modal models capable of understanding and generating images and text. The image tokenizer, \verb|chameleon-512|, is based on \citet{gafni2022make}, which is a modified VQGAN \cite{esser2021taming} model which adds perceptual losses to specific image regions such as faces and salient objects (in an attempt to improve the fidelty of generated images). The chameleon tokenizer is trained from scratch on a closed-source set of licensed images, and encodes images at a resolution of $512 \times 512$ into a discrete token codebook size of 8192 and dimension 256. Notably, when training the tokenizer the model up-samples the percentage of images with faces by two times to improve performance on human face generation (which may somewhat skew the performance of the tokenizer on non-face based images).  

\textbf{Stable Diffusion (Compvis) \citep{rombach2022high}:} Stable Diffusion is a latent text-to-image diffusion model, which learns a joint distribution over image and text representations in a discretized latent space. Similar to the chameleon tokenizer, these tokenizers are trained in an adversarial manner following \citet{esser2021taming} on OpenImages \cite{kuznetsova2020open}, such that a patch-based discriminator can differentiate original images from reconstructions. The stable diffusion tokenizers (\verb|compvis-vq-f8-64| and \verb|compvis-vq-f8-256|) have an image resolution of $384 \times 384$ with a crop-size of $256$, and use a codebook dimension of size 4, with a very high VQ quantization dimension of 16384. While these models were trained at a crop size of 256, for grammatical analysis, many of the generated sequences are much too long to solve using traditional methods. Thus, we additionally consider a model, \verb|compvis-vq-f8-64| which uses a $64 \times 64$ crop of the image, which produces linearized sequences of a more manageable length of 32, used in \autoref{sec:grammars}. The tokenizer \verb|compvis-vq-imagenet-f16-1024-256| (originally trained by \citet{esser2021taming}) uses the same training procedure as those in \citet{rombach2022high}, but was trained on the ImageNet dataset, with a codebook of dimension 256, and size 1024. 

\textbf{LlamaGen \citep{sun2024autoregressive}:} LlamaGen is a family of image-generation models that apply next-token prediction to perform iamge synthesis. The LlamaGen tokenizer, \verb|llamagen-vq-ds16-c2i| takes images of resolution $256 \times 256$, and uses a codebook of size 16384 and dimension 8.  \verb|llamagen-vq-ds16-c2i| is trained on the ImageNet training dataset.

\begin{table}[h]
    \scriptsize
    \centering
    \begin{tabularx}{\linewidth}{Xccccc}
    \toprule
    Tokenizer & R-FID & R-IS & PSNR & PSIM & SSIM \\
    \midrule
        \texttt{chameleon-512} & - & - & - & - & - \\
        \texttt{compvis-vq-f8-64} & - & - & - & - & - \\
        \texttt{compvis-vq-f8-256} & 1.14 & 201.92 & 23.07 & 1.17 & 0.650 \\
        \texttt{compvis-vq-imagenet-f16-1024-256} & 4.98 & - & - & - & -  \\
        \texttt{llamagen-vq-ds16-c2i} & 2.19 &  - & 20.79 & - & 0.675 \\
        \bottomrule
    \end{tabularx}
    \caption{Tokenizer Performance (As available in the original papers) - Evaluated on ImageNet 50K Validation dataset.}
    \label{tab:my_label}
\end{table}

\section{Datasets}
\label{app:datasets}

In this work, we explore the effects of tokenization across several datasets:

\textbf{Conceptual Captions (12M) \citep{sharma2018conceptual}:} Conceptual captions (12M, CC12M) is a dataset with approximately 12 million image-text pairs soruce from web alt-text, traditionally used for vision-language pre-training. 

\textbf{MS-COCO \citep{lin2014microsoft}:} The MS-COCO dataset is a dataset for image description containing 328K images, each with 5 ground truth descriptions in English. In addition to the standard annotations, we also leverage translated annotations from \citet{thapliyal2022crossmodal}, which provide machine translations into 36 languages for each of the MS-COCO images. 

\textbf{ImageNet (ILSVRC) \citep{deng2009imagenet}:} ImageNet contains approximately 1.2M images which are manually annotated to indicate the objects present in each image. These annotations are linked to the WordNet hierarchy, providing a rich set of object categories. The dataset covers 1,000 object classes for the classification task, including common objects like animals, vehicles, and household items.

\textbf{XM-3600 \citep{thapliyal2022crossmodal}:} The Crossmodal-3600/XM3600 dataset is a multilingual multimodal evaluation dataset designed to support image captioning tasks across 36 languages. It consists of 3600 geographically diverse images, each annotated with human-generated captions that are consistent across languages but not derived from direct translations, ensuring linguistic naturalness and cultural relevance. The images were selected from regions where these languages are spoken, drawn from the Open Images Dataset using a careful algorithm to ensure regional diversity. 

\textbf{SPIN \citep{myers2024spin}:} The SPIN (SubPartImageNet) dataset is a hierarchical semantic segmentation dataset designed to provide detailed annotations for natural images at multiple levels of granularity, specifically focusing on objects, parts, and subparts. SPIN builds on the PartImageNet dataset, expanding its scope by introducing over 106,000 subpart annotations across 203 subpart categories, covering 34 part categories from diverse objects such as animals, vehicles, and human figures. The dataset contains 10,387 images divided across 11 supercategories, including rigid objects like cars and non-rigid entities like animals.

\section{Limitations}
\label{app:limitations}
\label{app:scan-order}

While this paper does have significant empirical results, we want to recognize the several potential limitations/opportunities for future work:

\textbf{Tokenizer Selection:} While the paper does focus on a fairly wide range of common (and modern) visual tokenizers, there is a fairly large potential selection of additional tokenizers that could be compared. Indeed, a key limiting factor is that all of the tokenizers explored in this work are VQ-VAE based. As discussed in \autoref{sec:preliminaries}, a detailed analysis of continuous tokenizers (such as auto-encoders which are KL-regularized, CLIP-style encoders, or BERT-style encoders) would provide significant additional information. Directly applying natural language statistics to these continuous embeddings, however, is non-trival, as to understand ideas of ``token frequency'' or ``grammar'', such analyses would have to either (a) be extended to the continuous domain, or (b) the tokens themselves would have to be quantized to discrete representations. We believe that such extensions are highly interesting, but are worthy of detailed analysis and discussion which is outside the scope of this initial work.

\textbf{Dataset Coverage:} Another limiting factor of this research is the dataset coverage. While it is impossible to analyze all data, visual information is highly diverse, and domains such as medical imaging, geospatial imaging, or autonomous driving may have entirely different statistics. In general, however, we found that across the datasets that we did use (which represent a fairly general slice of traditional training data), the statistical representations were similar. For example, it is fairly challenging to distinguish any dataset-level patterns in \autoref{fig:zipf_full_ngrams}, which shows a per-dataset breakdown of the empirical token frequency distributions, or \autoref{fig:model_fits_xm3600_ngram} which shows the Yule-Simon fits for emprical token frequencies. \

\textbf{Scan Order of Images:} One of the notable limitations of this work is that we primarily investigate a linear row-wise scan order of the images. We primarily limit our experiments to this scan order as (1) this is the de-facto scan order used in all existing transformer-based tokenization schemes and (2) we do not want to introduce further confounding analytical axes in this work. Exploring non-row-wise scan orders is, however, an extremely interesting question. In our limited experimentation, we found that a row-wise scan order does not significantly impact the explorations in the paper, as the majority of the analyses are scan-order independent.

\textbf{Token Granularity and Semantic Understanding:} Although granularity analysis is insightful, a deeper examination of how well visual tokens capture complex semantic meaning in images (e.g., context, object relationships, or scene understanding) remains future research. We strongly believe that future research should explore how visual tokens represent not just parts of objects but also their roles in broader scenes or tasks requiring semantic understanding (e.g., visual reasoning, narrative generation), however such explorations would require signficant new labeled data, or novel statistical approaches.

\textbf{Visual Tokens in Video Data:} In tasks like video understanding or motion tracking, the temporal relationships between visual tokens might reveal additional complexities not captured in static image analysis. Future research could explore how the behavior of visual tokens changes in sequential or temporal data settings and whether current statistical patterns hold when accounting for time.

\section{Zipf's Law}
\label{app:zipf}

As discussed in \autoref{sec:token_frequency}, Zipf’s Law \citep{kingsley1932selected}, describes a power-law relationship between the frequency of words and their rank in a language where a small number of high-frequency words dominate natural language, while the majority of words occur infrequently. Formally, Zipf's law states that:
\begin{equation}
    f(r) \propto {r^{\alpha + \sigma Z}}
\end{equation}
where $f(r)$ is the frequency of the element with rank $r$ and $\alpha/\sigma$ parameterize a learned Gaussian distribution (close to 1/0 in many natural languages).

\begin{figure}[htbp]
    \captionsetup[subfigure]{labelfont=tiny, textfont=tiny, skip=3pt} %
    \begin{minipage}{0.15\textwidth}
        \raggedleft
        \textbf{cc12m} %
    \end{minipage}%
    \begin{minipage}{0.80\textwidth}
        \begin{subfigure}[b]{0.19\textwidth}
            \includegraphics[width=\textwidth]{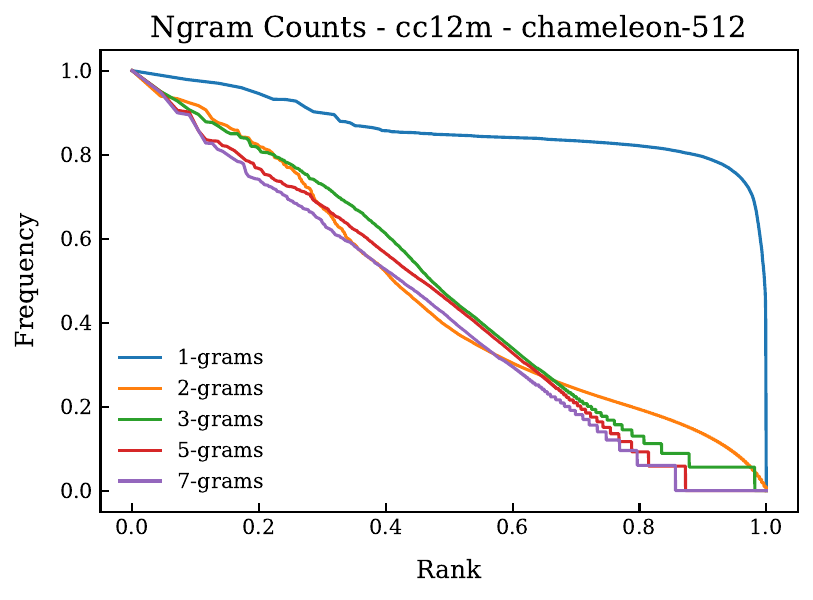}
            \caption{\tiny Chameleon}
        \end{subfigure}
        \begin{subfigure}[b]{0.19\textwidth}
            \includegraphics[width=\textwidth]{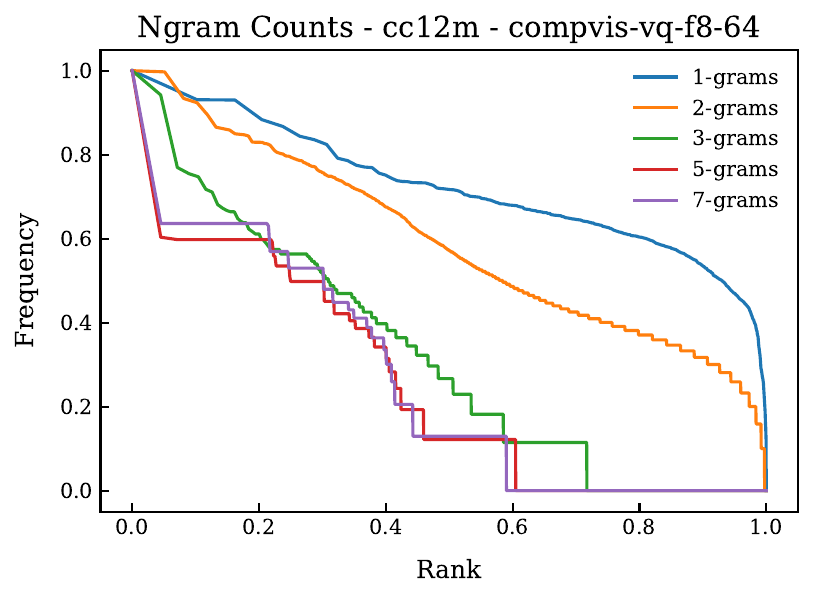}
            \caption{\tiny VQ-VAE (64)}
        \end{subfigure}
        \begin{subfigure}[b]{0.19\textwidth}
            \includegraphics[width=\textwidth]{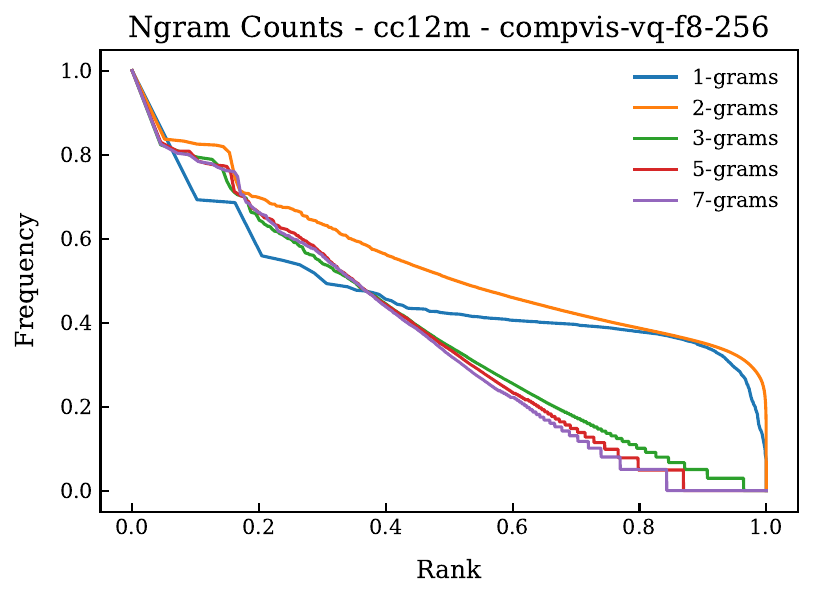}
            \caption{\tiny VQ-VAE (256)}
        \end{subfigure}
        \begin{subfigure}[b]{0.19\textwidth}
            \includegraphics[width=\textwidth]{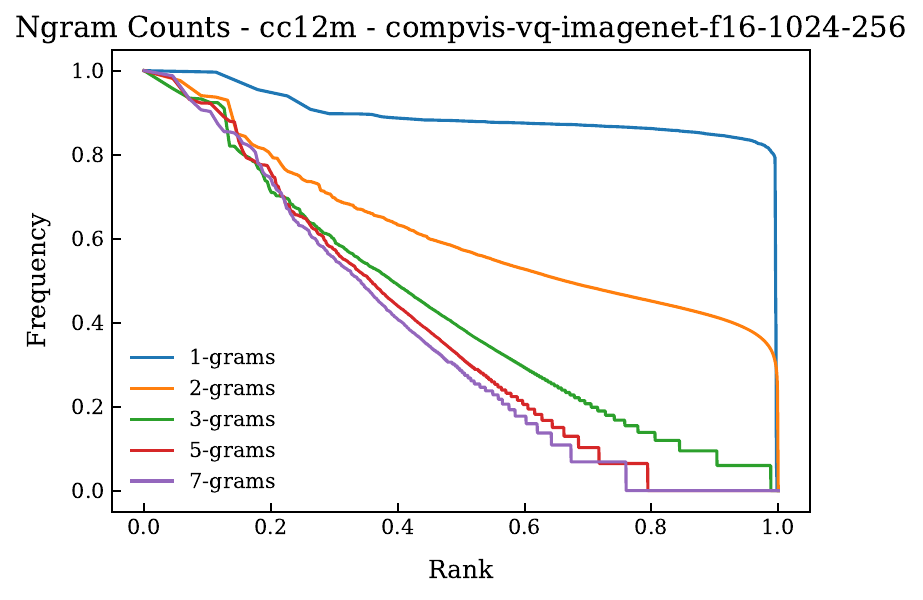}
            \caption{\tiny VQ-VAE (ImageNet)}
        \end{subfigure}
        \begin{subfigure}[b]{0.19\textwidth}
            \includegraphics[width=\textwidth]{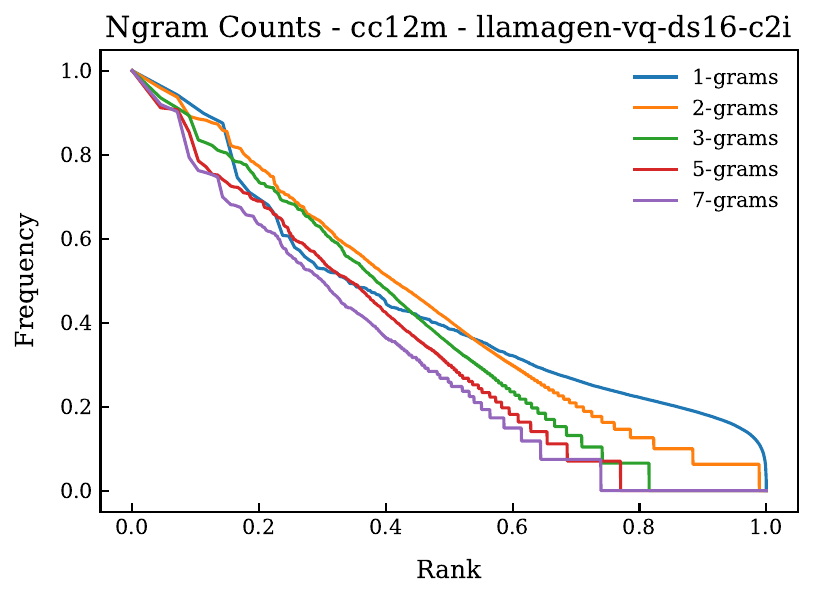}
            \caption{\tiny LlamaGen}
        \end{subfigure}
    \end{minipage}
    \vspace{0.5cm} %

    \begin{minipage}{0.15\textwidth}
        \raggedleft
        \textbf{cc12m-full} %
    \end{minipage}%
    \begin{minipage}{0.80\textwidth}
        \begin{subfigure}[b]{0.19\textwidth}
            \includegraphics[width=\textwidth]{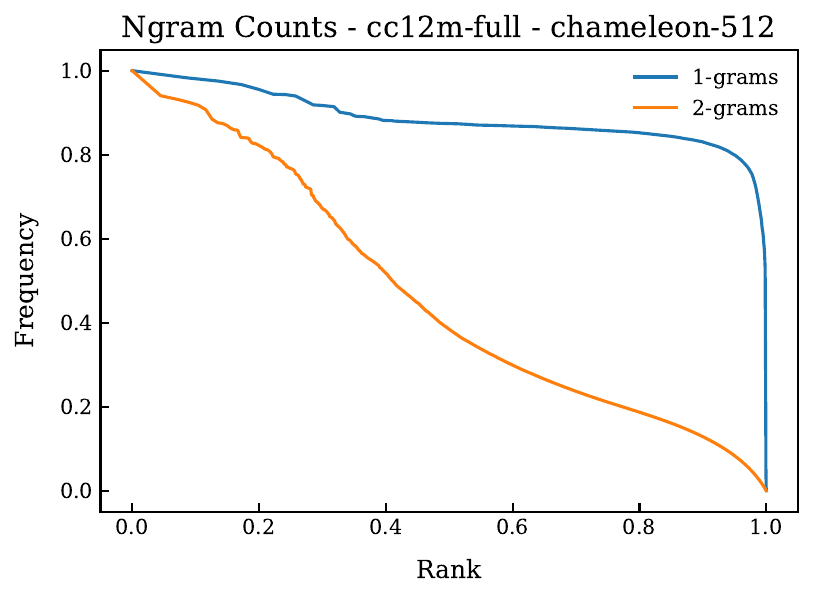}
            \caption{\tiny Chameleon}
        \end{subfigure}
        \begin{subfigure}[b]{0.19\textwidth}
            \includegraphics[width=\textwidth]{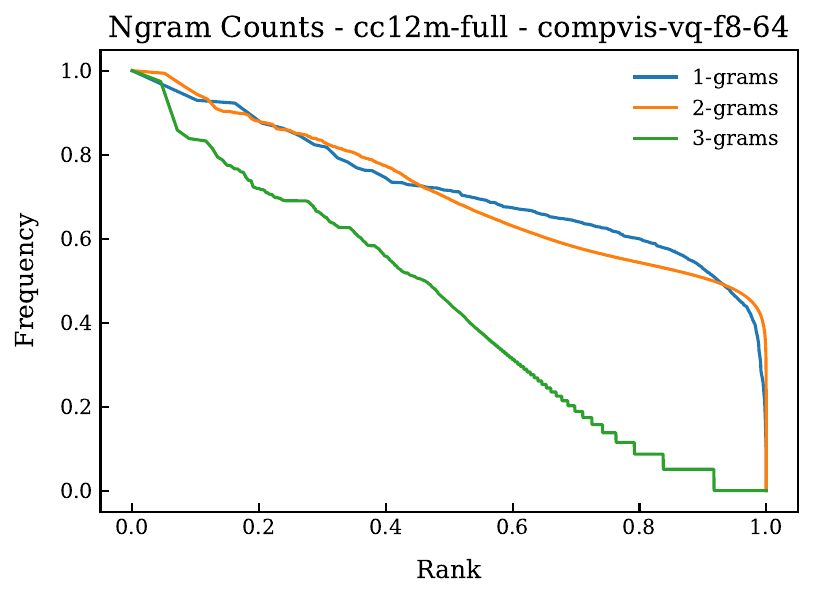}
            \caption{\tiny VQ-VAE (64)}
        \end{subfigure}
        \begin{subfigure}[b]{0.19\textwidth}
            {\centering \textcolor{red}{\tiny $\quad$*** Memory Error ***}}
            \caption{\tiny VQ-VAE (256)}
        \end{subfigure}
        \begin{subfigure}[b]{0.19\textwidth}
            \includegraphics[width=\textwidth]{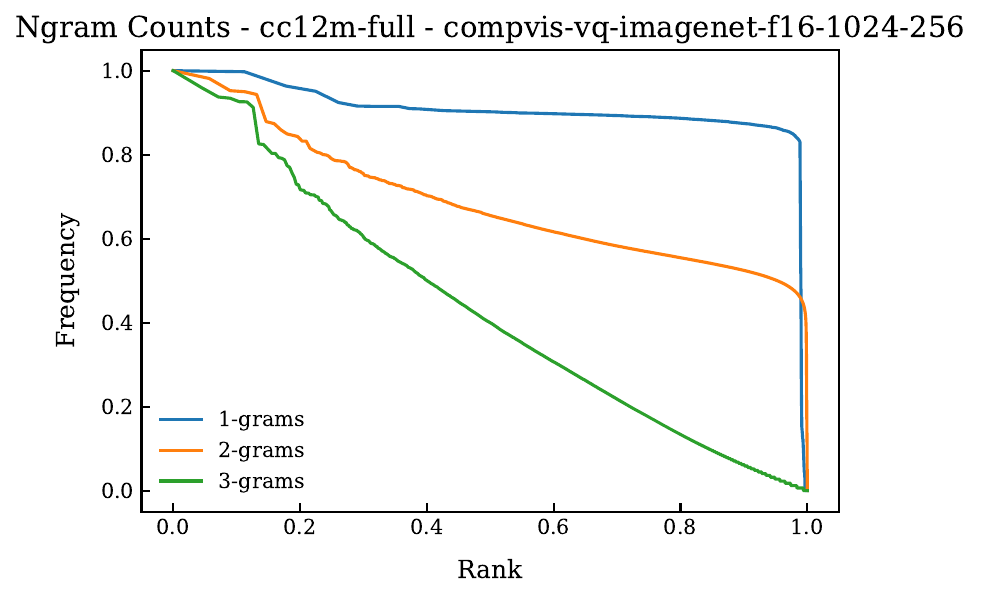}
            \caption{\tiny VQ-VAE (ImageNet)}
        \end{subfigure}
        \begin{subfigure}[b]{0.19\textwidth}
            \includegraphics[width=\textwidth]{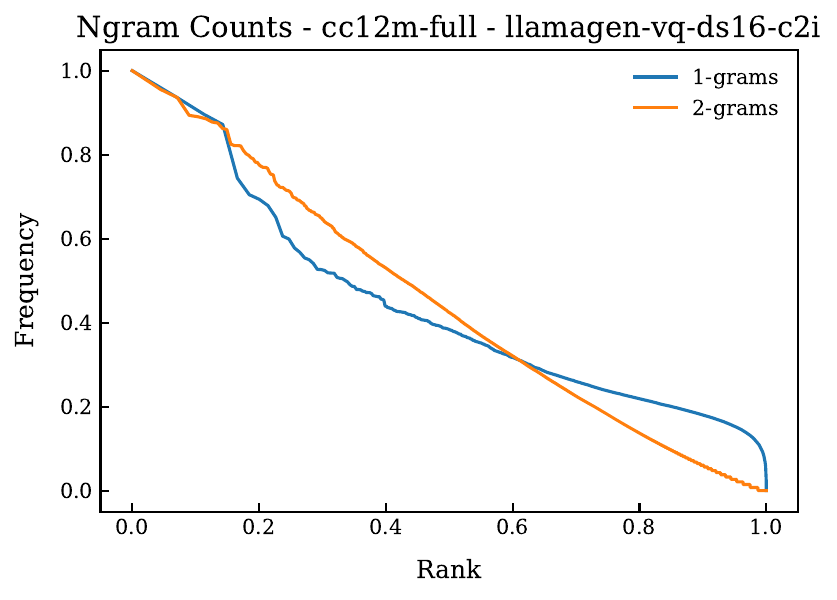}
            \caption{\tiny LlamaGen}
        \end{subfigure}
        \hfill
    \end{minipage}
    \vspace{0.5cm} %

    \begin{minipage}{0.15\textwidth}
        \raggedleft
        \textbf{coco} %
    \end{minipage}%
    \begin{minipage}{0.80\textwidth}
        \begin{subfigure}[b]{0.19\textwidth}
            \includegraphics[width=\textwidth]{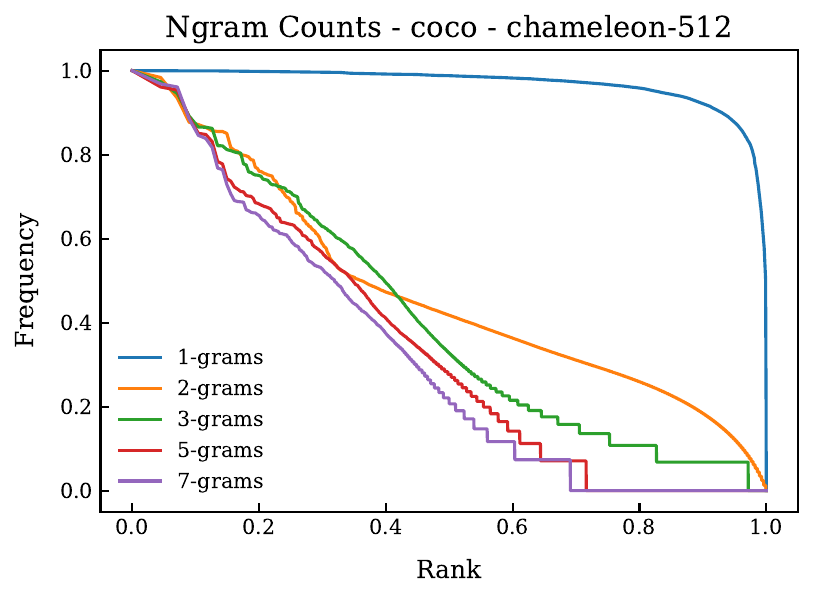}
            \caption{\tiny Chameleon}
        \end{subfigure}
        \begin{subfigure}[b]{0.19\textwidth}
            \includegraphics[width=\textwidth]{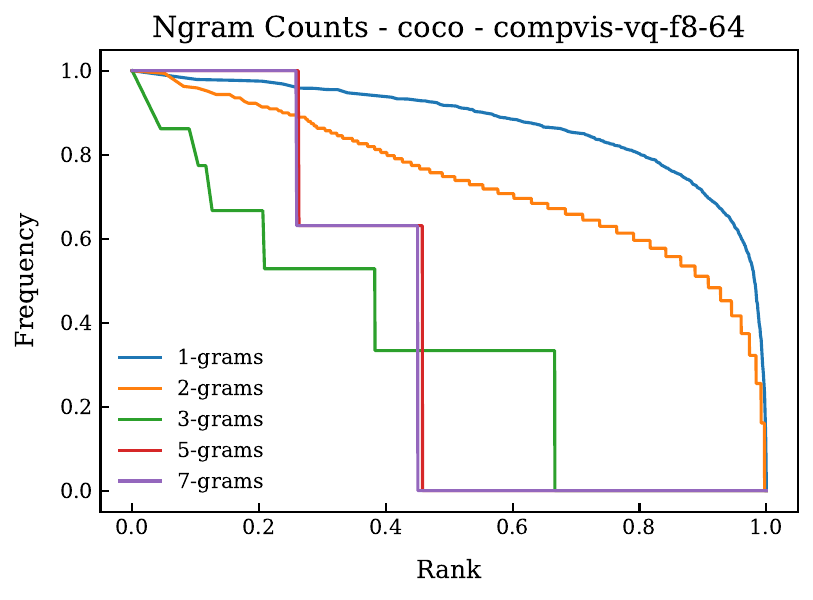}
            \caption{\tiny VQ-VAE (64)}
        \end{subfigure}
        \begin{subfigure}[b]{0.19\textwidth}
            \includegraphics[width=\textwidth]{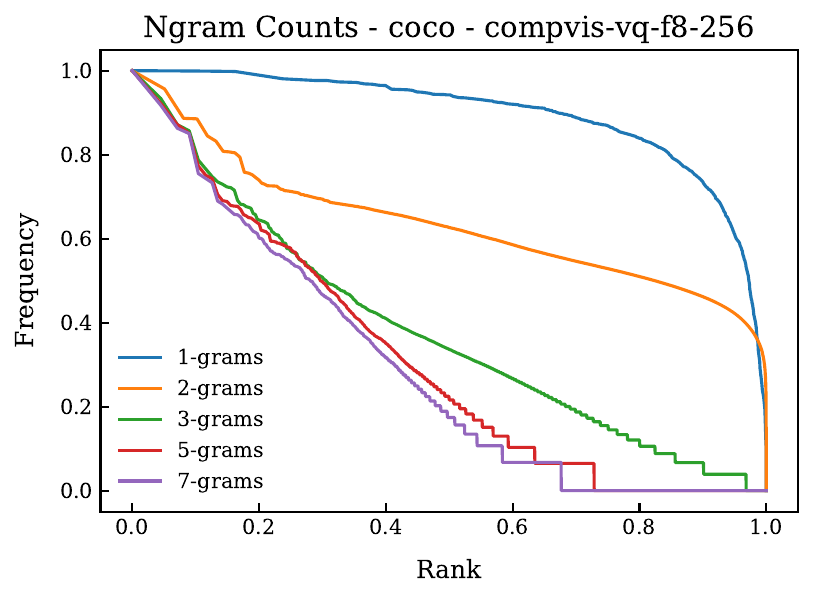}
            \caption{\tiny VQ-VAE (256)}
        \end{subfigure}
        \begin{subfigure}[b]{0.19\textwidth}
            \includegraphics[width=\textwidth]{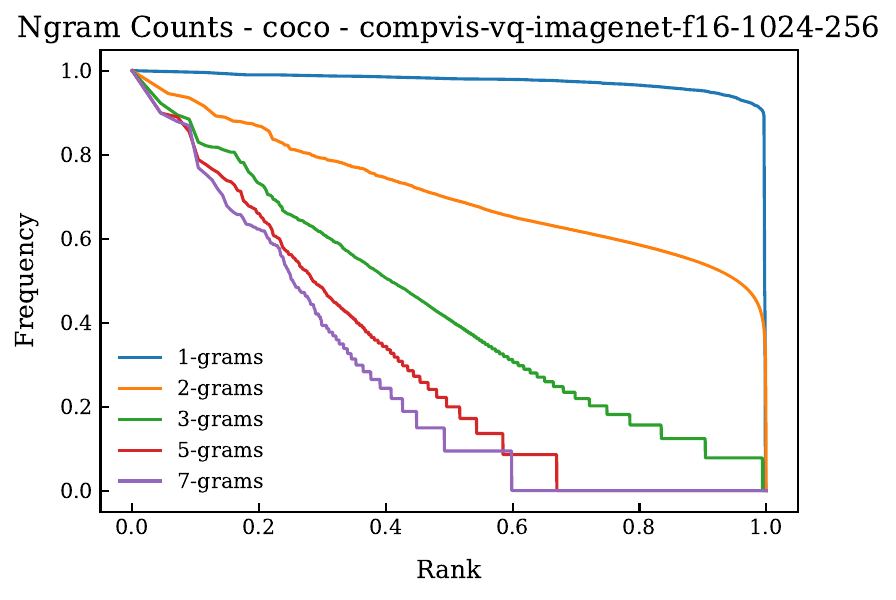}
            \caption{\tiny VQ-VAE (ImageNet)}
        \end{subfigure}
        \begin{subfigure}[b]{0.19\textwidth}
            \includegraphics[width=\textwidth]{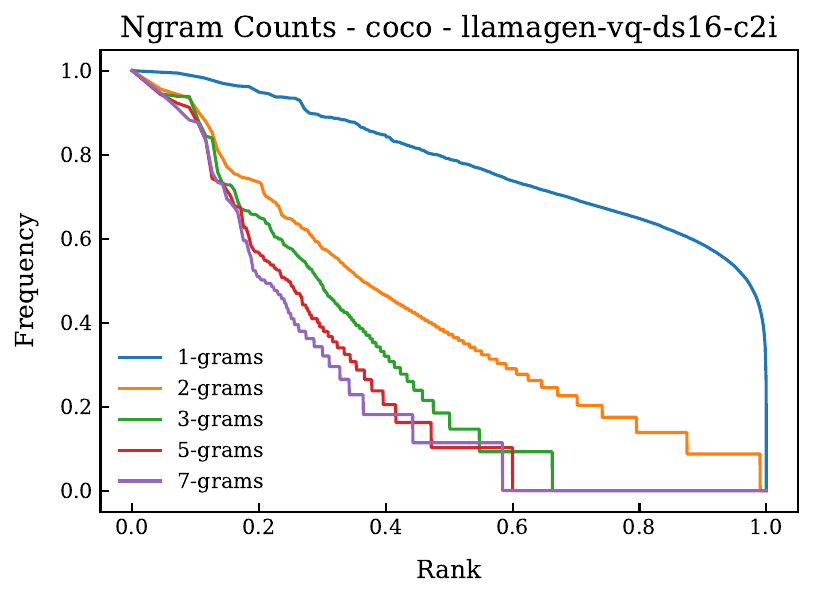}
            \caption{\tiny LlamaGen}
        \end{subfigure}
    \end{minipage}
    \vspace{0.5cm} %

    \begin{minipage}{0.15\textwidth}
        \raggedleft
        \textbf{ilsvrc} %
    \end{minipage}%
    \begin{minipage}{0.80\textwidth}
        \begin{subfigure}[b]{0.19\textwidth}
            \includegraphics[width=\textwidth]{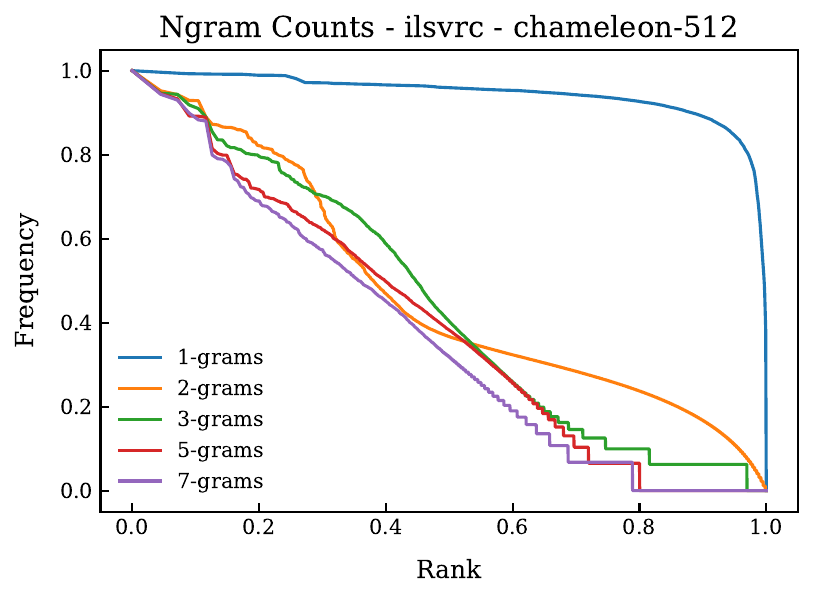}
            \caption{\tiny Chameleon}
        \end{subfigure}
        \begin{subfigure}[b]{0.19\textwidth}
            \includegraphics[width=\textwidth]{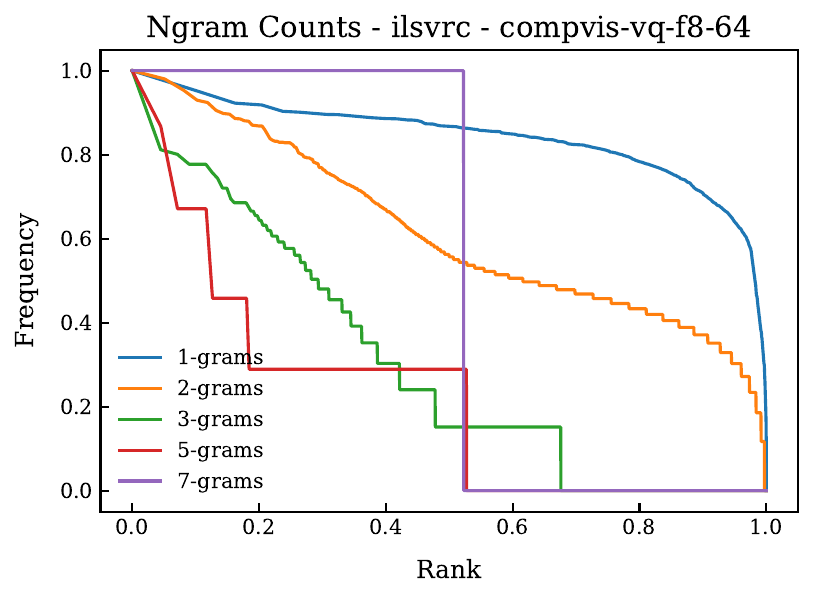}
            \caption{\tiny VQ-VAE (64)}
        \end{subfigure}
        \begin{subfigure}[b]{0.19\textwidth}
            \includegraphics[width=\textwidth]{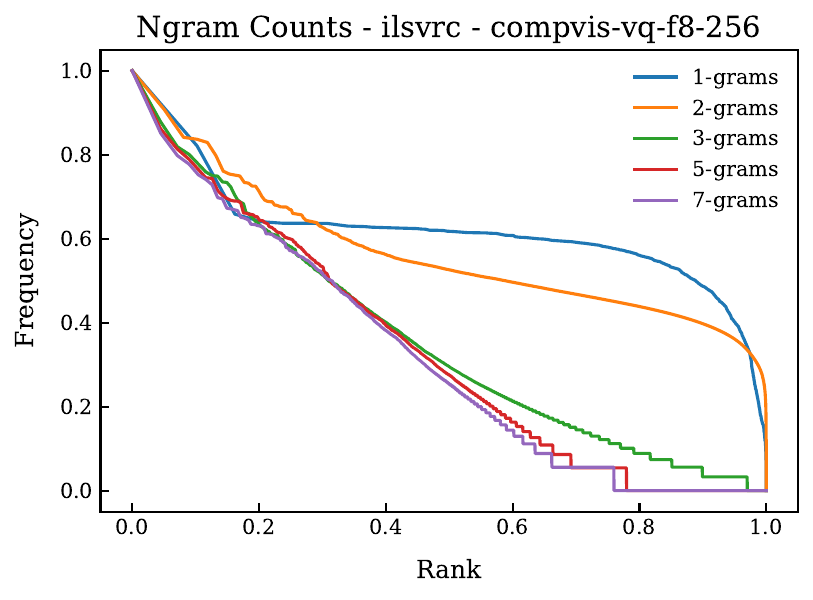}
            \caption{\tiny VQ-VAE (256)}
        \end{subfigure}
        \begin{subfigure}[b]{0.19\textwidth}
            \includegraphics[width=\textwidth]{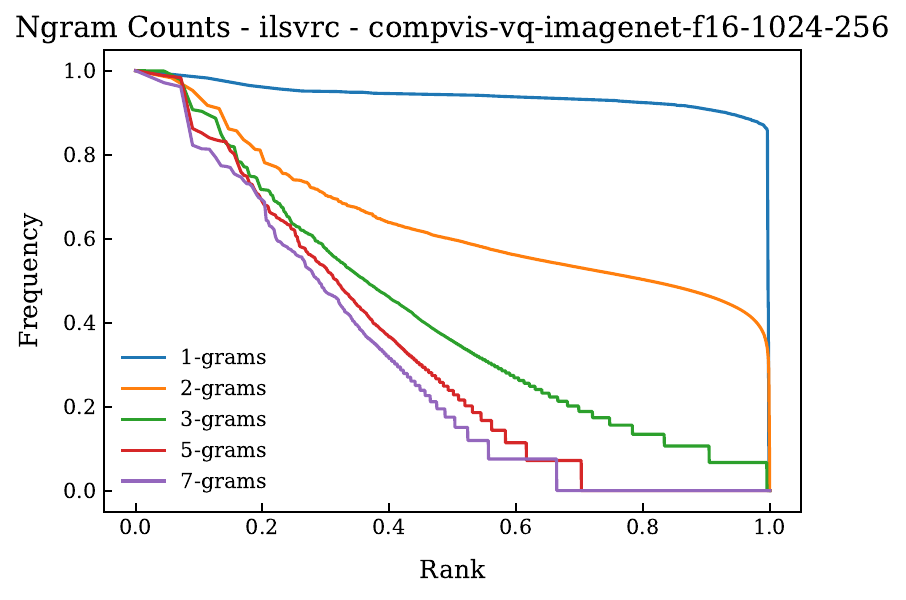}
            \caption{\tiny VQ-VAE (ImageNet)}
        \end{subfigure}
        \begin{subfigure}[b]{0.19\textwidth}
            \includegraphics[width=\textwidth]{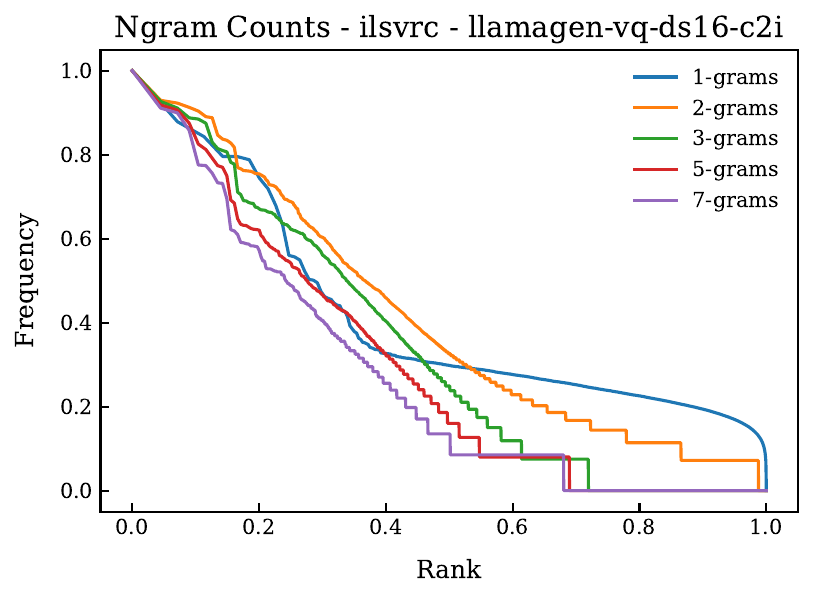}
            \caption{\tiny LlamaGen}
        \end{subfigure}
    \end{minipage}
    \vspace{0.5cm} %

    \begin{minipage}{0.15\textwidth}
        \raggedleft
        \textbf{spin} %
    \end{minipage}%
    \begin{minipage}{0.80\textwidth}
        \begin{subfigure}[b]{0.19\textwidth}
            \includegraphics[width=\textwidth]{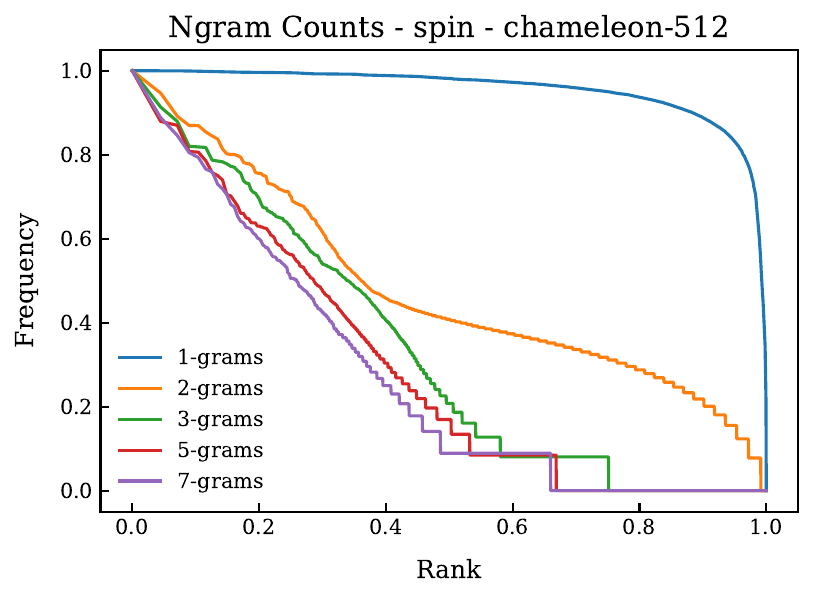}
            \caption{\tiny Chameleon}
        \end{subfigure}
        \begin{subfigure}[b]{0.19\textwidth}
            \includegraphics[width=\textwidth]{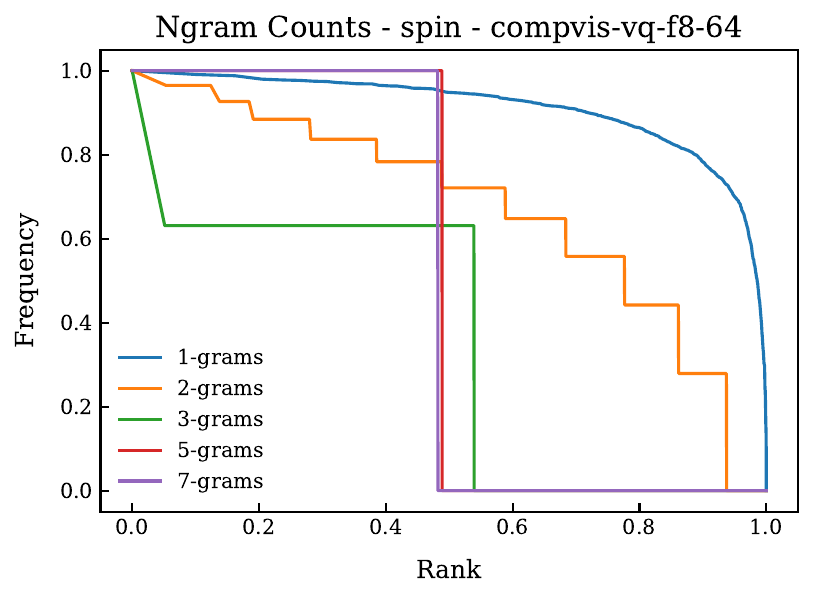}
            \caption{\tiny VQ-VAE (64)}
        \end{subfigure}
        \begin{subfigure}[b]{0.19\textwidth}
            \includegraphics[width=\textwidth]{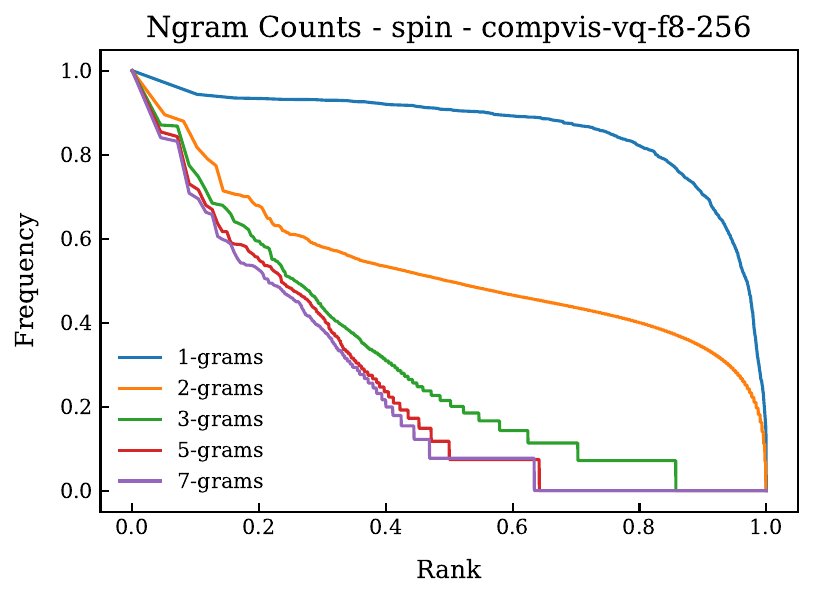}
            \caption{\tiny VQ-VAE (256)}
        \end{subfigure}
        \begin{subfigure}[b]{0.19\textwidth}
            \includegraphics[width=\textwidth]{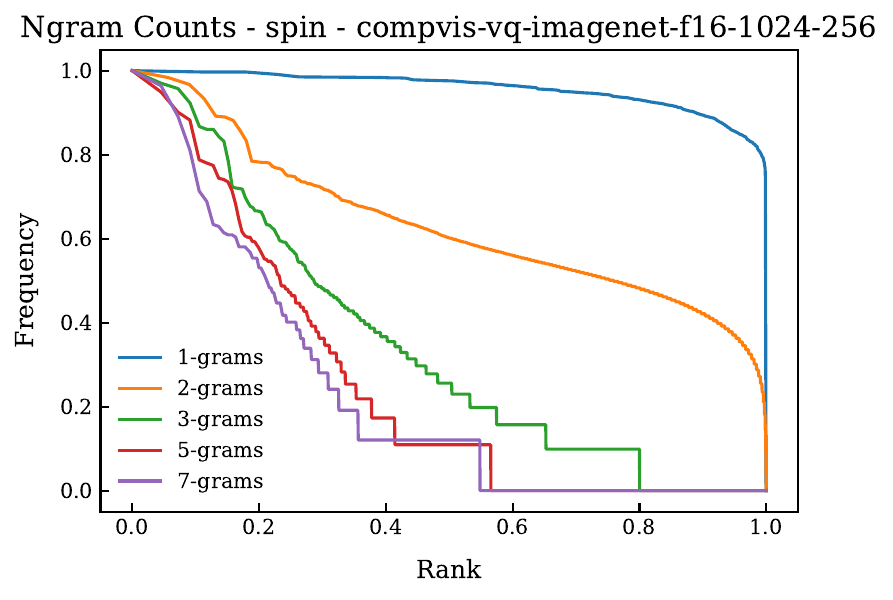}
            \caption{\tiny VQ-VAE (ImageNet)}
        \end{subfigure}
        \begin{subfigure}[b]{0.19\textwidth}
            \includegraphics[width=\textwidth]{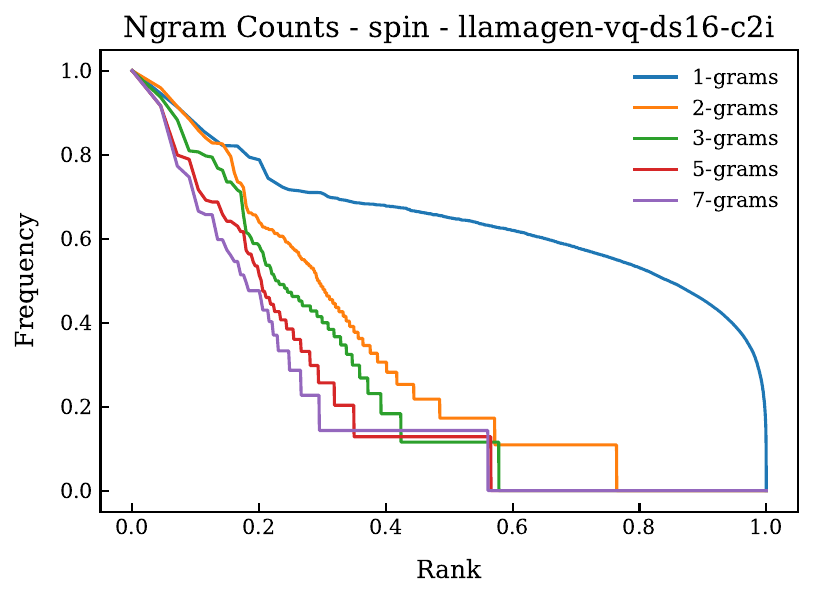}
            \caption{\tiny LlamaGen}
        \end{subfigure}
    \end{minipage}
    \vspace{0.5cm} %

    \begin{minipage}{0.15\textwidth}
        \raggedleft
        \textbf{xm3600} %
    \end{minipage}%
    \begin{minipage}{0.80\textwidth}
        \begin{subfigure}[b]{0.19\textwidth}
            \includegraphics[width=\textwidth]{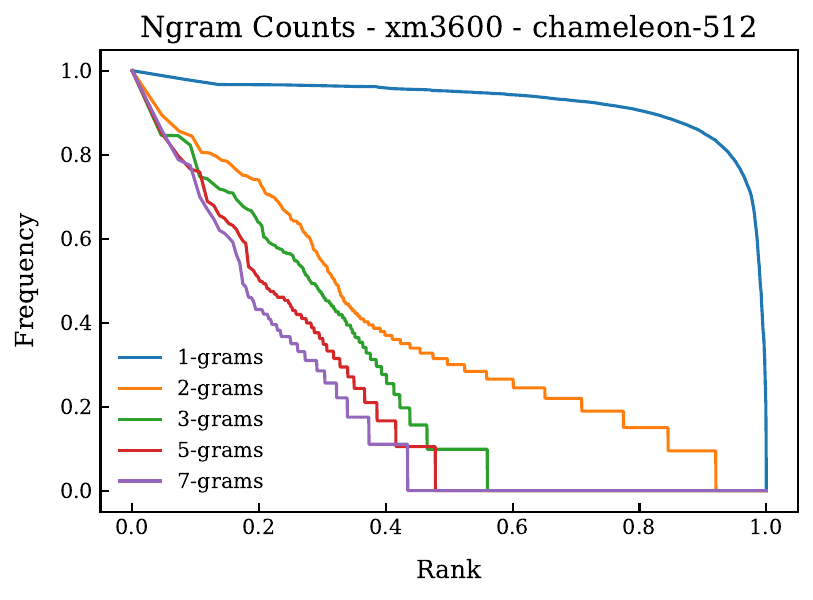}
            \caption{\tiny Chameleon}
        \end{subfigure}
        \begin{subfigure}[b]{0.19\textwidth}
            \includegraphics[width=\textwidth]{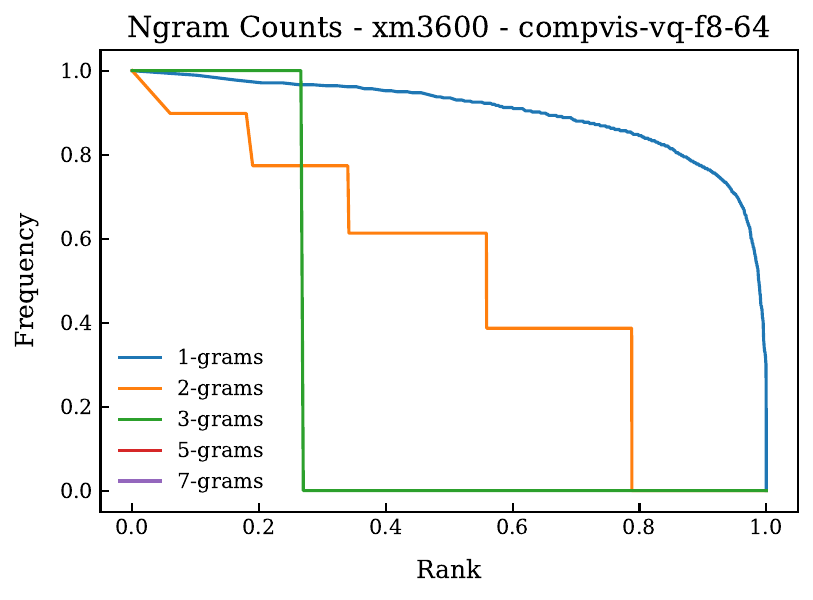}
            \caption{\tiny VQ-VAE (64)}
        \end{subfigure}
        \begin{subfigure}[b]{0.19\textwidth}
            \includegraphics[width=\textwidth]{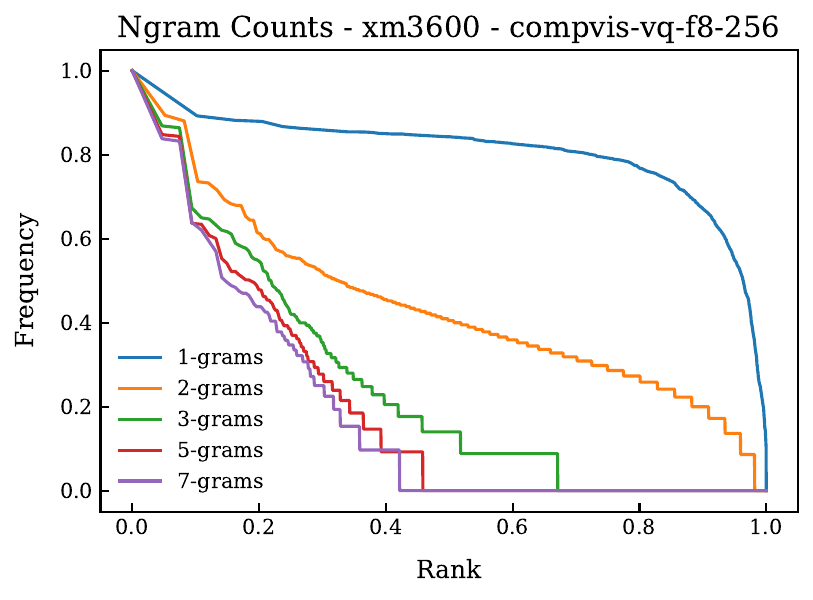}
            \caption{\tiny VQ-VAE (256)}
        \end{subfigure}
        \begin{subfigure}[b]{0.19\textwidth}
            \includegraphics[width=\textwidth]{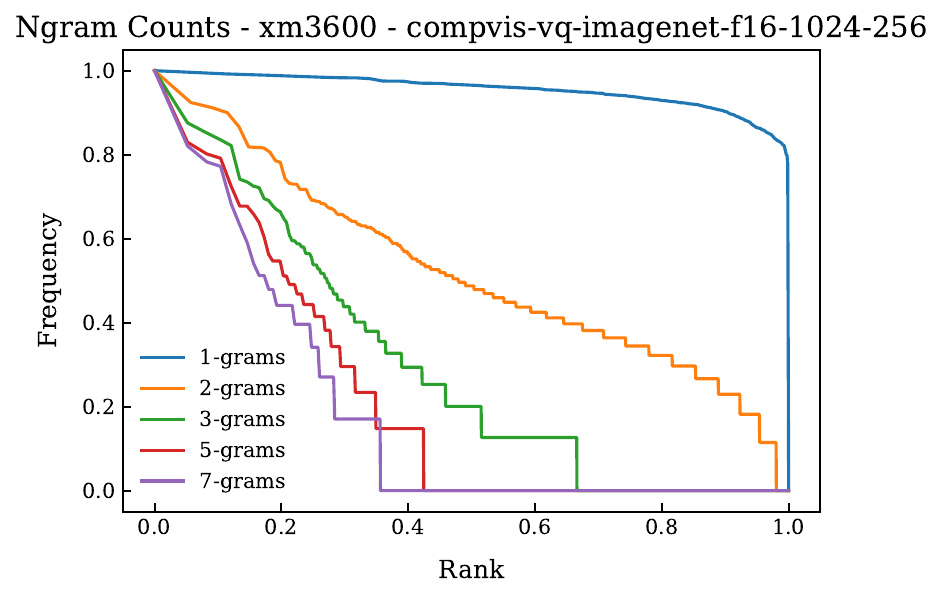}
            \caption{\tiny VQ-VAE (ImageNet)}
        \end{subfigure}
        \begin{subfigure}[b]{0.19\textwidth}
            \includegraphics[width=\textwidth]{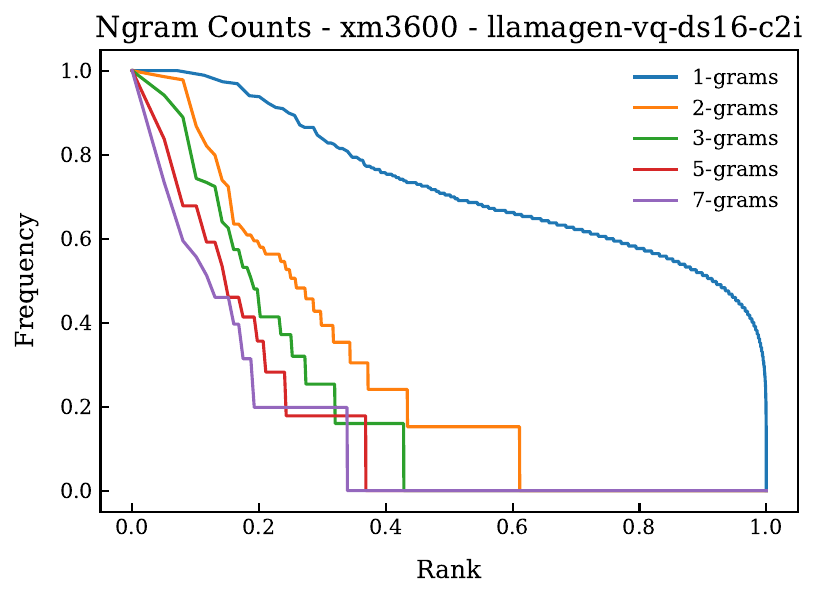}
            \caption{\tiny LlamaGen}
        \end{subfigure}
    \end{minipage}

    \caption{Empirical N-gram distributions for different datasets comparing normalized log-rank against normalized log-frequency. In general, visual languages do not achieve power-law distributions, and when they do, it is at high levels of N, and fairly steep slopes (compared to natural langauges).}
    \label{fig:zipf_full_ngrams}
\end{figure}

For each dataset and tokenizer, to compute the power law fit, we leverage the method/code in \citet{alstott2014powerlaw}. When fitting the power laws, because of computational limits, we limit the number of processed N-grams to 5M, and on CC12M and ILSVRC, unless otherwise noted, we compute the n-grams on only a subset of the full dataset consisting of a randomly sub-sampled 200K image set). Results broken down by N-gram are shown in \autoref{fig:zipf_n}, while results broken down by model/dataset are given in \autoref{fig:zipf_full_ngrams}

\section{Heaps'/Herdan's Law}
\label{app:heaps}

Heaps' law (also referred to as Herdan's law) is an empirical rule that describes the relationship between the size of a corpus  and the number of unique word in the corpus \citep{heaps1978information,herdan1964quantitative}. Specifically, it predicts that as the size of a text grows, the number of unique words increases, but at a decreasing rate. 

Mathematically, the law is described by:
\begin{equation}
    V(N) = kN^\beta
\end{equation}
where V(N) is the number of distinct words (the vocabulary size), N is the total number of words, and $k$ and $\beta$ are parameters, $0 < \beta < 1$. Heaps' law reflects the fact that even as new text is added to a corpus, the frequency of newly introduced words diminishes, meaning a large corpus doesn't proportionally expand its vocabulary. 

\begin{figure}[t]
    \centering
    \begin{subfigure}[b]{0.49\textwidth}
        \centering
        \includegraphics[width=\textwidth]{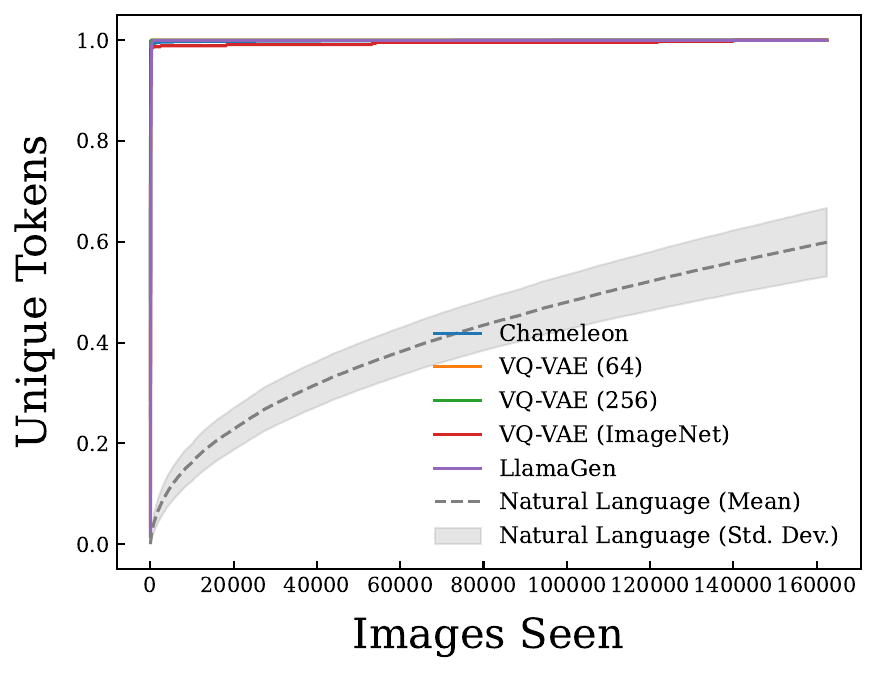}
        \caption{1-grams}
        \label{fig:heaps_ngrams1_coco}
    \end{subfigure}
    \hfill
    \begin{subfigure}[b]{0.49\textwidth}
        \centering
        \includegraphics[width=\textwidth]{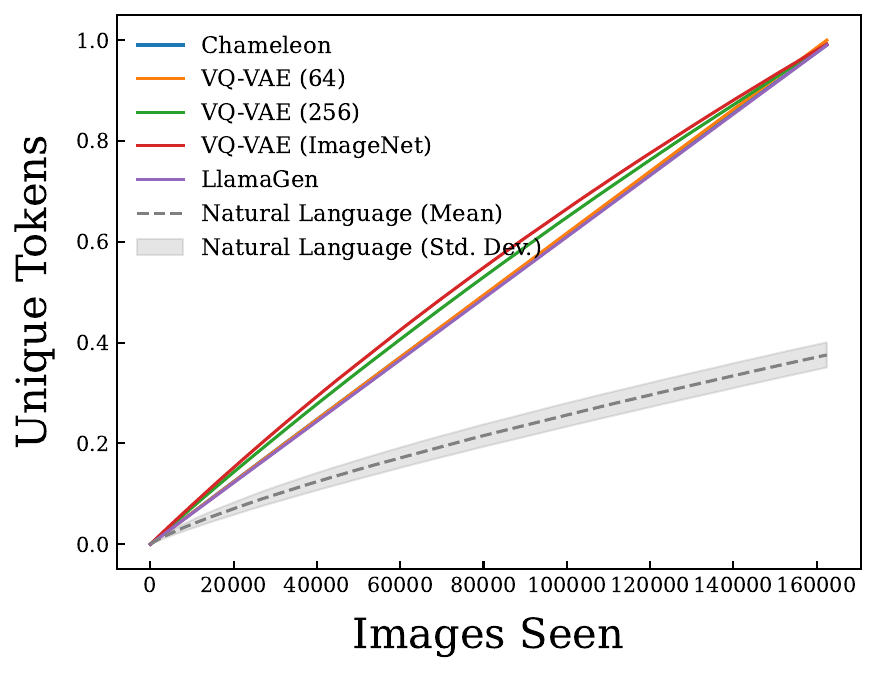}
        \caption{2-grams}
        \label{fig:heaps_ngrams2_coco}
    \end{subfigure}
    \hfill
    \begin{subfigure}[b]{0.49\textwidth}
        \centering
        \includegraphics[width=\textwidth]{sections/appendix/heaps/heaps_law_coco_3.pdf}
        \caption{3-grams}
        \label{fig:heaps_ngrams3_coco}
    \end{subfigure}
    \hfill
    \begin{subfigure}[b]{0.49\textwidth}
        \centering
        \includegraphics[width=\textwidth]{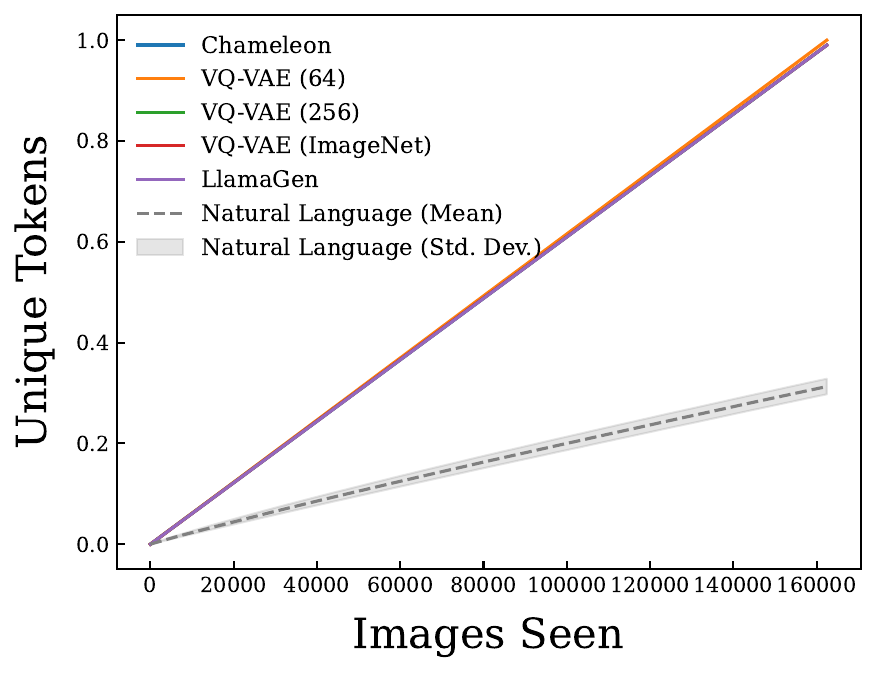}
        \caption{5-grams}
        \label{fig:heaps_ngrams4_coco}
    \end{subfigure}
    
    \caption{Comparison of unique tokens as a function of images seen on the MS-COCO dataset for different N-grams. }
    \label{fig:heaps_coco}
\end{figure}

Plots for unique tokens vs. images seen on XM-3600 are given in \autoref{fig:heaps_xm3600}, with those for MS-COCO given in \autoref{fig:heaps_coco}.

\section{Yule-Simon Distribution}
\label{app:yule-simon}

The Yule-Simon distribution \citep{willis1922some} is a model often used to describe processes where new elements (in this case, tokens) are introduced over time with a probability that decreases as the existing set of elements grows. Specifically, for a sequence of tokens, the Yule-Simon distribution describes the probability of the $k$-th token occurring $m$ times as:

\begin{equation}
P(m) = \alpha \, \mathrm{B}(m, \alpha+1)
\end{equation}

where $\mathrm{B}(\cdot, \cdot)$ is the Beta function. This captures the balance between token reuse and token innovation, and the shape parameter $\alpha$ reflects the likelihood of encountering a novel token versus reusing an existing one.

\subsection{Experimental Design}

For each dataset and tokenizer configuration on the COCO and XM-3600 datasets, we fit the Yule-Simon distribution to the observed token frequency distributions by minimizing the negative log-likelihood using the L-BFGS-B optimization algorithm. This method is selected due to its ability to handle the bound constraints placed on the parameter $\alpha$, ensuring that $\alpha > 0$. The optimization starts with an initial guess of $\alpha = 1.0$, and the negative log-likelihood is computed based on the observed token frequencies. The optimization process continues until convergence, with the final $\alpha$ value corresponding to the best-fit parameter for the Yule-Simon distribution. Invalid $\alpha$ values are penalized by assigning an infinite log-likelihood to ensure feasible solutions. Once the optimal $\alpha$ is found, we compute the empirical PMF from the frequency distributions by normalizing the observed token counts. In parallel, the theoretical PMF is computed using the fitted $\alpha$ value.

\subsection{Additional Experimental Results}

\begin{figure}
    \centering
    \tiny
    \setlength{\tabcolsep}{1pt} %
    \renewcommand{\arraystretch}{1.2} %
    \captionsetup[subfigure]{labelfont=scriptsize, textfont=scriptsize, skip=3pt} %
    \begin{tabular}{cccccc}
        \subcaptionbox{\tiny Arabic (ar)\vspace{1em}}{\includegraphics[width=0.15\textwidth]{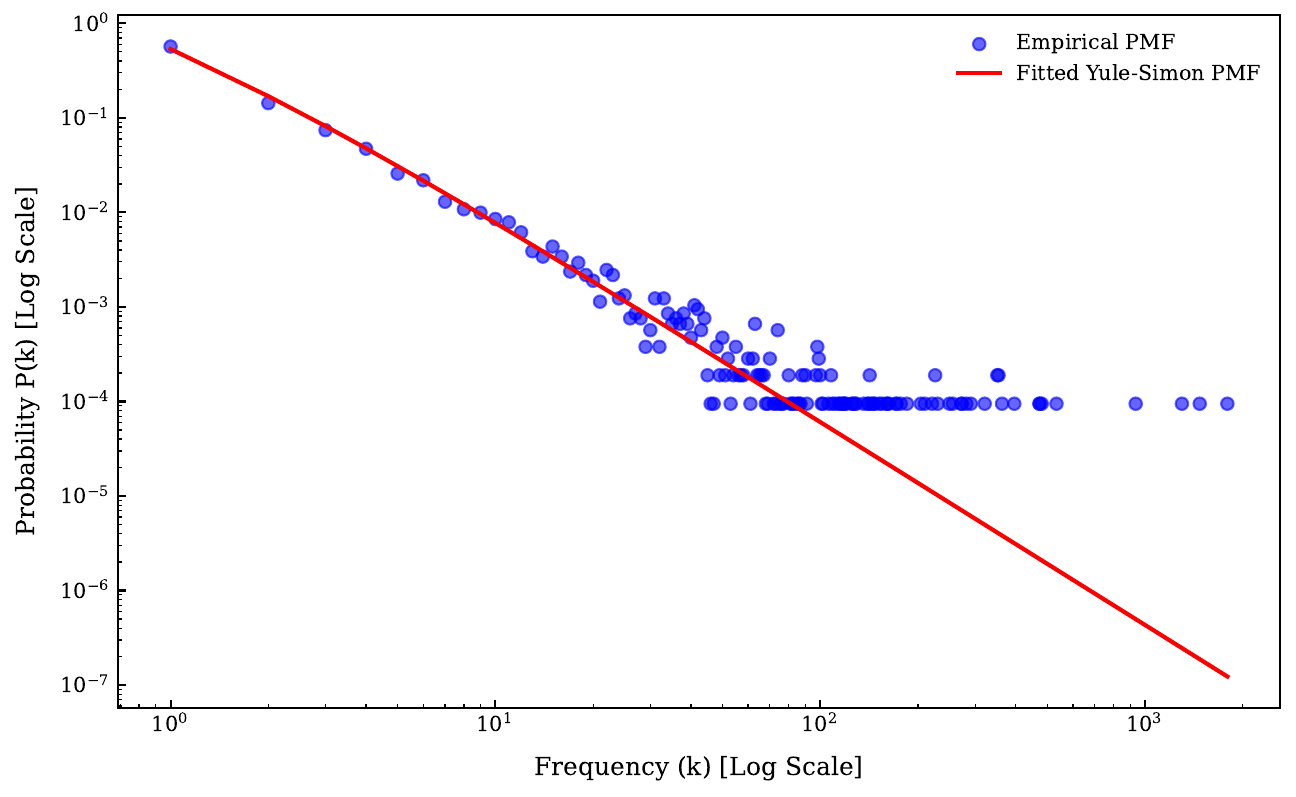}} &
        \subcaptionbox{\tiny Bengali (bn)}{\includegraphics[width=0.15\textwidth]{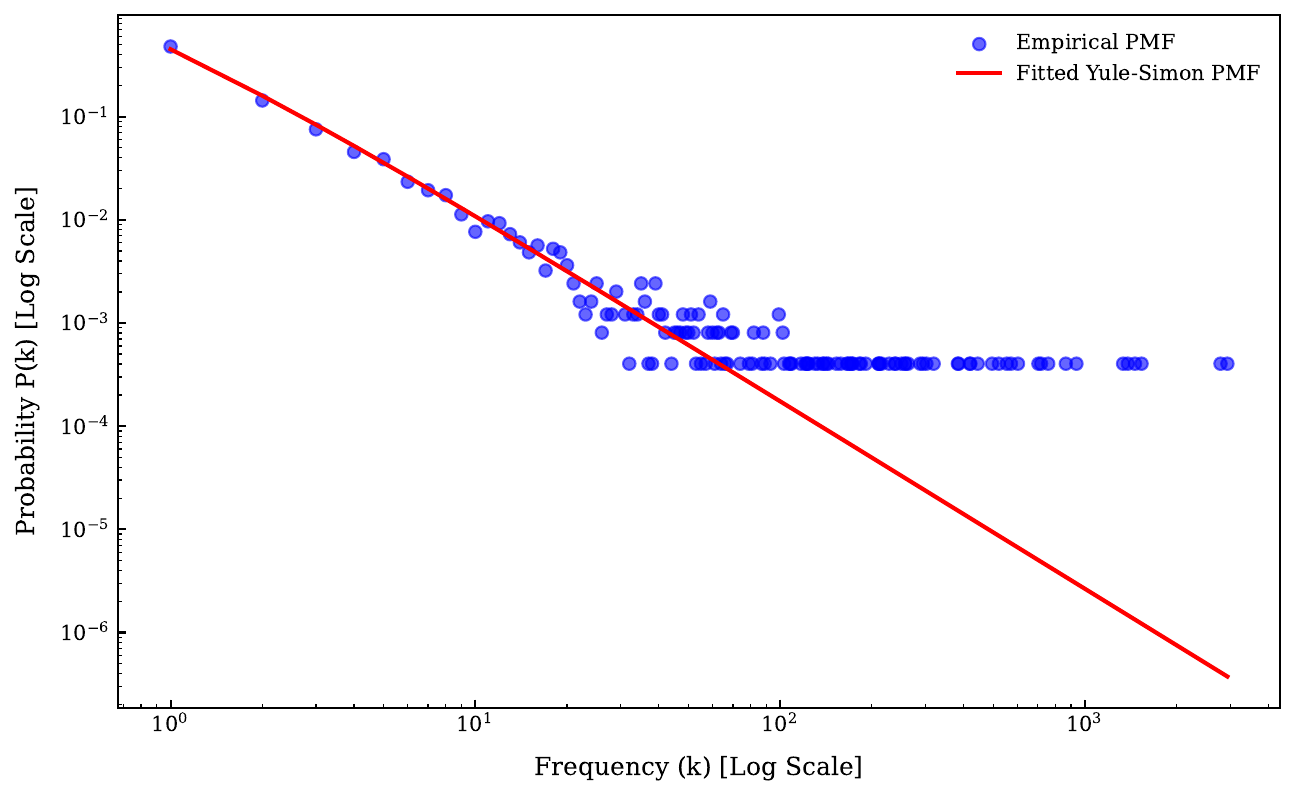}} &
        \subcaptionbox{\tiny Czech (cs)}{\includegraphics[width=0.15\textwidth]{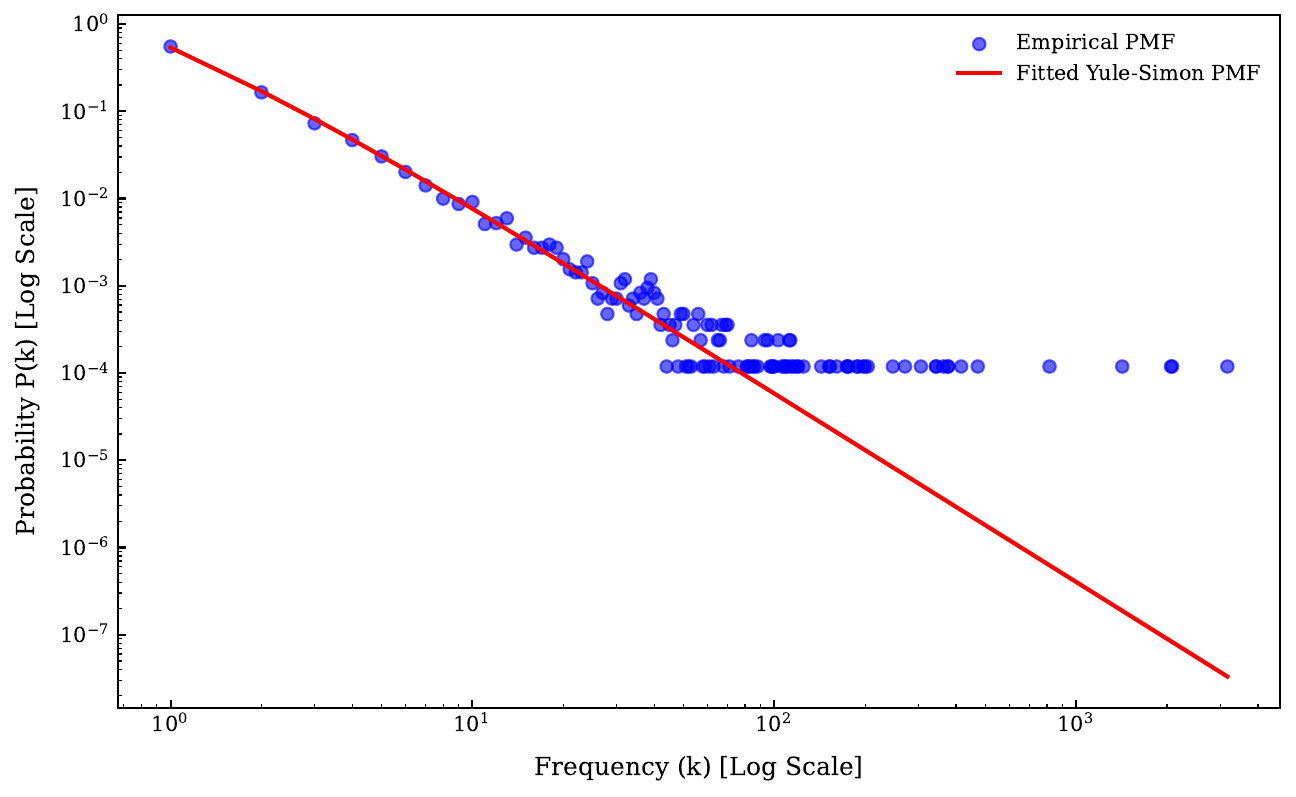}} &
        \subcaptionbox{\tiny Danish (da)}{\includegraphics[width=0.15\textwidth]{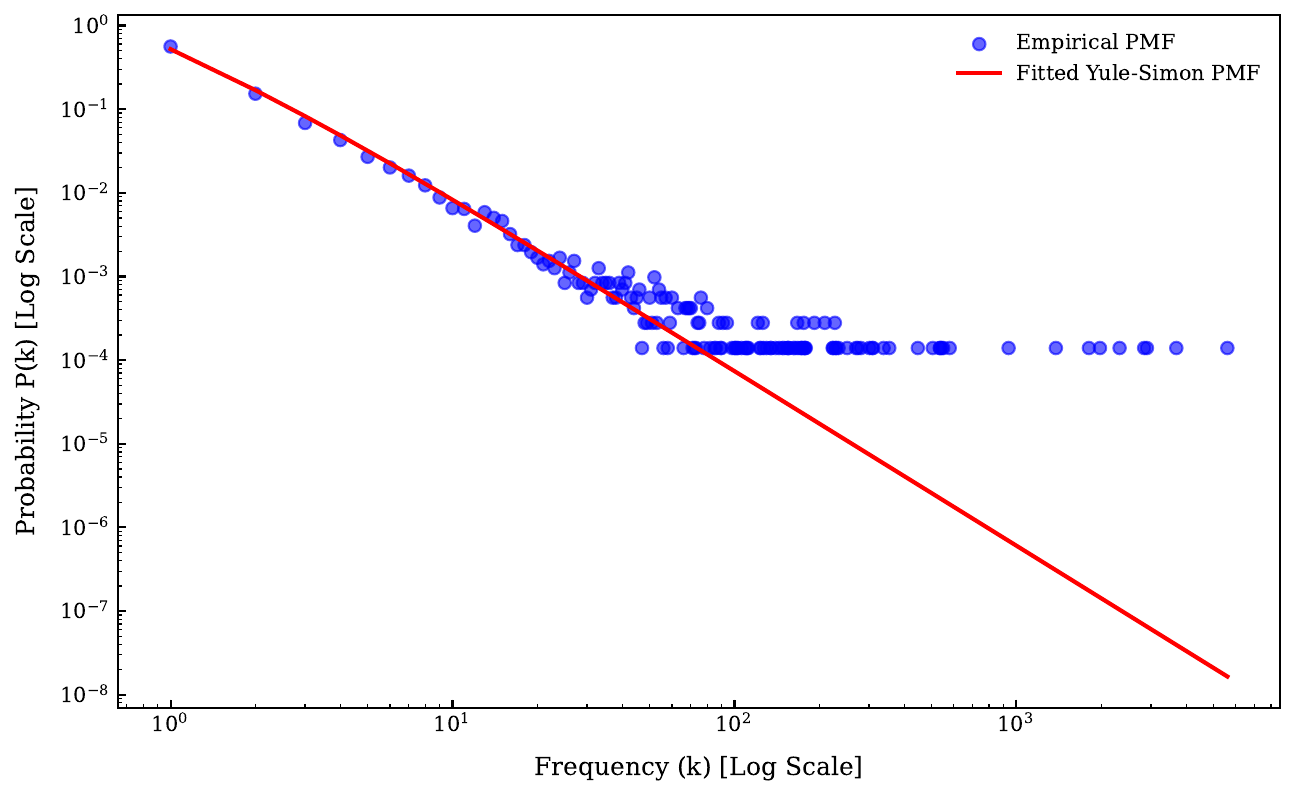}} &
        \subcaptionbox{\tiny German (de)}{\includegraphics[width=0.15\textwidth]{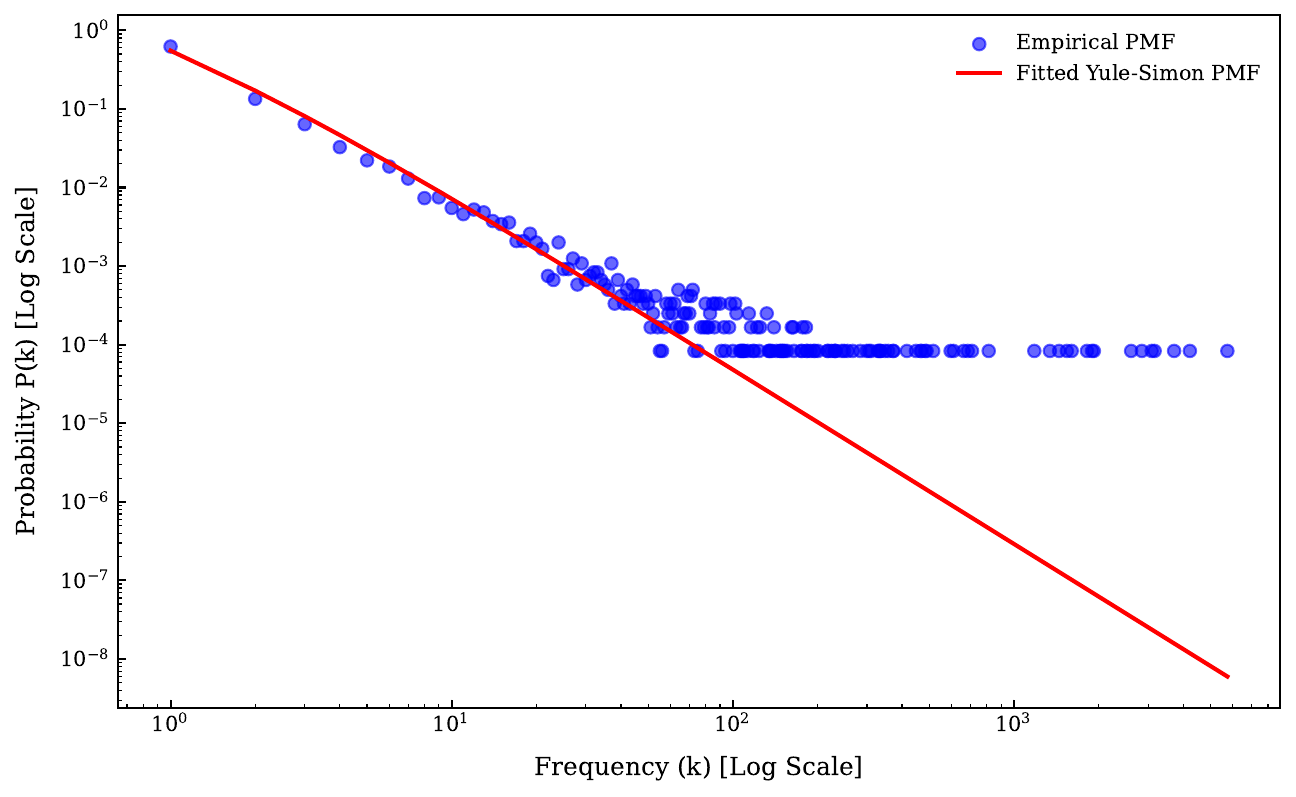}} &
        \subcaptionbox{\tiny Greek (el)}{\includegraphics[width=0.15\textwidth]{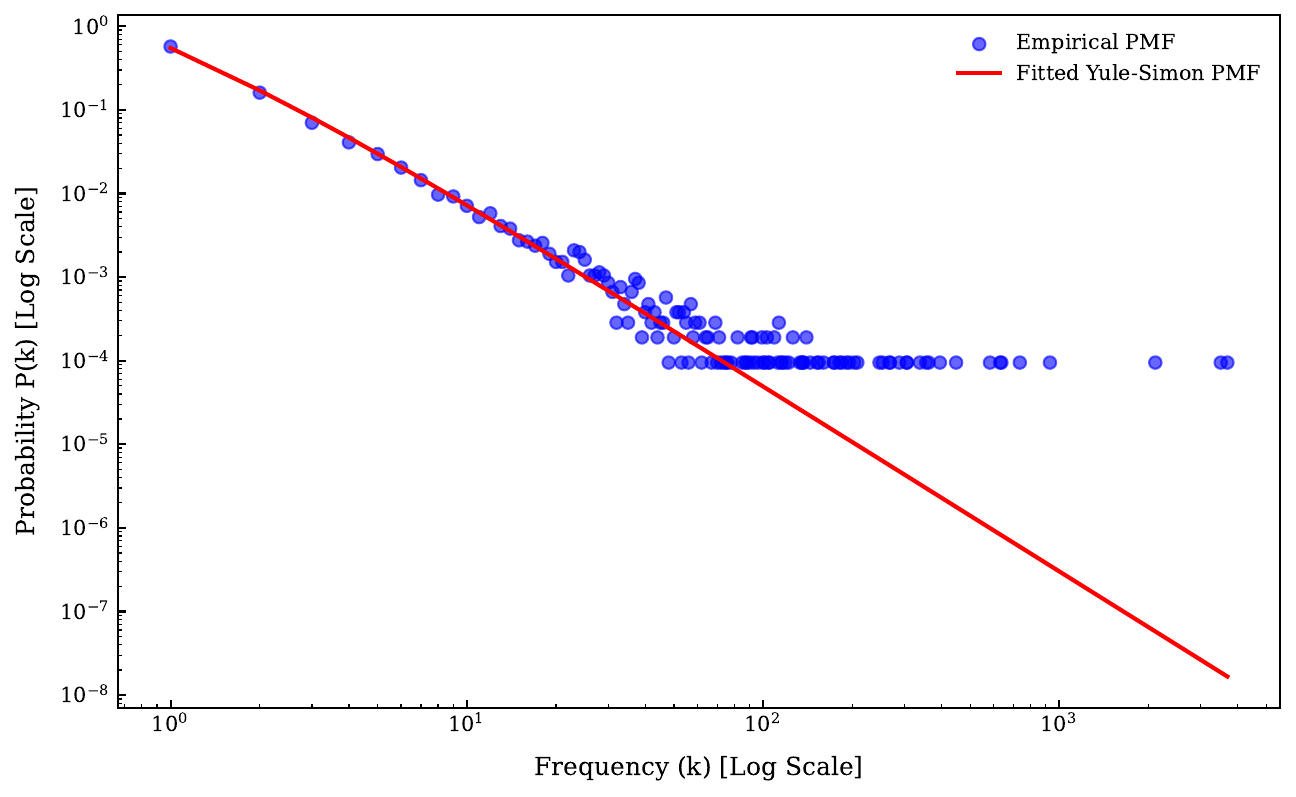}} \\
        
        \subcaptionbox{\tiny Spanish (es)\vspace{1em}}{\includegraphics[width=0.15\textwidth]{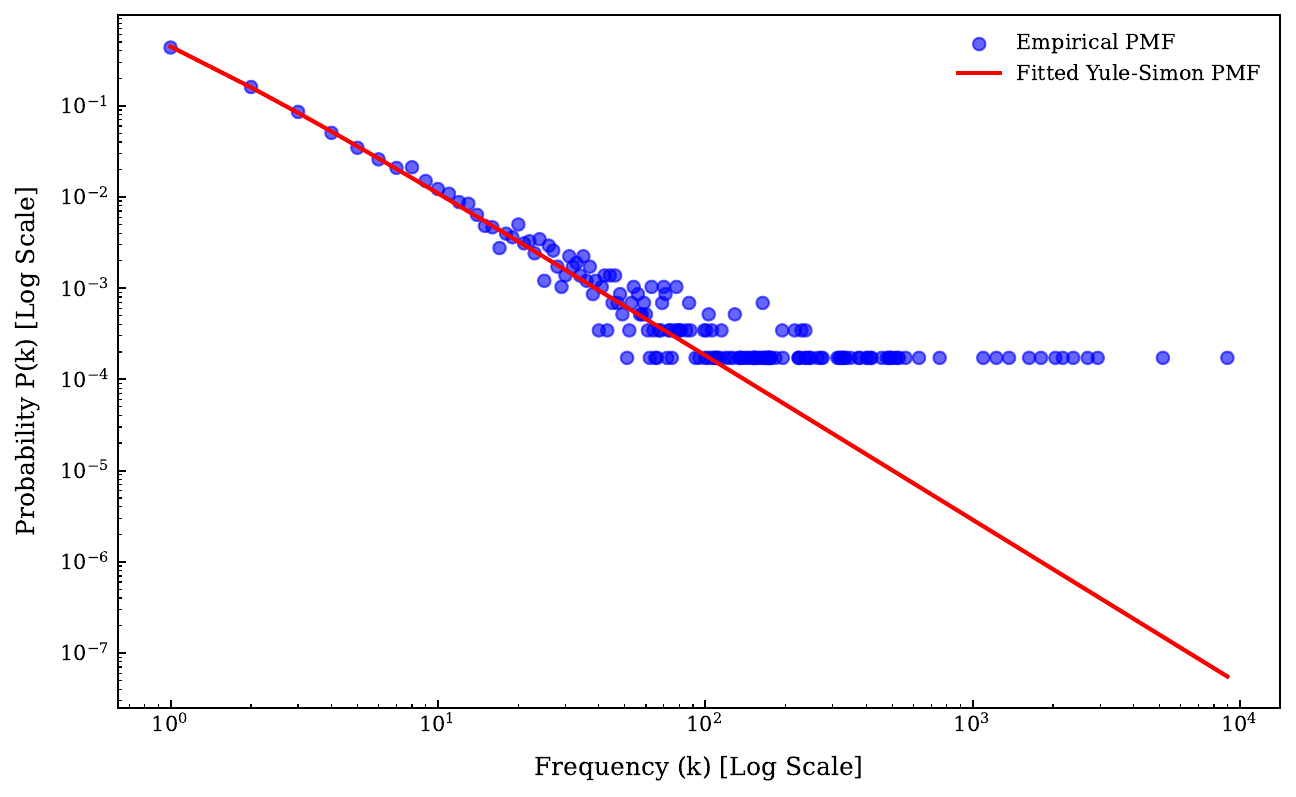}} &
        \subcaptionbox{\tiny English (en)}{\includegraphics[width=0.15\textwidth]{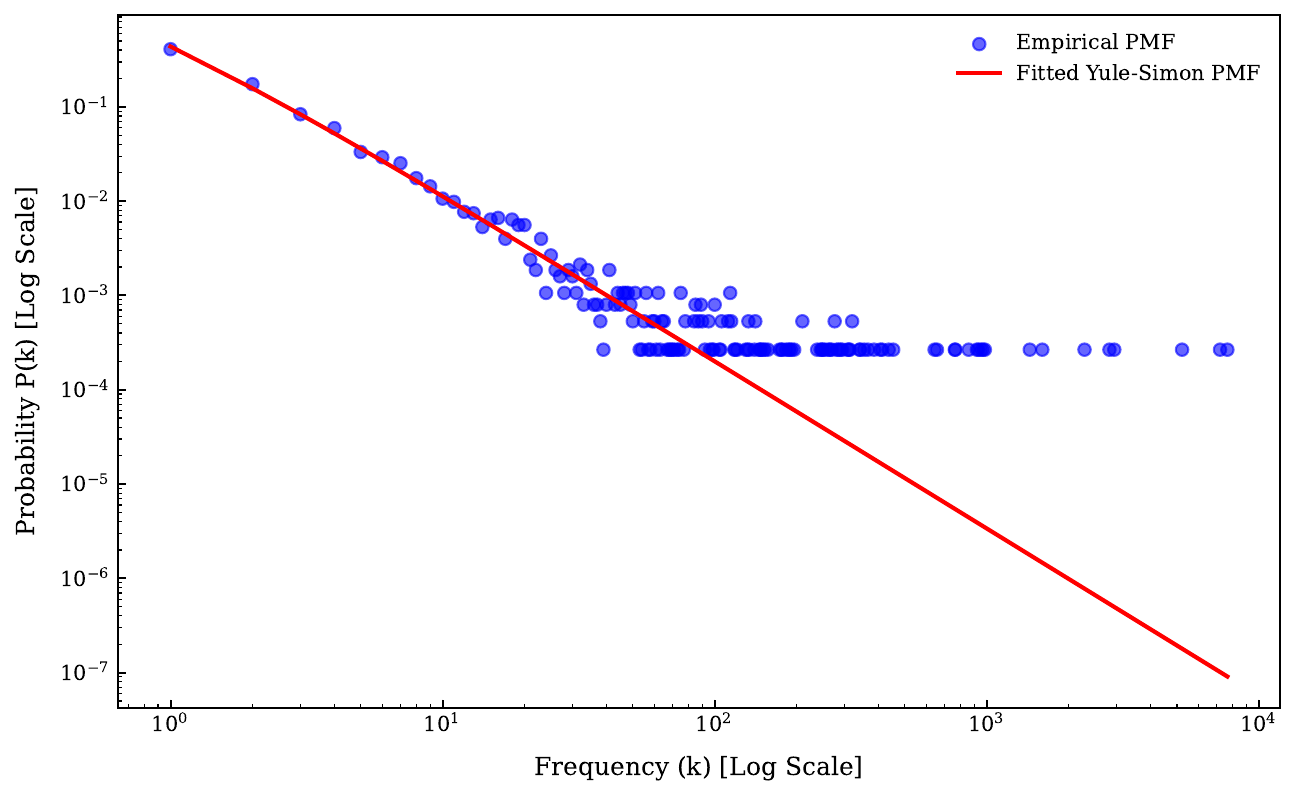}} &
        \subcaptionbox{\tiny Persian (fa)}{\includegraphics[width=0.15\textwidth]{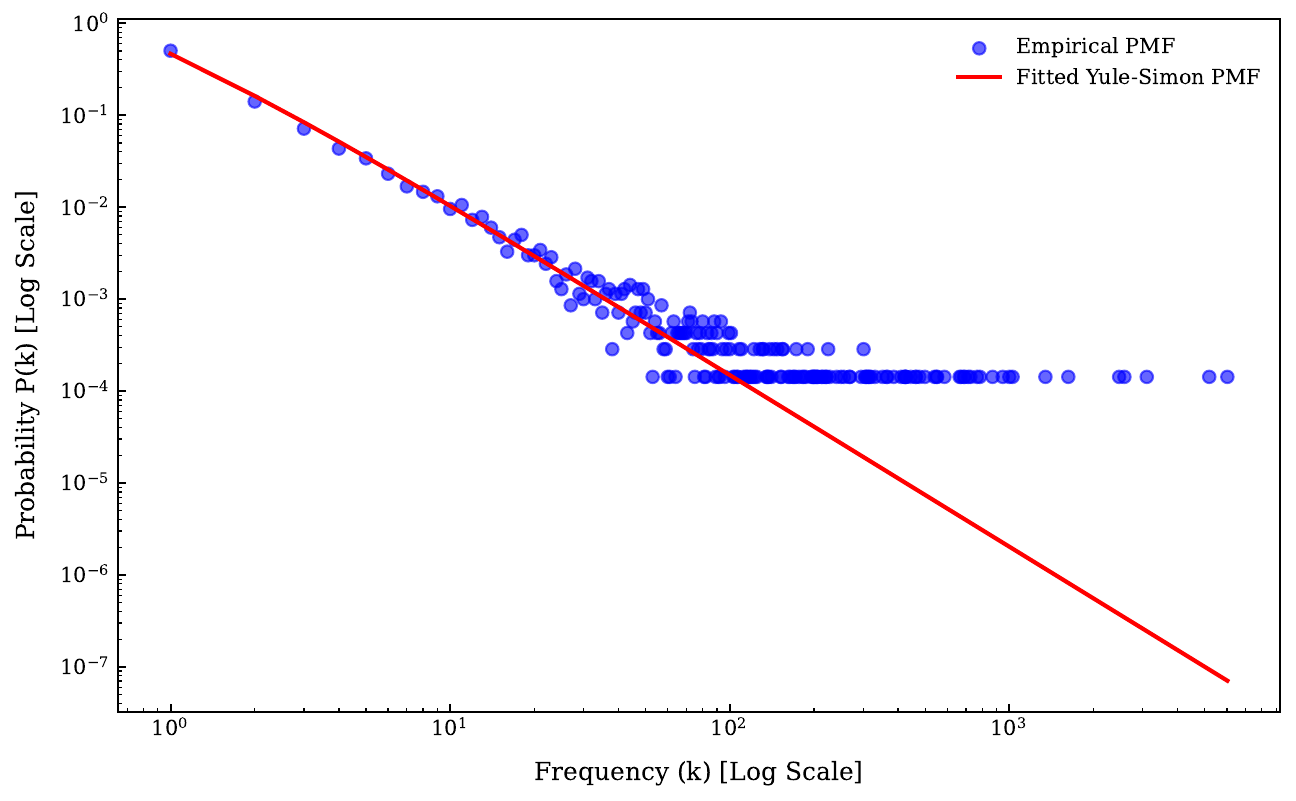}} &
        \subcaptionbox{\tiny Finnish (fi)}{\includegraphics[width=0.15\textwidth]{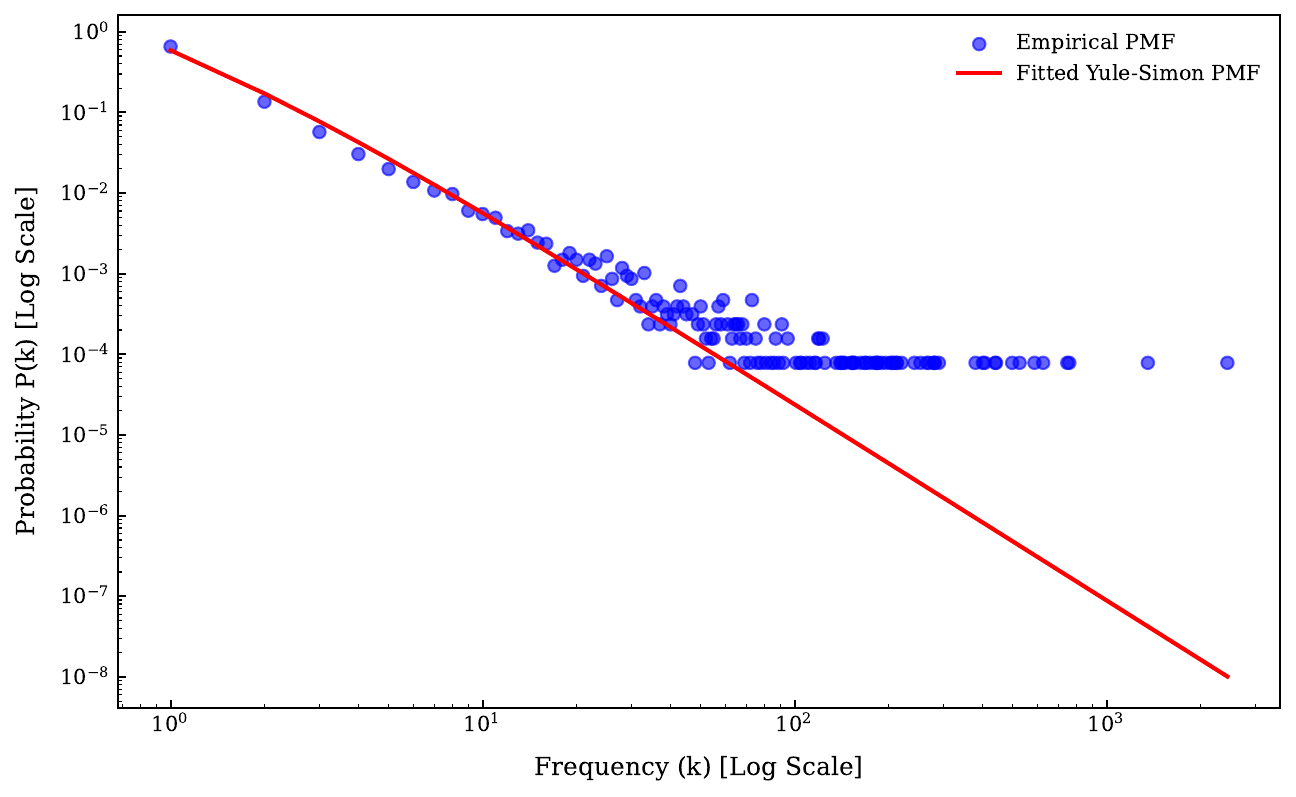}} &
        \subcaptionbox{\tiny Filipino (fil)}{\includegraphics[width=0.15\textwidth]{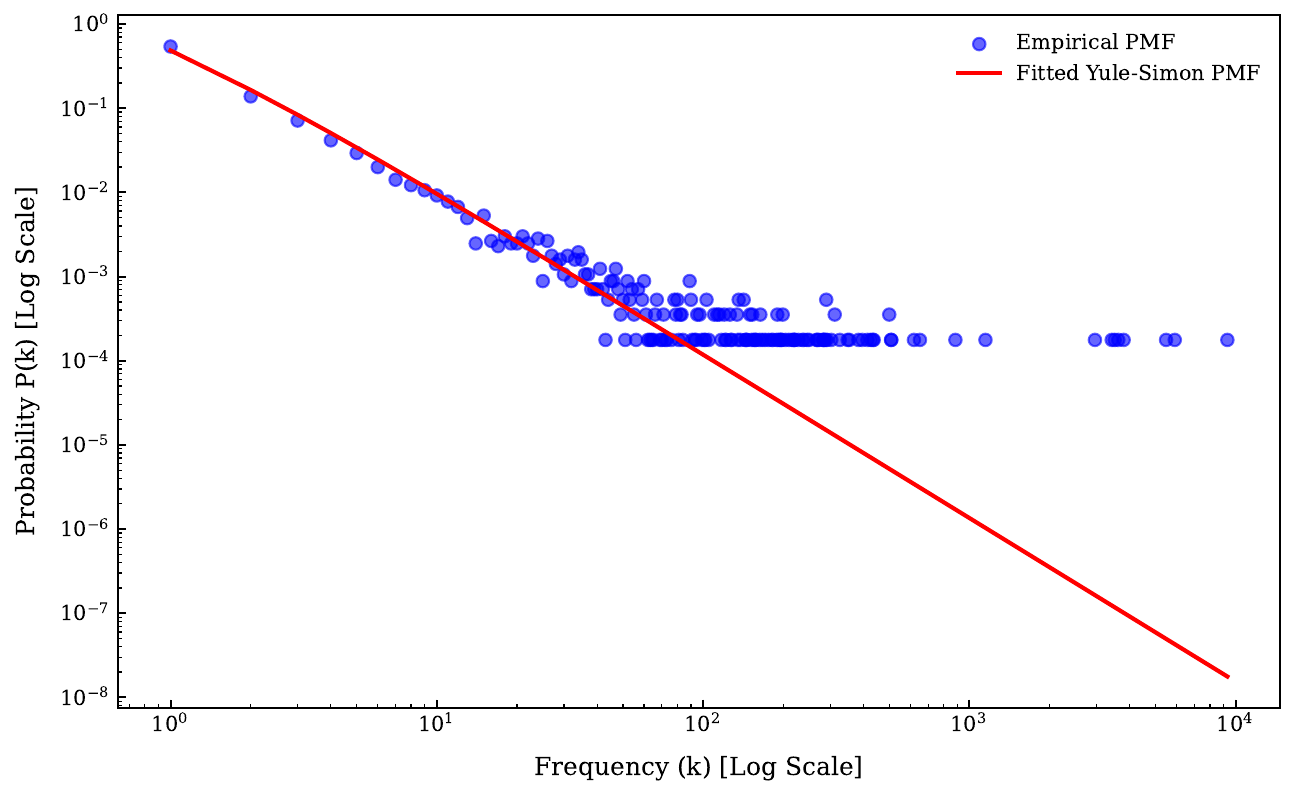}} &
        \subcaptionbox{\tiny French (fr)}{\includegraphics[width=0.15\textwidth]{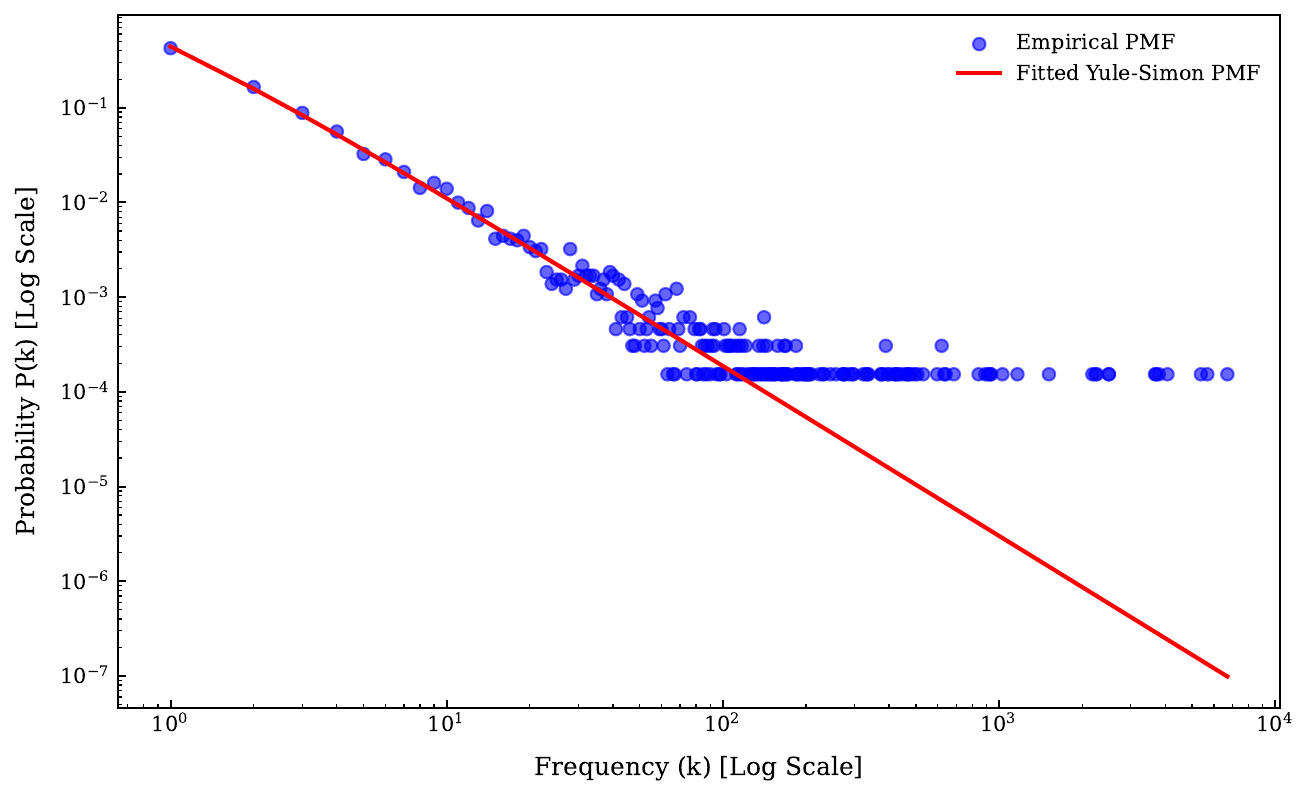}} \\
        
        \subcaptionbox{\tiny Hindi (hi)\vspace{1em}}{\includegraphics[width=0.15\textwidth]{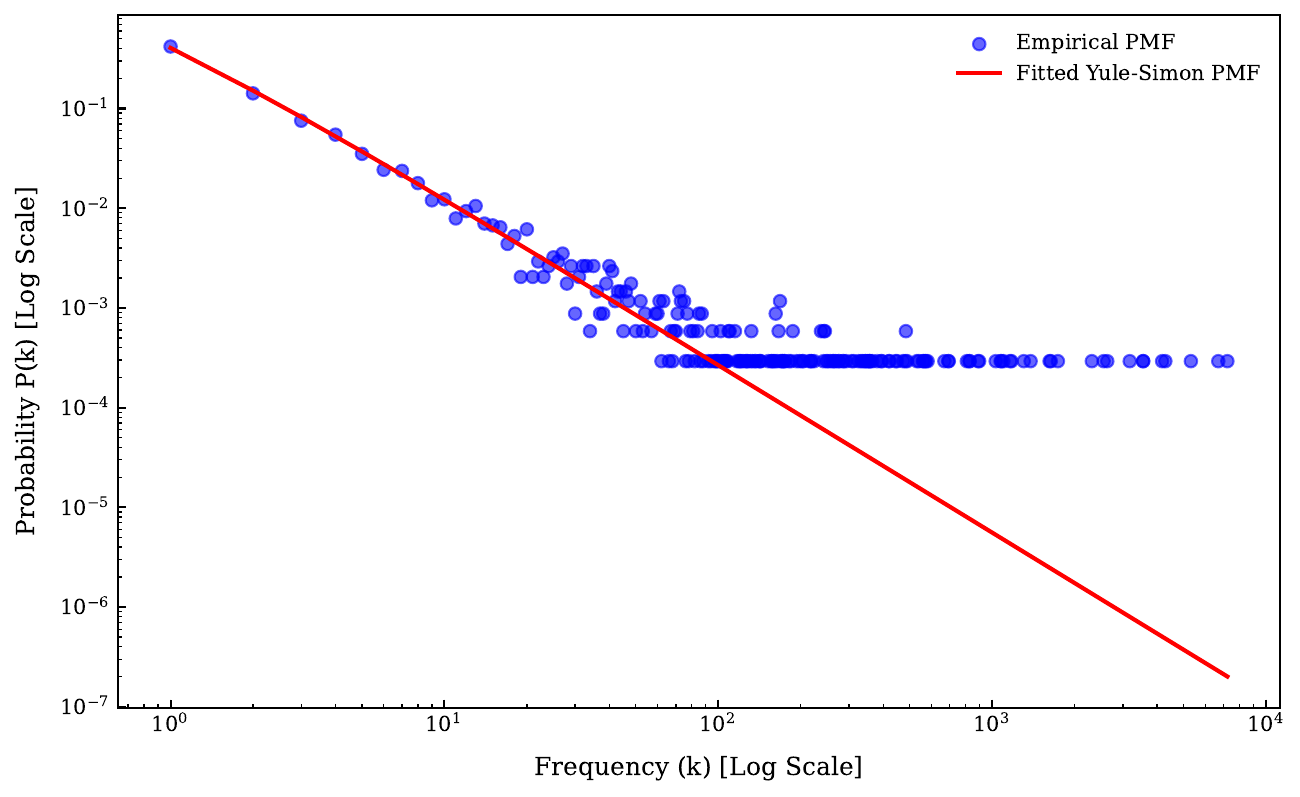}} &
        \subcaptionbox{\tiny Croatian (hr)}{\includegraphics[width=0.15\textwidth]{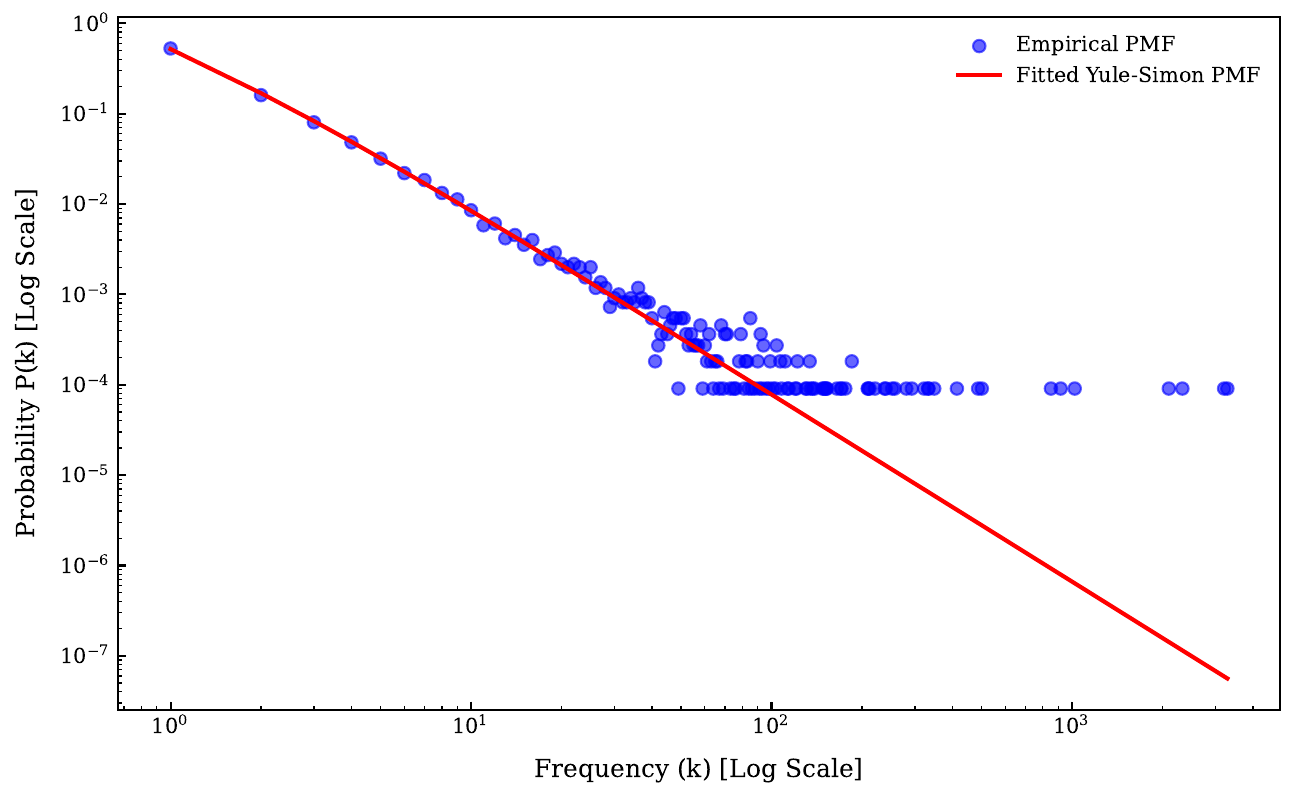}} &
        \subcaptionbox{\tiny Hungarian (hu)}{\includegraphics[width=0.15\textwidth]{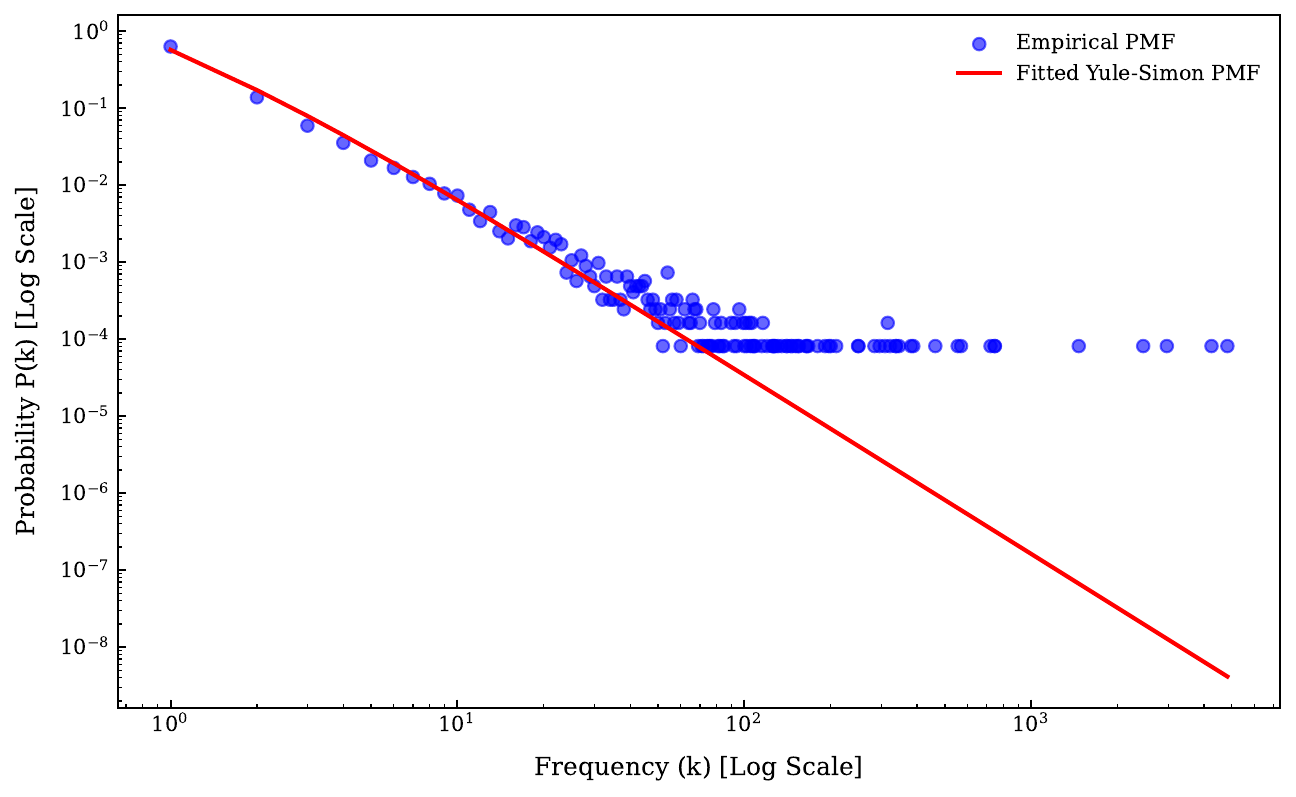}} &
        \subcaptionbox{\tiny Indonesian (id)}{\includegraphics[width=0.15\textwidth]{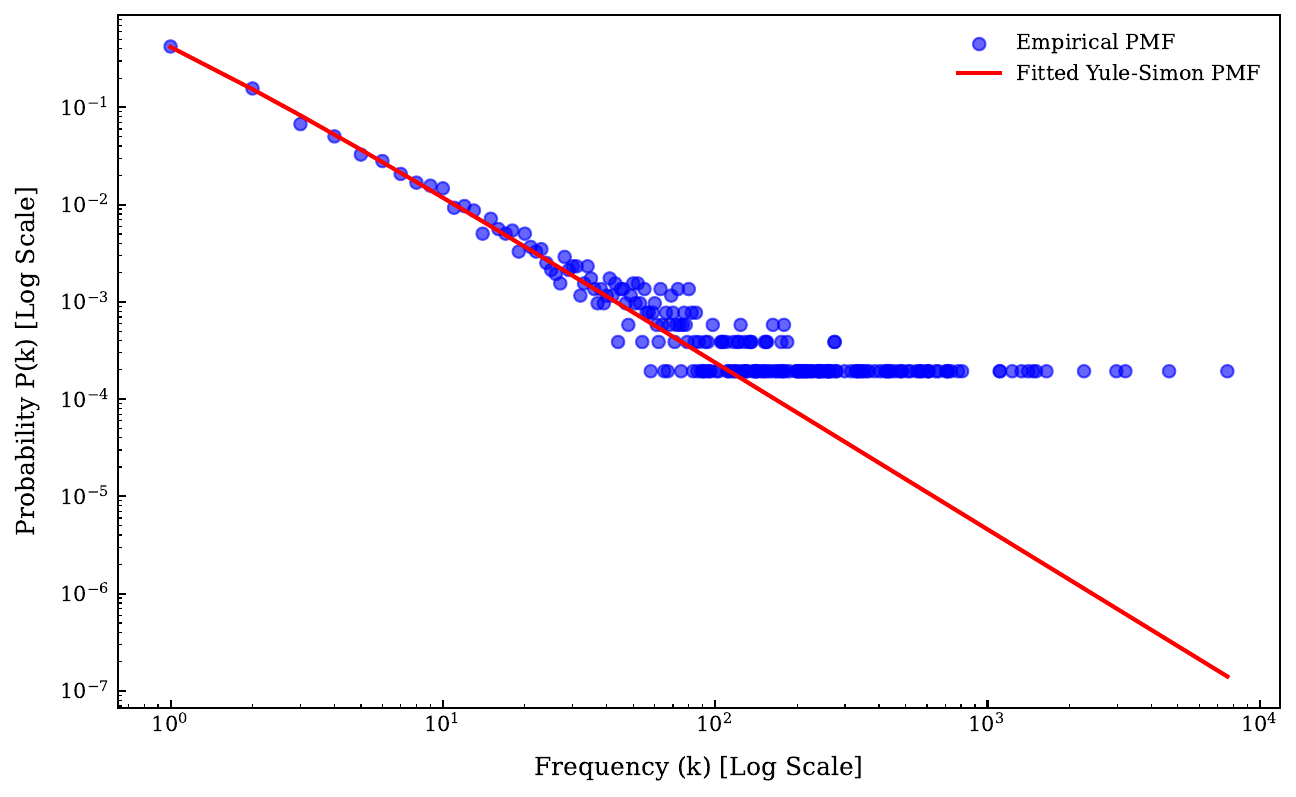}} &
        \subcaptionbox{\tiny Italian (it)}{\includegraphics[width=0.15\textwidth]{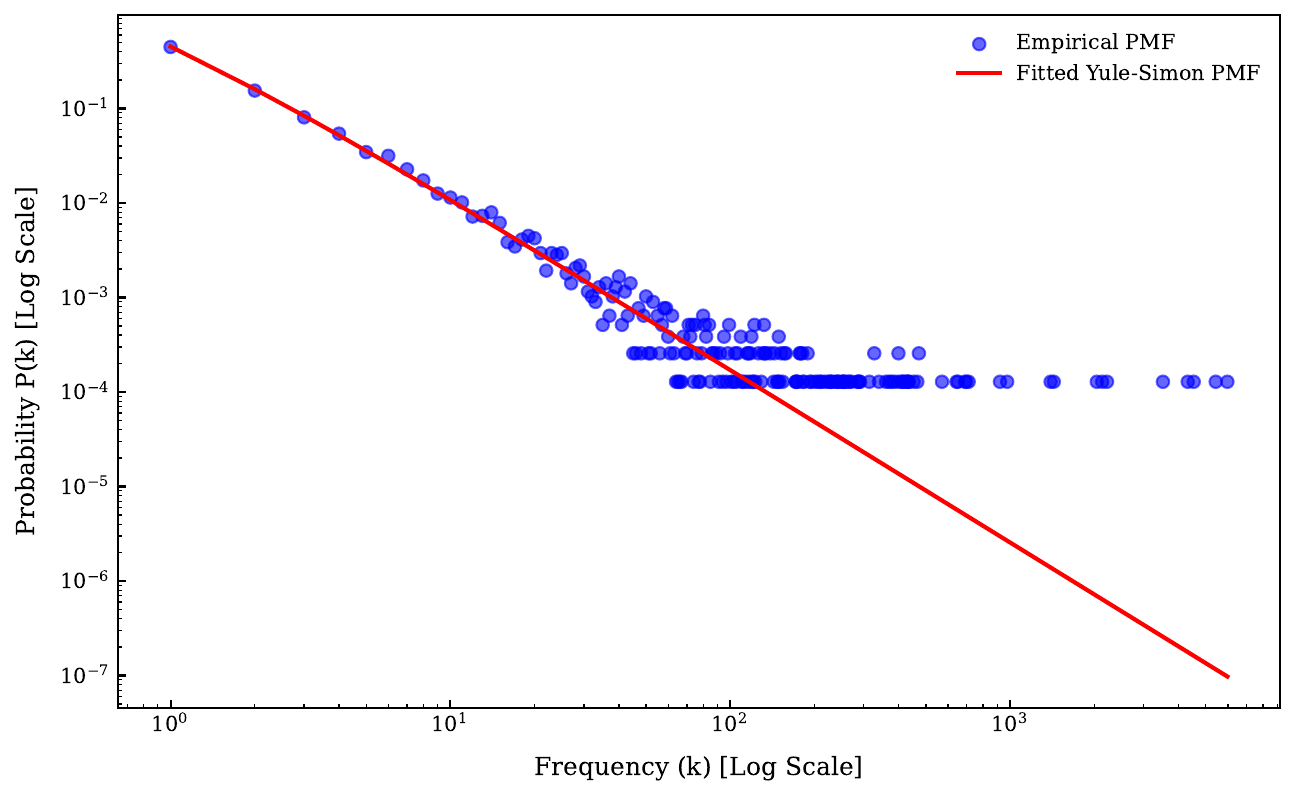}} &
        \subcaptionbox{\tiny Hebrew (he)}{\includegraphics[width=0.15\textwidth]{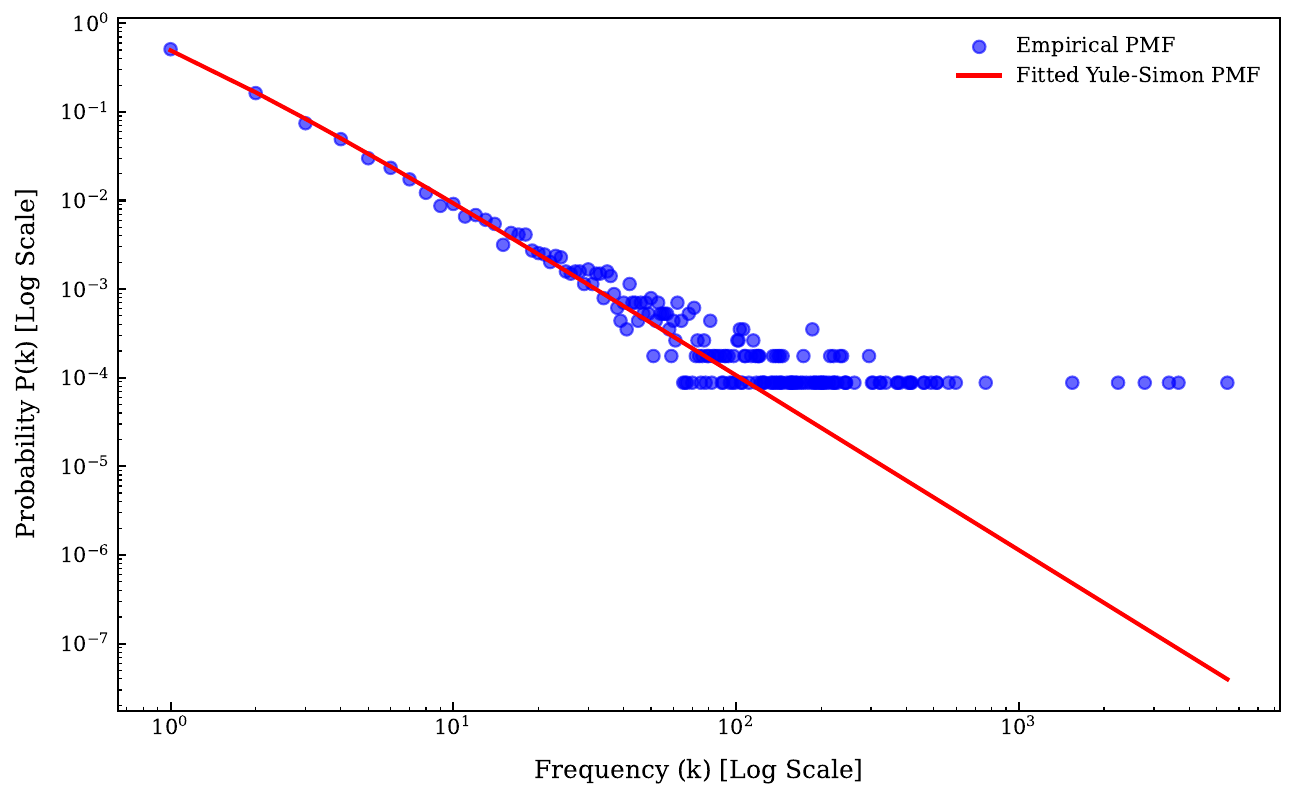}} \\
        
        \subcaptionbox{\tiny Japanese (ja)\vspace{1em}}{\includegraphics[width=0.15\textwidth]{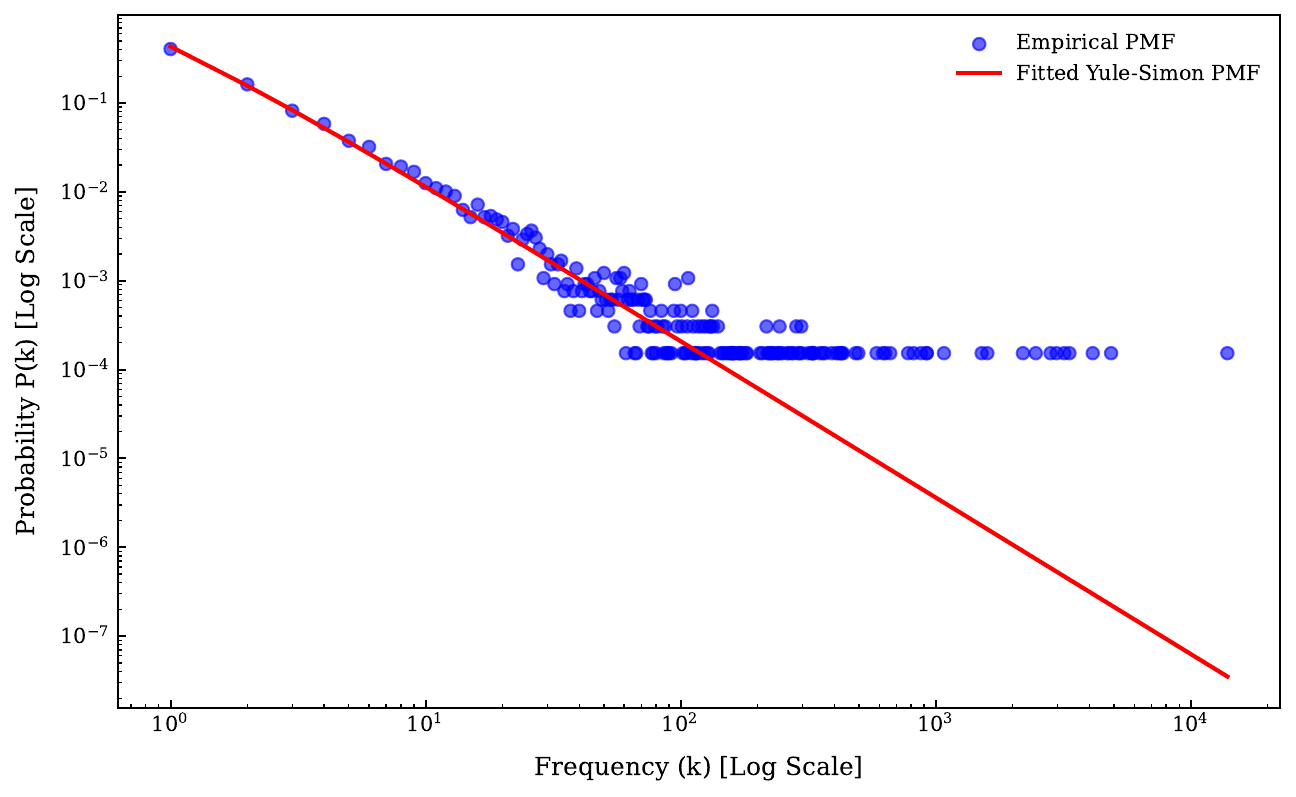}} &
        \subcaptionbox{\tiny Korean (ko)}{\includegraphics[width=0.15\textwidth]{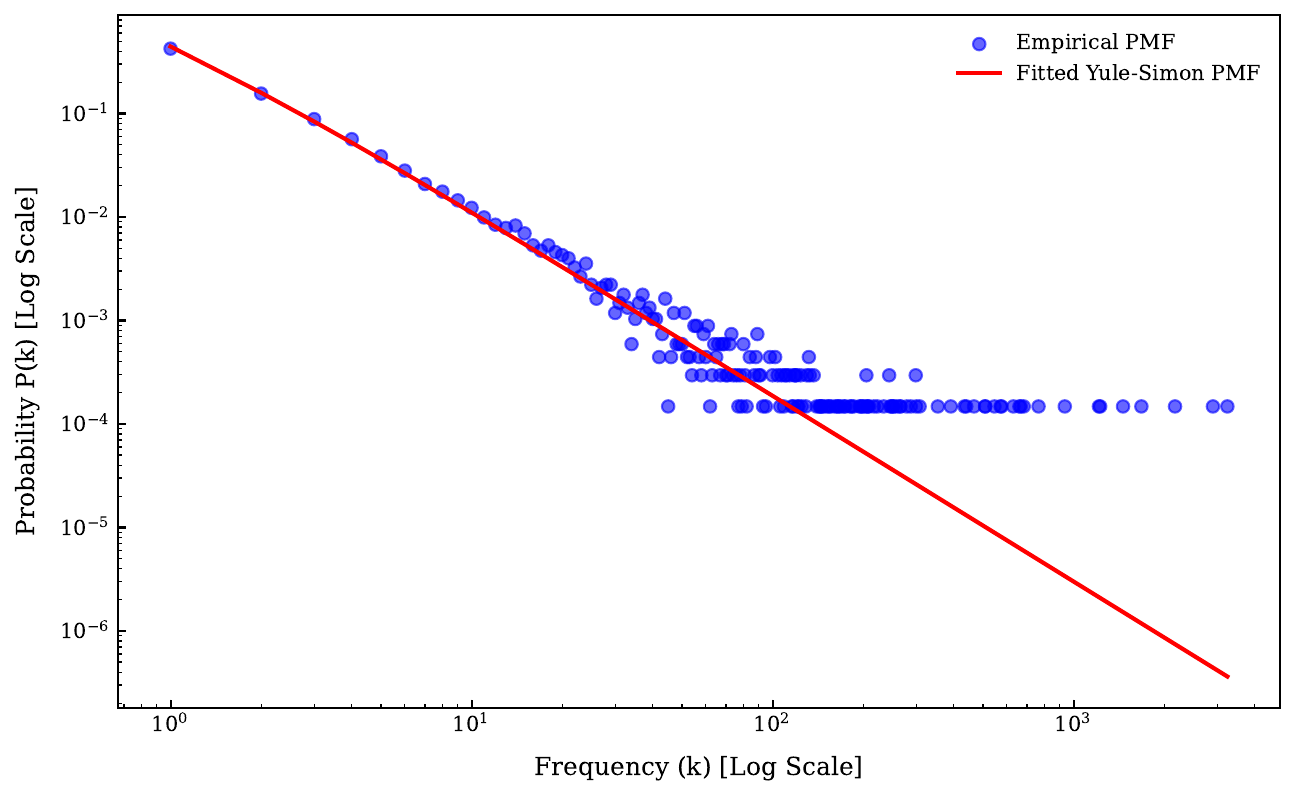}} &
        \subcaptionbox{\tiny Maori (mi)}{\includegraphics[width=0.15\textwidth]{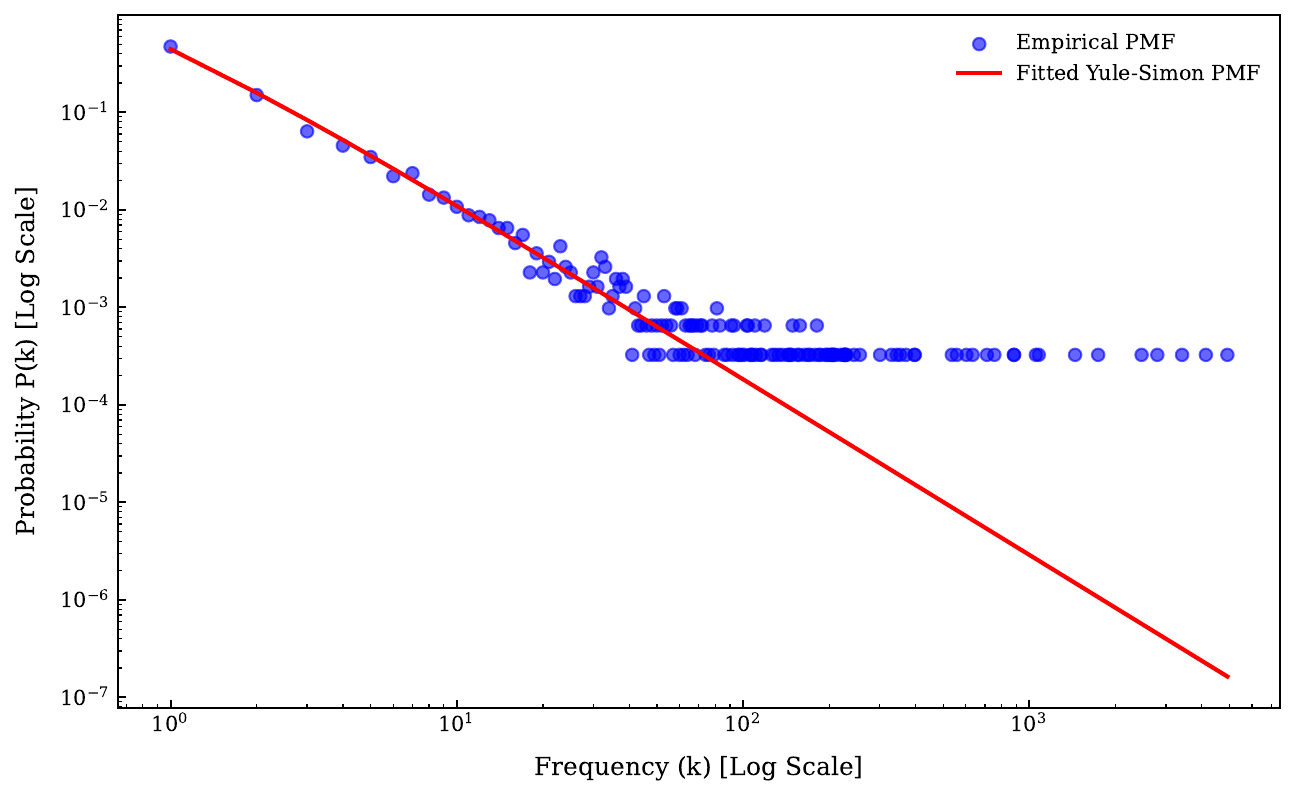}} &
        \subcaptionbox{\tiny Dutch (nl)}{\includegraphics[width=0.15\textwidth]{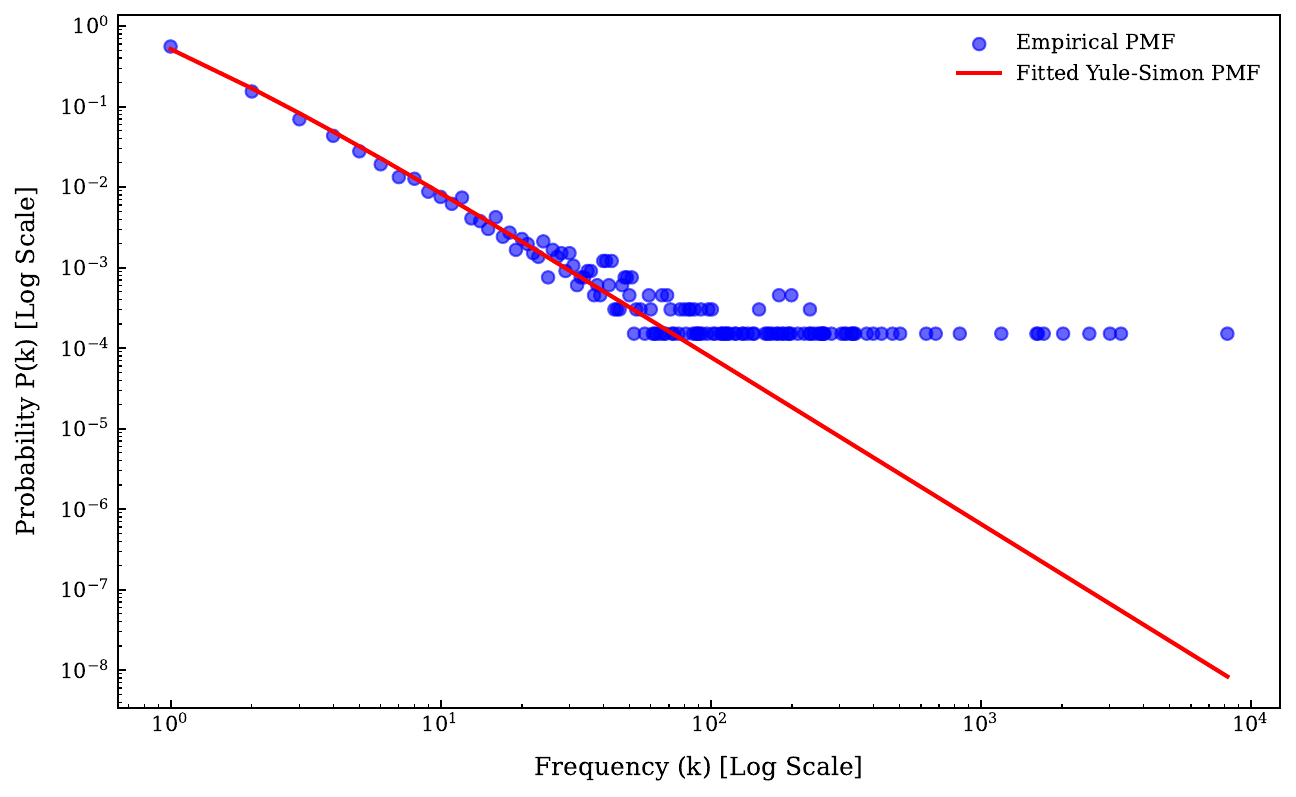}} &
        \subcaptionbox{\tiny Norwegian (no)}{\includegraphics[width=0.15\textwidth]{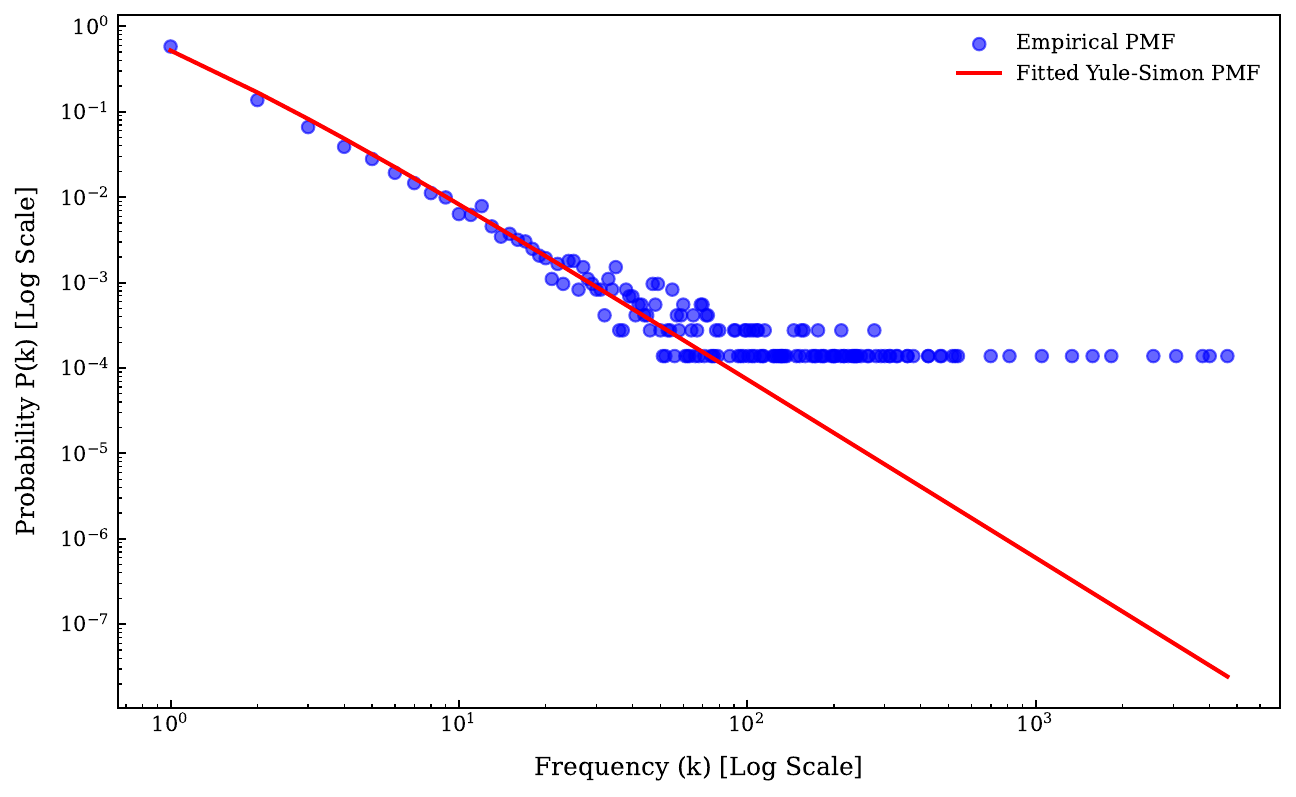}} &
        \subcaptionbox{\tiny Polish (pl)}{\includegraphics[width=0.15\textwidth]{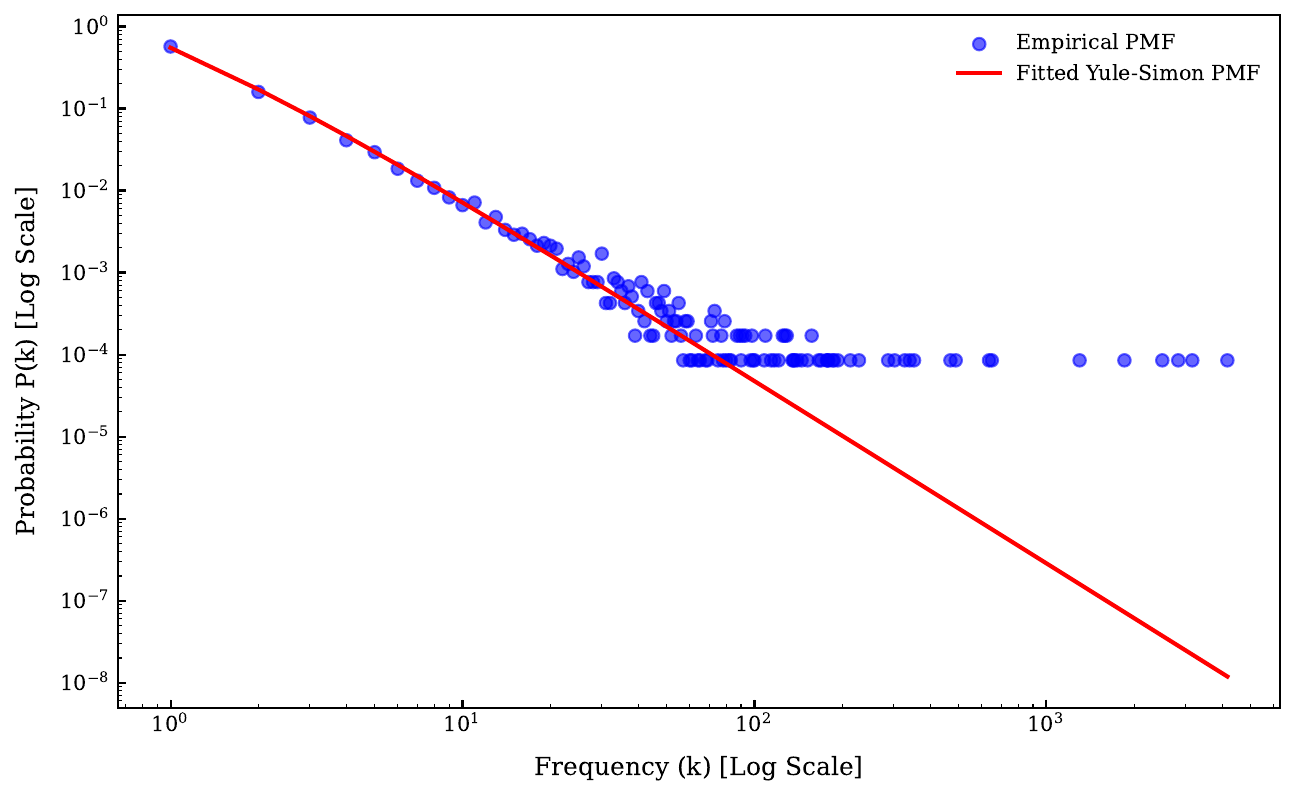}} \\
        
        \subcaptionbox{\tiny Portuguese (pt)\vspace{1em}}{\includegraphics[width=0.15\textwidth]{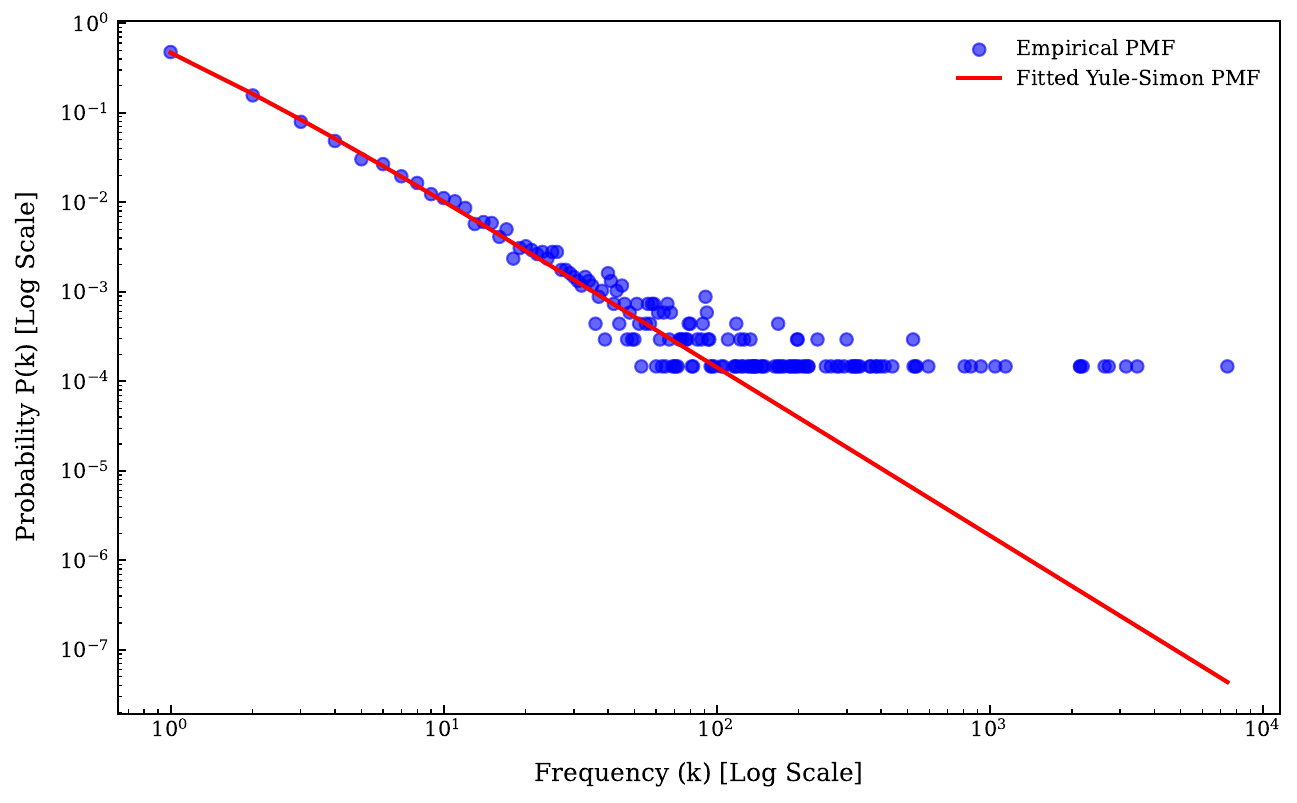}} &
        \subcaptionbox{\tiny Romanian (ro)}{\includegraphics[width=0.15\textwidth]{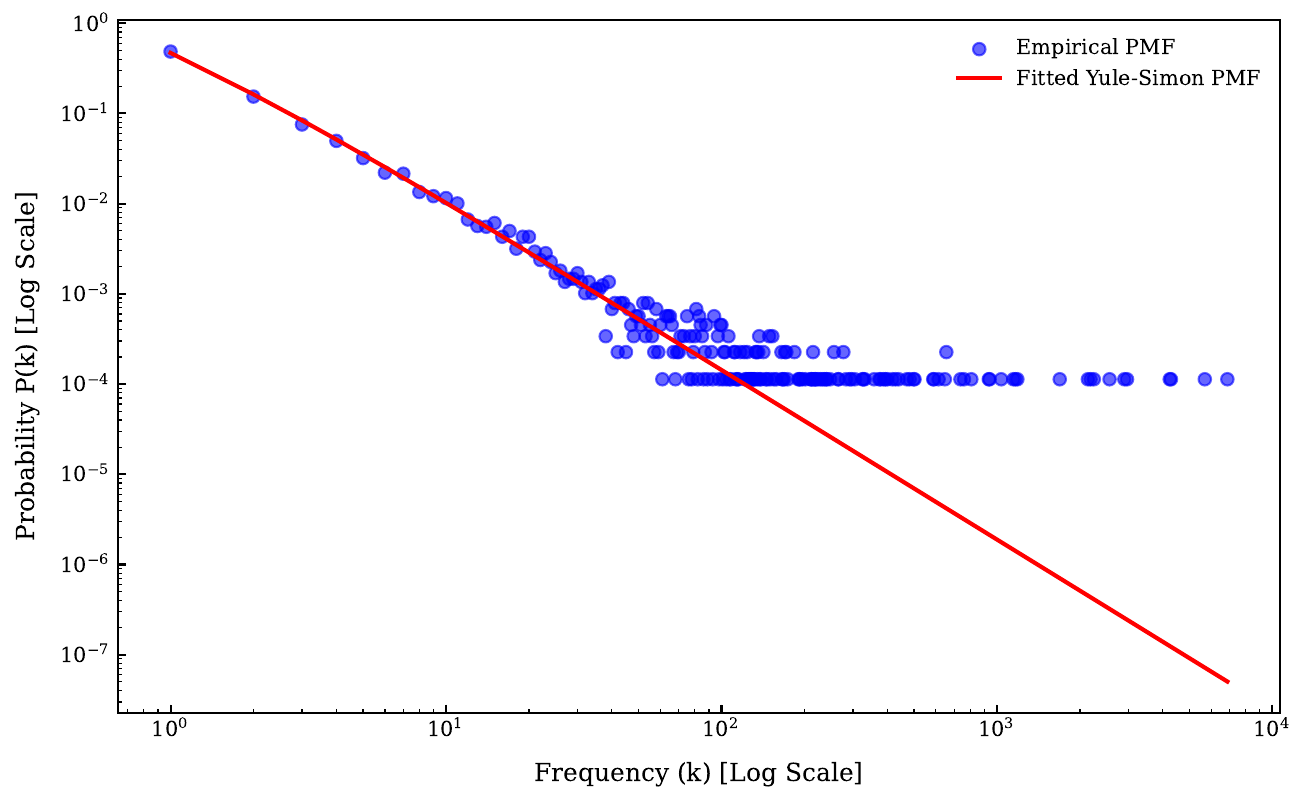}} &
        \subcaptionbox{\tiny Russian (ru)}{\includegraphics[width=0.15\textwidth]{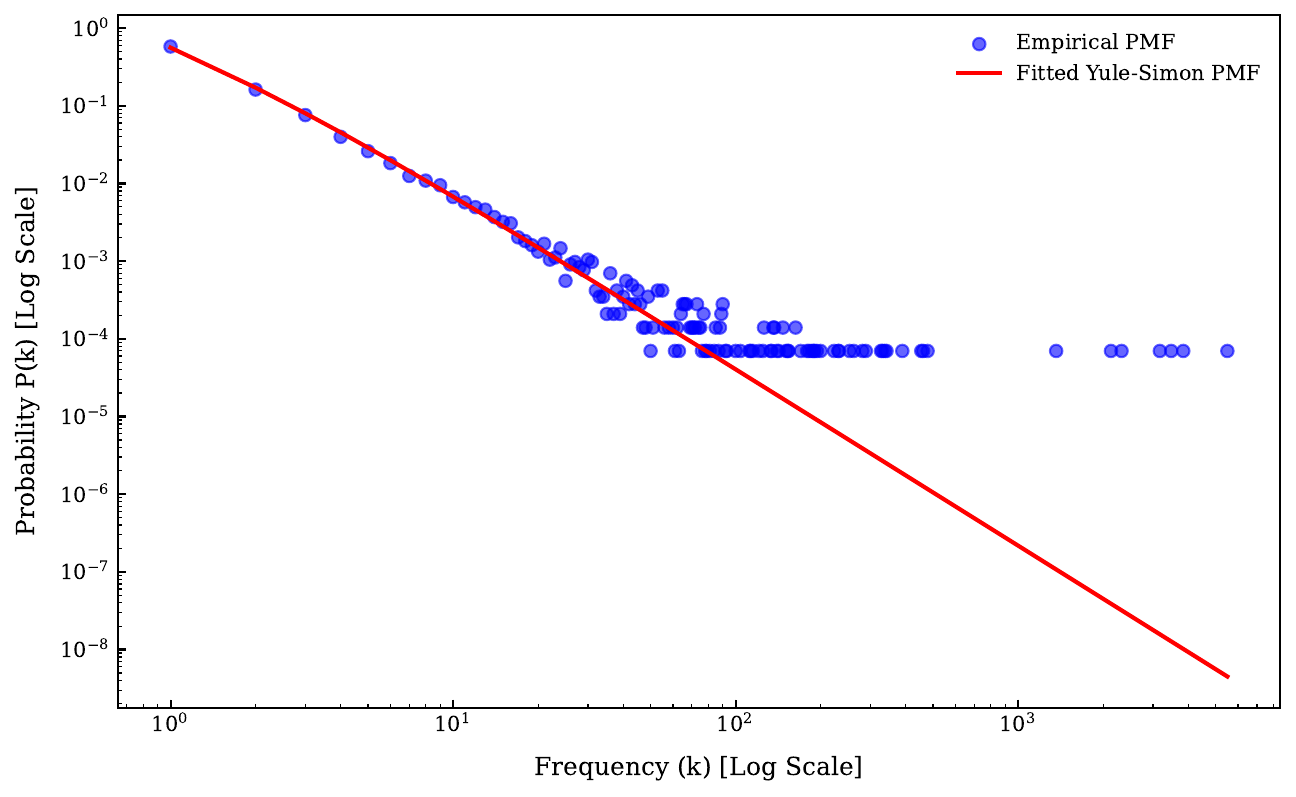}} &
        \subcaptionbox{\tiny Swedish (sv)}{\includegraphics[width=0.15\textwidth]{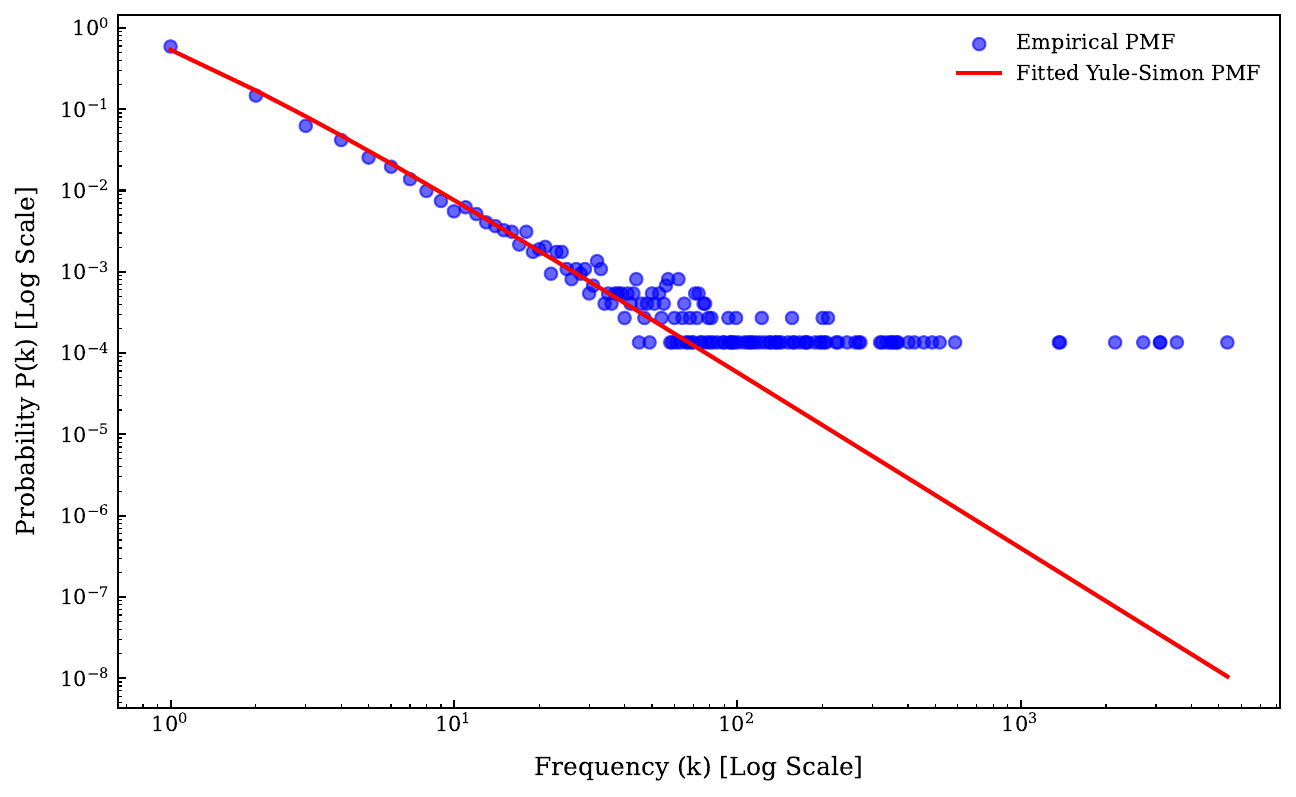}} &
        \subcaptionbox{\tiny Swahili (sw)}{\includegraphics[width=0.15\textwidth]{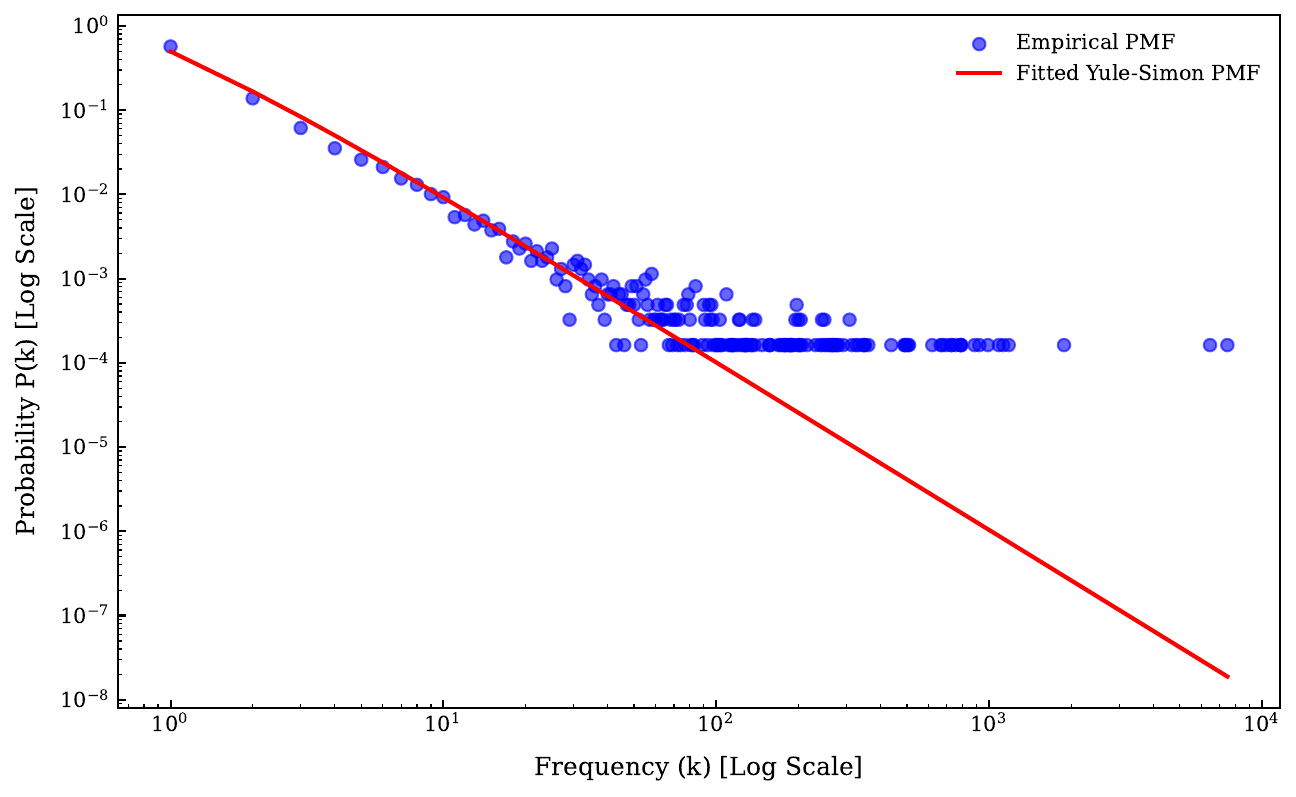}} &
        \subcaptionbox{\tiny Telugu (te)}{\includegraphics[width=0.15\textwidth]{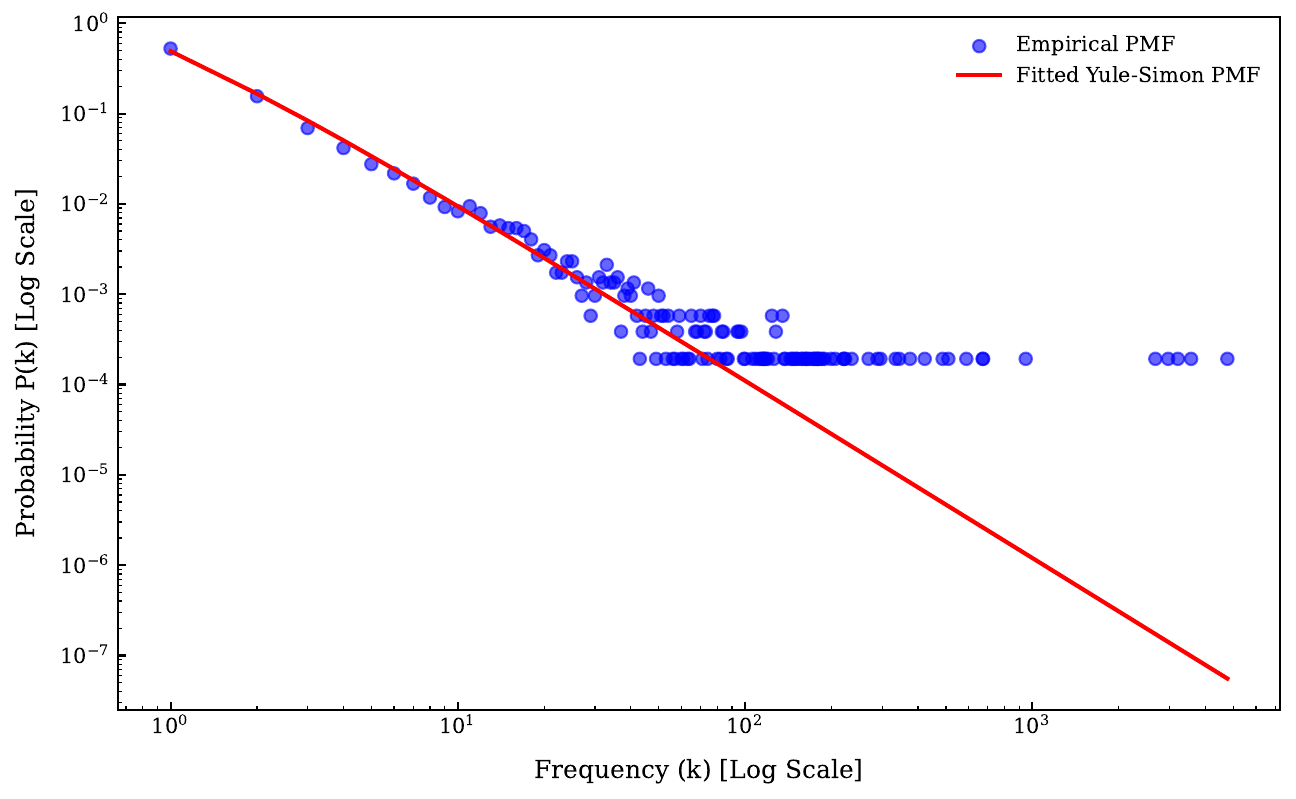}} \\
        
        \subcaptionbox{\tiny Thai (th)\vspace{1em}}{\includegraphics[width=0.15\textwidth]{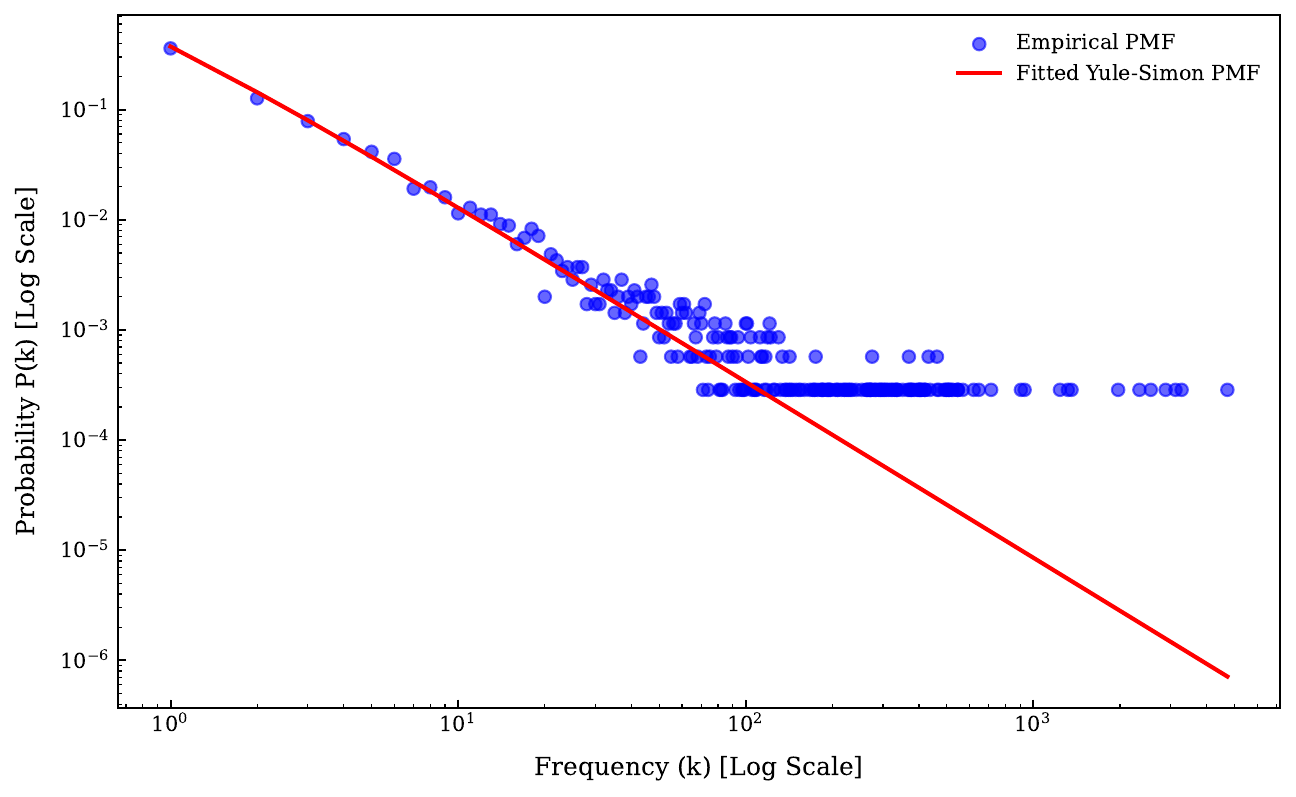}} &
        \subcaptionbox{\tiny Turkish (tr)}{\includegraphics[width=0.15\textwidth]{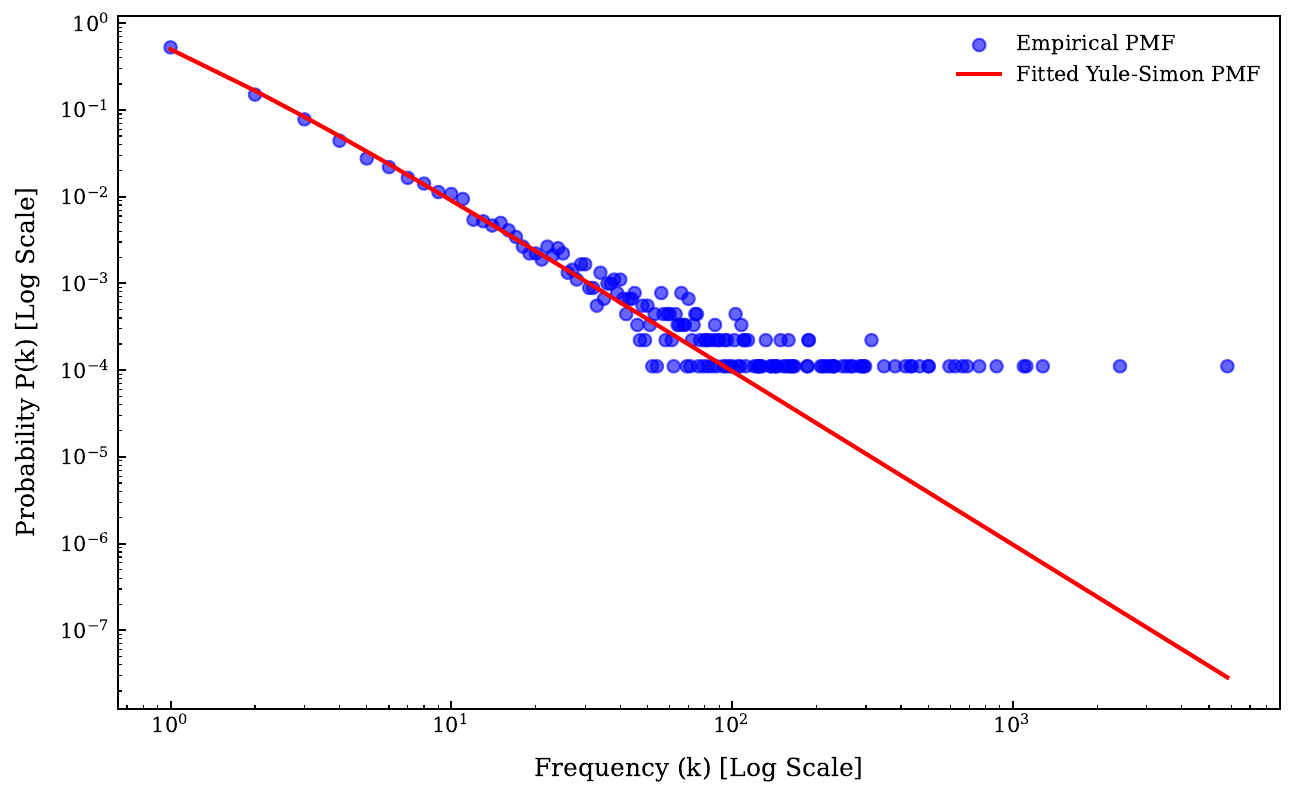}} &
        \subcaptionbox{\tiny Ukrainian (uk)}{\includegraphics[width=0.15\textwidth]{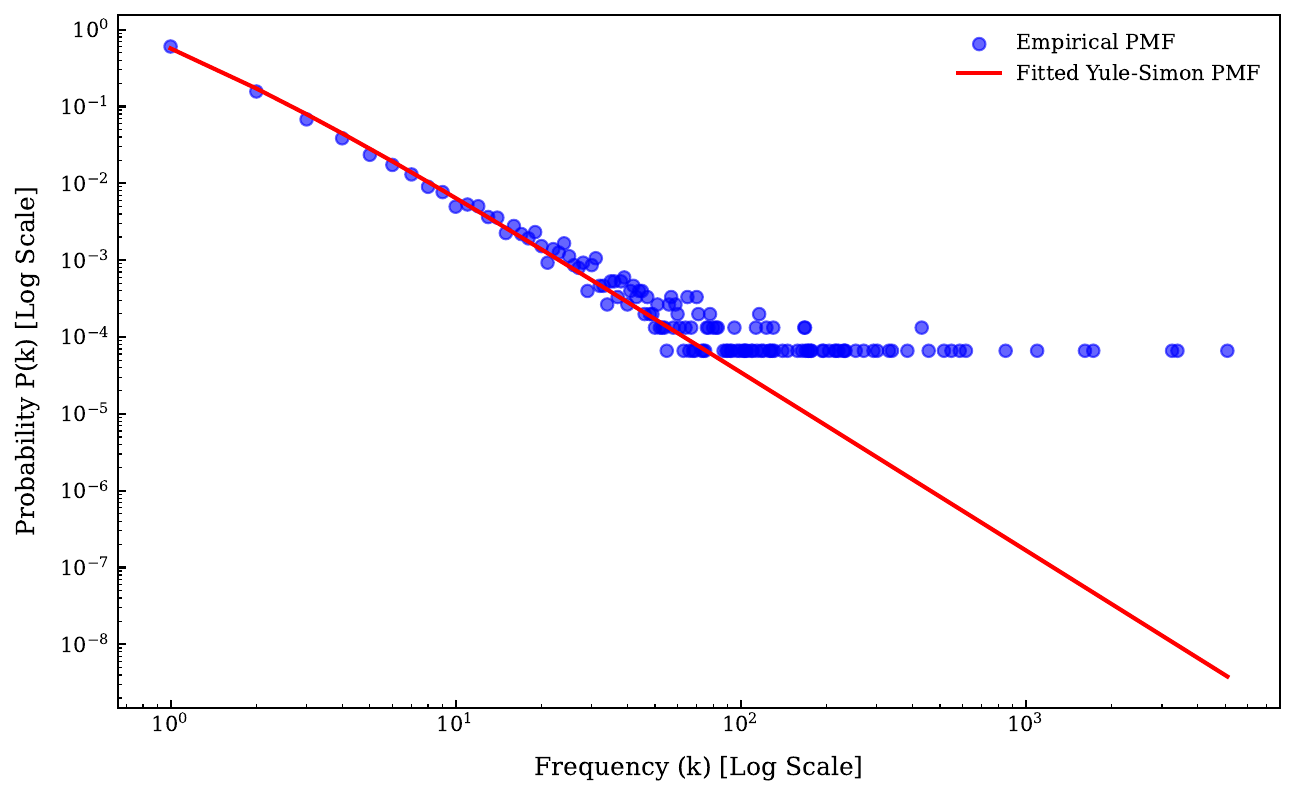}} &
        \subcaptionbox{\tiny Vietnamese (vi)}{\includegraphics[width=0.15\textwidth]{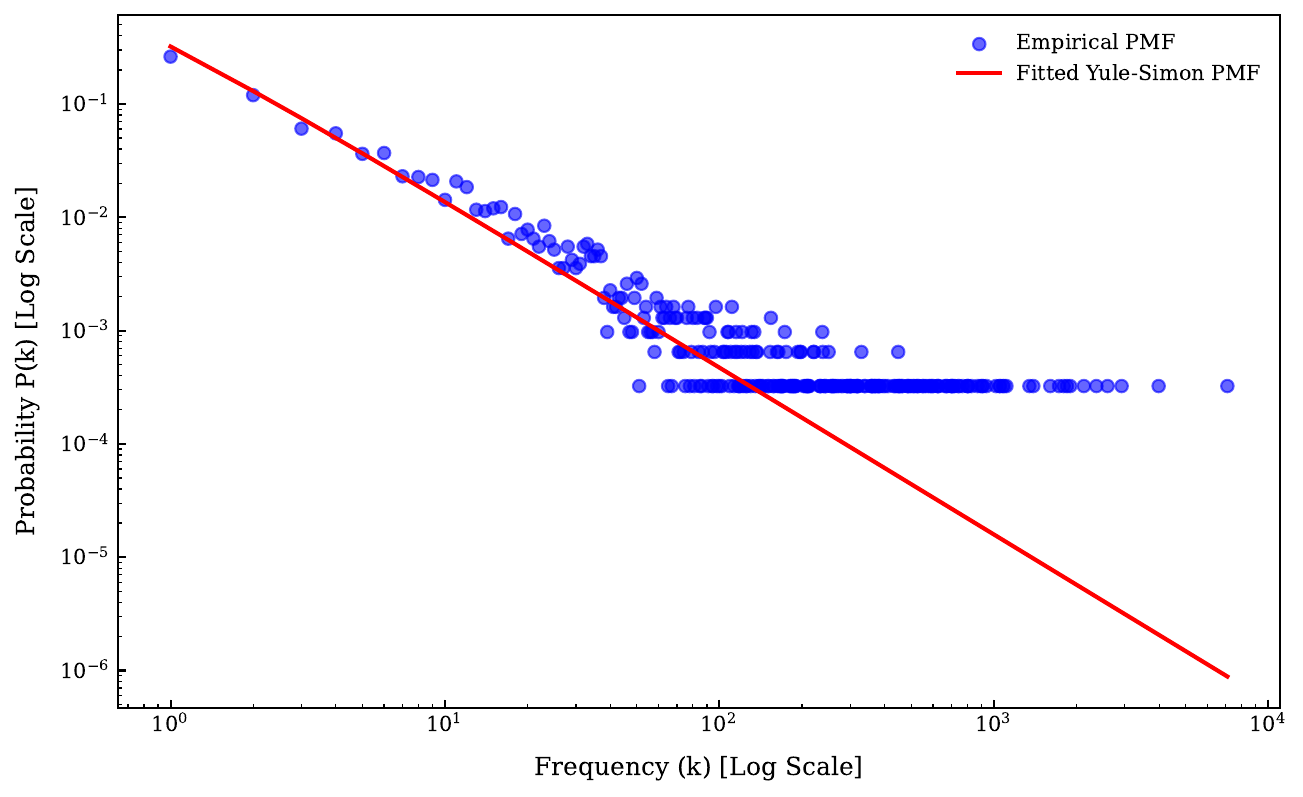}} &
        \subcaptionbox{\tiny Chinese (zh)}{\includegraphics[width=0.15\textwidth]{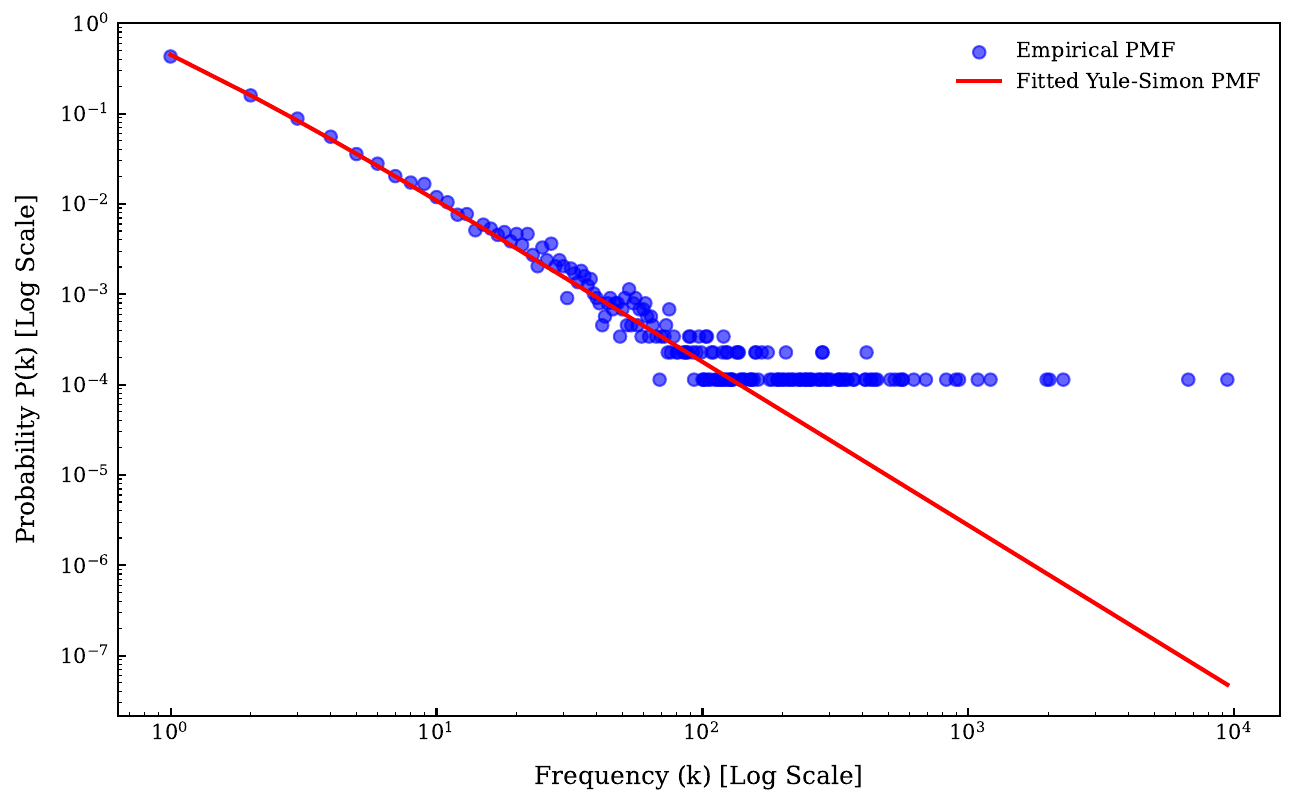}} &
    \end{tabular}
    \caption{Log-log fits on XM-3600 for various languages (Simon model, $n=1$)}
    \label{fig:locale_fits_xm3600}
\end{figure}

\begin{figure}
    \centering
    \tiny
    \setlength{\tabcolsep}{1pt} %
    \renewcommand{\arraystretch}{1.2} %
    \captionsetup[subfigure]{labelfont=scriptsize, textfont=scriptsize, skip=3pt} %
    \begin{tabular}{ccc}
        \subcaptionbox{\tiny Chameleon-512 (1-gram)\vspace{2em}}{\textcolor{red}{***** Convergence Error *****}} &
        \subcaptionbox{\tiny Chameleon-512 (2-gram)}{\includegraphics[width=0.3\textwidth]{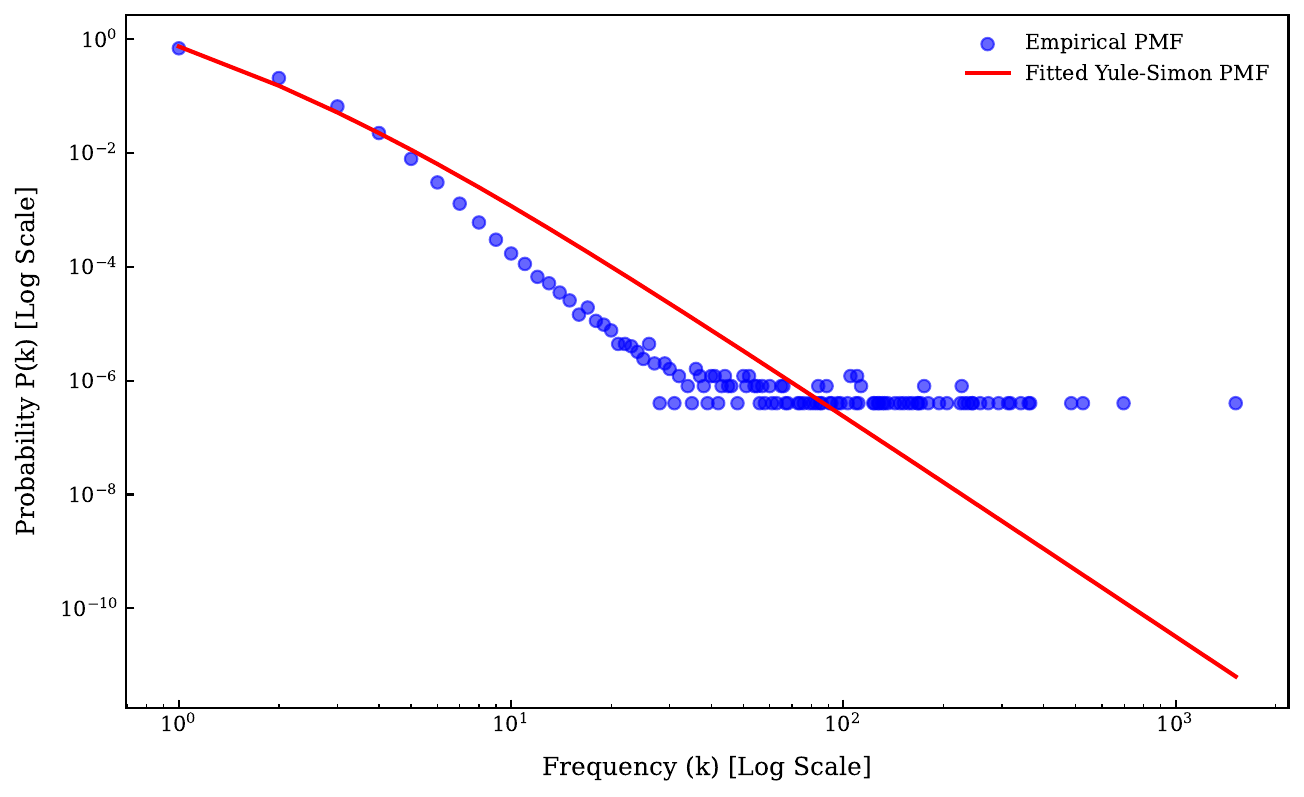}} &
        \subcaptionbox{\tiny Chameleon-512 (3-gram)}{\includegraphics[width=0.3\textwidth]{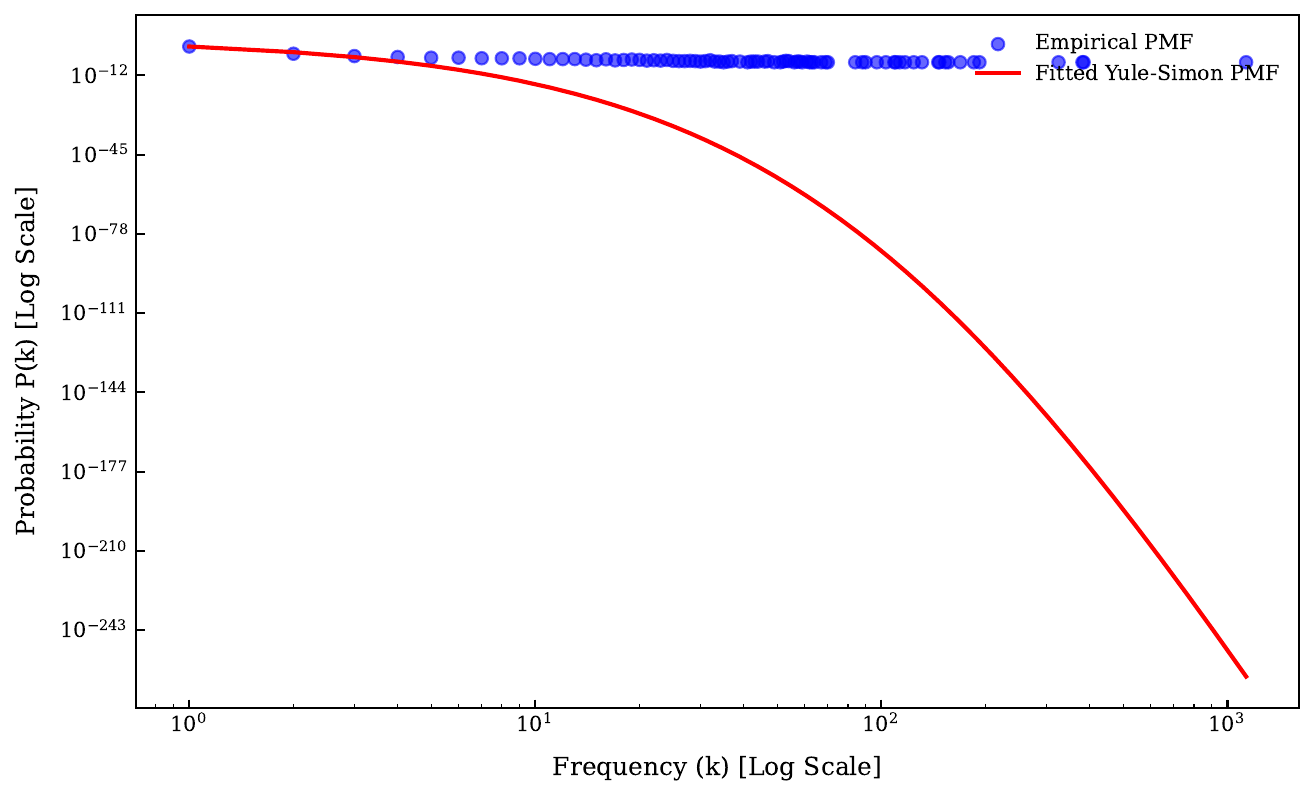}} \\
        
        \subcaptionbox{\tiny VQ-F8-64 (1-gram)\vspace{2em}}{\includegraphics[width=0.3\textwidth]{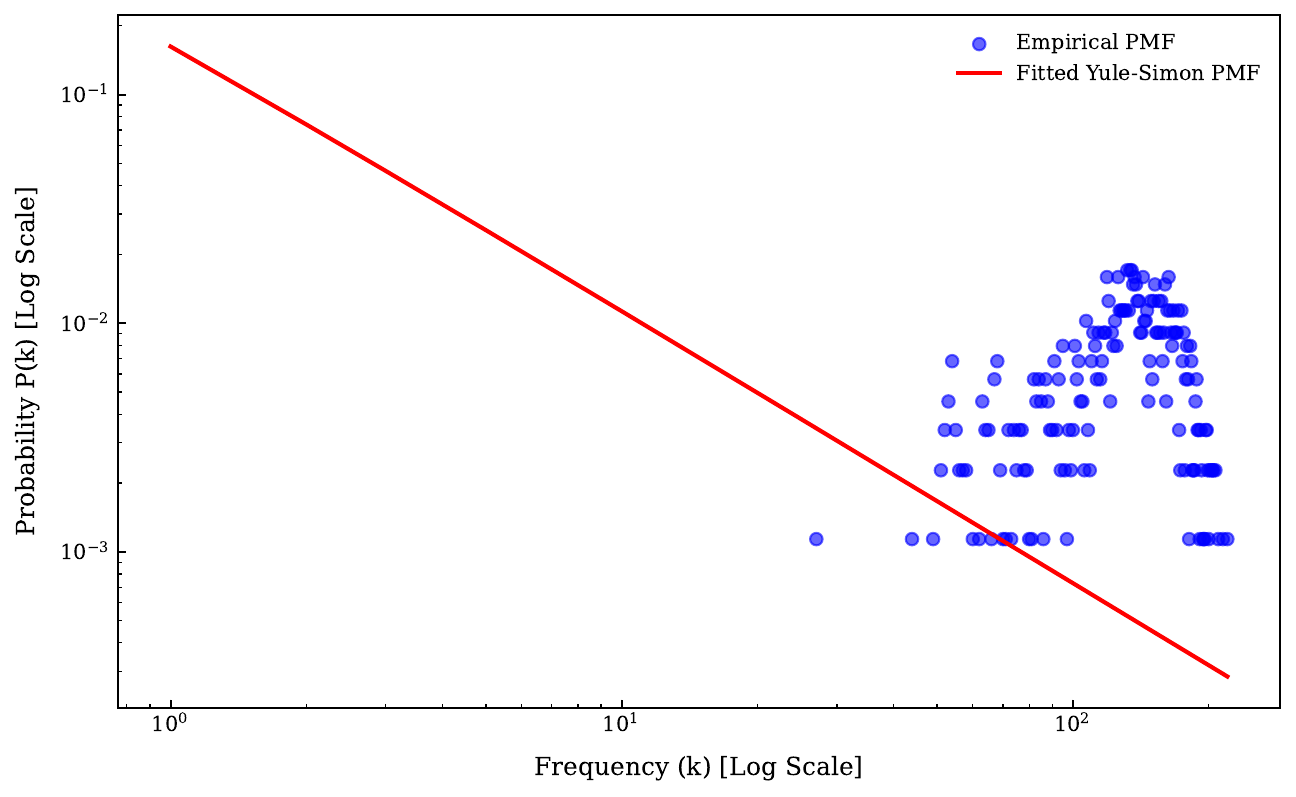}} &
        \subcaptionbox{\tiny VQ-F8-64 (2-gram)}{\includegraphics[width=0.3\textwidth]{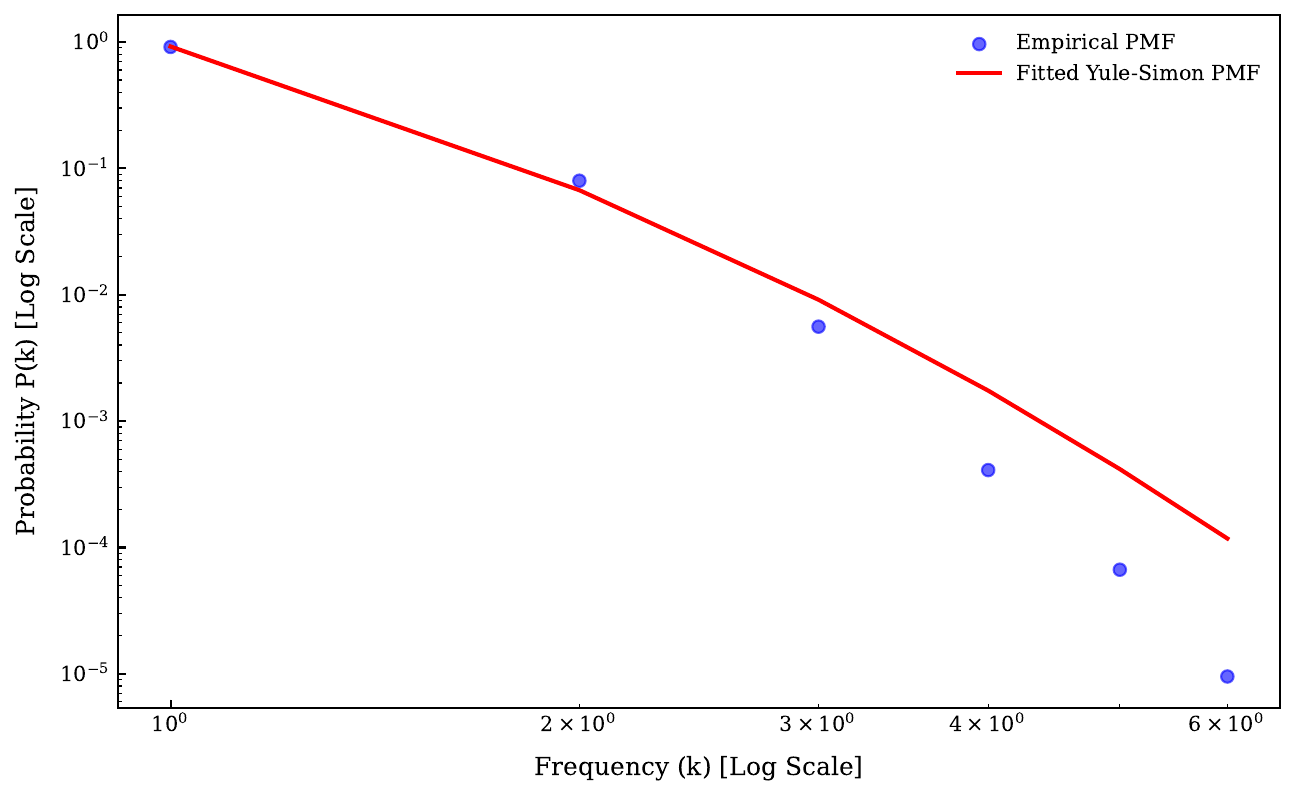}} &
        \subcaptionbox{\tiny VQ-F8-64 (3-gram)}{\textcolor{red}{***** Convergence Error *****}} \\
        
        \subcaptionbox{\tiny VQ-F8-256 (1-gram)\vspace{2em}}{\includegraphics[width=0.3\textwidth]{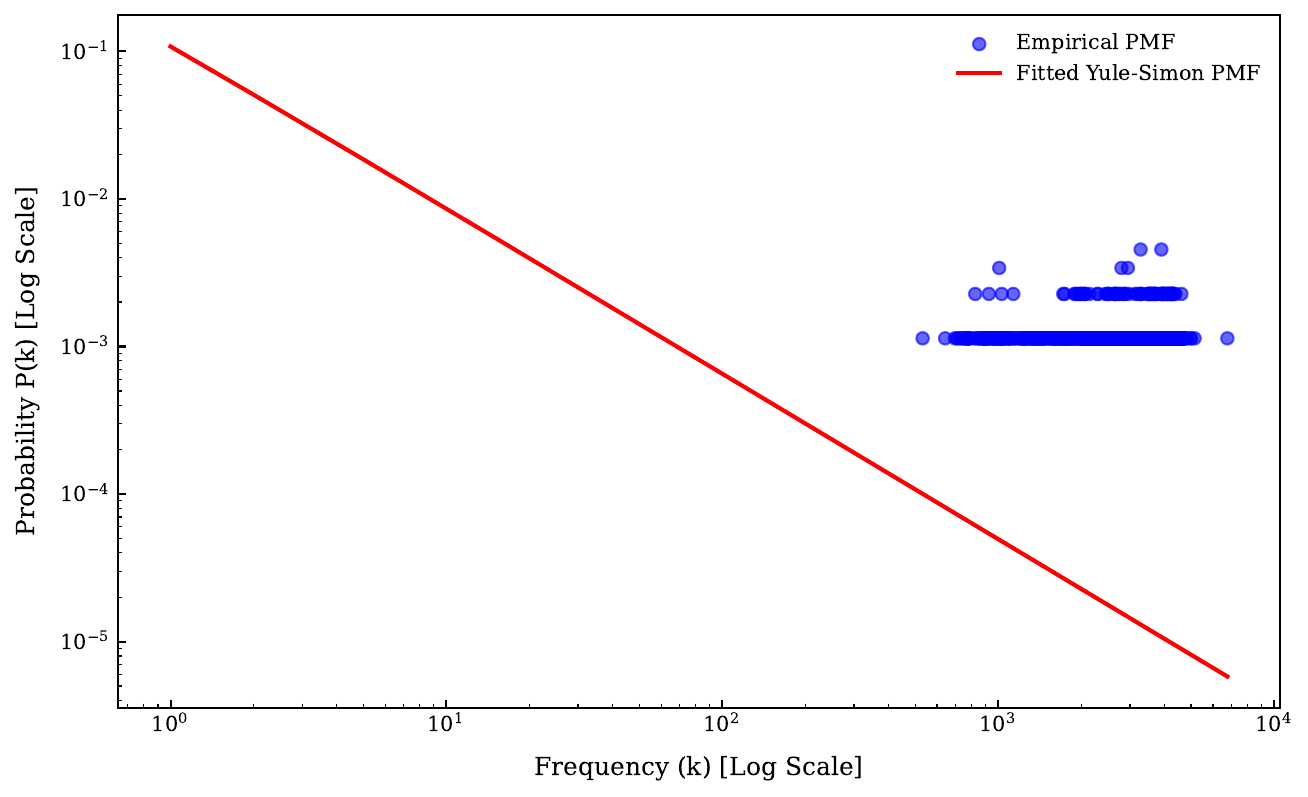}} &
        \subcaptionbox{\tiny VQ-F8-256 (2-gram)}{\includegraphics[width=0.3\textwidth]{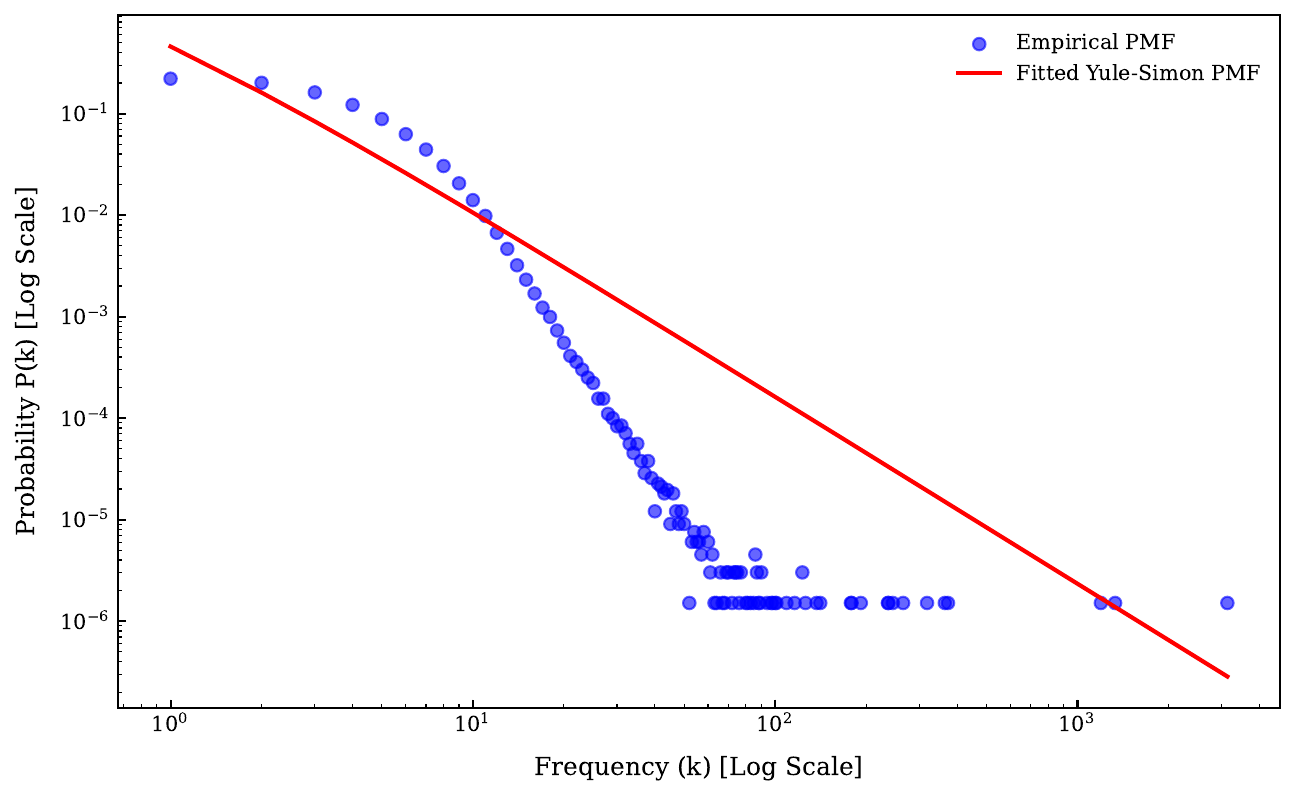}} &
        \subcaptionbox{\tiny VQ-F8-256 (3-gram)}{\textcolor{red}{***** Convergence Error *****}} \\
        
        \subcaptionbox{\tiny VQ-Imagenet-F16-1024-256 (1-gram)\vspace{2em}}{\includegraphics[width=0.3\textwidth]{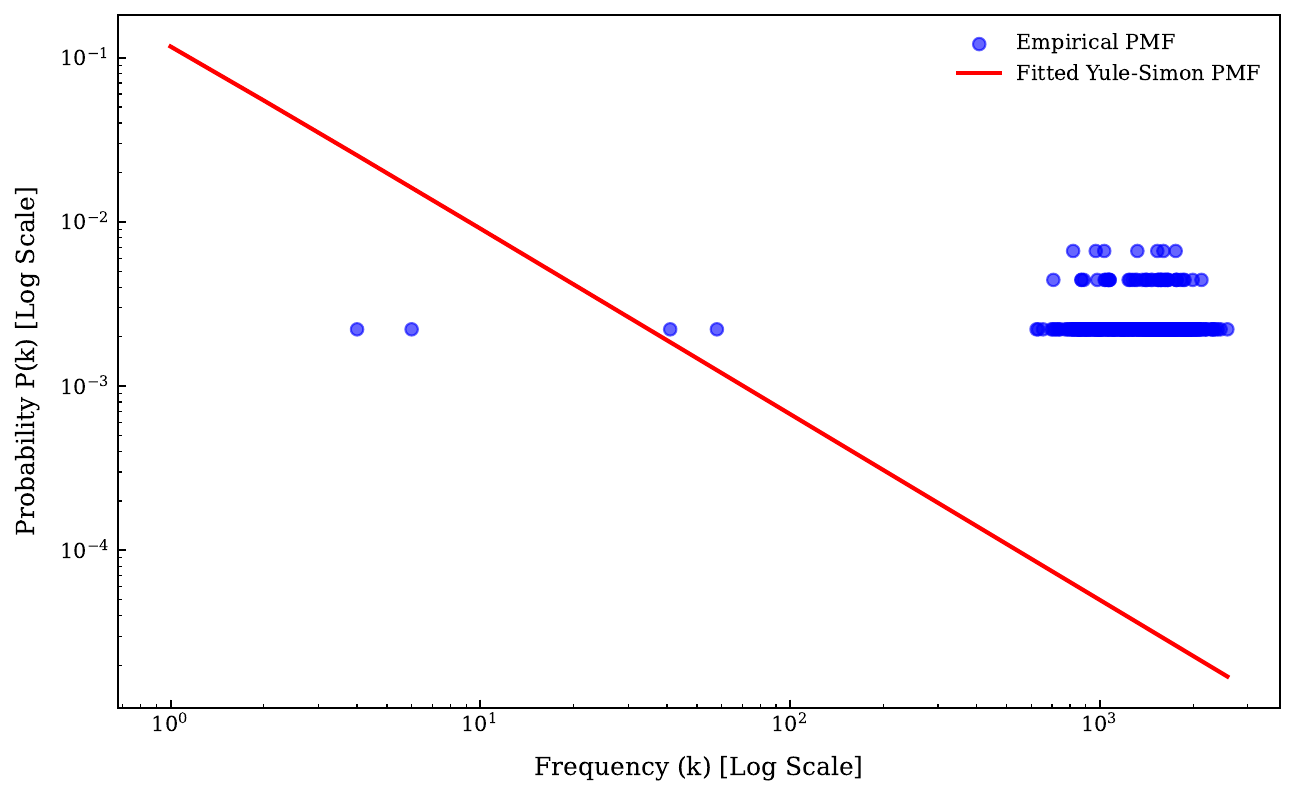}} &
        \subcaptionbox{\tiny VQ-Imagenet-F16-1024-256 (2-gram)}{\includegraphics[width=0.3\textwidth]{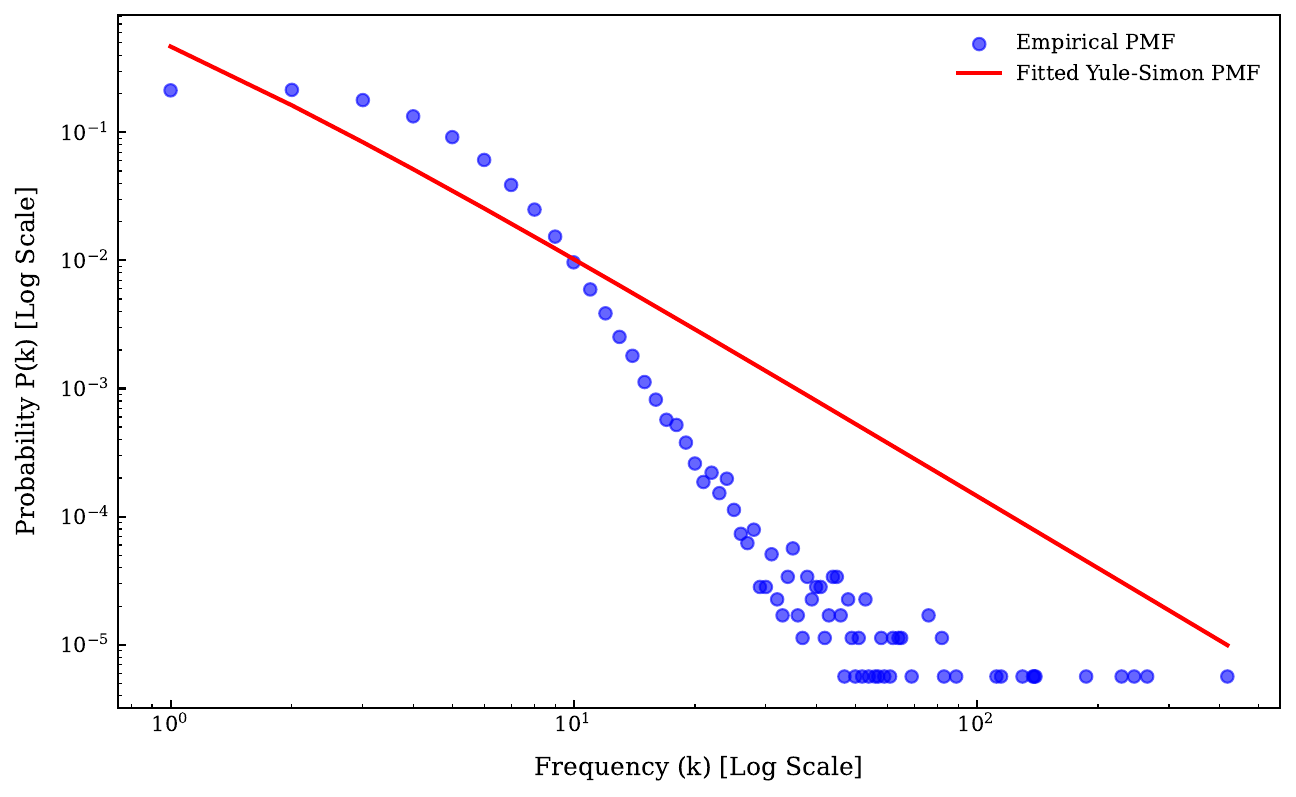}} &
        \subcaptionbox{\tiny VQ-Imagenet-F16-1024-256 (3-gram)}{\includegraphics[width=0.3\textwidth]{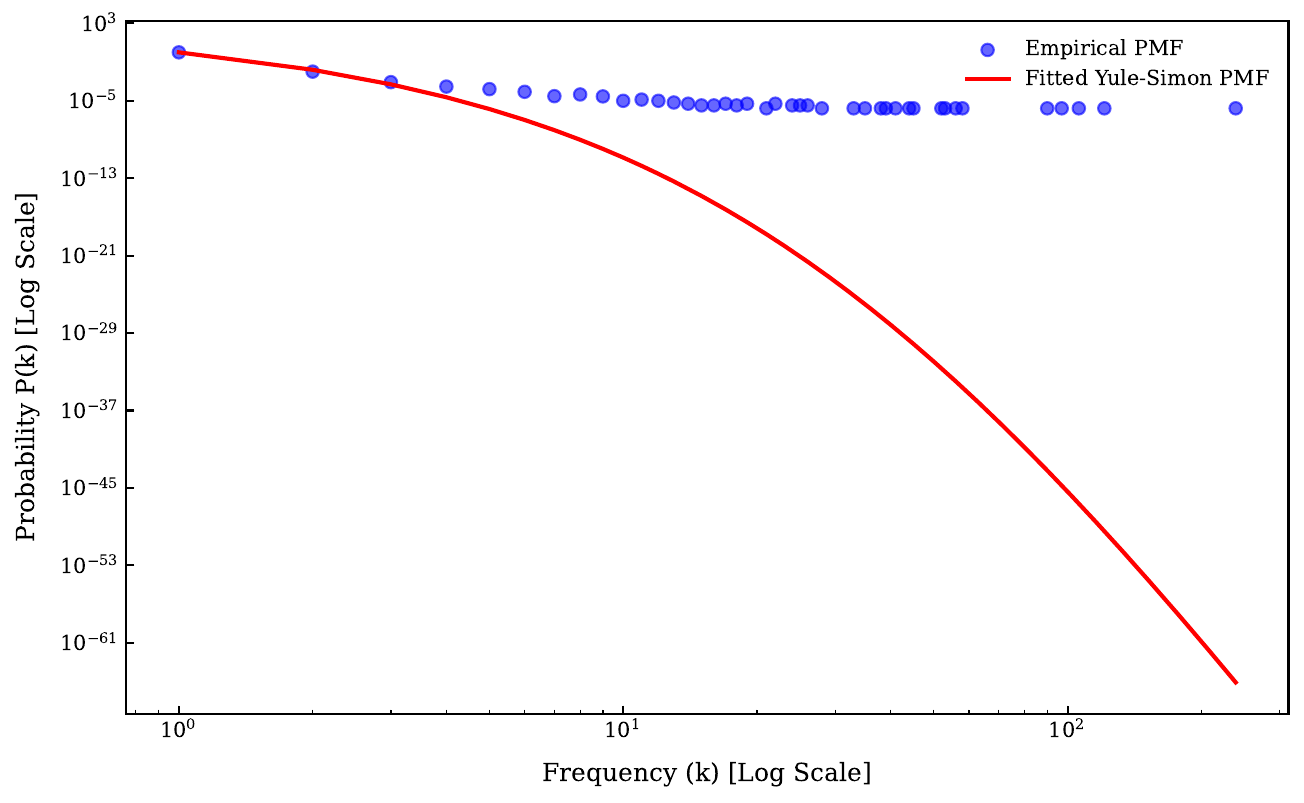}} \\
        
        \subcaptionbox{\tiny LlamaGen-VQ-DS16-C2I (1-gram)\vspace{2em}}{\includegraphics[width=0.3\textwidth]{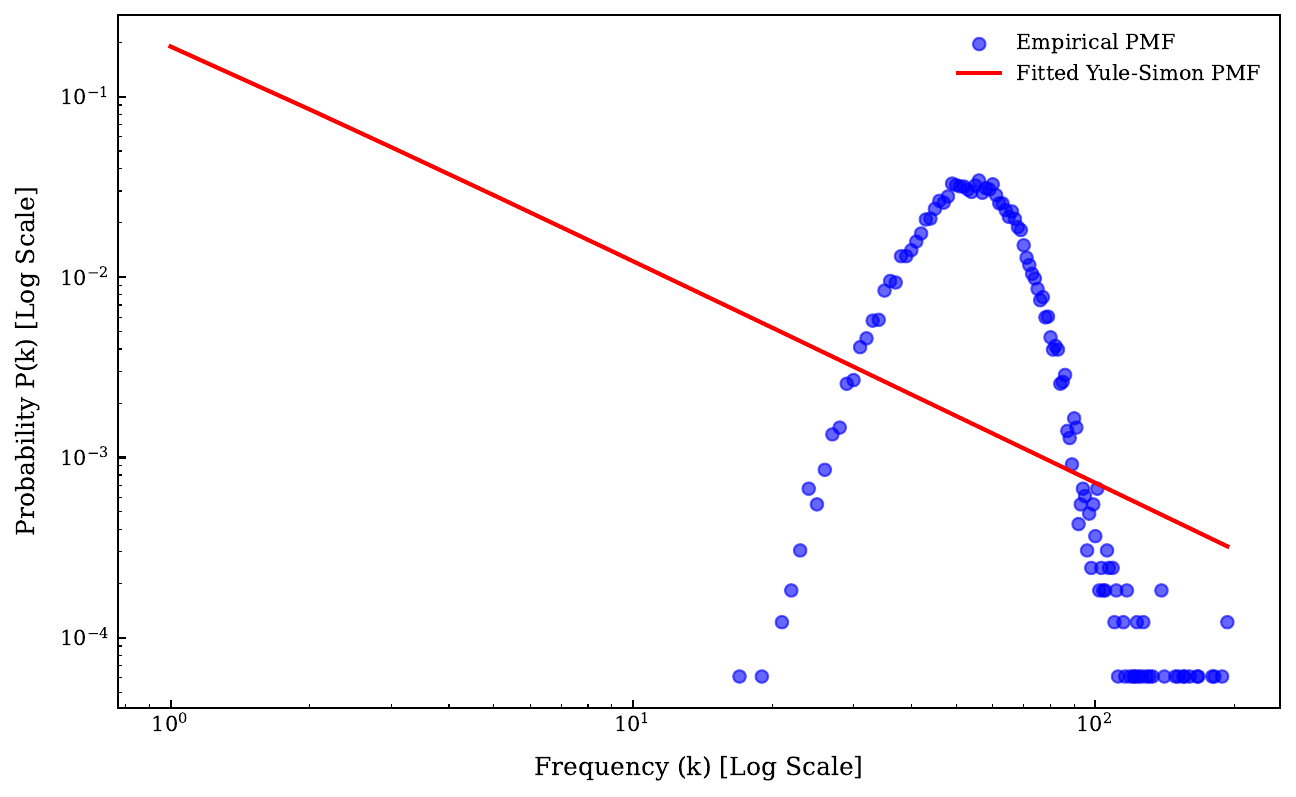}} &
        \subcaptionbox{\tiny LlamaGen-VQ-DS16-C2I (2-gram)}{\includegraphics[width=0.3\textwidth]{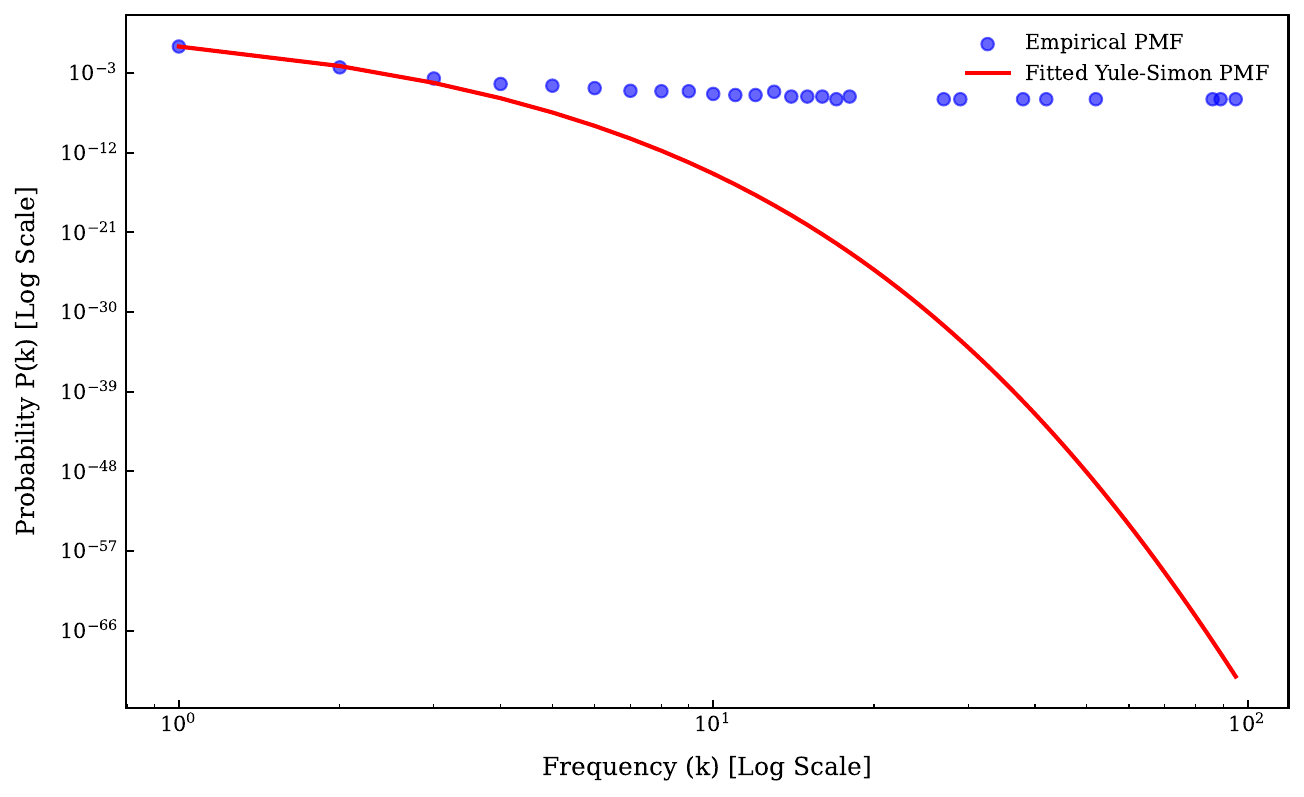}} &
        \subcaptionbox{\tiny LlamaGen-VQ-DS16-C2I (3-gram)}{\includegraphics[width=0.3\textwidth]{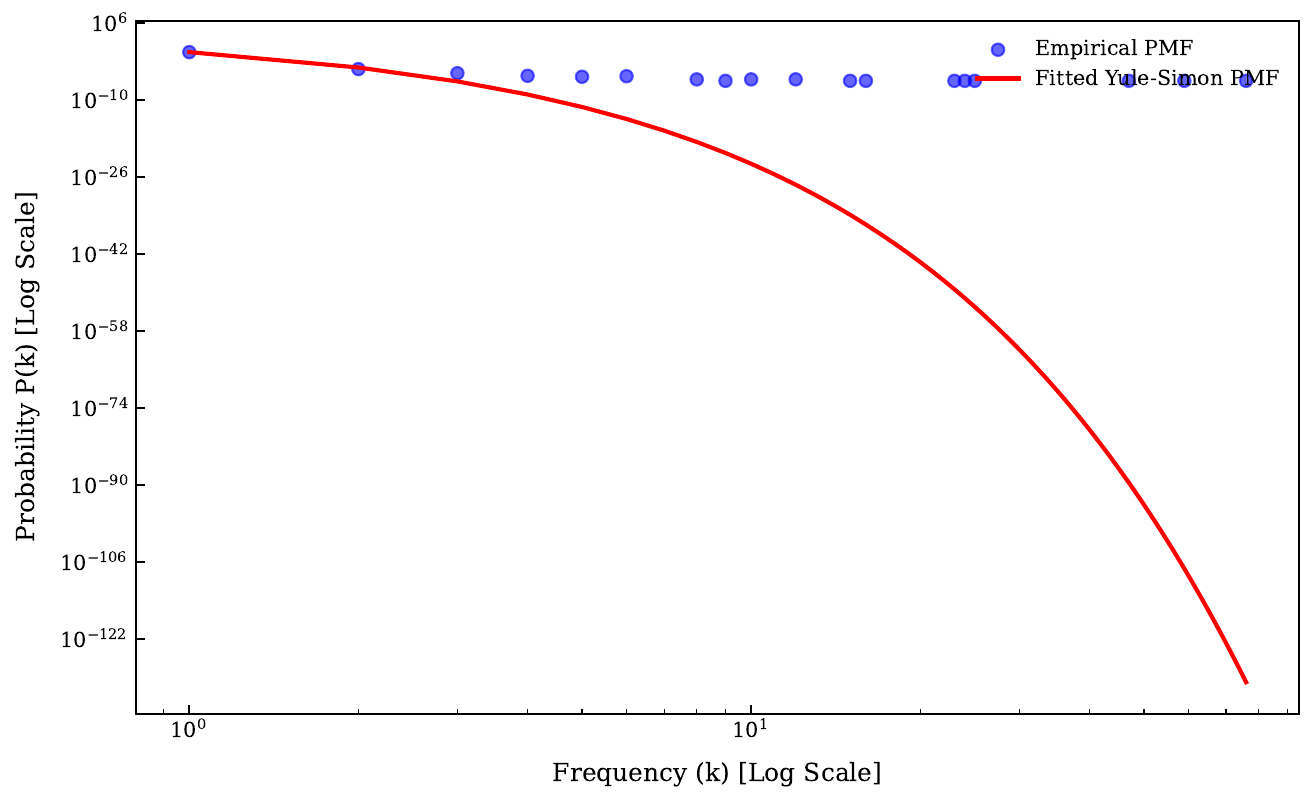}} \\
    \end{tabular}
    \caption{N-gram analysis on various models (Simon model on XM-3600)}
    \label{fig:model_fits_xm3600_ngram}
\end{figure}

\begin{figure}
    \centering
    \tiny
    \setlength{\tabcolsep}{1pt} %
    \renewcommand{\arraystretch}{1.2} %
    \captionsetup[subfigure]{labelfont=scriptsize, textfont=scriptsize, skip=3pt} %
    \begin{tabular}{cccccc}
        \subcaptionbox{\tiny Arabic (ar)\vspace{1em}}{\includegraphics[width=0.15\textwidth]{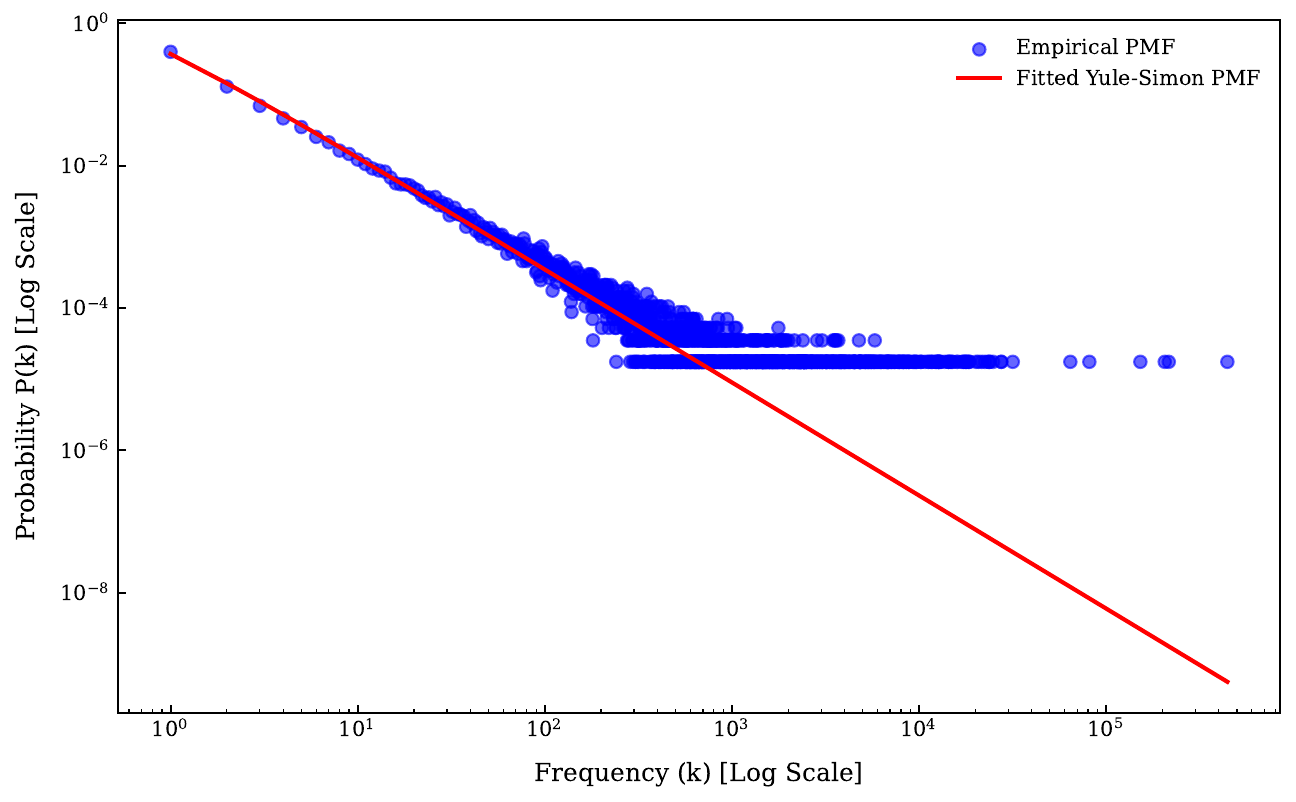}} &
        \subcaptionbox{\tiny Bengali (bn)}{\includegraphics[width=0.15\textwidth]{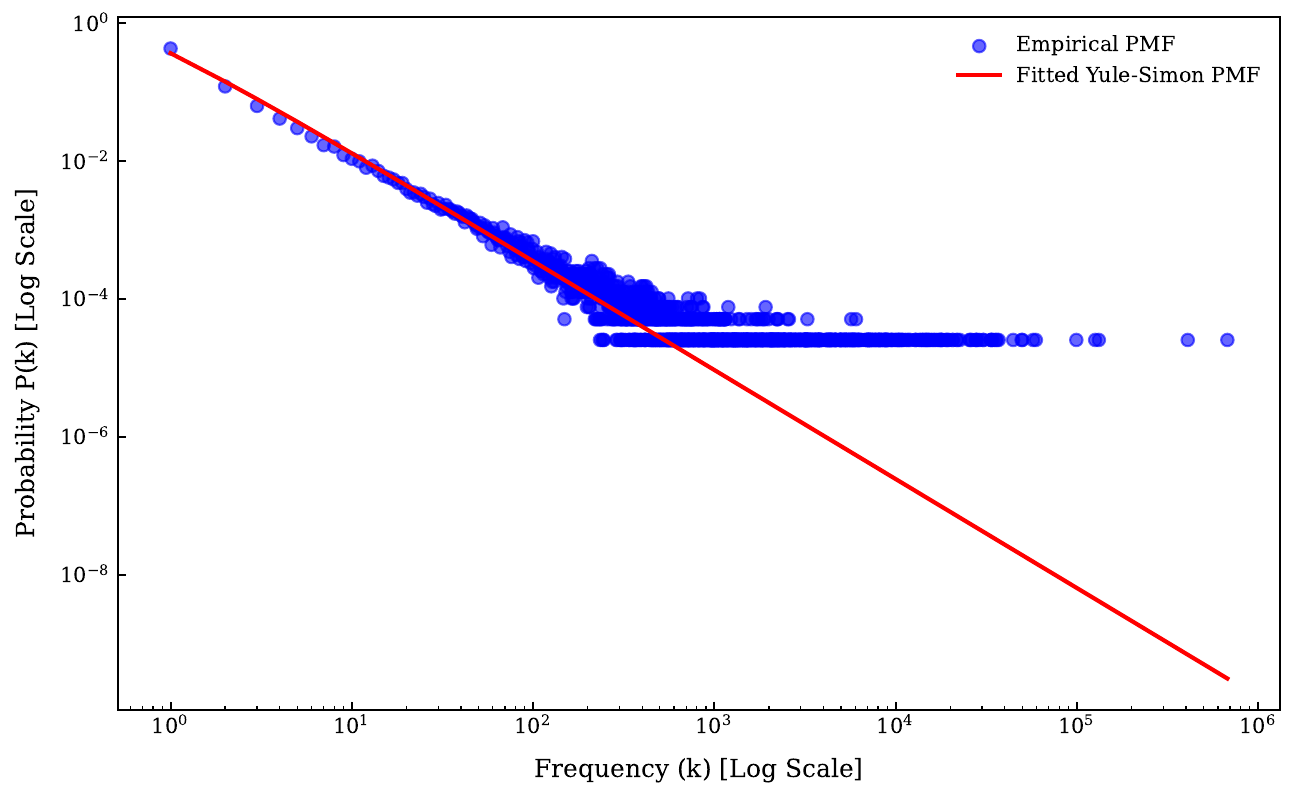}} &
        \subcaptionbox{\tiny Czech (cs)}{\includegraphics[width=0.15\textwidth]{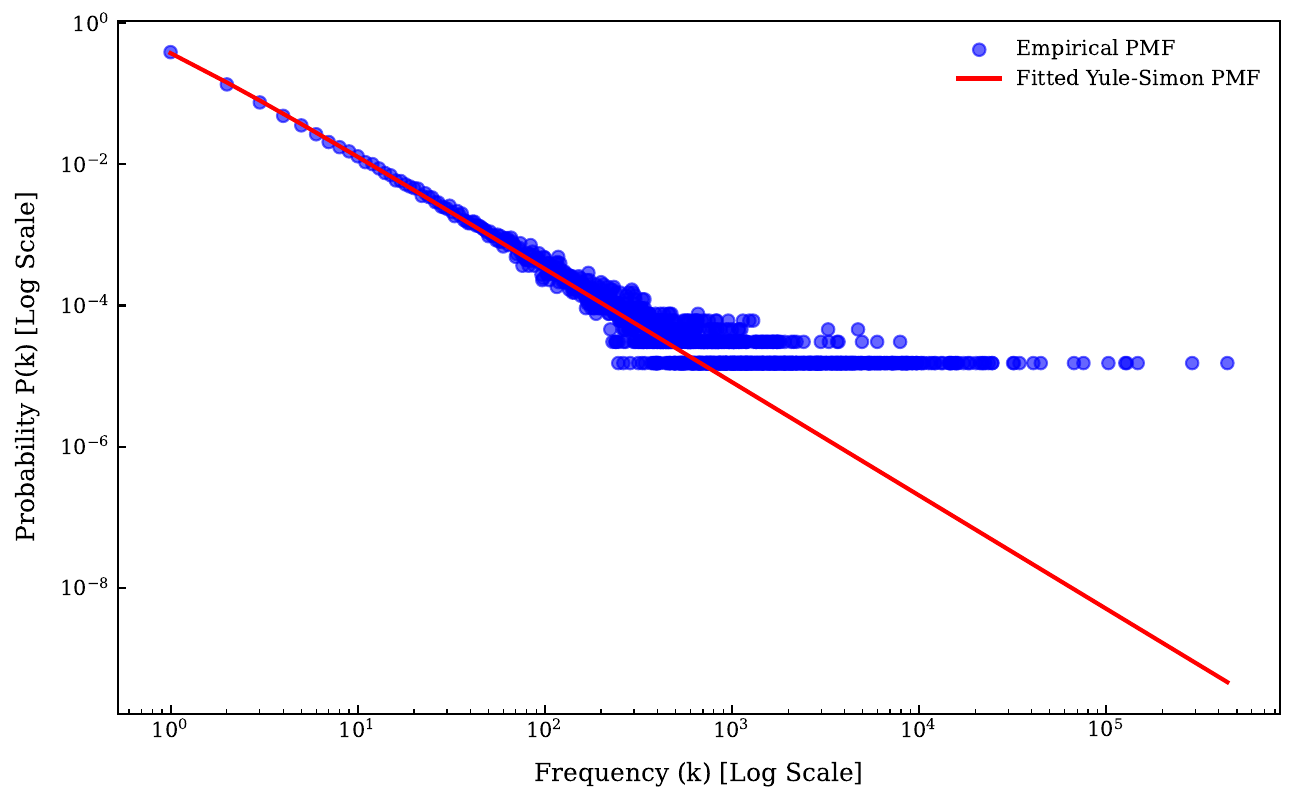}} &
        \subcaptionbox{\tiny Danish (da)}{\includegraphics[width=0.15\textwidth]{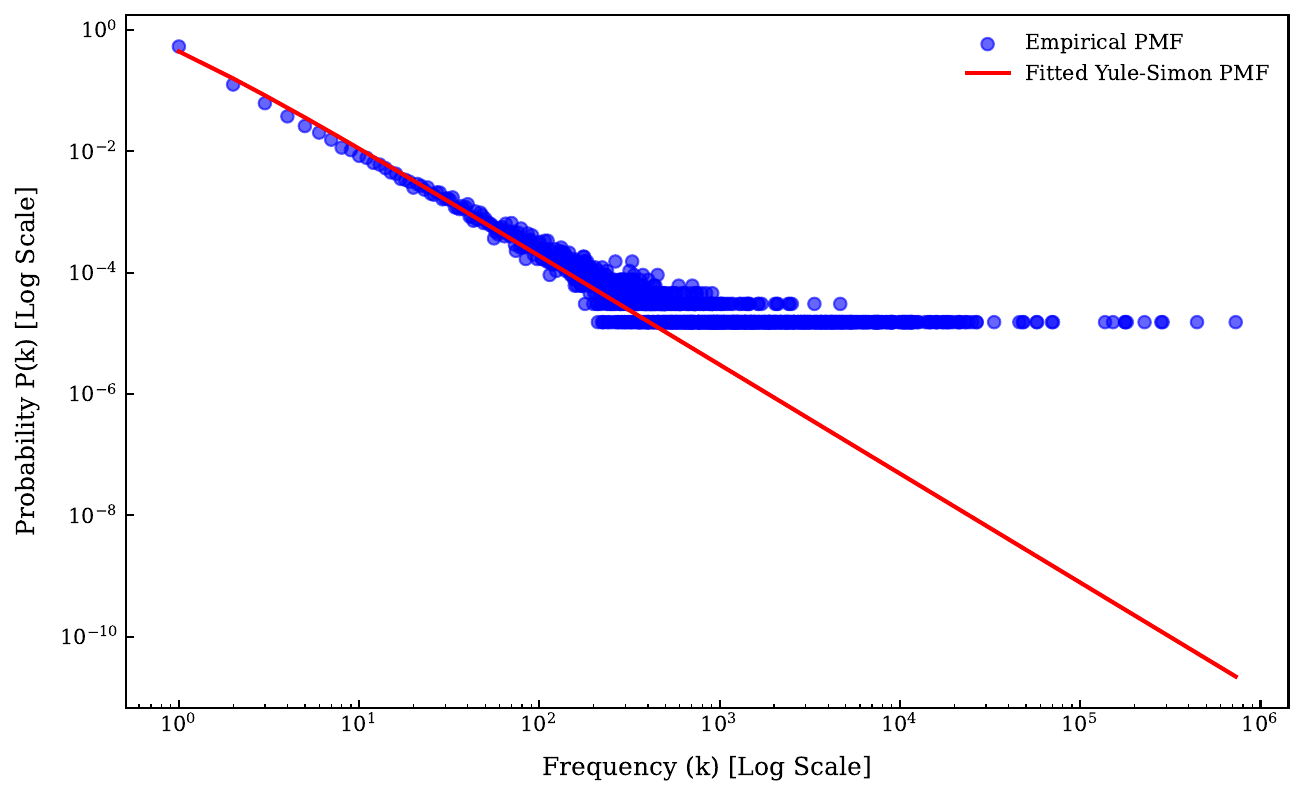}} &
        \subcaptionbox{\tiny German (de)}{\includegraphics[width=0.15\textwidth]{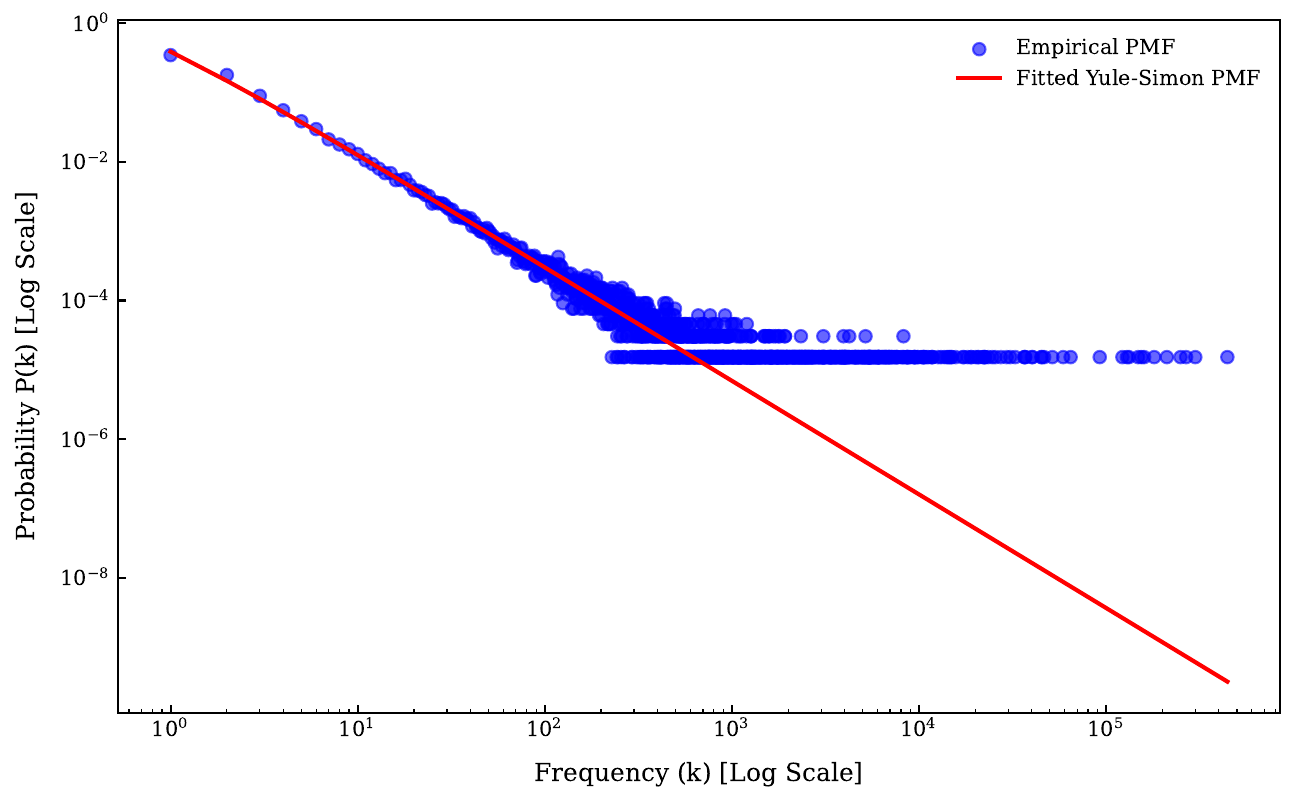}} &
        \subcaptionbox{\tiny Greek (el)}{\includegraphics[width=0.15\textwidth]{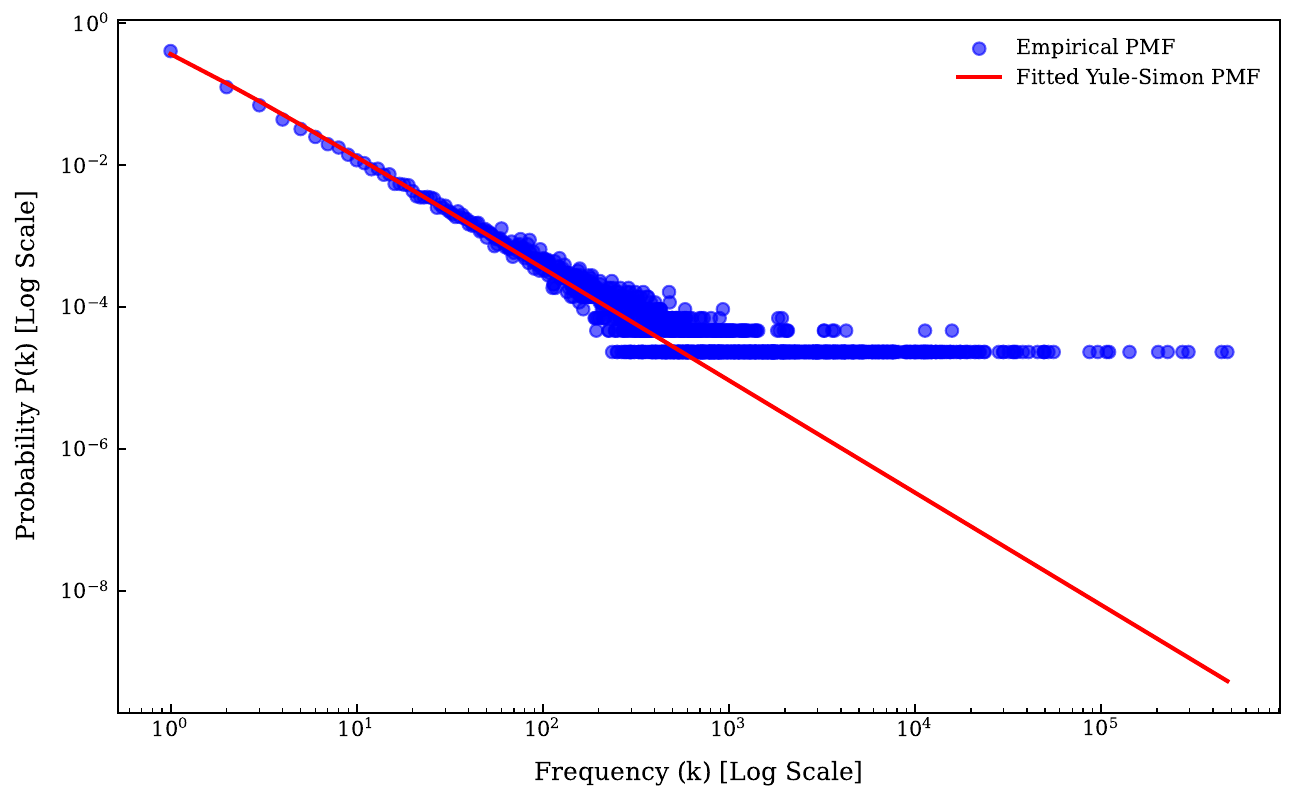}} \\
        
        \subcaptionbox{\tiny Spanish (es)\vspace{1em}}{\includegraphics[width=0.15\textwidth]{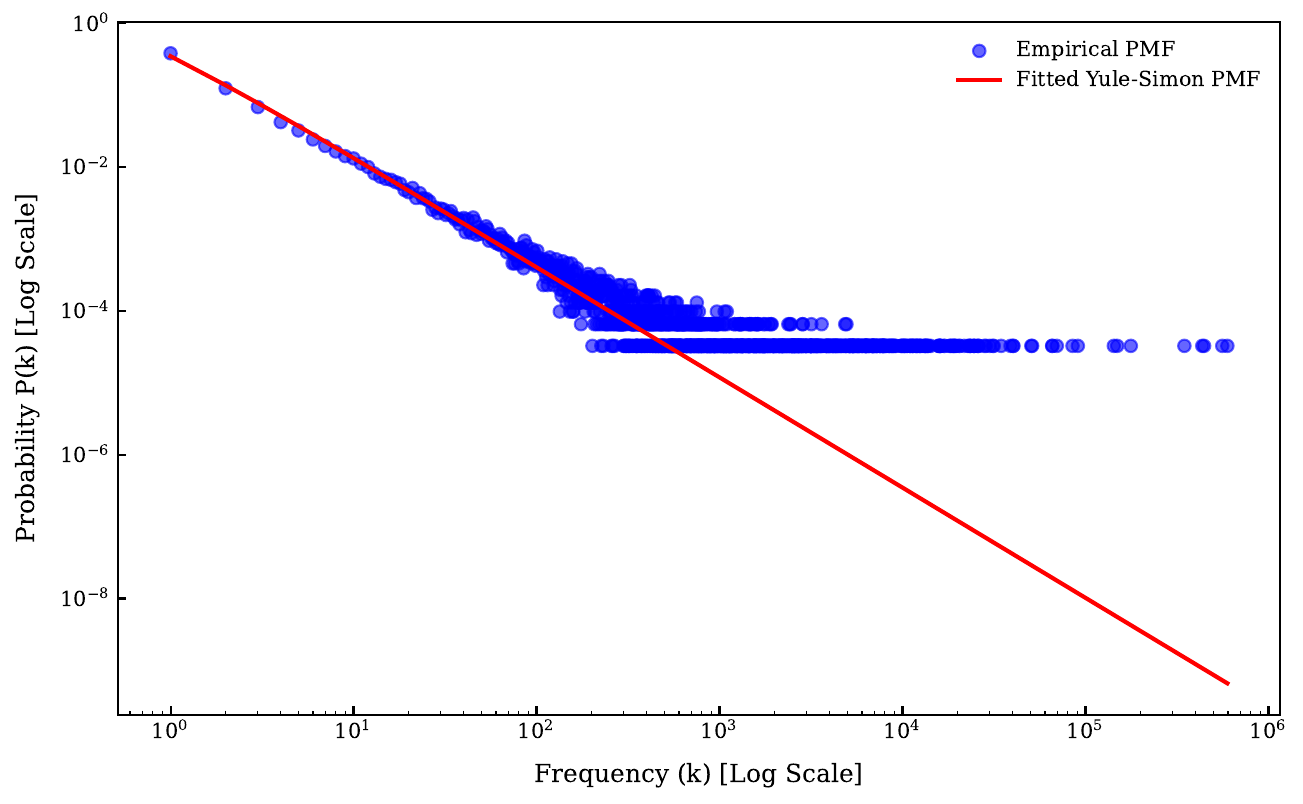}} &
        \subcaptionbox{\tiny English (en)}{\includegraphics[width=0.15\textwidth]{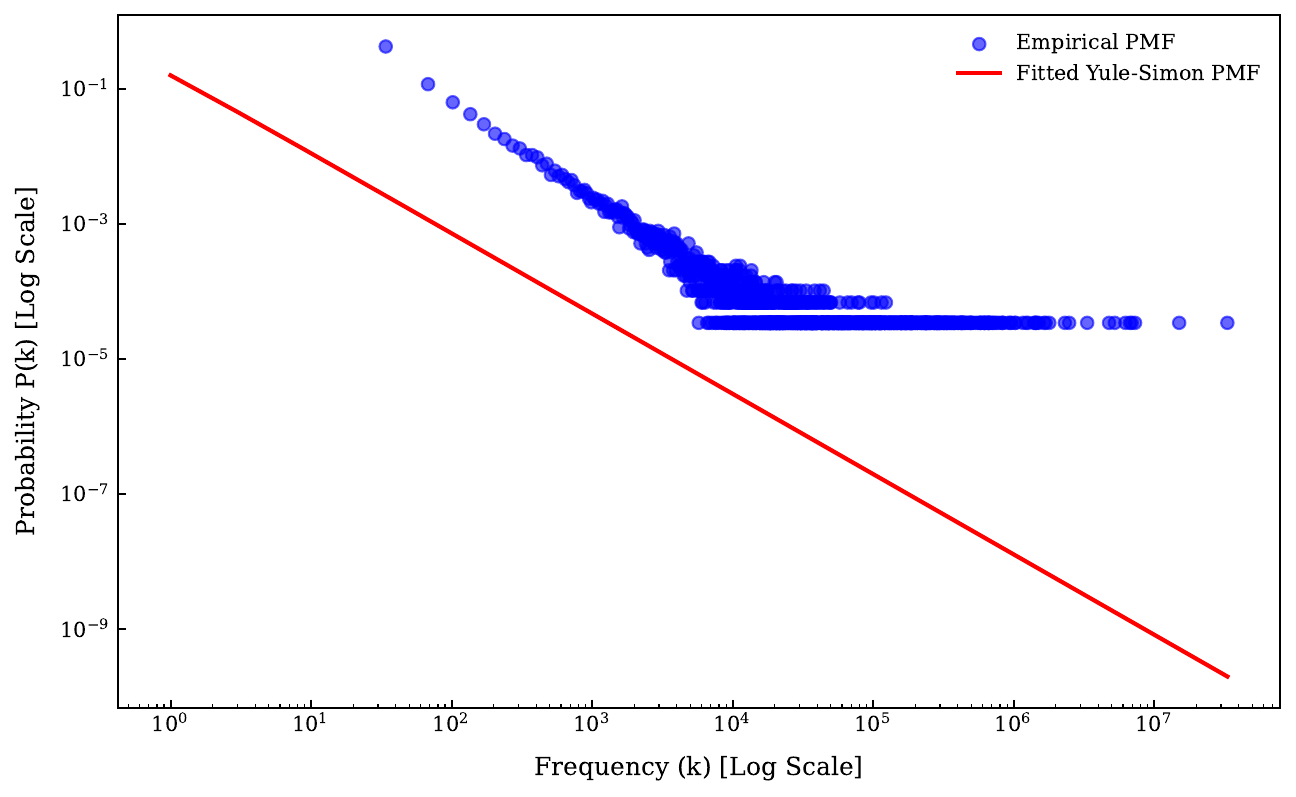}} &
        \subcaptionbox{\tiny Persian (fa)}{\includegraphics[width=0.15\textwidth]{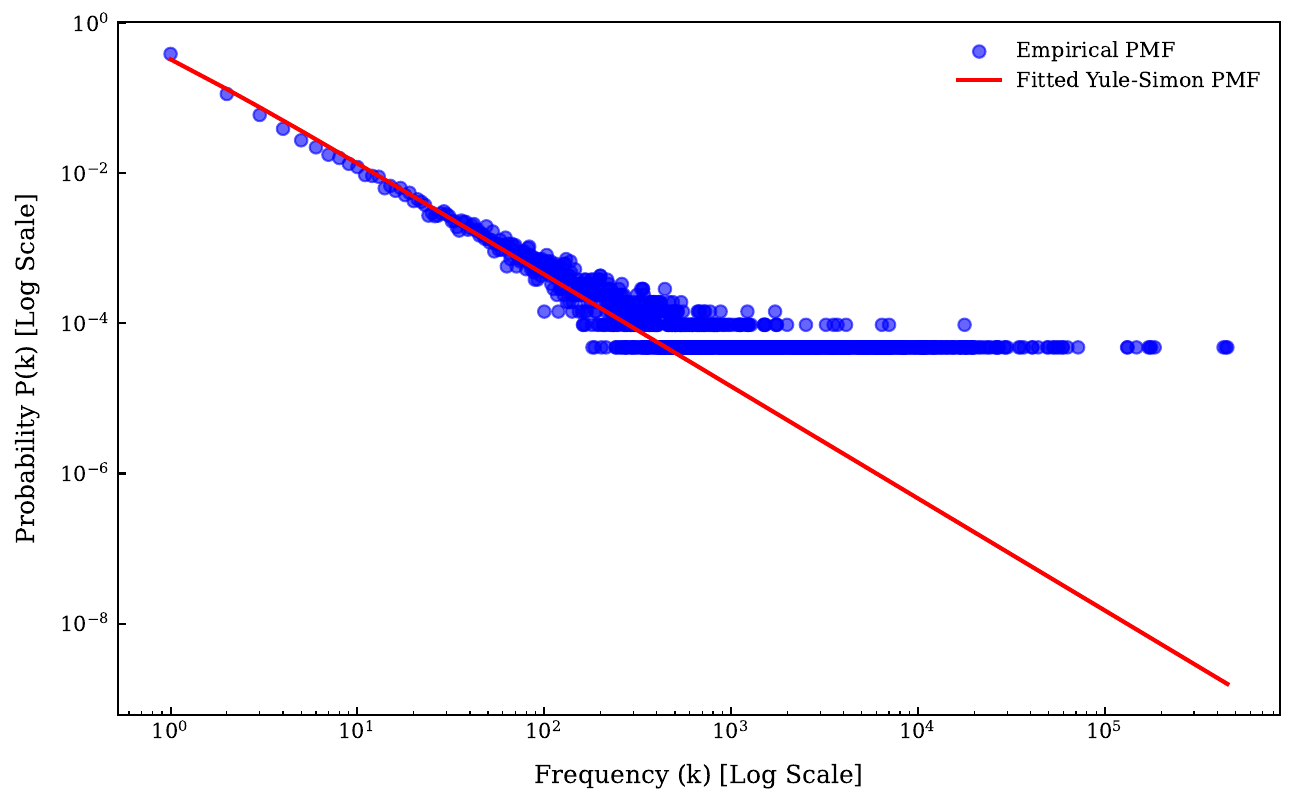}} &
        \subcaptionbox{\tiny Finnish (fi)}{\includegraphics[width=0.15\textwidth]{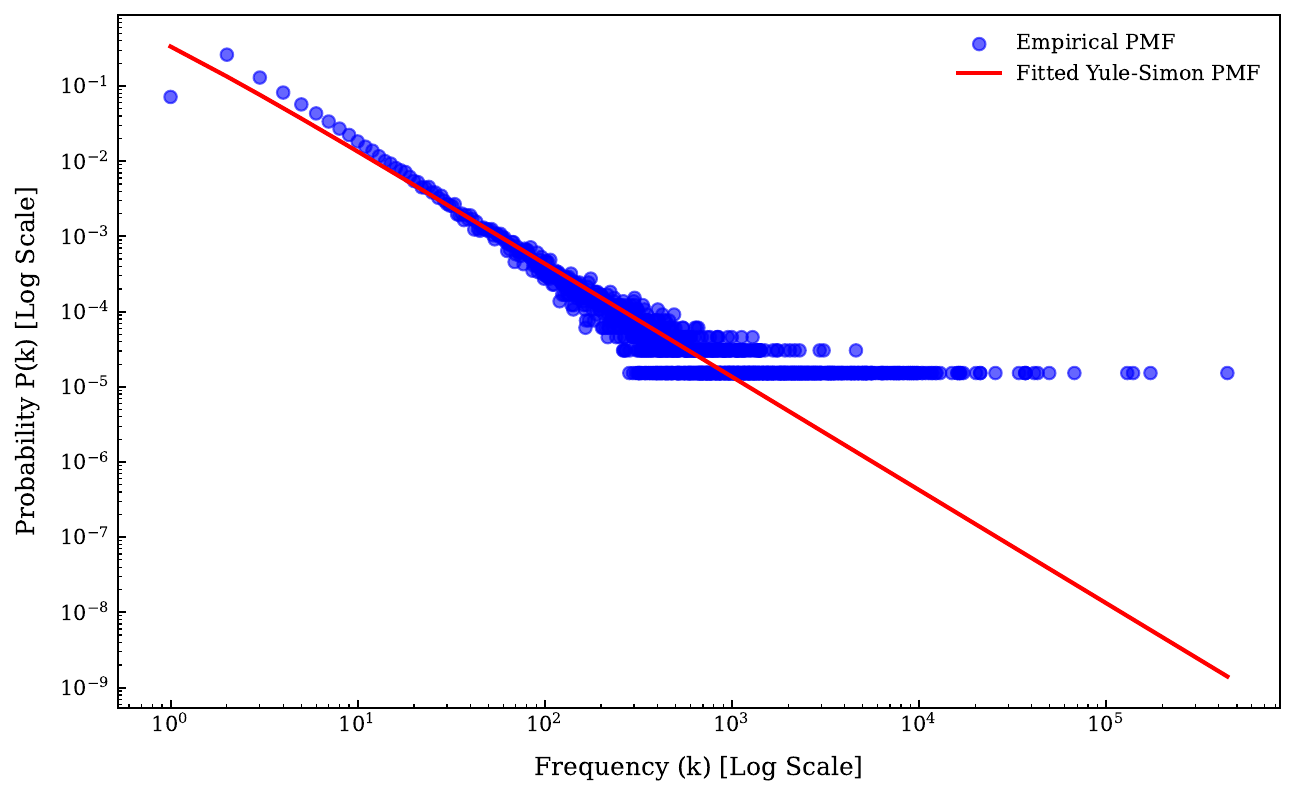}} &
        \subcaptionbox{\tiny Filipino (fil)}{\includegraphics[width=0.15\textwidth]{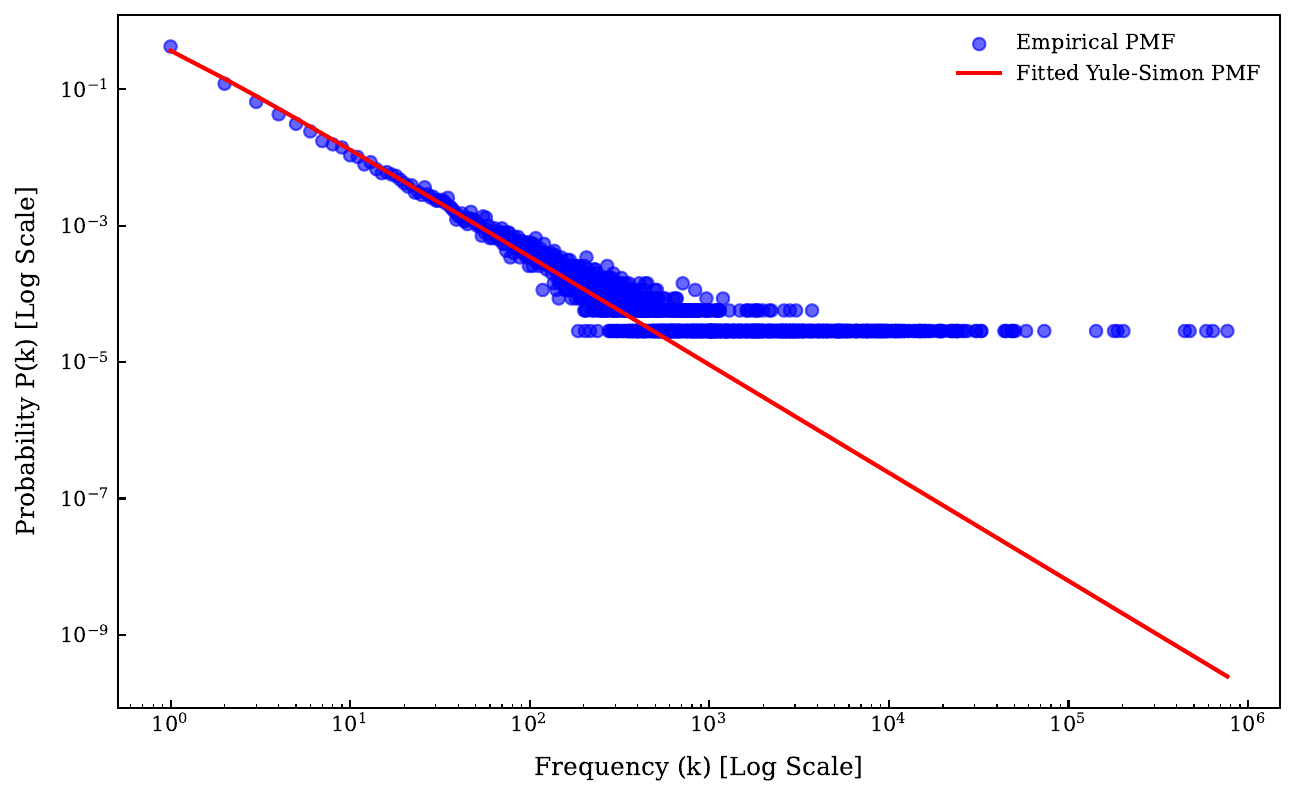}} &
        \subcaptionbox{\tiny French (fr)}{\includegraphics[width=0.15\textwidth]{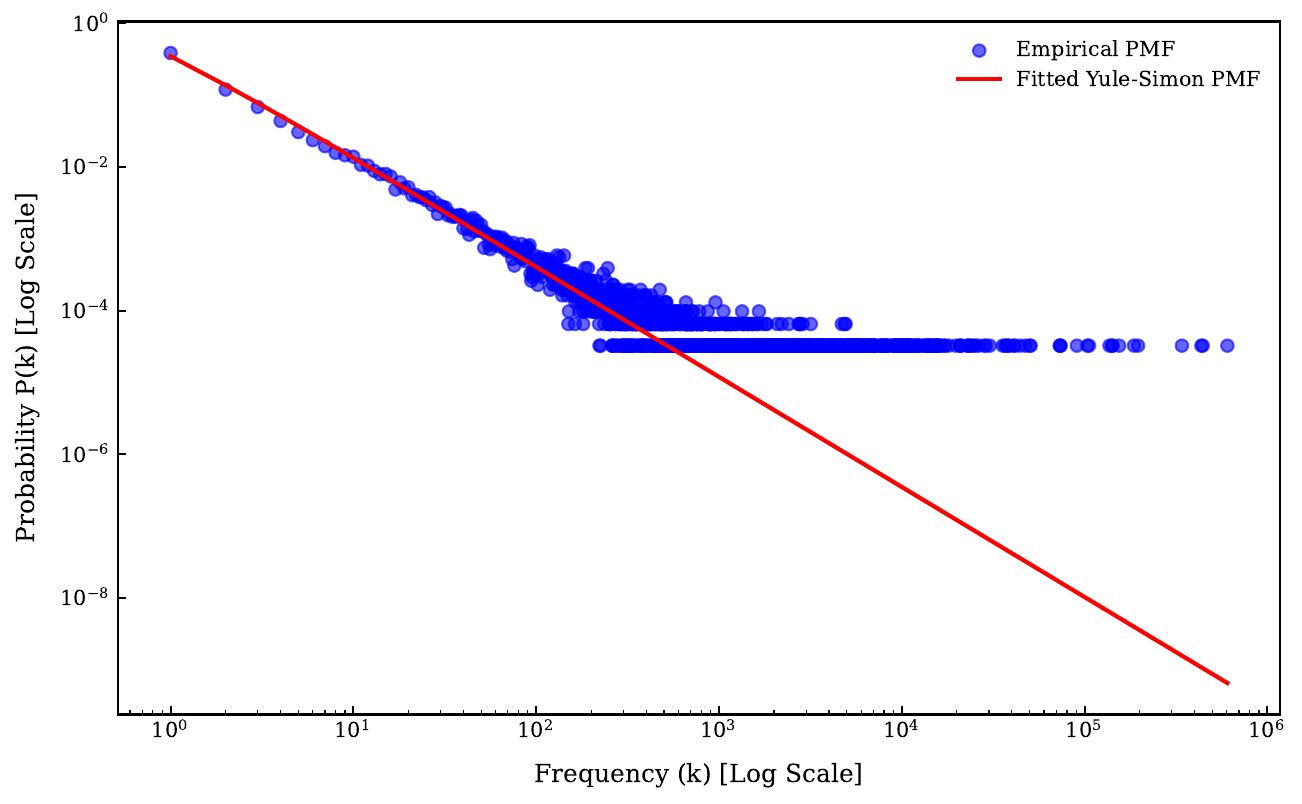}} \\
        
        \subcaptionbox{\tiny Hindi (hi)\vspace{1em}}{\includegraphics[width=0.15\textwidth]{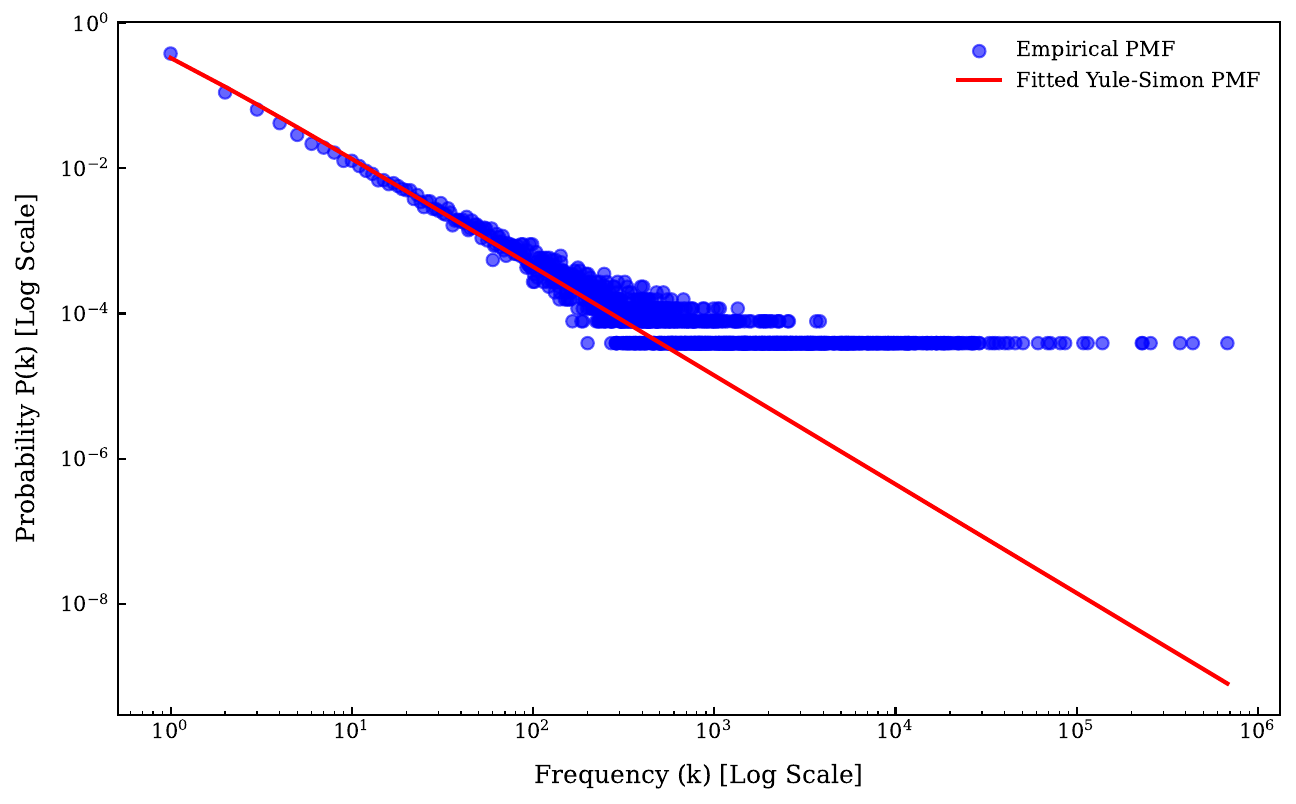}} &
        \subcaptionbox{\tiny Croatian (hr)}{\includegraphics[width=0.15\textwidth]{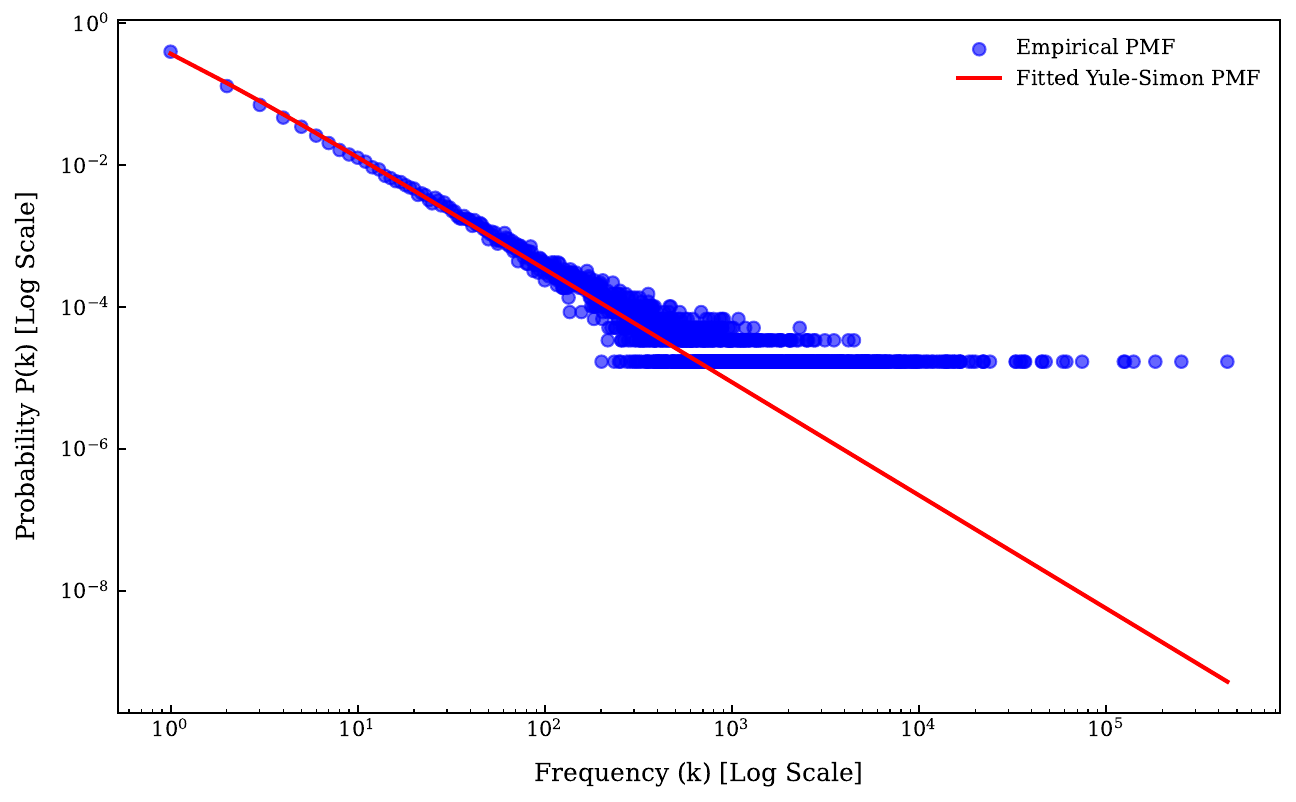}} &
        \subcaptionbox{\tiny Hungarian (hu)}{\includegraphics[width=0.15\textwidth]{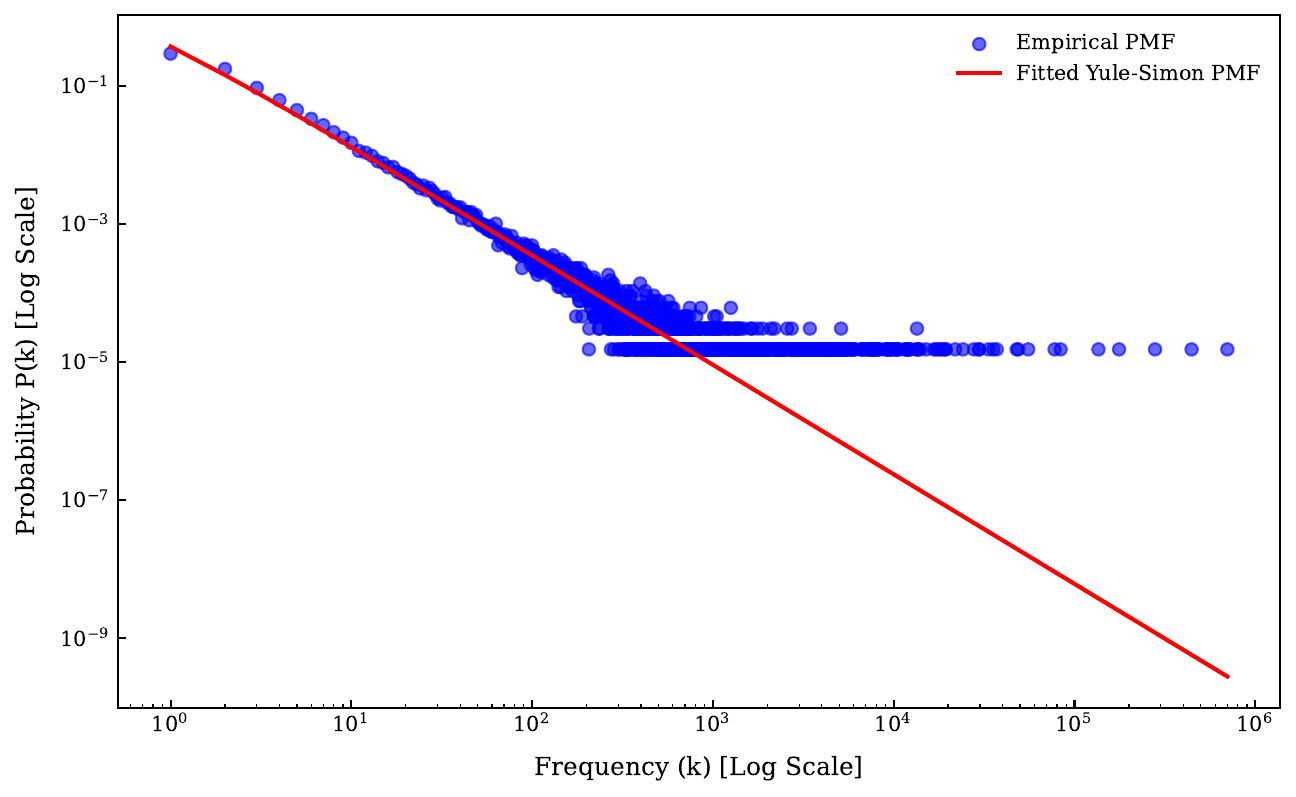}} &
        \subcaptionbox{\tiny Indonesian (id)}{\includegraphics[width=0.15\textwidth]{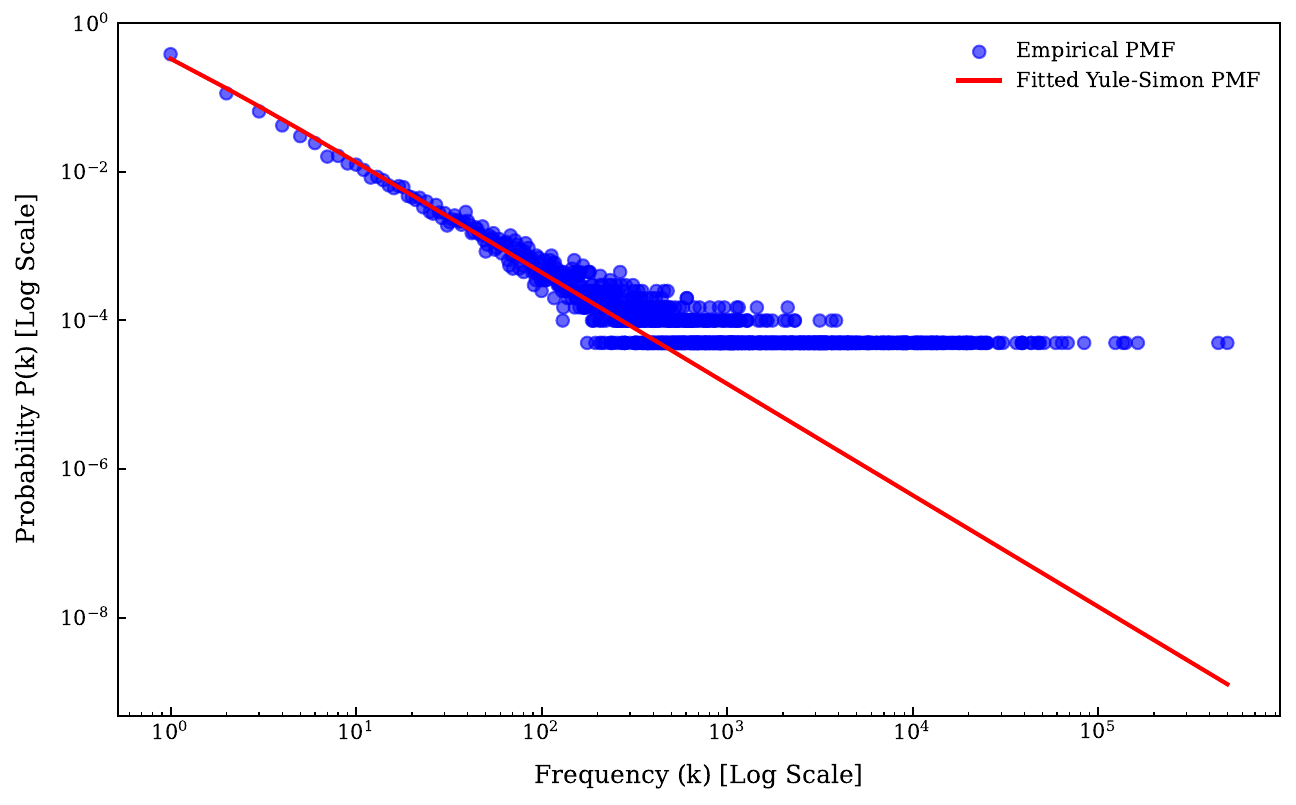}} &
        \subcaptionbox{\tiny Italian (it)}{\includegraphics[width=0.15\textwidth]{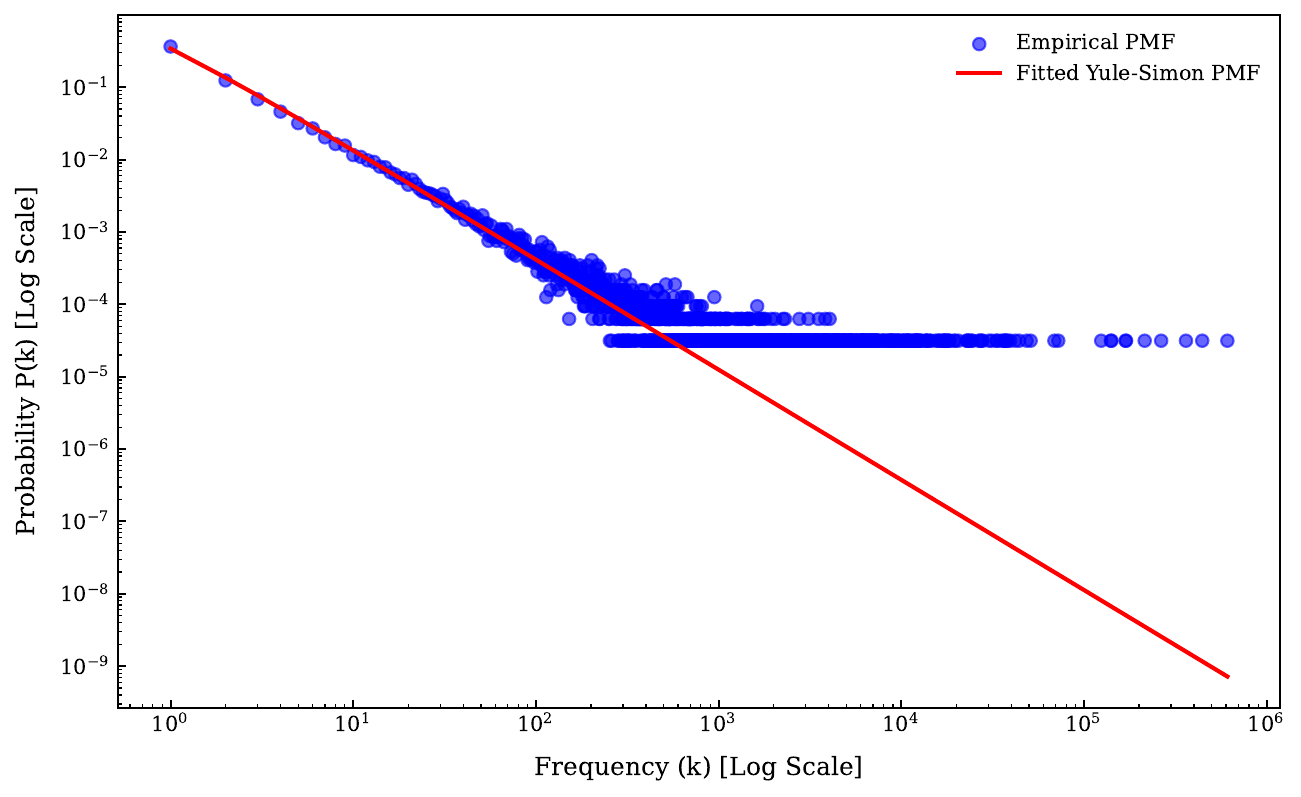}} &
        \subcaptionbox{\tiny Hebrew (he)}{\includegraphics[width=0.15\textwidth]{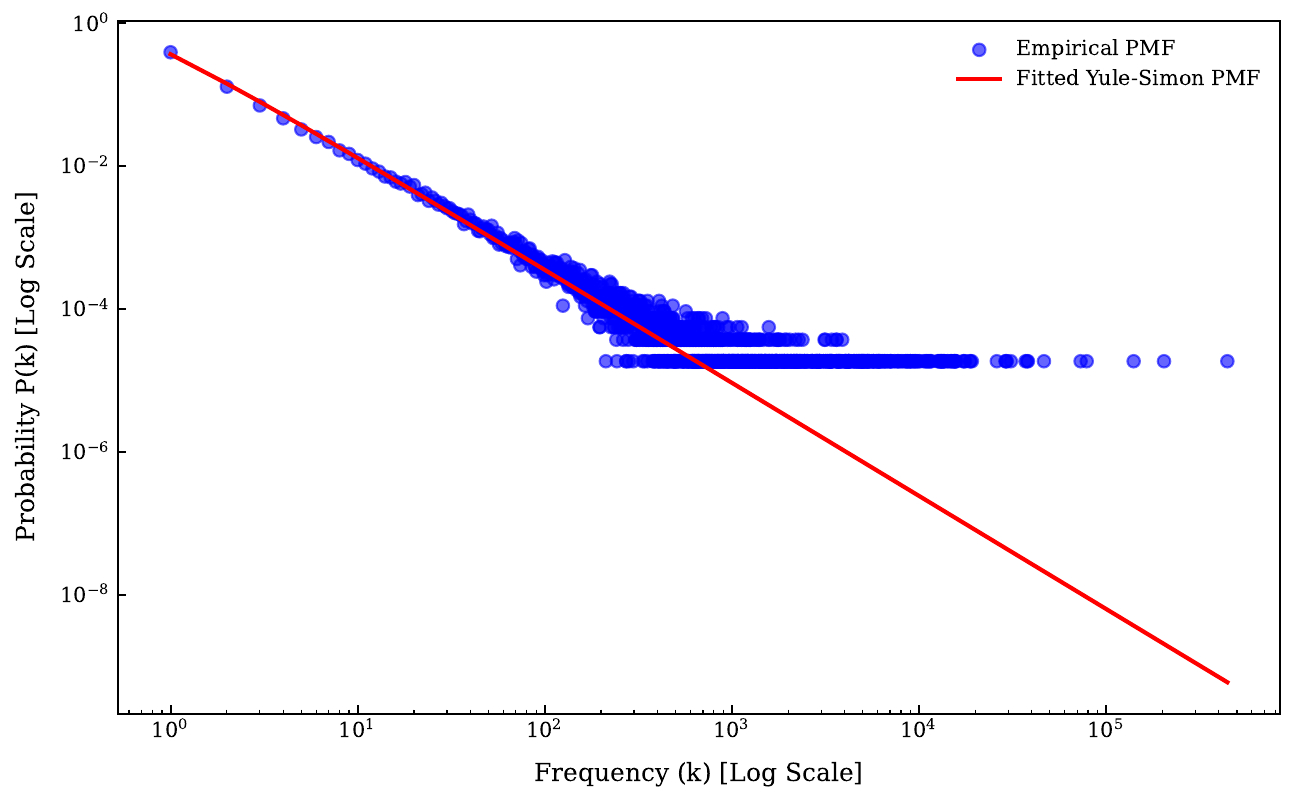}} \\
        
        \subcaptionbox{\tiny Japanese (ja)\vspace{1em}}{\includegraphics[width=0.15\textwidth]{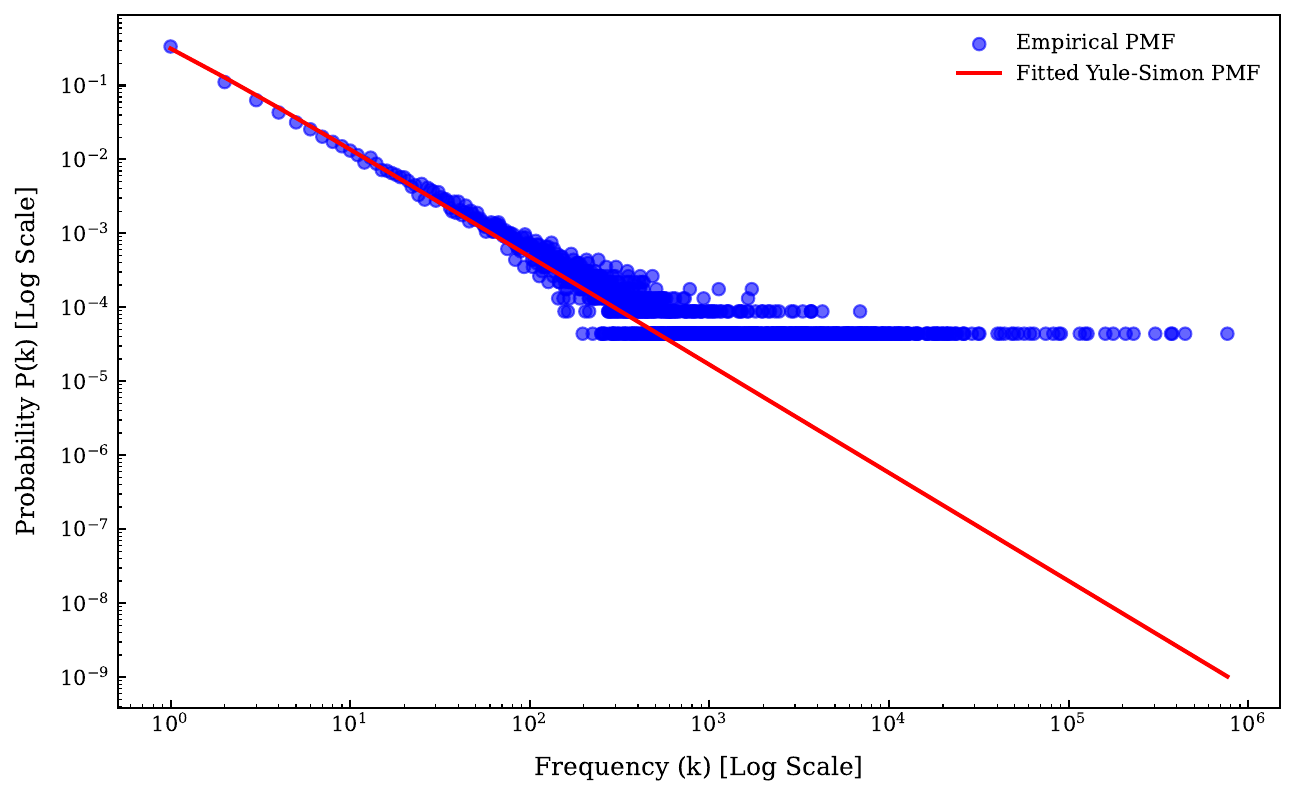}} &
        \subcaptionbox{\tiny Korean (ko)}{\includegraphics[width=0.15\textwidth]{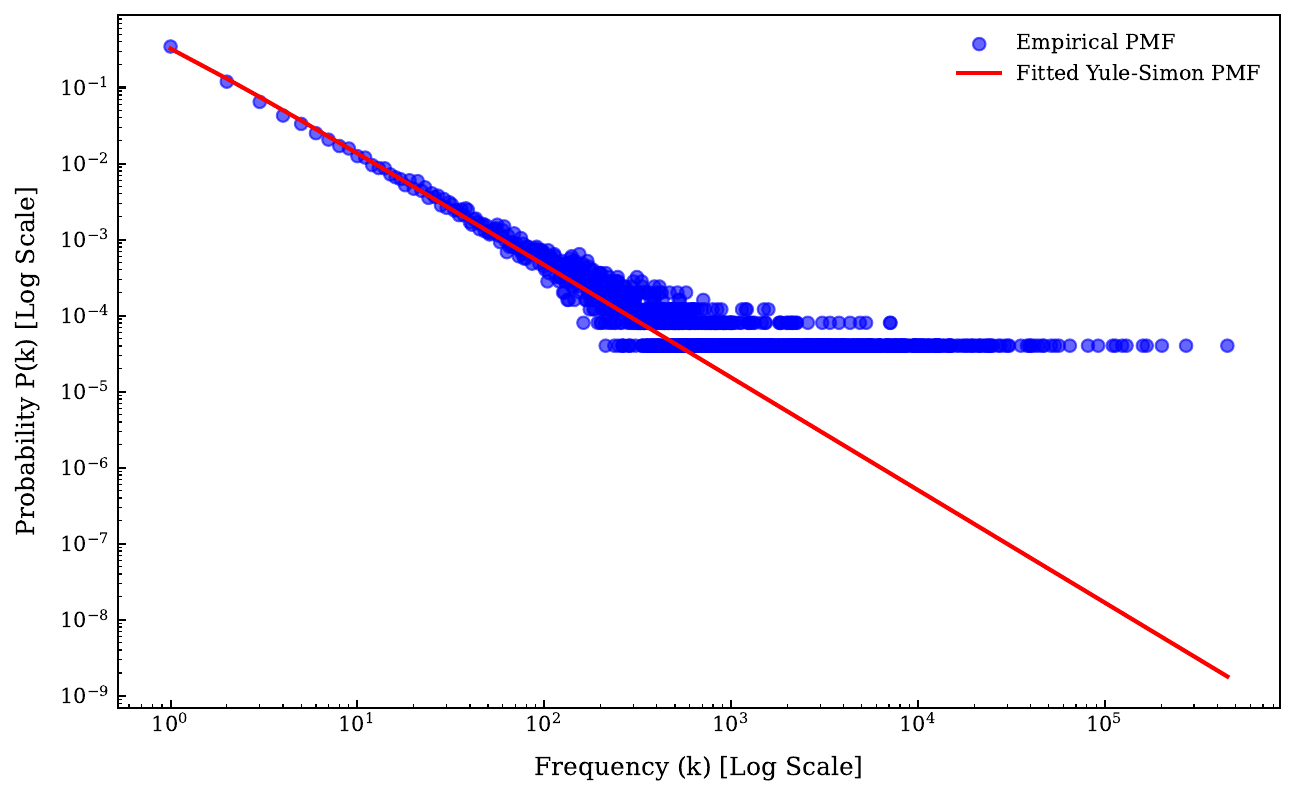}} &
        \subcaptionbox{\tiny Maori (mi)}{\includegraphics[width=0.15\textwidth]{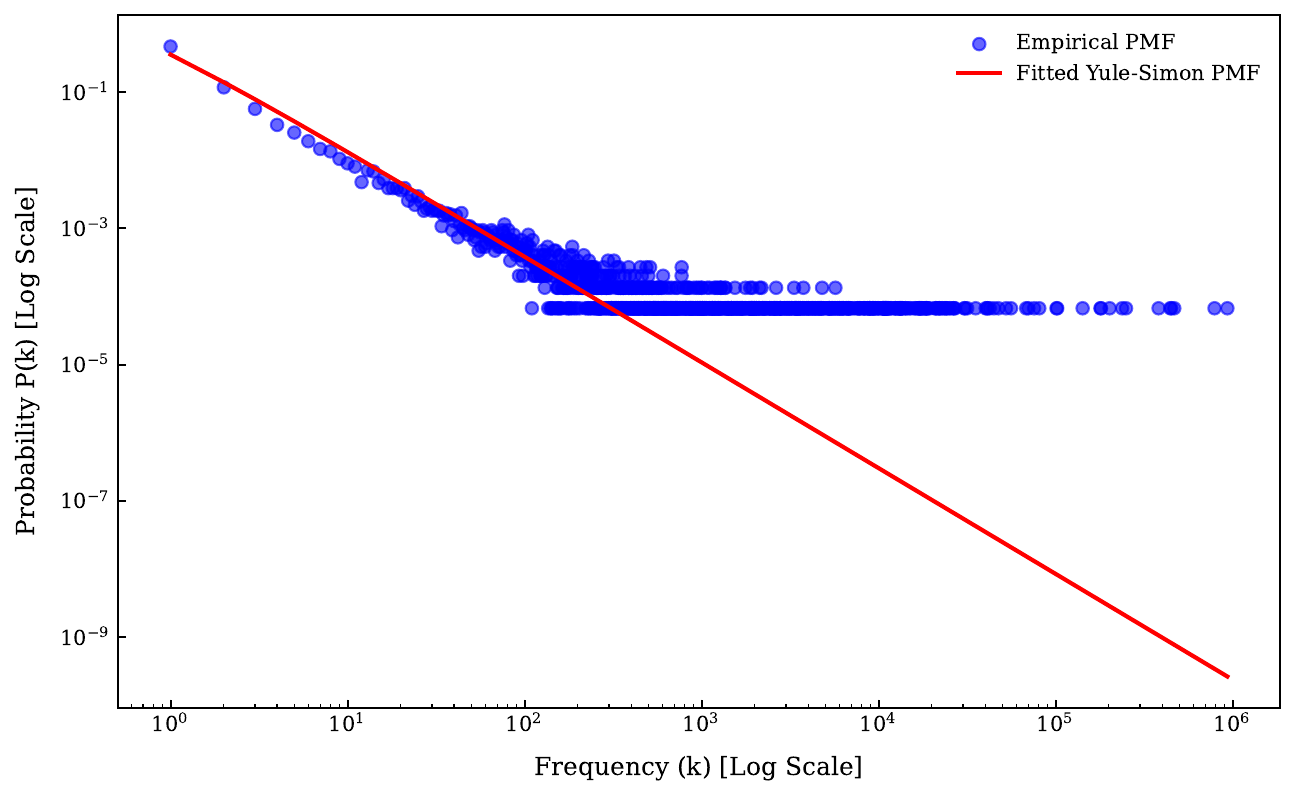}} &
        \subcaptionbox{\tiny Dutch (nl)}{\includegraphics[width=0.15\textwidth]{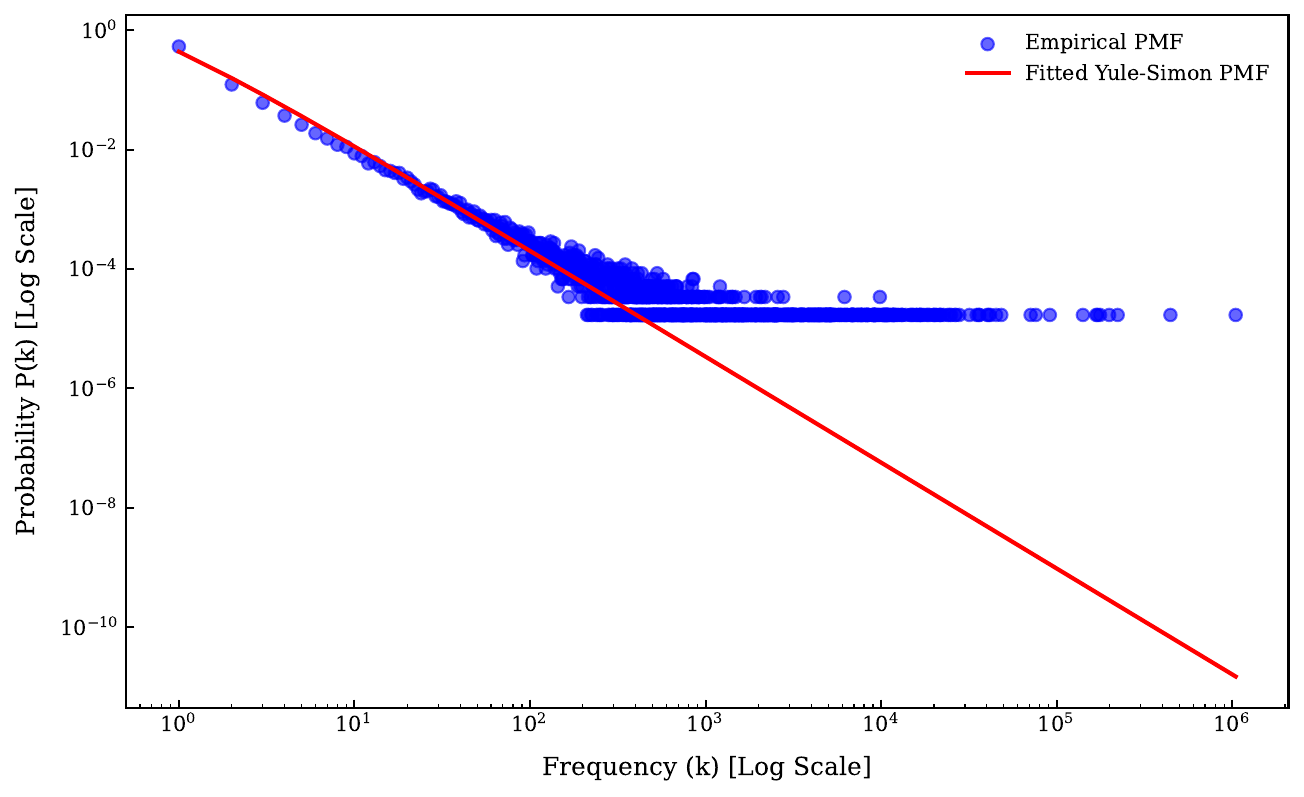}} &
        \subcaptionbox{\tiny Norwegian (no)}{\includegraphics[width=0.15\textwidth]{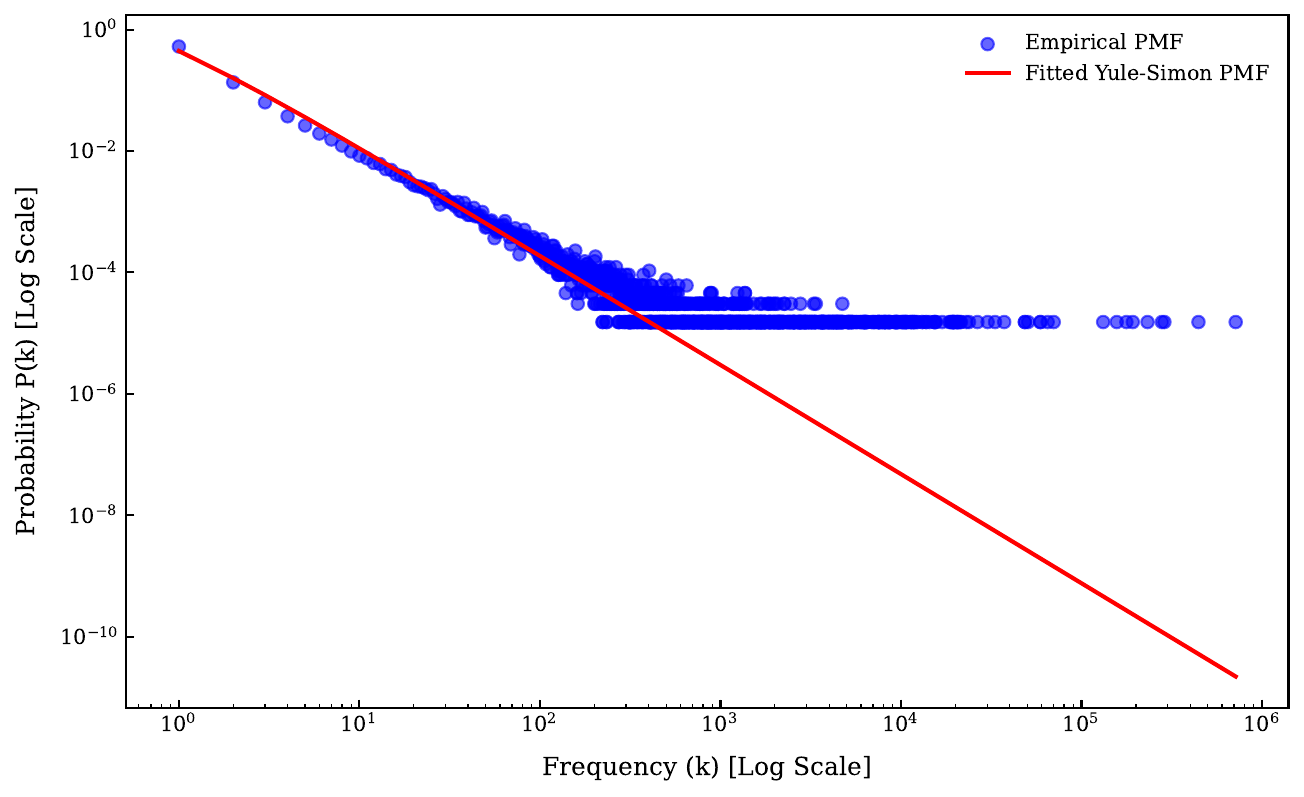}} &
        \subcaptionbox{\tiny Polish (pl)}{\includegraphics[width=0.15\textwidth]{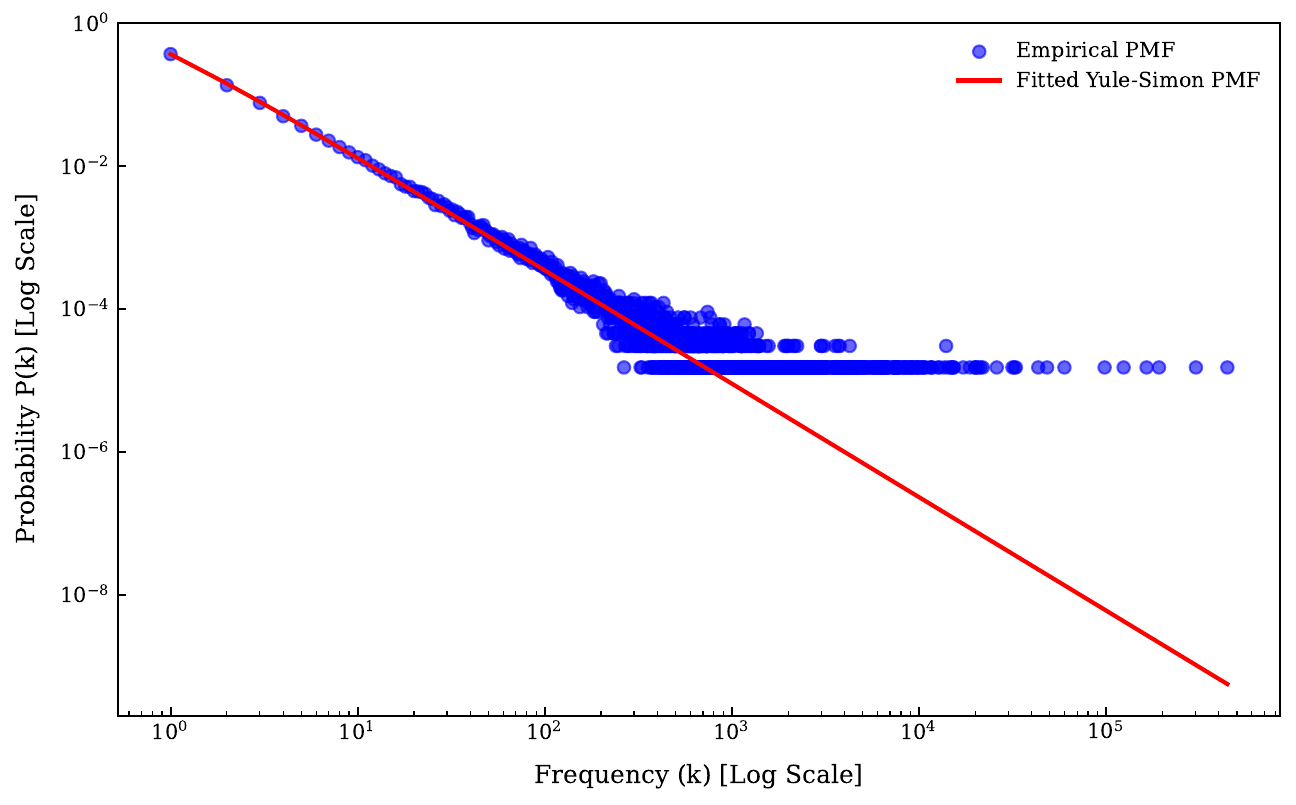}} \\
        
        \subcaptionbox{\tiny Portuguese (pt)\vspace{1em}}{\includegraphics[width=0.15\textwidth]{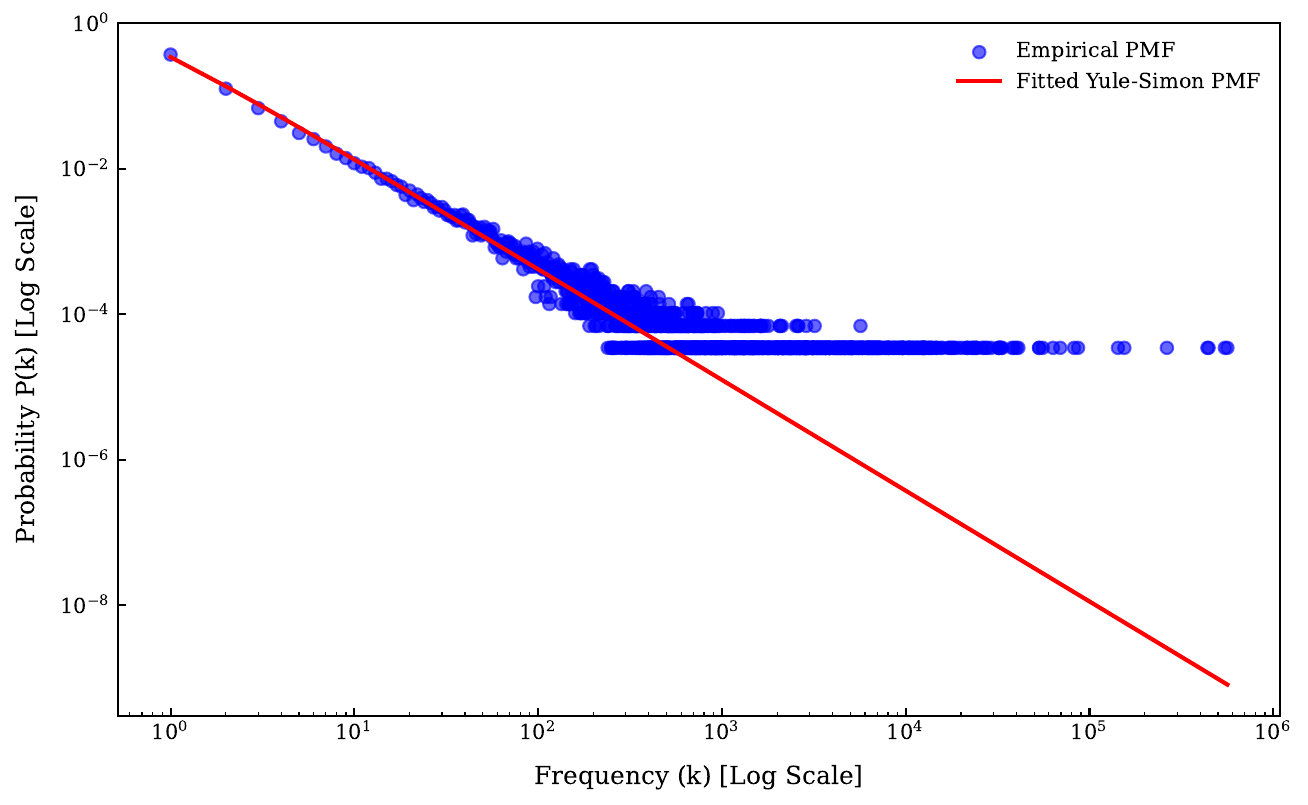}} &
        \subcaptionbox{\tiny Romanian (ro)}{\includegraphics[width=0.15\textwidth]{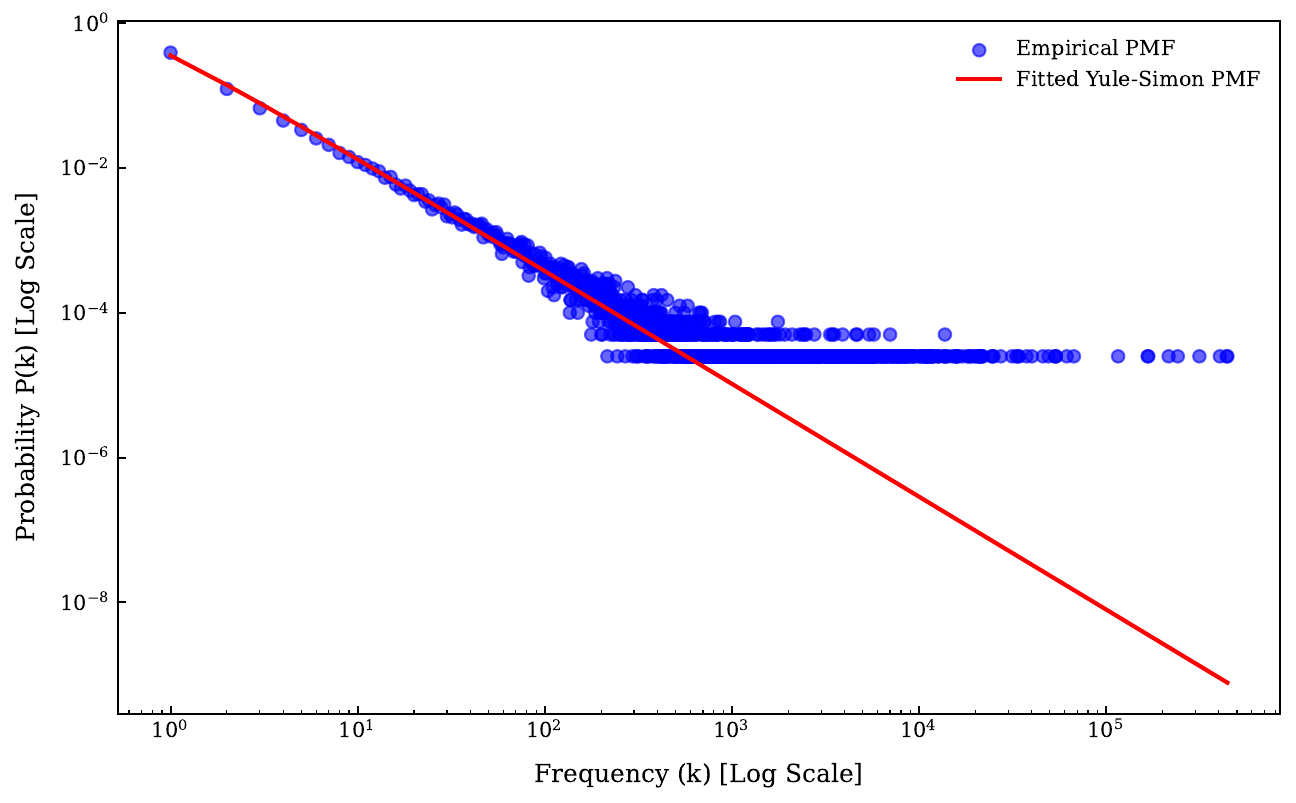}} &
        \subcaptionbox{\tiny Russian (ru)}{\includegraphics[width=0.15\textwidth]{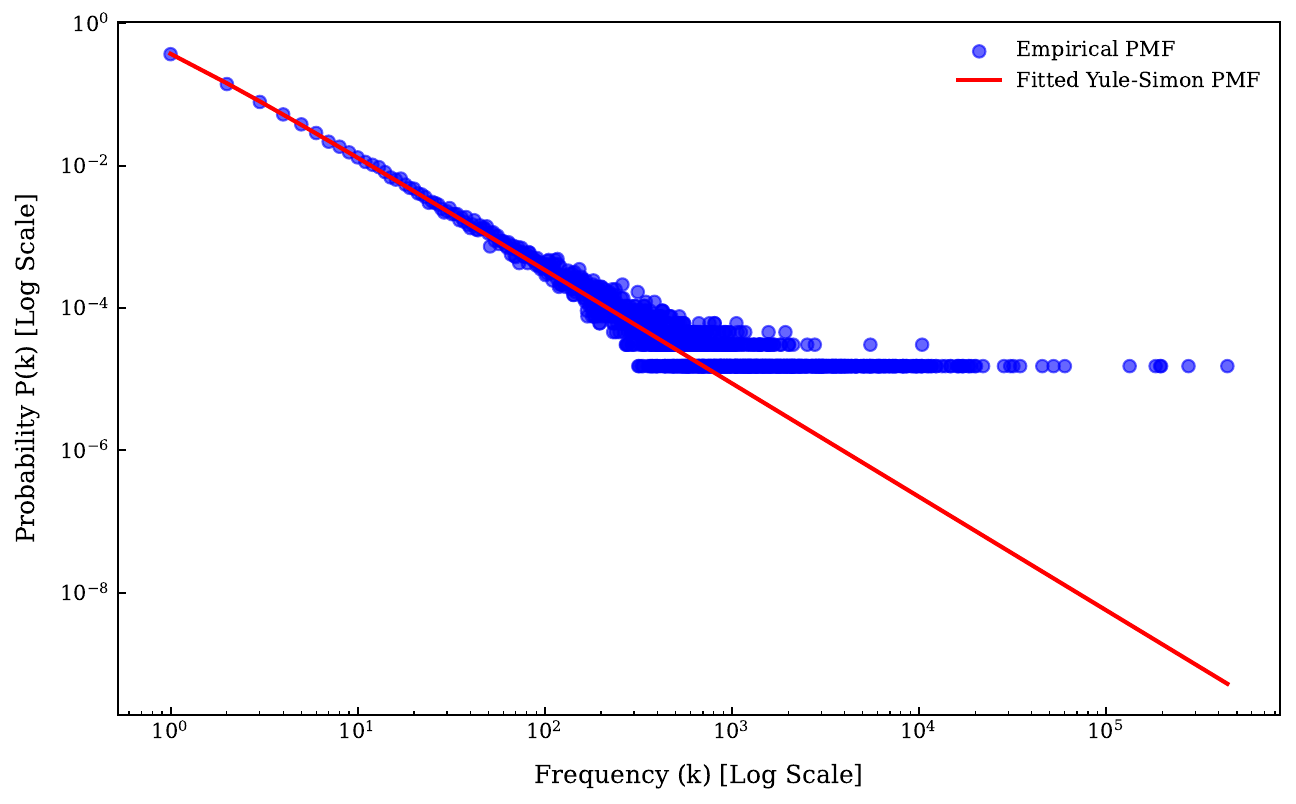}} &
        \subcaptionbox{\tiny Swedish (sv)}{\includegraphics[width=0.15\textwidth]{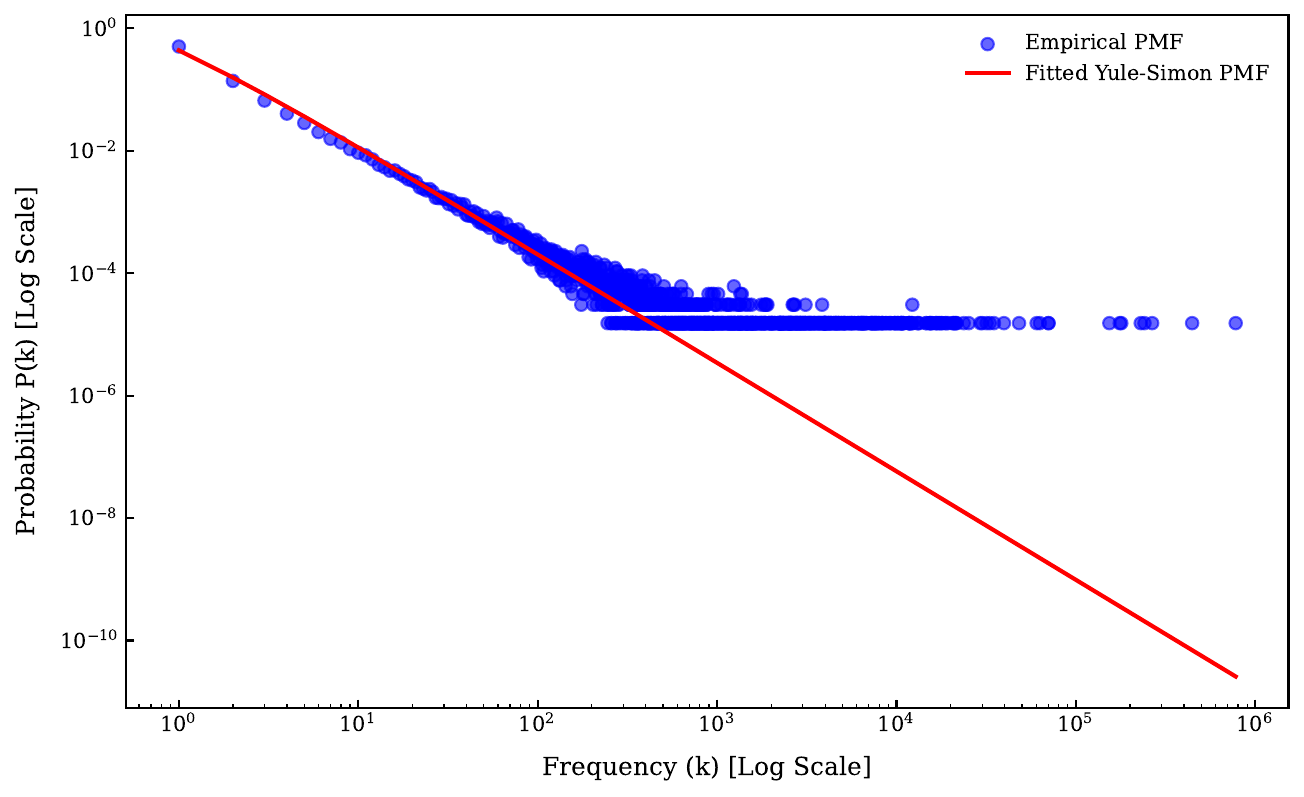}} &
        \subcaptionbox{\tiny Swahili (sw)}{\includegraphics[width=0.15\textwidth]{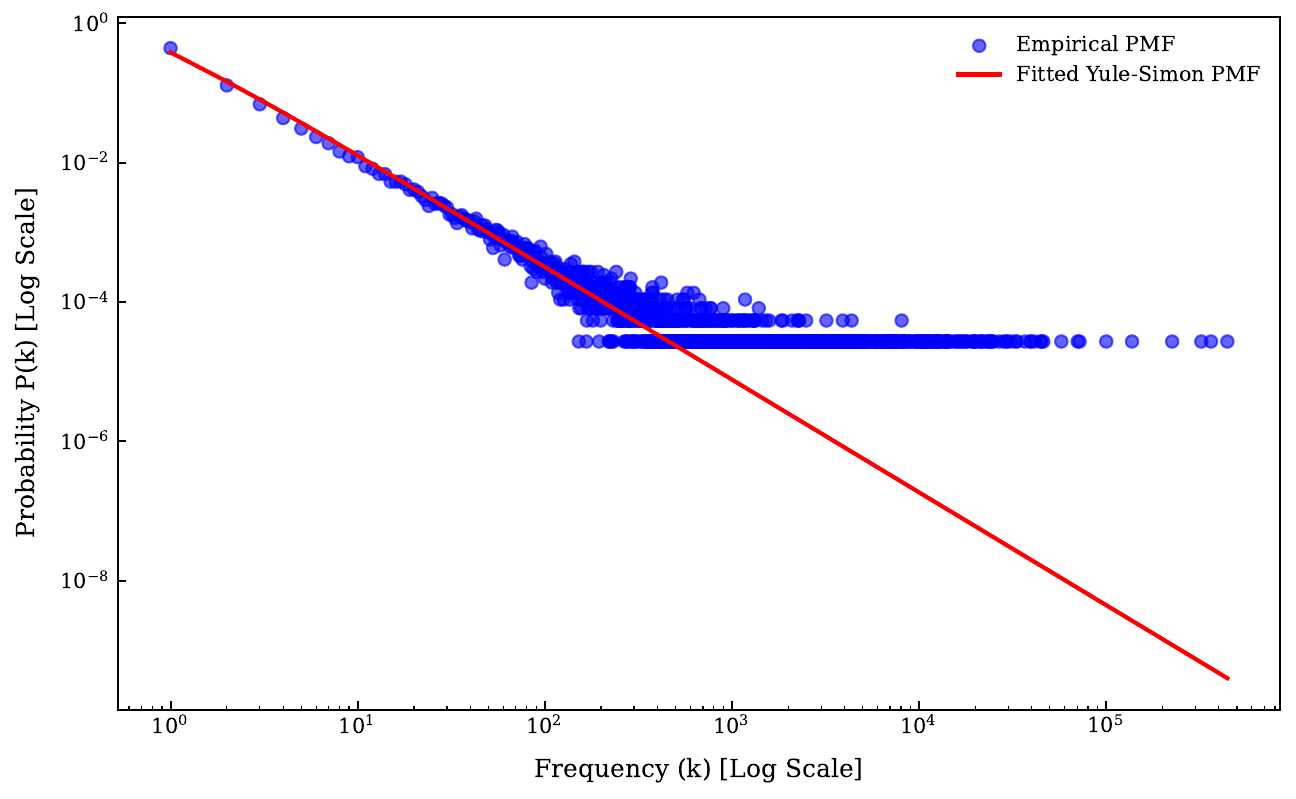}} &
        \subcaptionbox{\tiny Telugu (te)}{\includegraphics[width=0.15\textwidth]{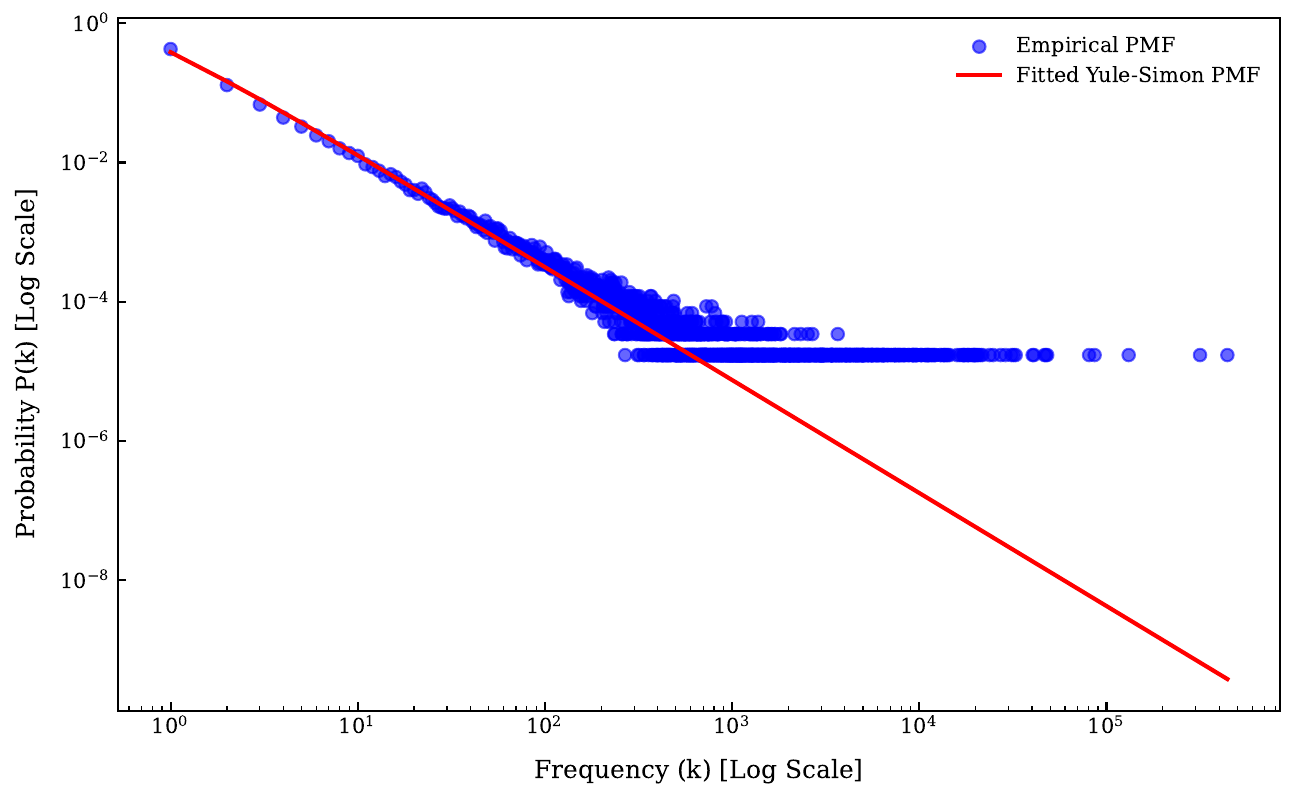}} \\
        
        \subcaptionbox{\tiny Thai (th)\vspace{1em}}{\includegraphics[width=0.15\textwidth]{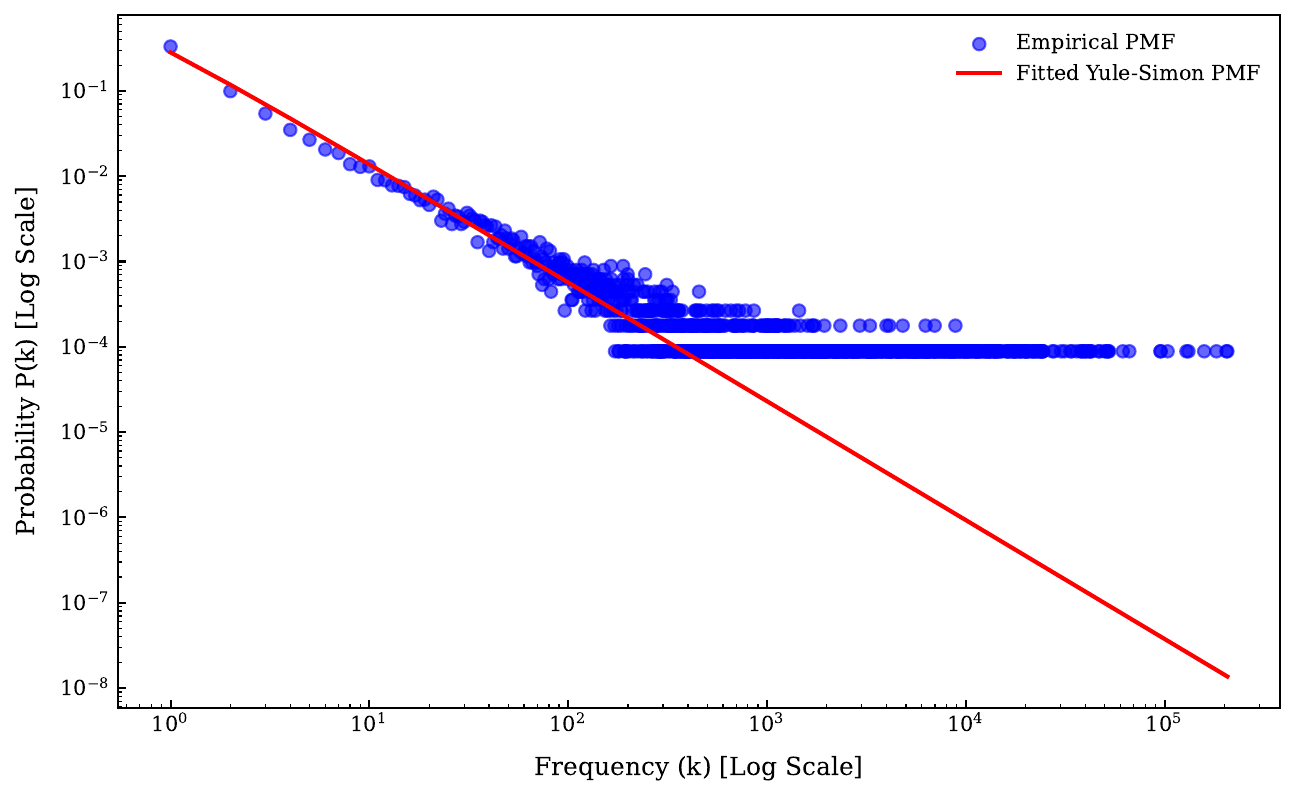}} &
        \subcaptionbox{\tiny Turkish (tr)}{\includegraphics[width=0.15\textwidth]{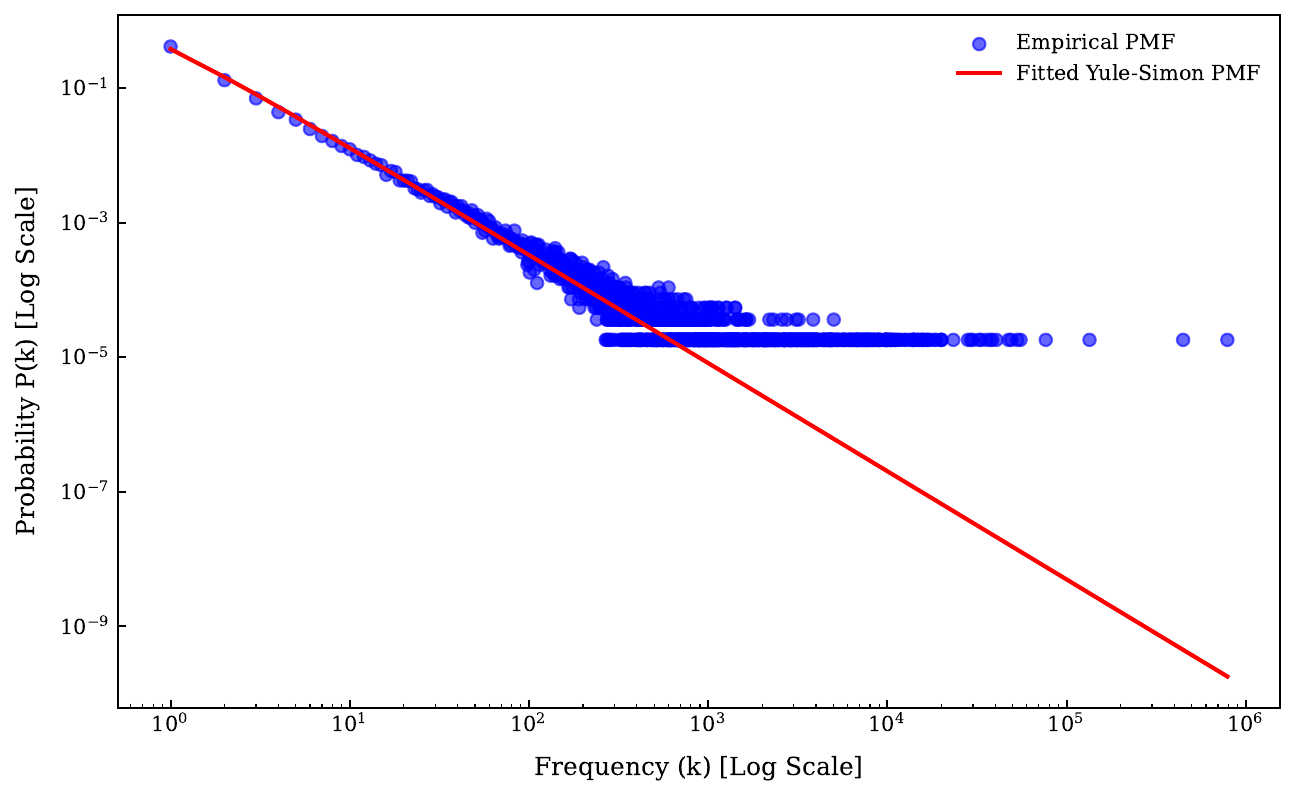}} &
        \subcaptionbox{\tiny Ukrainian (uk)}{\includegraphics[width=0.15\textwidth]{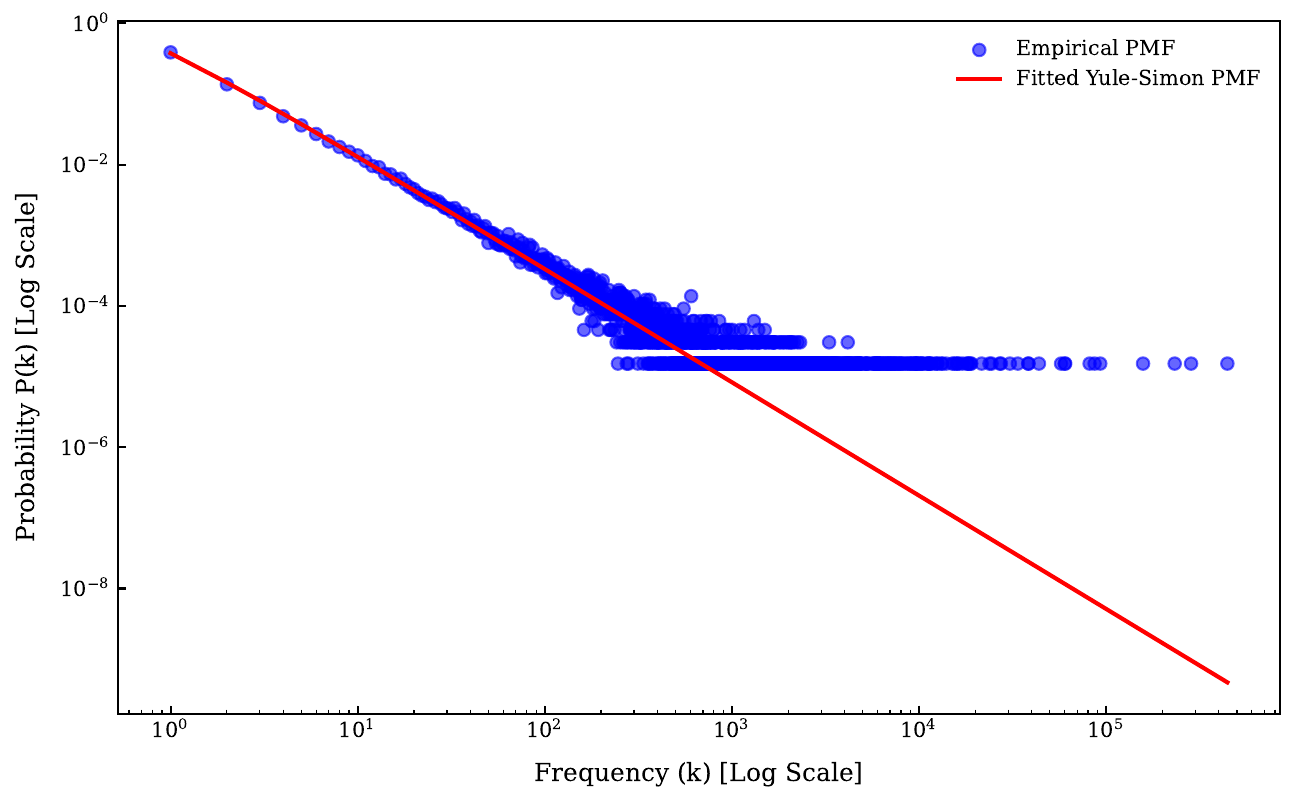}} &
        \subcaptionbox{\tiny Vietnamese (vi)}{\includegraphics[width=0.15\textwidth]{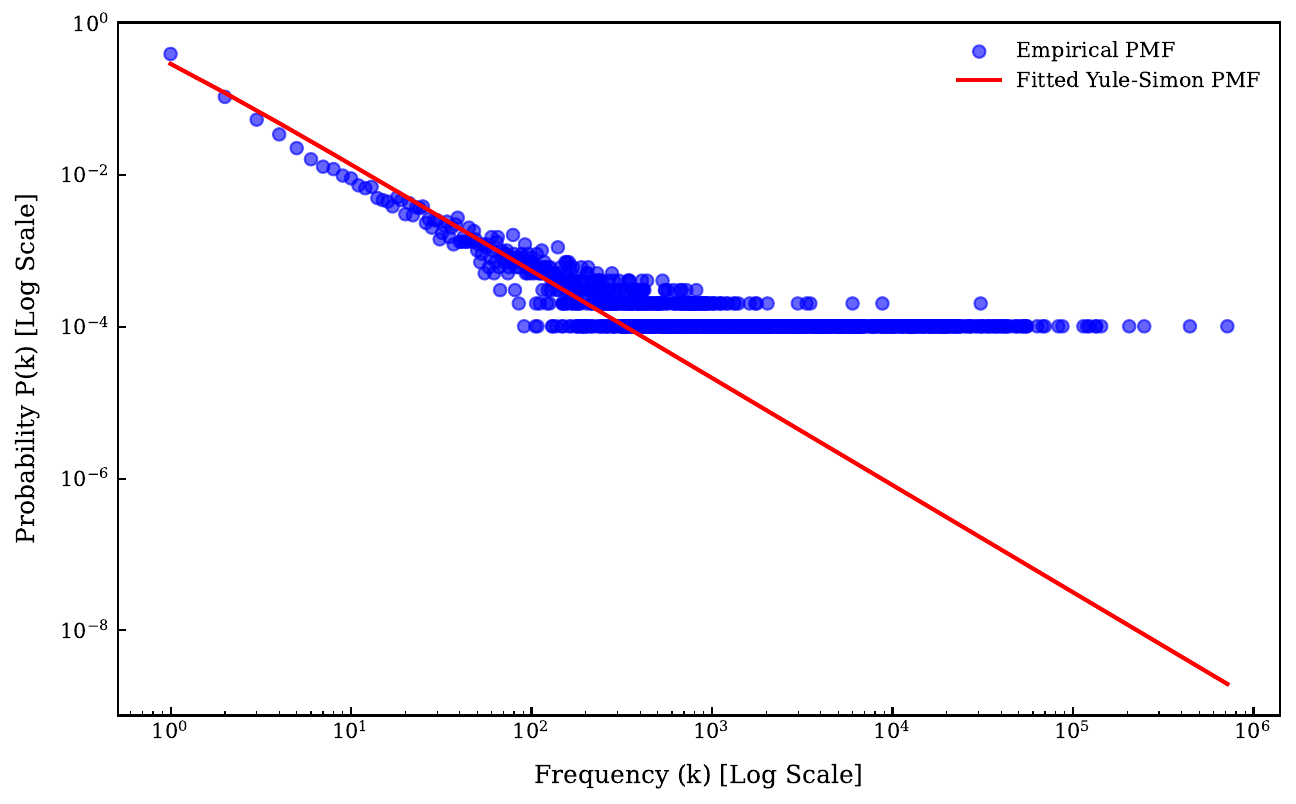}} &
        \subcaptionbox{\tiny Chinese (zh)}{\includegraphics[width=0.15\textwidth]{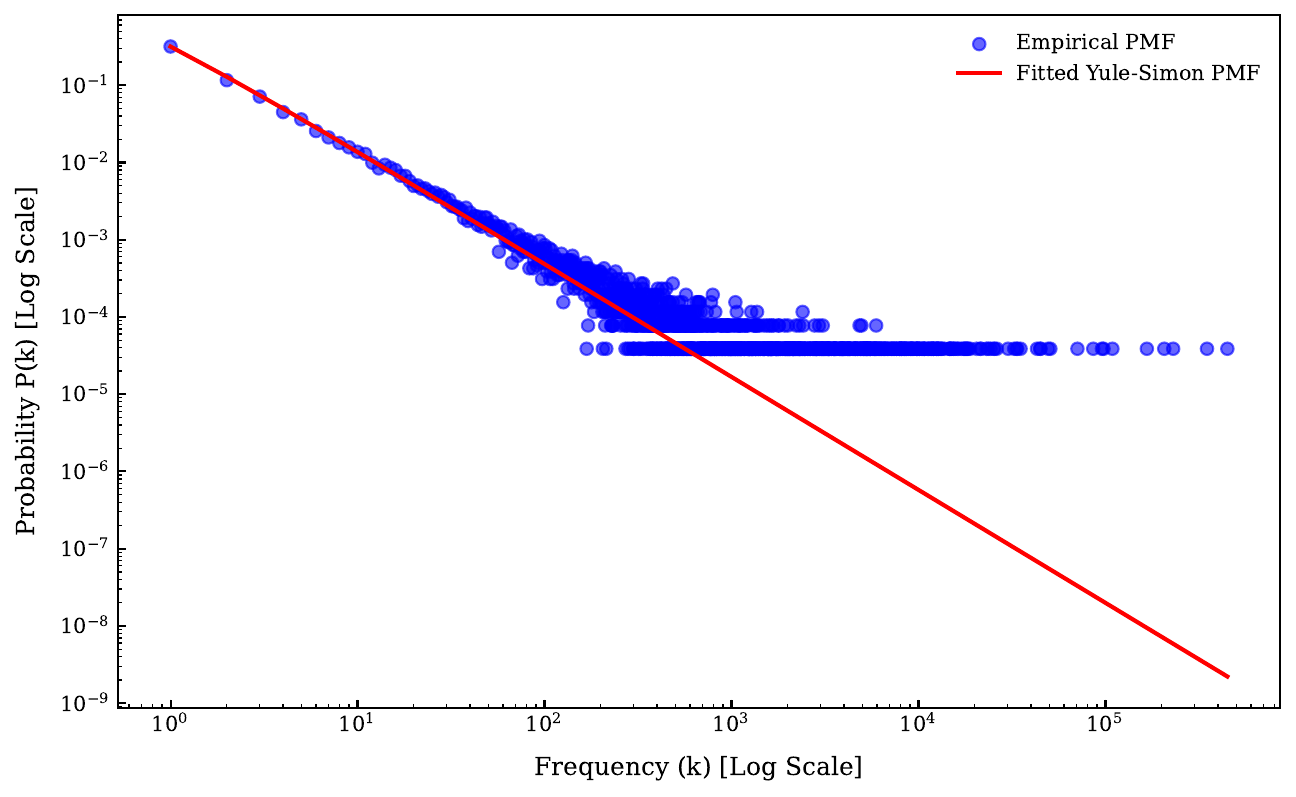}} &
    \end{tabular}
    \caption{Log-log fits on COCO for various languages (Simon model, $n=1$). The horizontal shift in the English frequencies is likely caused by duplicate unfiltered captions in the empirical distribution.}
    \label{fig:locale_fits_coco}
\end{figure}

\begin{figure}
    \centering
    \tiny
    \setlength{\tabcolsep}{1pt} %
    \renewcommand{\arraystretch}{1.2} %
    \captionsetup[subfigure]{labelfont=scriptsize, textfont=scriptsize, skip=3pt} %
    \begin{tabular}{ccc}
        \subcaptionbox{\tiny Chameleon-512 (1-gram)\vspace{2em}}{\includegraphics[width=0.3\textwidth]{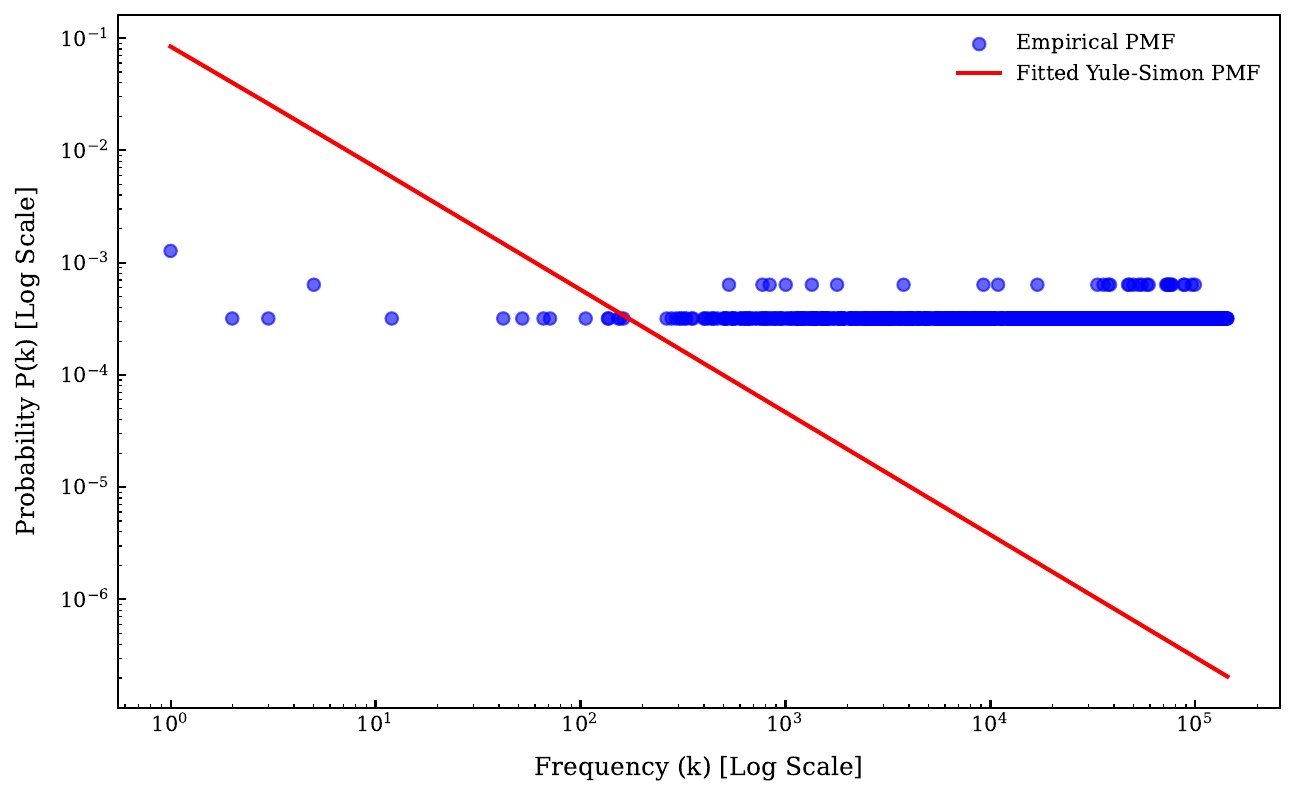}} &
        \subcaptionbox{\tiny Chameleon-512 (2-gram)}{\includegraphics[width=0.3\textwidth]{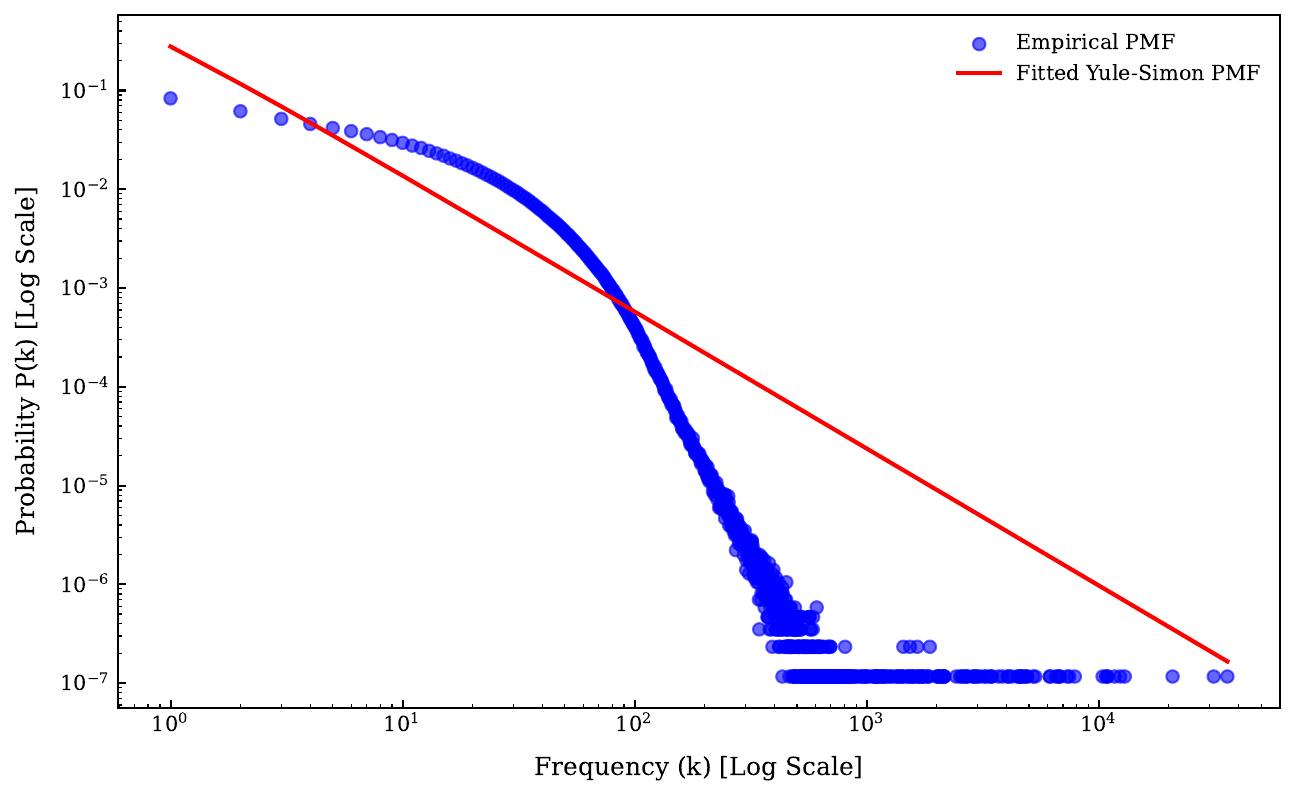}} &
        \subcaptionbox{\tiny Chameleon-512 (3-gram)}{\includegraphics[width=0.3\textwidth]{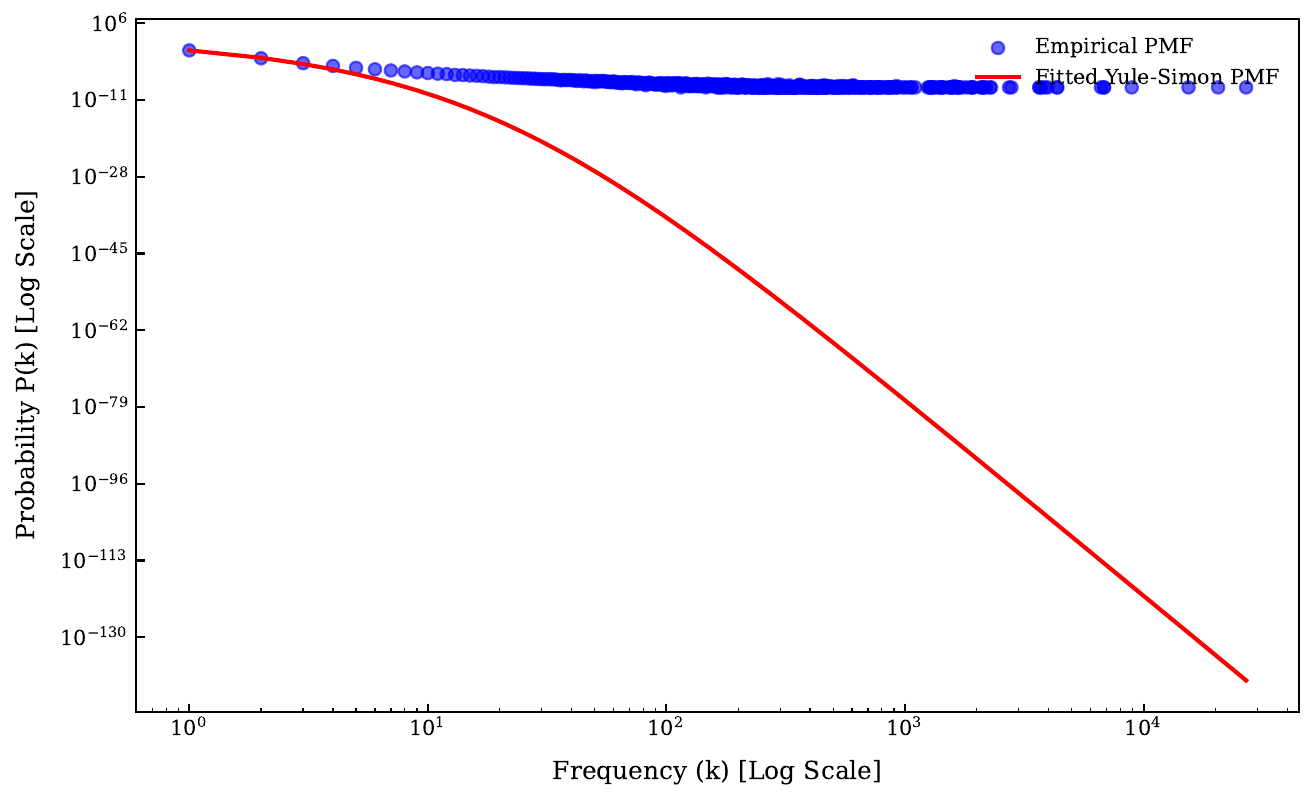}} \\
        
        \subcaptionbox{\tiny VQ-F8-64 (1-gram)\vspace{2em}}{\includegraphics[width=0.3\textwidth]{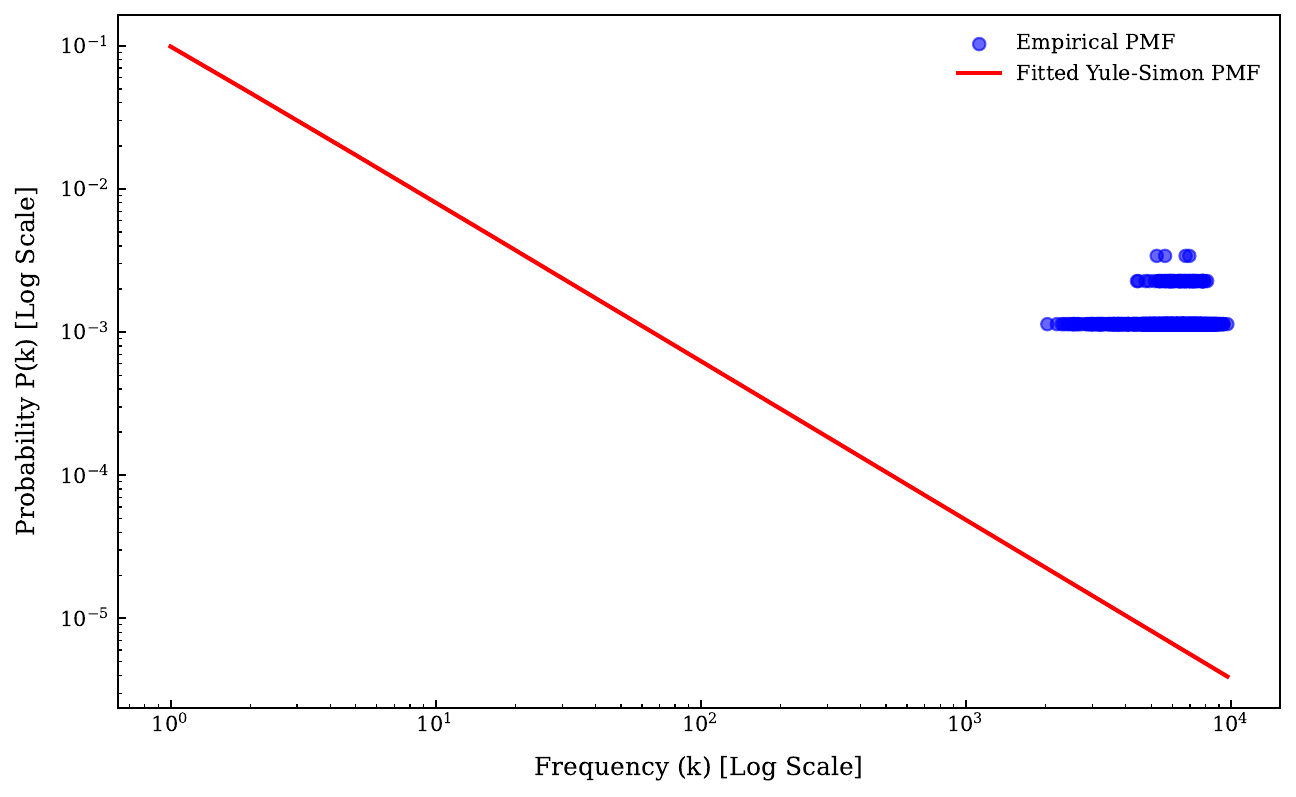}} &
        \subcaptionbox{\tiny VQ-F8-64 (2-gram)}{\includegraphics[width=0.3\textwidth]{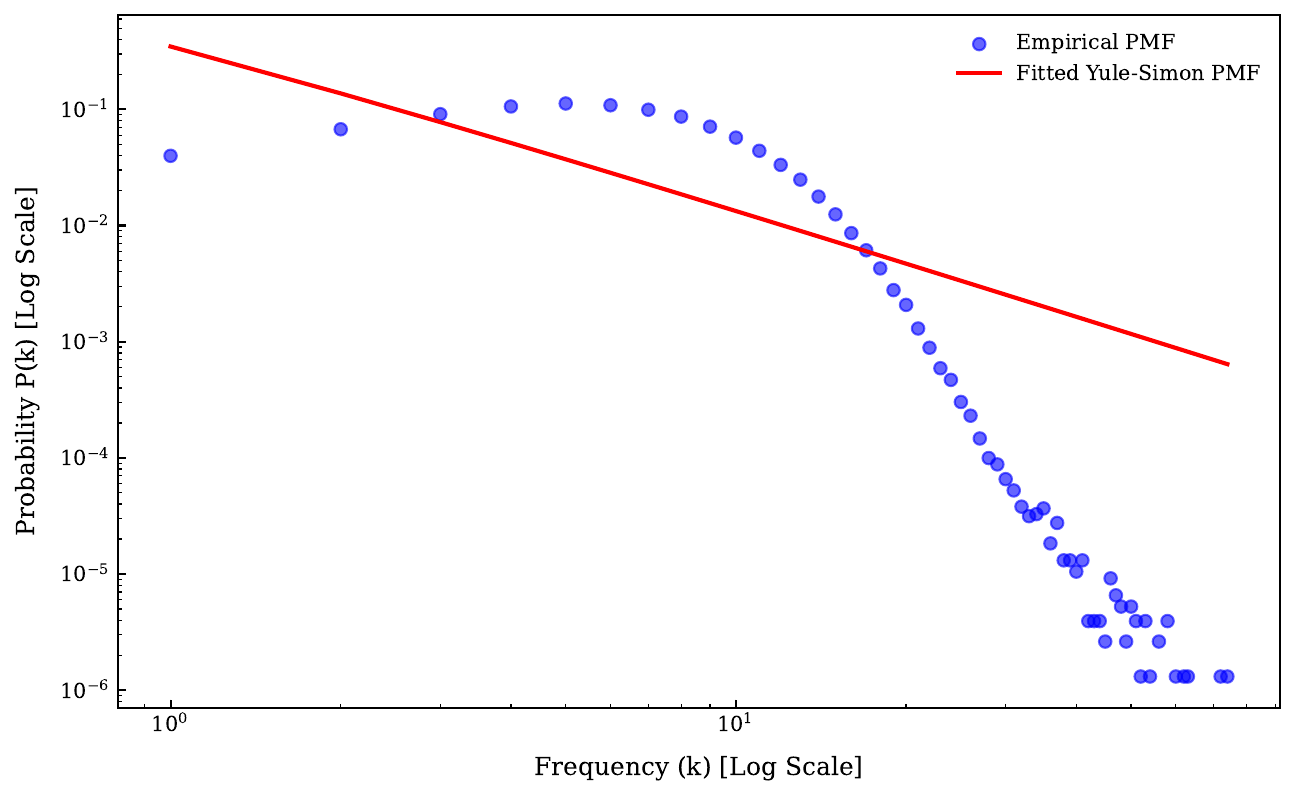}} &
        \subcaptionbox{\tiny VQ-F8-64 (3-gram)}{\includegraphics[width=0.3\textwidth]{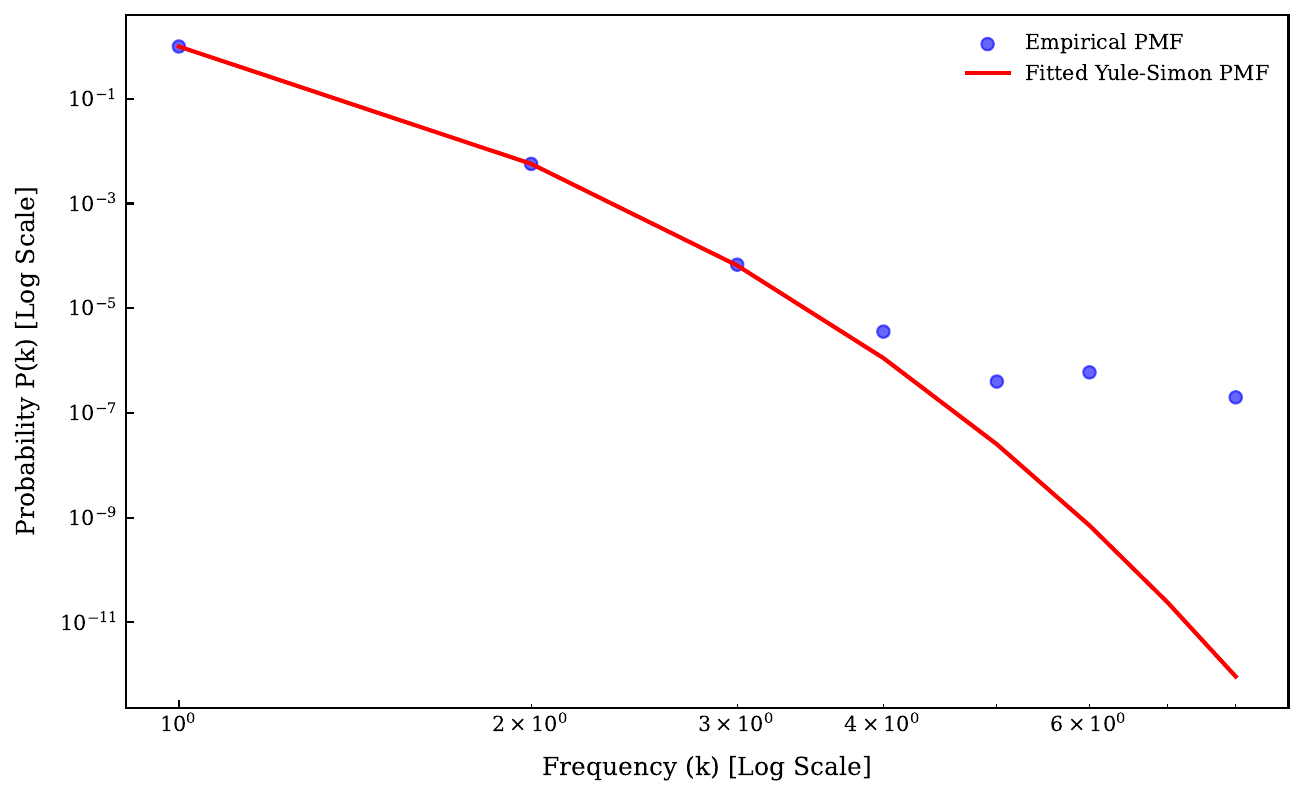}} \\
        
        \subcaptionbox{\tiny VQ-F8-256 (1-gram)\vspace{2em}}{\includegraphics[width=0.3\textwidth]{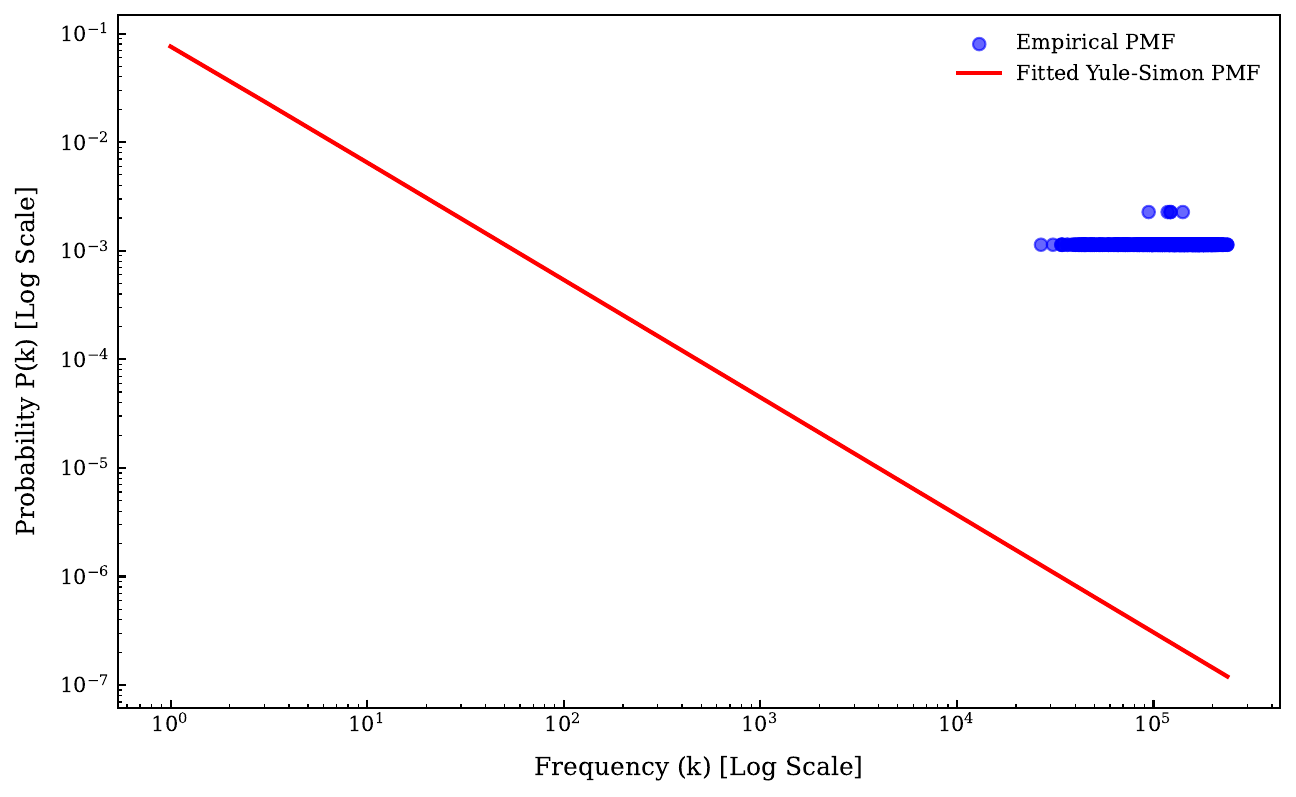}} &
        \subcaptionbox{\tiny VQ-F8-256 (2-gram)}{\includegraphics[width=0.3\textwidth]{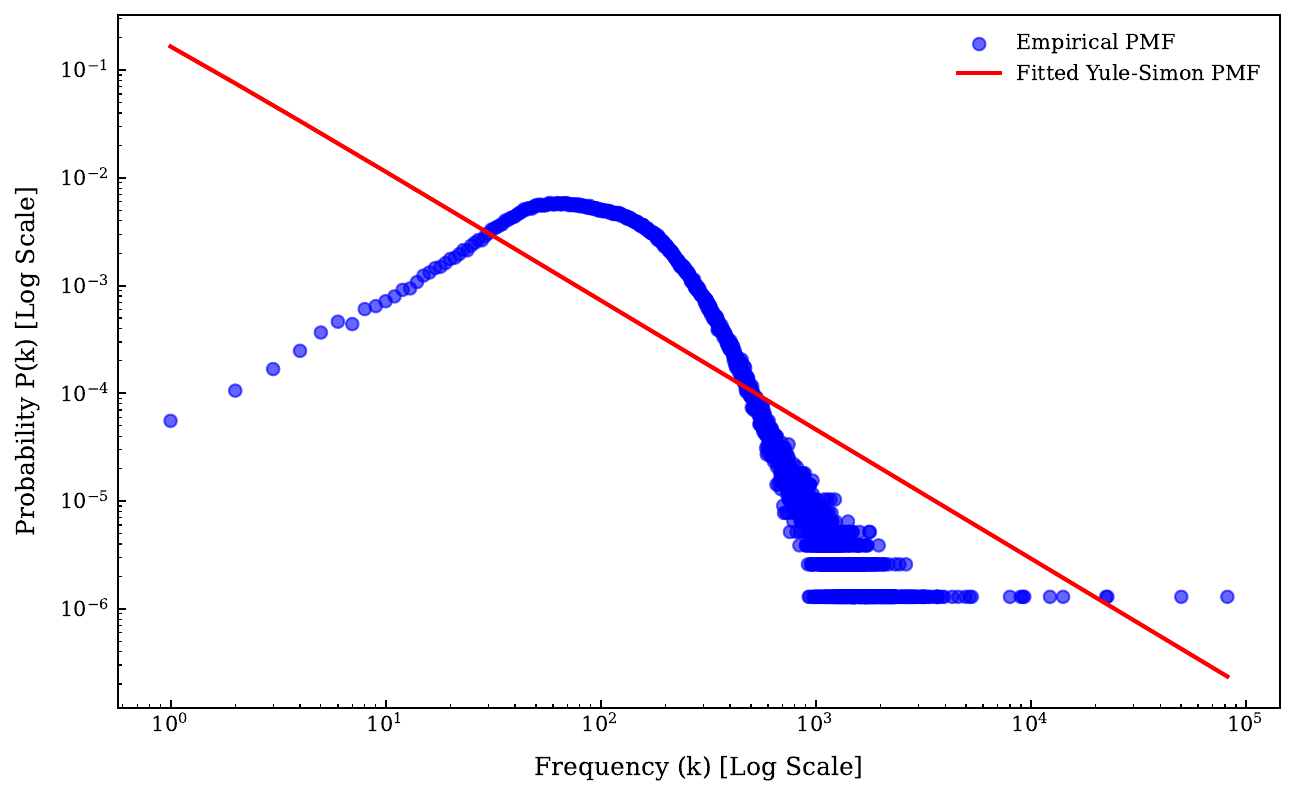}} &
        \subcaptionbox{\tiny VQ-F8-256 (3-gram)}{\includegraphics[width=0.3\textwidth]{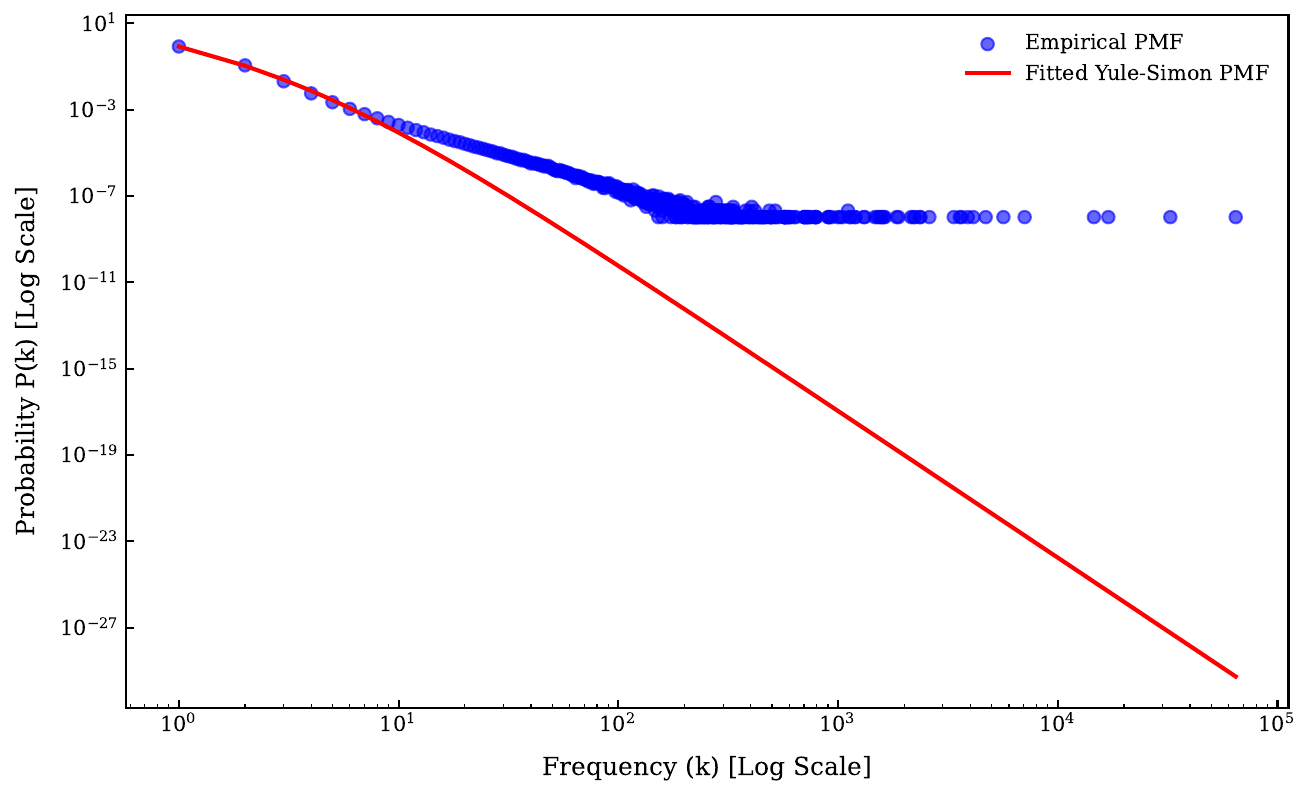}} \\
        
        \subcaptionbox{\tiny VQ-Imagenet-F16-1024-256 (1-gram)\vspace{2em}}{\includegraphics[width=0.3\textwidth]{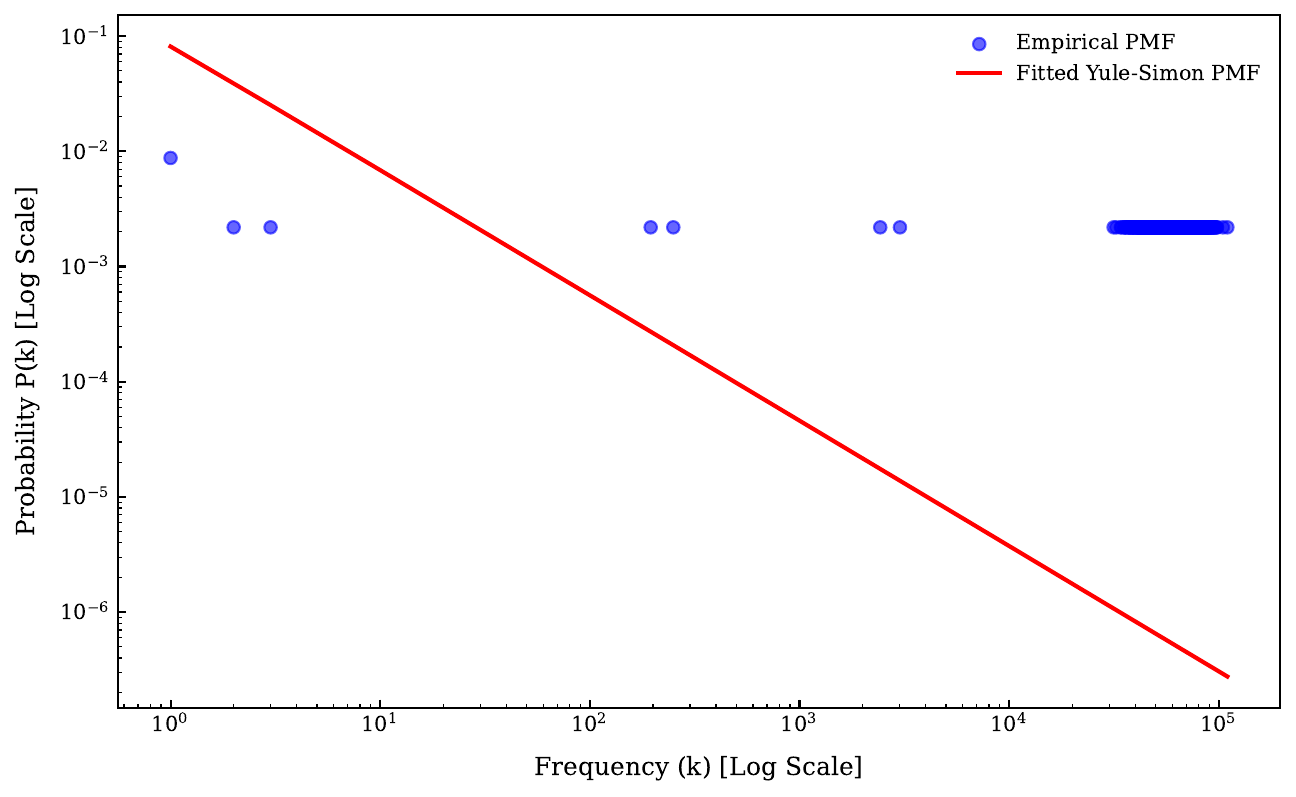}} &
        \subcaptionbox{\tiny VQ-Imagenet-F16-1024-256 (2-gram)}{\includegraphics[width=0.3\textwidth]{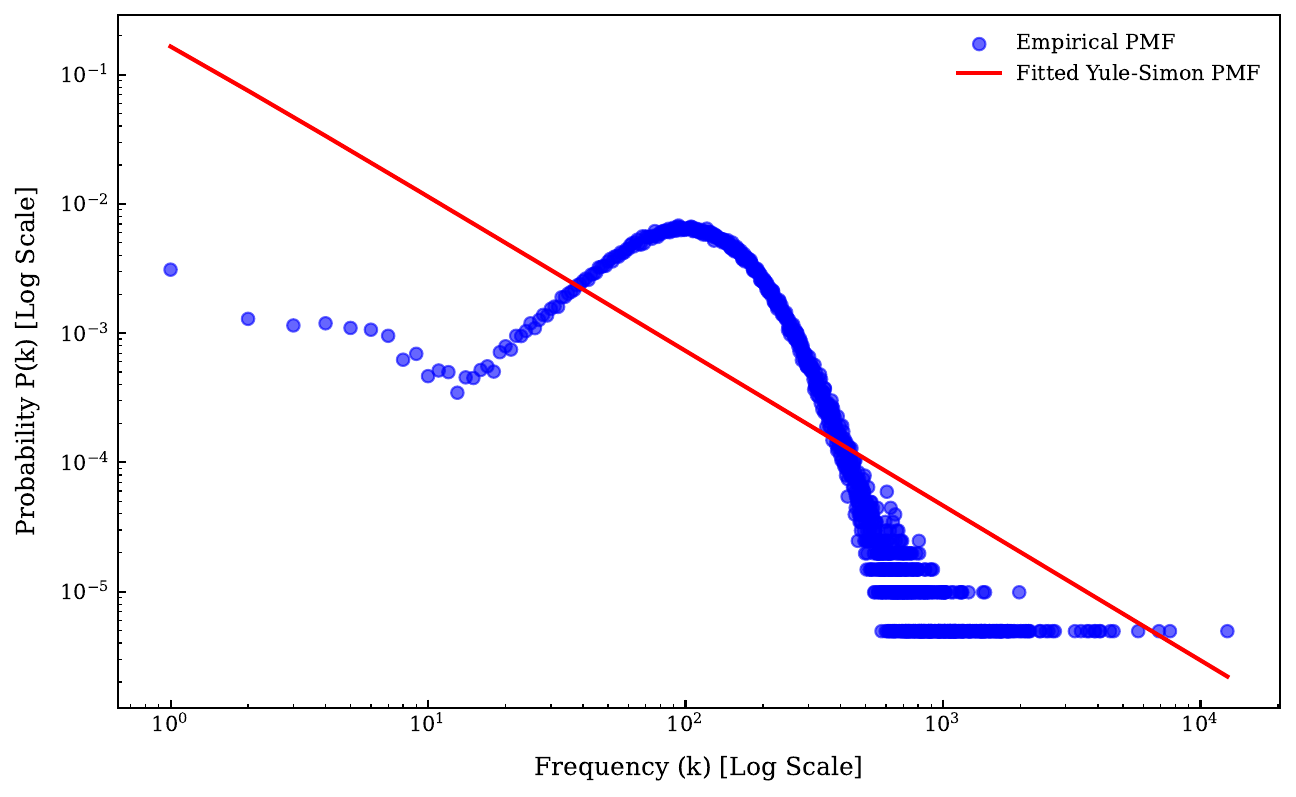}} &
        \subcaptionbox{\tiny VQ-Imagenet-F16-1024-256 (3-gram)}{\includegraphics[width=0.3\textwidth]{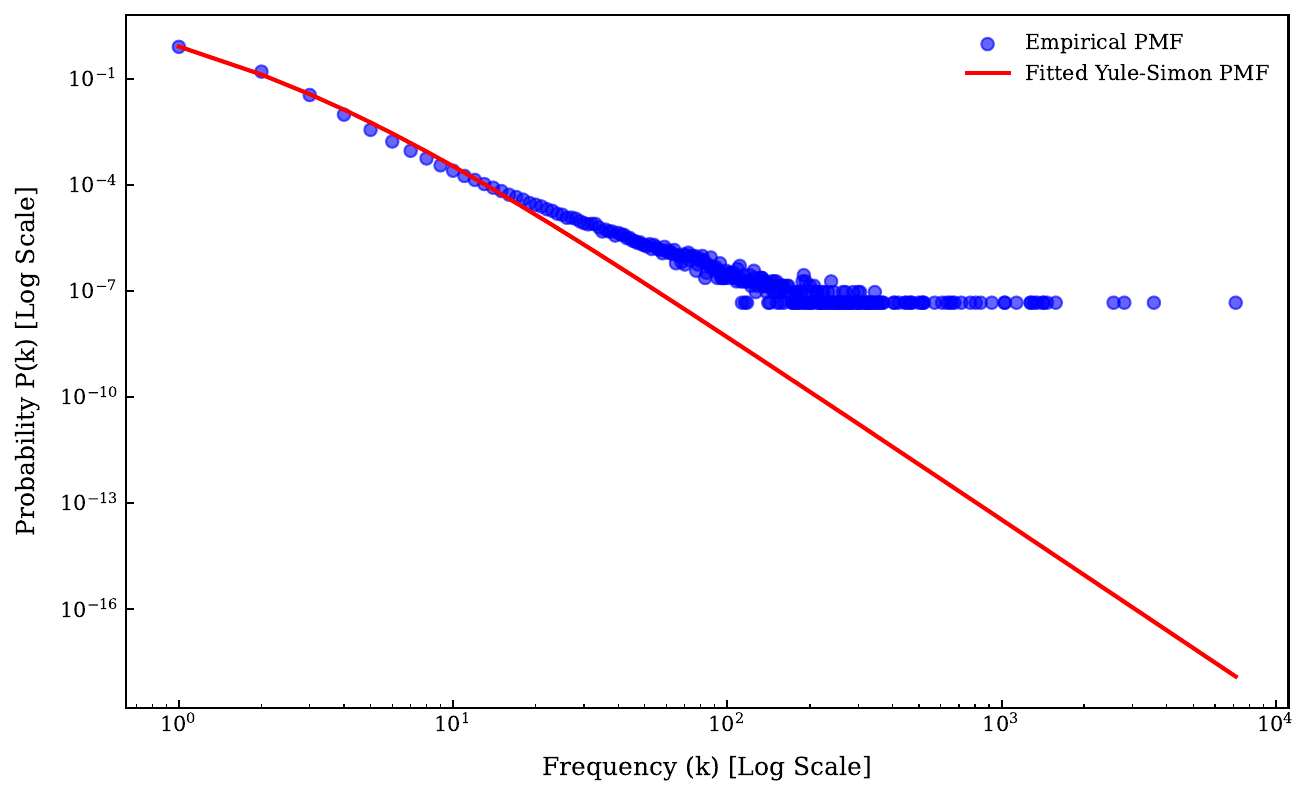}} \\
        
        \subcaptionbox{\tiny LlamaGen-VQ-DS16-C2I (1-gram)\vspace{2em}}{\includegraphics[width=0.3\textwidth]{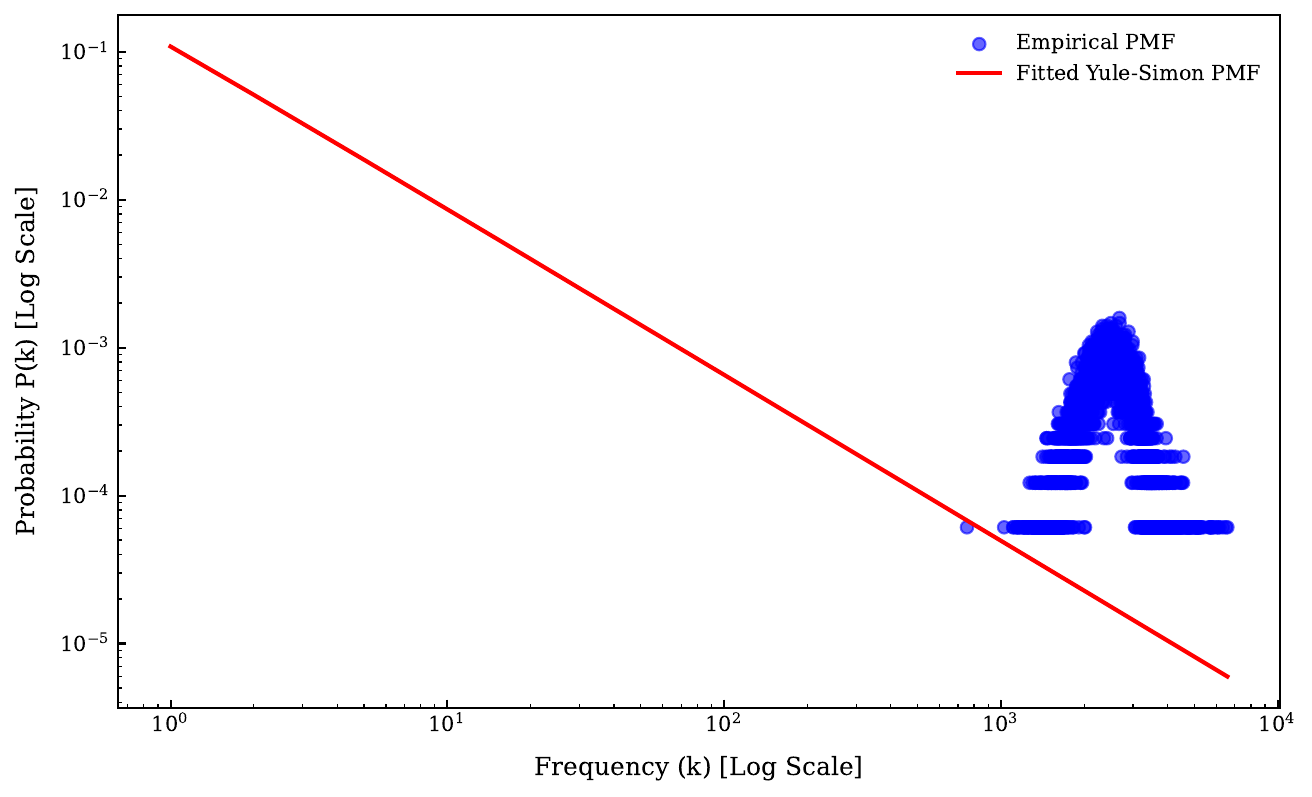}} &
        \subcaptionbox{\tiny LlamaGen-VQ-DS16-C2I (2-gram)}{\includegraphics[width=0.3\textwidth]{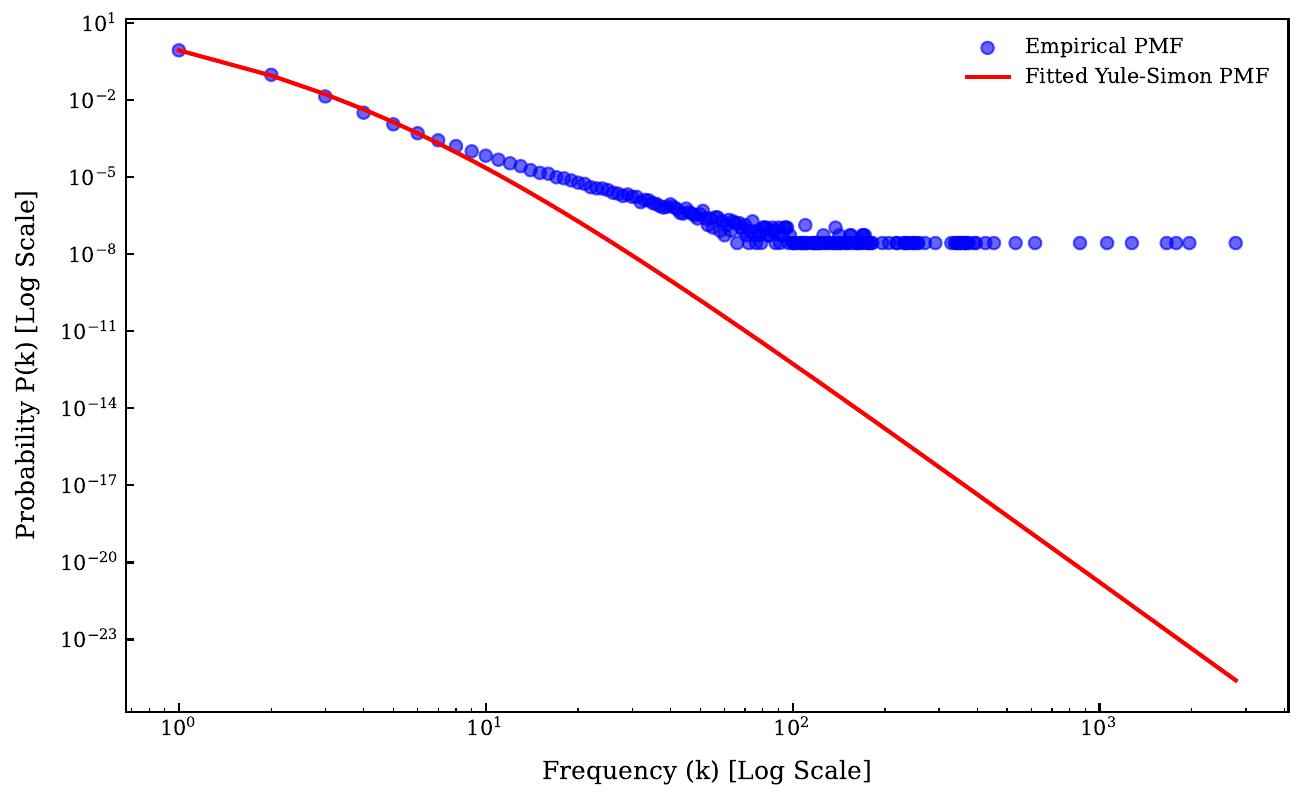}} &
        \subcaptionbox{\tiny LlamaGen-VQ-DS16-C2I (3-gram)}{\textcolor{red}{***** Convergence Error *****}} \\
    \end{tabular}
    \caption{N-gram analysis on various models (Simon model on XM-3600)}
    \label{fig:model_fits_coco_ngram}
\end{figure}

The full experimental results on text data for the XM-3600 dataset are shown in \autoref{fig:locale_fits_xm3600}, with model data shown in \autoref{fig:model_fits_xm3600_ngram}. The text results for COCO are shown in \autoref{fig:locale_fits_coco}, with COCO model convergence shown in \autoref{fig:model_fits_coco_ngram}.


\section{Benefords Law}
\label{app:benfords}

Benford's Law \citep{benford1938law} describes the distribution of leading digits in many naturally occurring datasets, where smaller digits are more likely to appear as the first digit. Specifically, the probability $P(d)$ of a digit $d$ (where $d$ is between 1 and 9) being the leading digit is given by:

\begin{equation}
P(d) = \log_{10} \left( 1 + \frac{1}{d} \right)
\end{equation}

According to this law, the number 1 appears as the first digit around 30\% of the time, while larger digits like 9 appear less frequently, around 5\% of the time. 

For each dataset and tokenization configuration, we extract n-grams (with n = 1, 2, and 3) from tokenized text and image data. We aggregate the token frequencies by computing the distribution of the first digits of these counts. Specifically, the first digits of each token frequency are extracted, and their occurrences are counted to form a first-digit distribution. In cases where natural language data is available, we also compute aggregate distributions across multiple locales for text-based tokenizations. The aggregated text distributions include the mean, standard deviation, minimum, and maximum values for each first-digit count across different locales.

\begin{figure}
    \centering
    \tiny
    \setlength{\tabcolsep}{1pt} %
    \renewcommand{\arraystretch}{1.2} %
    \captionsetup[subfigure]{labelfont=scriptsize, textfont=scriptsize, skip=3pt} %
    \begin{tabular}{cccc}
        \subcaptionbox{\tiny XM3600 (N=1)\vspace{1em}}{\includegraphics[width=0.23\textwidth]{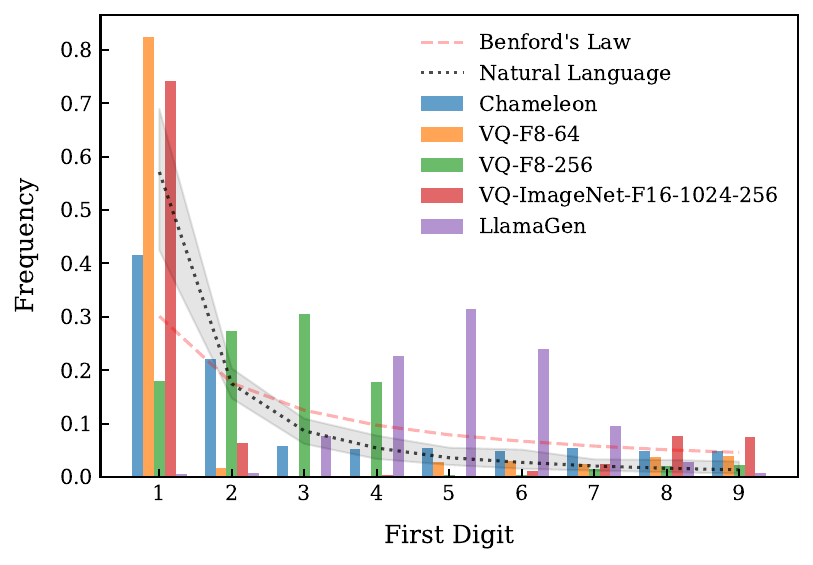}} &
        \subcaptionbox{\tiny XM3600 (N=2)}{\includegraphics[width=0.23\textwidth]{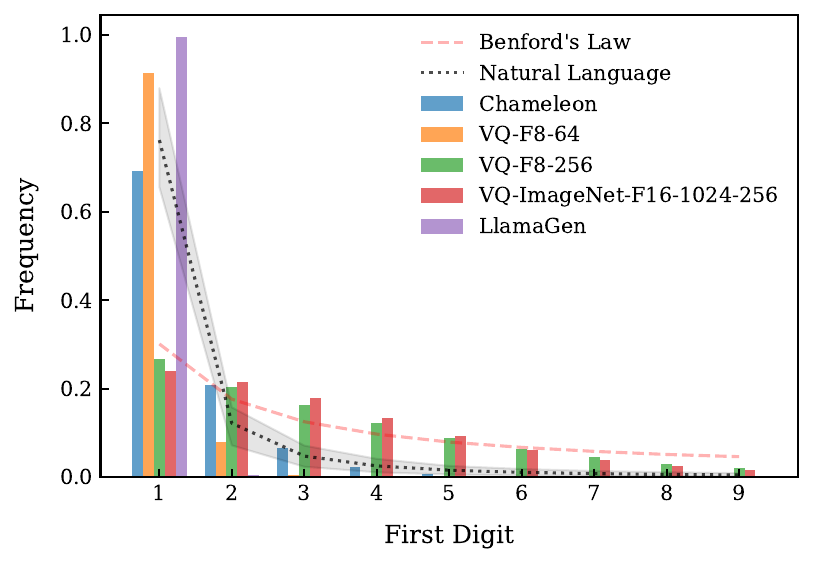}} &
        \subcaptionbox{\tiny XM3600 (N=3)}{\includegraphics[width=0.23\textwidth]{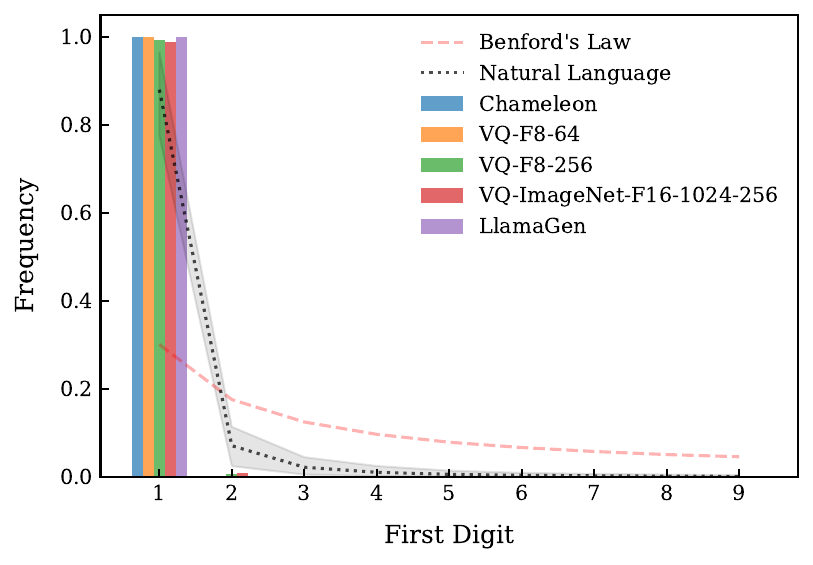}} &
        \subcaptionbox{\tiny XM3600 (N=5)}{\includegraphics[width=0.23\textwidth]{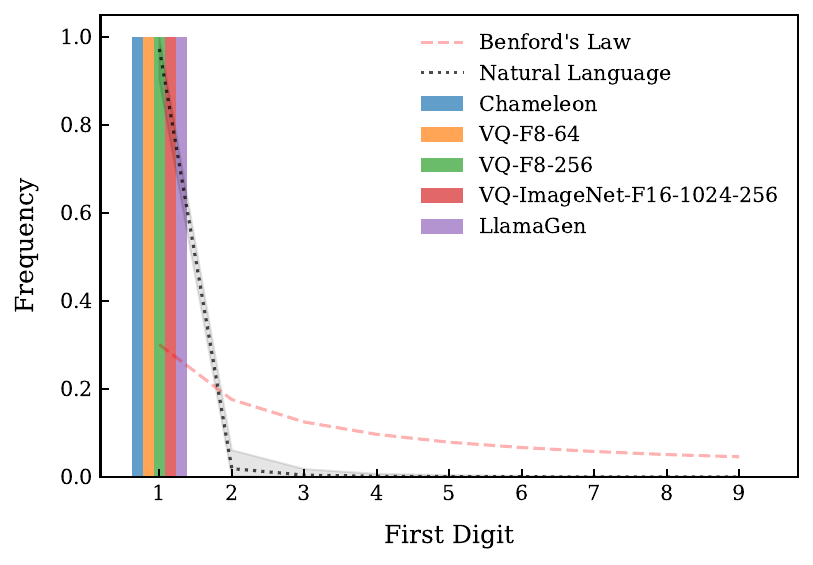}} \\
        
        \subcaptionbox{\tiny CC12M (N=1)\vspace{1em}}{\includegraphics[width=0.23\textwidth]{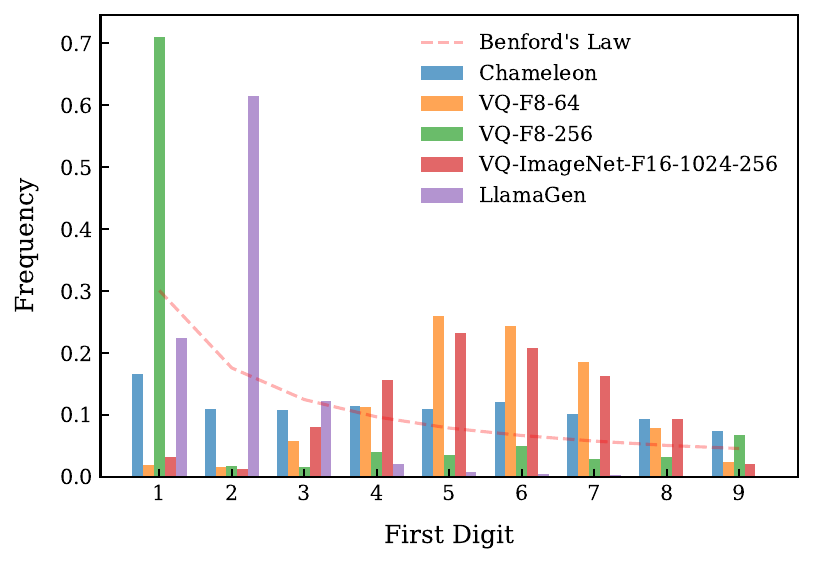}} &
        \subcaptionbox{\tiny CC12M (N=2)}{\includegraphics[width=0.23\textwidth]{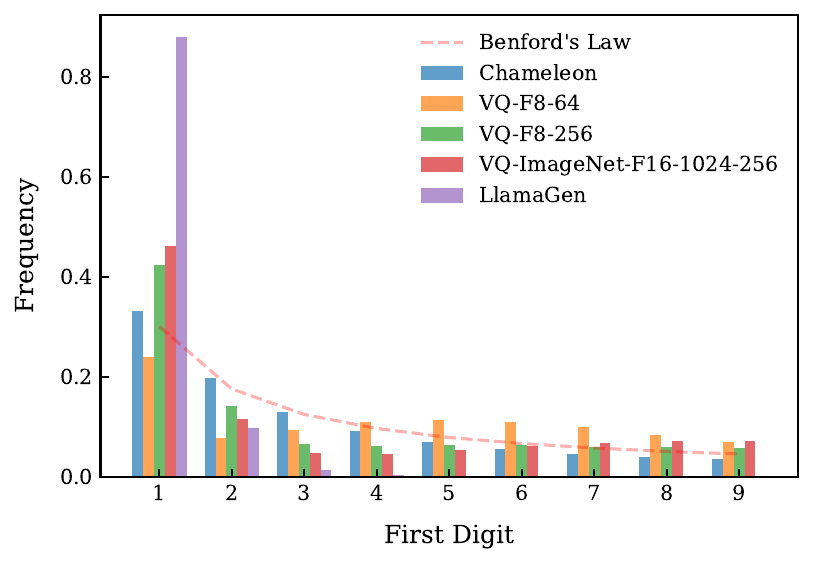}} &
        \subcaptionbox{\tiny CC12M (N=3)}{\includegraphics[width=0.23\textwidth]{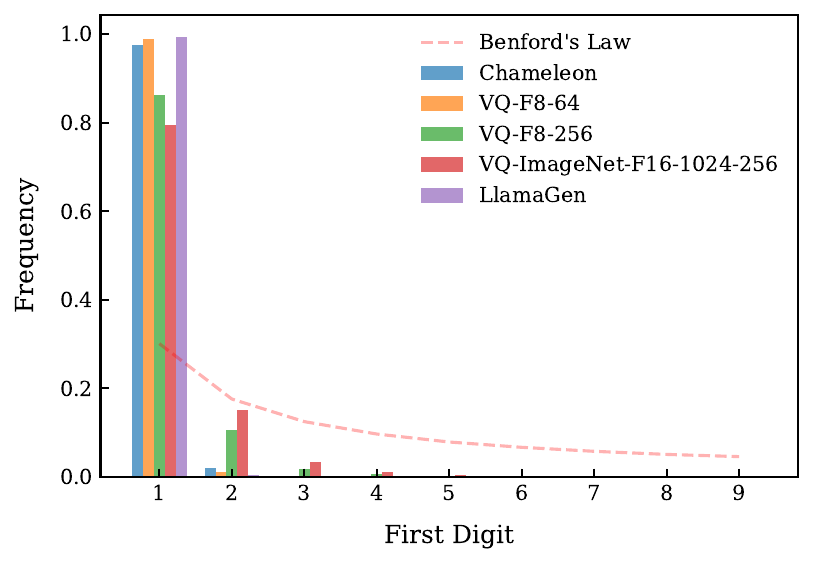}} &
        \subcaptionbox{\tiny CC12M (N=5)}{\includegraphics[width=0.23\textwidth]{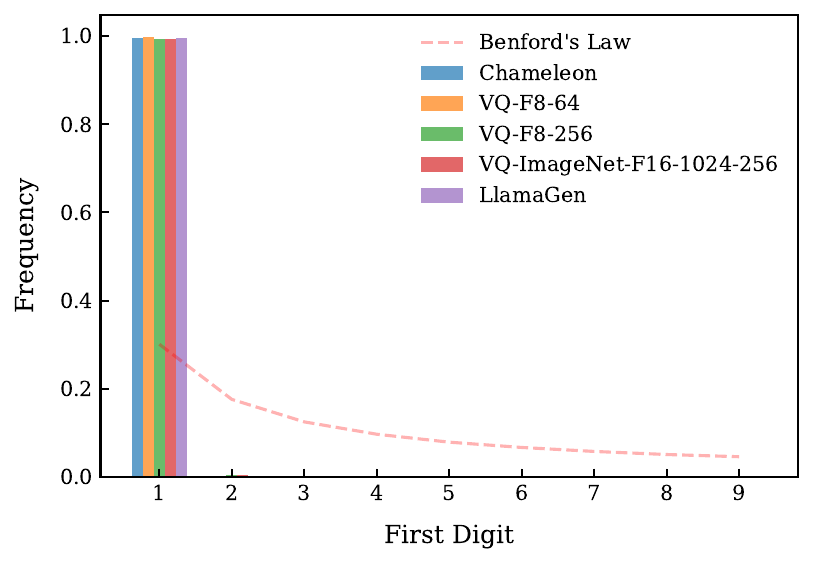}} \\
        
        \subcaptionbox{\tiny COCO (N=1)\vspace{1em}}{\includegraphics[width=0.23\textwidth]{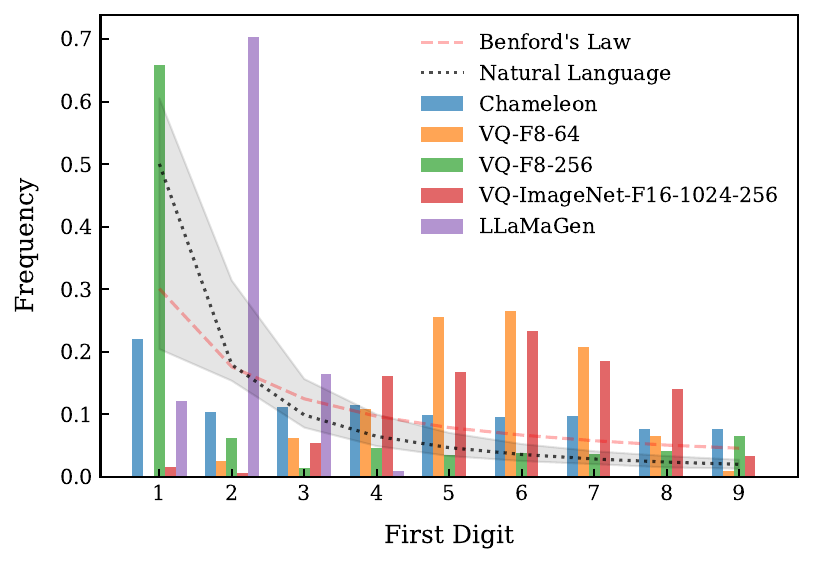}} &
        \subcaptionbox{\tiny COCO (N=2)}{\includegraphics[width=0.23\textwidth]{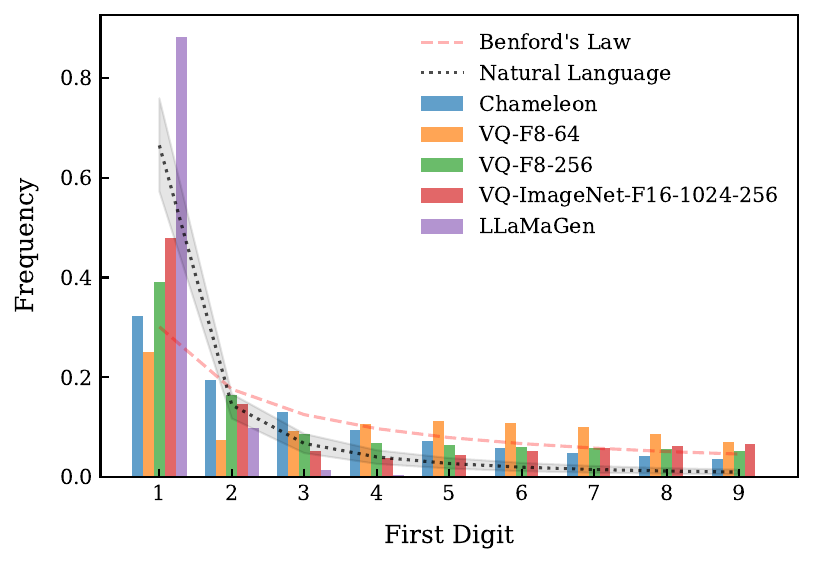}} &
        \subcaptionbox{\tiny COCO (N=3)}{\includegraphics[width=0.23\textwidth]{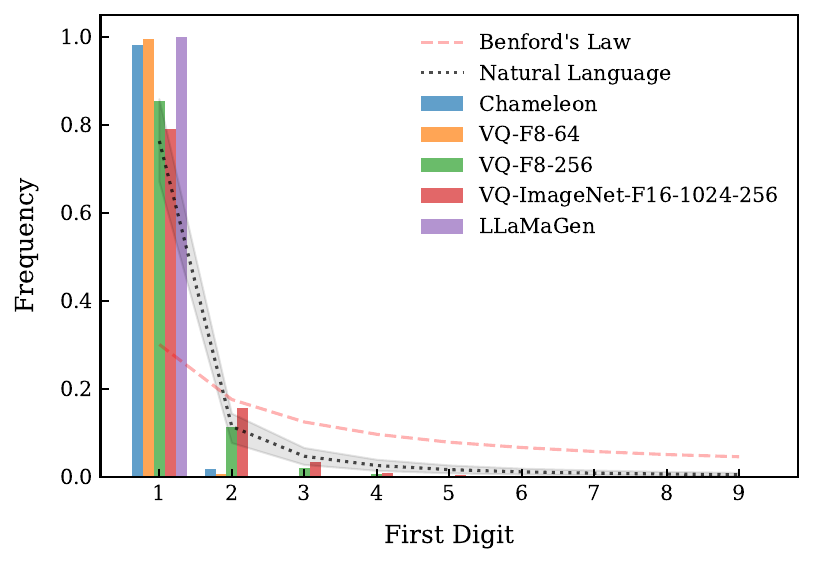}} &
        \subcaptionbox{\tiny COCO (N=5)}{\includegraphics[width=0.23\textwidth]{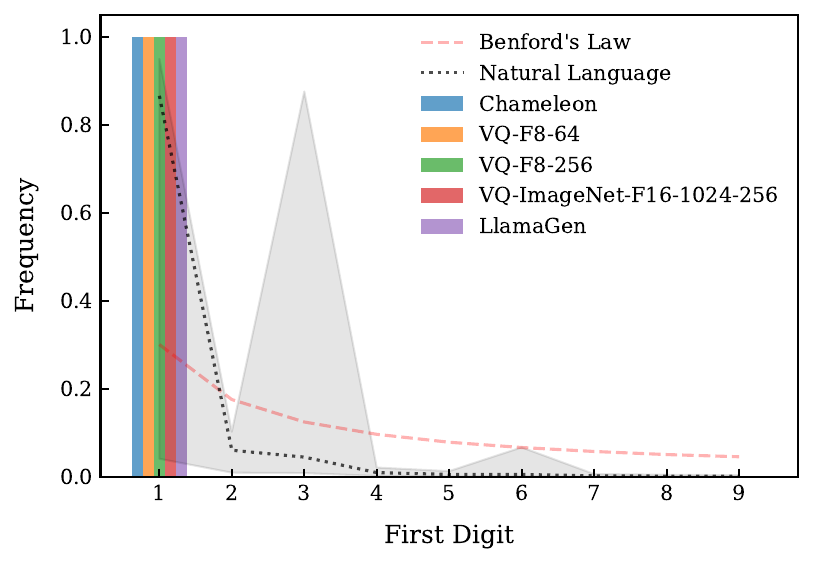}} \\
        
        \subcaptionbox{\tiny ILSVRC (N=1)\vspace{1em}}{\includegraphics[width=0.23\textwidth]{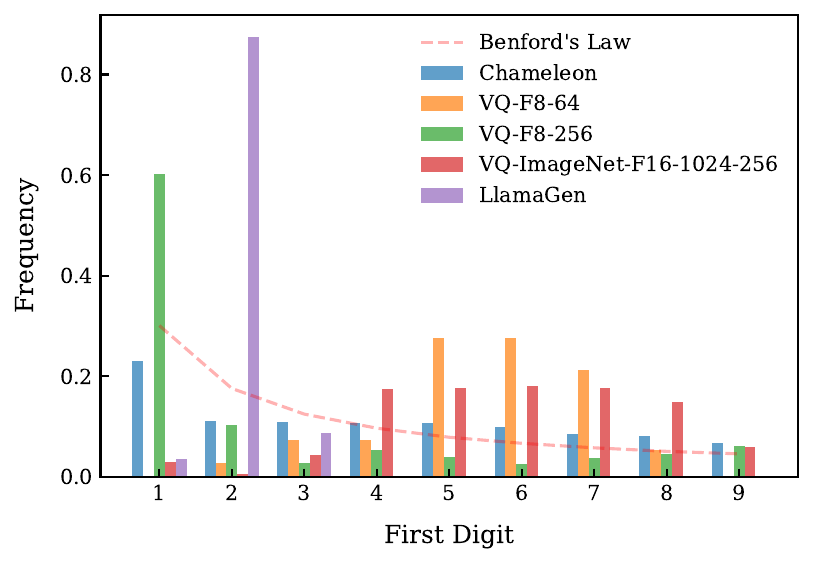}} &
        \subcaptionbox{\tiny ILSVRC (N=2)}{\includegraphics[width=0.23\textwidth]{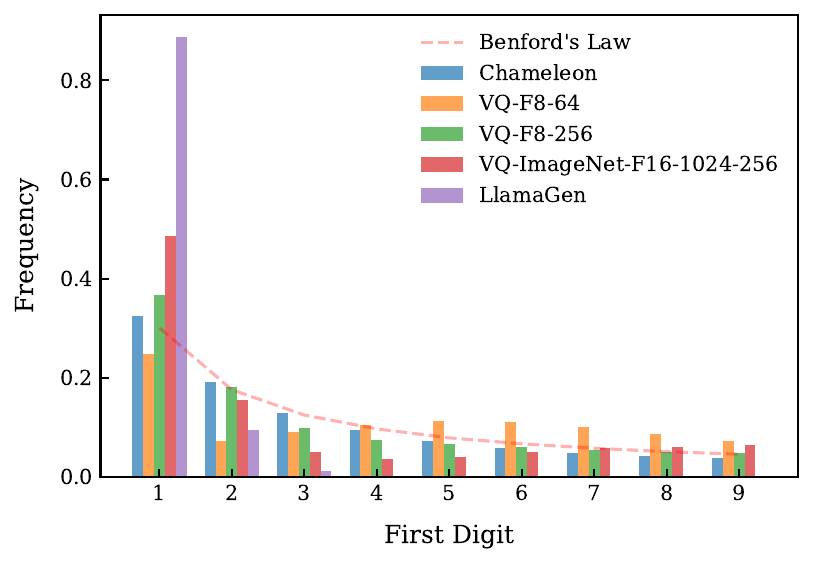}} &
        \subcaptionbox{\tiny ILSVRC (N=3)}{\includegraphics[width=0.23\textwidth]{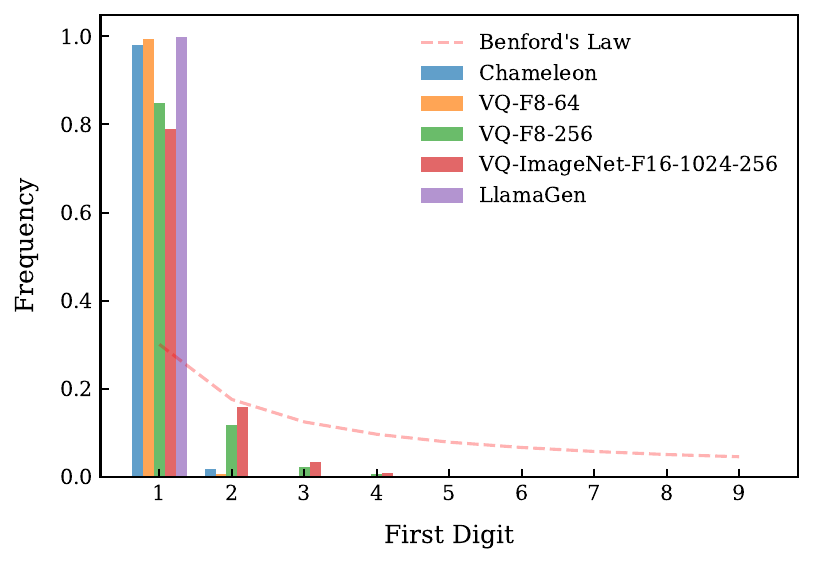}} &
        \subcaptionbox{\tiny ILSVRC (N=5)}{\includegraphics[width=0.23\textwidth]{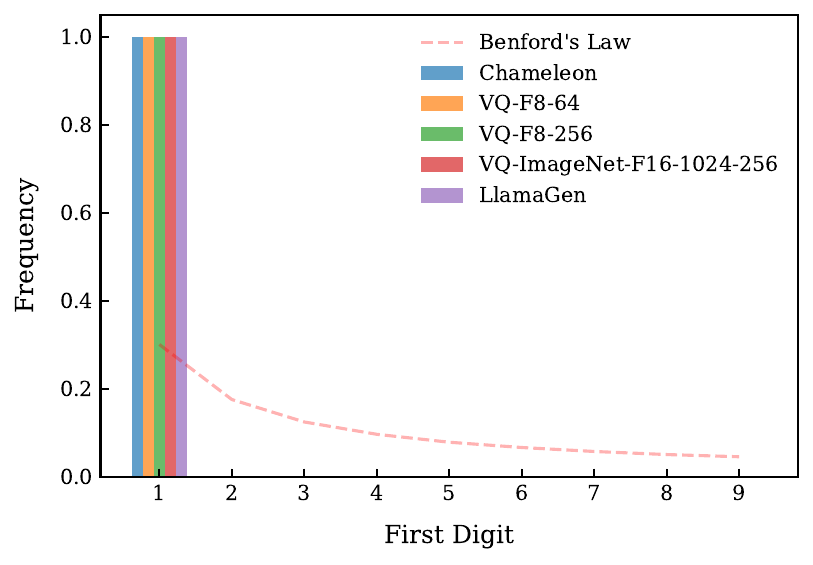}} \\
        
        \subcaptionbox{\tiny SPIN (N=1)\vspace{1em}}{\includegraphics[width=0.23\textwidth]{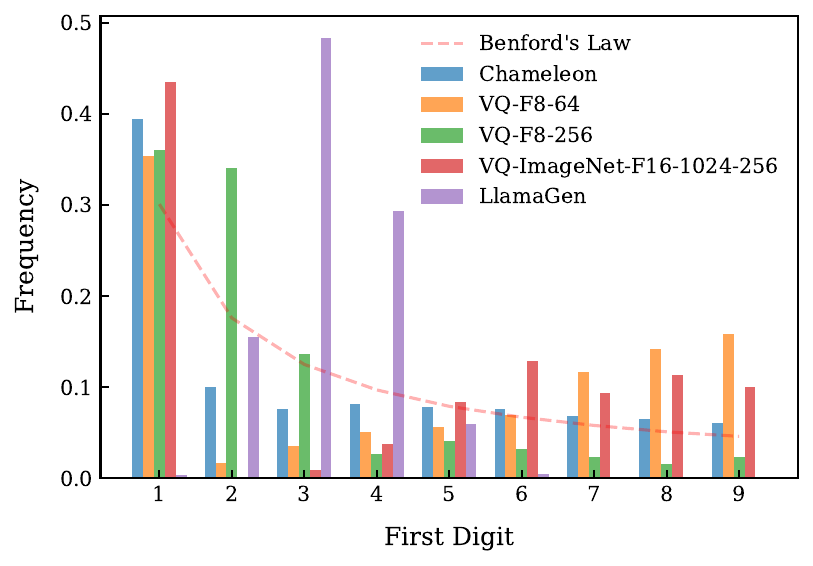}} &
        \subcaptionbox{\tiny SPIN (N=2)}{\includegraphics[width=0.23\textwidth]{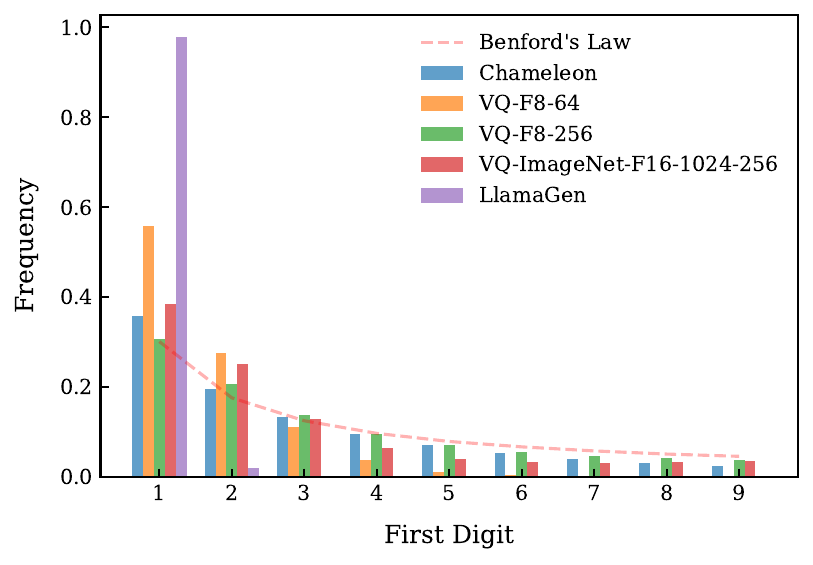}} &
        \subcaptionbox{\tiny SPIN (N=3)}{\includegraphics[width=0.23\textwidth]{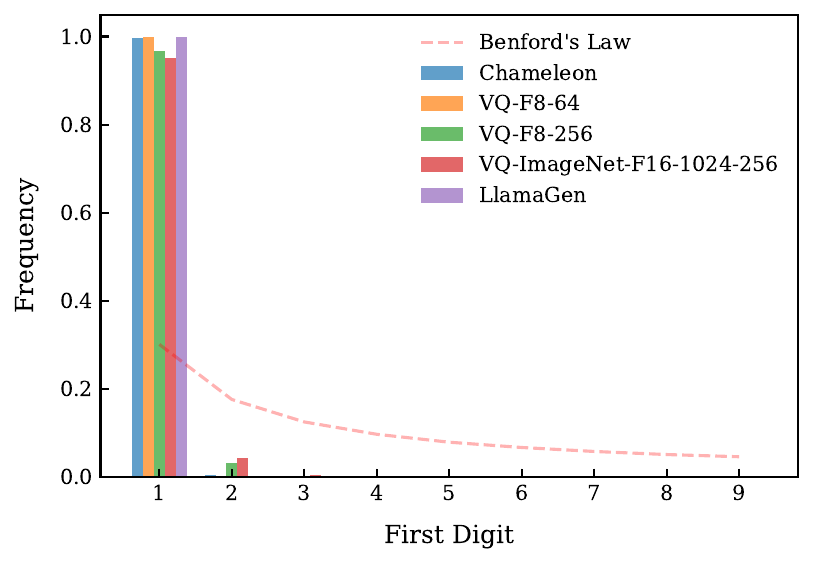}} &
        \subcaptionbox{\tiny SPIN (N=5)}{\includegraphics[width=0.23\textwidth]{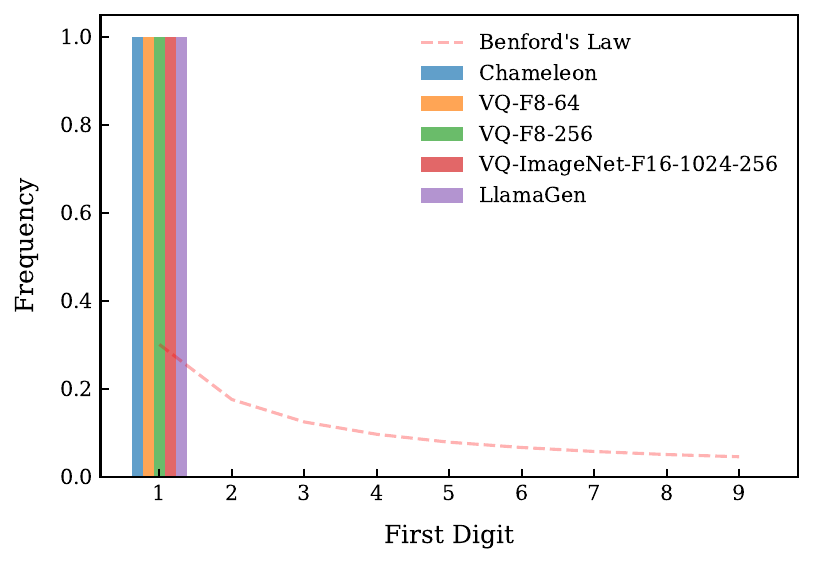}} \\
    \end{tabular}
    \caption{Benford's Law on XM-3600, CC12M, COCO, ILSVRC, and SPIN.}
    \label{fig:benford_plots}
\end{figure}

The full results for each of the datasets (XM-3600, CC12M, COCO, ILSVRC and SPIN) is given in \autoref{fig:benford_plots}.

\section{Huffman Encoding / Entropy}
\label{app:huffman}

Huffman encoding is a widely-used algorithm for lossless data compression, which assigns variable-length codes to tokens based on their frequencies. The core idea is to minimize the total number of bits required to represent the token stream by assigning shorter codes to more frequent tokens and longer codes to less frequent ones. This is achieved by constructing a binary tree where each token is a leaf, and its depth (or code length) corresponds to its frequency. The encoding process ensures that the total number of bits, $ L_{\text{Huffman}} $, needed to encode a stream of tokens is reduced compared to fixed-length encoding, where each token would require $\lceil \log_2(n) \rceil$ bits, with $n$ being the number of unique tokens.

Entropy, denoted as $H(X)$, represents the theoretical limit on the average number of bits needed to encode the token stream, and is calculated using Shannon’s entropy formula:

\begin{equation}
\label{eq:entropy}
H(X) = - \sum_{x \in X} P(x) \log_2 P(x)
\end{equation}

where $P(x)$ is the empirical probability of token $x$ in the stream. In this experiment, entropy serves as a benchmark for comparing the performance of Huffman encoding. The closer the average code length of the Huffman encoding is to the entropy, the more efficient the compression. By evaluating the compression rate and percentage reduction, we can quantify how effectively Huffman encoding reduces the bit length compared to the fixed-length encoding, with the goal of approaching the entropy limit.

\subsection{Experimental Design}

In our experiments, for each dataset/tokenizer combination (both $N=1$ (unigrams) and $N=2$ (bigrams)), we extract the first 500,000 tokens as a token stream, and then apply Huffman encoding to compress this token stream. We extract several key metrics from the resulting encoding:
\begin{itemize}
    \item Average Code Length: The weighted average of the lengths of Huffman codes for all tokens.
    \item Entropy: The theoretical minimum average code length for the specified token distribution (See \autoref{eq:entropy}).
    \item Fixed Code Length: The length of the fixed-length codes used for comparison.
    \item Original Bits: The number of bits required for fixed-length encoding of the token stream.
    \item Huffman bits: The number of bits required after applying Huffman encoding.
    \item Compression Rate: The ratio of the original bits to the Huffman bits.
    \item Percentage Reduction: The percent reduction in the total number of bits after applying Huffman encoding.
\end{itemize}

\subsection{Further Experimental Results}

The full experimental results for the Huffman coding experiment are given in \autoref{tab:huffman_full}. A surprising detailed result is that the chameleon tokenizer, the most effective of the tokenizers, is also the most compressible representation of them, with almost twice the percentage reduction compared to other models. Llama-gen is the least compressible, and indeed, is almost completely incompressible, suggesting it has very efficient token use but does not contain any repeatable structure. 
\begin{tiny}
\begin{longtable}{l@{\extracolsep{\fill}}lcccccccc}
\caption{Full huffman coding results for N=1 and N=2. ACL:Average Code Length, E: Entropy, FC: Fixed Code Length, OB: Original Bits (MB), HB: Huffman Bits (MB), CR: Compression Rate, PR: Percentage Reduction }\label{tab:huffman_full}\\    %
\toprule
Dataset & Model & N & ACL & E & FCL & OB & HB & CR & PR \\
\midrule
\endfirsthead
\toprule
Dataset & Model & N & ACL & E & FCL & OB & HB & CR & PR \\
\midrule
\endhead
\midrule
\multicolumn{10}{r}{Continued on next page} \\
\midrule
\endfoot
\bottomrule
\endlastfoot
coco & chameleon-512 & 1 & 11.30 & 11.27 & 12 & 6.00 & 5.65 & 1.06 & 5.86 \\
coco & compvis-vq-f8-64 & 1 & 9.78 & 9.74 & 10 & 5.00 & 4.89 & 1.02 & 2.25 \\
coco & compvis-vq-f8-256 & 1 & 9.71 & 9.67 & 10 & 5.00 & 4.85 & 1.03 & 2.93 \\
coco & compvis-vq-imagenet-f16-1024-256 & 1 & 8.80 & 8.76 & 9 & 4.50 & 4.40 & 1.02 & 2.22 \\
coco & llamagen-vq-ds16-c2i & 1 & 13.98 & 13.95 & 14 & 7.00 & 6.99 & 1.00 & 0.18 \\
coco & text-ar & 1 & 9.85 & 9.82 & 15 & 7.50 & 4.93 & 1.52 & 34.31 \\
coco & text-bn & 1 & 8.91 & 8.88 & 14 & 7.00 & 4.46 & 1.57 & 36.36 \\
coco & text-cs & 1 & 9.80 & 9.77 & 15 & 7.50 & 4.90 & 1.53 & 34.64 \\
coco & text-da & 1 & 8.09 & 8.06 & 14 & 7.00 & 4.05 & 1.73 & 42.21 \\
coco & text-de & 1 & 8.70 & 8.67 & 15 & 7.50 & 4.35 & 1.72 & 41.98 \\
coco & text-el & 1 & 8.55 & 8.53 & 14 & 7.00 & 4.28 & 1.64 & 38.90 \\
coco & text-es & 1 & 7.98 & 7.95 & 14 & 7.00 & 3.99 & 1.75 & 42.99 \\
coco & text-fa & 1 & 8.20 & 8.17 & 13 & 6.50 & 4.10 & 1.59 & 36.94 \\
coco & text-fi & 1 & 10.14 & 10.11 & 16 & 8.00 & 5.07 & 1.58 & 36.65 \\
coco & text-fil & 1 & 7.47 & 7.44 & 14 & 7.00 & 3.74 & 1.87 & 46.63 \\
coco & text-fr & 1 & 8.12 & 8.09 & 14 & 7.00 & 4.06 & 1.72 & 42.02 \\
coco & text-hi & 1 & 8.22 & 8.19 & 14 & 7.00 & 4.11 & 1.70 & 41.31 \\
coco & text-hr & 1 & 9.66 & 9.63 & 15 & 7.50 & 4.83 & 1.55 & 35.58 \\
coco & text-hu & 1 & 8.89 & 8.87 & 15 & 7.50 & 4.45 & 1.69 & 40.70 \\
coco & text-id & 1 & 8.33 & 8.30 & 13 & 6.50 & 4.17 & 1.56 & 35.91 \\
coco & text-it & 1 & 8.33 & 8.30 & 14 & 7.00 & 4.17 & 1.68 & 40.50 \\
coco & text-he & 1 & 9.90 & 9.87 & 15 & 7.50 & 4.95 & 1.51 & 33.98 \\
coco & text-ja & 1 & 7.87 & 7.83 & 14 & 7.00 & 3.93 & 1.78 & 43.81 \\
coco & text-ko & 1 & 8.70 & 8.67 & 14 & 7.00 & 4.35 & 1.61 & 37.87 \\
coco & text-mi & 1 & 6.83 & 6.80 & 13 & 6.50 & 3.42 & 1.90 & 47.45 \\
coco & text-nl & 1 & 7.96 & 7.93 & 14 & 7.00 & 3.98 & 1.76 & 43.11 \\
coco & text-no & 1 & 8.12 & 8.09 & 15 & 7.50 & 4.06 & 1.85 & 45.87 \\
coco & text-pl & 1 & 9.84 & 9.80 & 15 & 7.50 & 4.92 & 1.53 & 34.43 \\
coco & text-pt & 1 & 8.05 & 8.02 & 14 & 7.00 & 4.02 & 1.74 & 42.52 \\
coco & text-ro & 1 & 8.49 & 8.47 & 14 & 7.00 & 4.24 & 1.65 & 39.36 \\
coco & text-ru & 1 & 9.70 & 9.67 & 15 & 7.50 & 4.85 & 1.55 & 35.30 \\
coco & text-sv & 1 & 8.18 & 8.15 & 15 & 7.50 & 4.09 & 1.83 & 45.48 \\
coco & text-sw & 1 & 8.63 & 8.60 & 14 & 7.00 & 4.32 & 1.62 & 38.36 \\
coco & text-te & 1 & 9.67 & 9.64 & 15 & 7.50 & 4.84 & 1.55 & 35.53 \\
coco & text-th & 1 & 8.67 & 8.64 & 13 & 6.50 & 4.33 & 1.50 & 33.31 \\
coco & text-tr & 1 & 9.05 & 9.02 & 15 & 7.50 & 4.53 & 1.66 & 39.64 \\
coco & text-uk & 1 & 9.72 & 9.68 & 15 & 7.50 & 4.86 & 1.54 & 35.23 \\
coco & text-vi & 1 & 8.15 & 8.12 & 12 & 6.00 & 4.08 & 1.47 & 32.07 \\
coco & text-zh & 1 & 8.80 & 8.77 & 14 & 7.00 & 4.40 & 1.59 & 37.15 \\
xm3600 & chameleon-512 & 1 & 11.27 & 11.24 & 12 & 6.00 & 5.63 & 1.06 & 6.09 \\
xm3600 & compvis-vq-f8-64 & 1 & 9.77 & 9.73 & 10 & 1.18 & 1.16 & 1.02 & 2.34 \\
xm3600 & compvis-vq-f8-256 & 1 & 9.69 & 9.66 & 10 & 5.00 & 4.85 & 1.03 & 3.05 \\
xm3600 & compvis-vq-imagenet-f16-1024-256 & 1 & 8.79 & 8.75 & 9 & 4.50 & 4.39 & 1.02 & 2.38 \\
xm3600 & llamagen-vq-ds16-c2i & 1 & 13.98 & 13.95 & 14 & 7.00 & 6.99 & 1.00 & 0.17 \\
xm3600 & text-ar & 1 & 10.91 & 10.89 & 14 & 0.80 & 0.62 & 1.28 & 22.05 \\
xm3600 & text-bn & 1 & 7.64 & 7.62 & 12 & 0.49 & 0.31 & 1.57 & 36.29 \\
xm3600 & text-cs & 1 & 10.13 & 10.09 & 14 & 0.66 & 0.48 & 1.38 & 27.65 \\
xm3600 & text-da & 1 & 8.76 & 8.73 & 13 & 0.85 & 0.58 & 1.48 & 32.60 \\
xm3600 & text-de & 1 & 9.31 & 9.29 & 14 & 1.44 & 0.96 & 1.50 & 33.48 \\
xm3600 & text-el & 1 & 10.32 & 10.30 & 14 & 0.80 & 0.59 & 1.36 & 26.28 \\
xm3600 & text-es & 1 & 8.53 & 8.49 & 13 & 1.14 & 0.75 & 1.52 & 34.41 \\
xm3600 & text-fa & 1 & 9.08 & 9.06 & 13 & 1.21 & 0.84 & 1.43 & 30.14 \\
xm3600 & text-fi & 1 & 11.03 & 11.01 & 14 & 0.78 & 0.62 & 1.27 & 21.18 \\
xm3600 & text-fil & 1 & 7.82 & 7.79 & 13 & 1.14 & 0.69 & 1.66 & 39.86 \\
xm3600 & text-fr & 1 & 8.44 & 8.41 & 13 & 1.51 & 0.98 & 1.54 & 35.08 \\
xm3600 & text-hi & 1 & 7.54 & 7.52 & 12 & 1.38 & 0.87 & 1.59 & 37.14 \\
xm3600 & text-hr & 1 & 10.55 & 10.53 & 14 & 0.95 & 0.72 & 1.33 & 24.62 \\
xm3600 & text-hu & 1 & 10.08 & 10.05 & 14 & 0.96 & 0.69 & 1.39 & 28.01 \\
xm3600 & text-id & 1 & 8.74 & 8.71 & 13 & 1.33 & 0.89 & 1.49 & 32.76 \\
xm3600 & text-it & 1 & 9.12 & 9.09 & 13 & 1.37 & 0.96 & 1.43 & 29.86 \\
xm3600 & text-he & 1 & 10.25 & 10.22 & 14 & 1.33 & 0.97 & 1.37 & 26.81 \\
xm3600 & text-ja & 1 & 8.48 & 8.45 & 13 & 1.44 & 0.94 & 1.53 & 34.81 \\
xm3600 & text-ko & 1 & 9.75 & 9.73 & 13 & 0.99 & 0.74 & 1.33 & 24.97 \\
xm3600 & text-mi & 1 & 7.54 & 7.51 & 12 & 0.68 & 0.43 & 1.59 & 37.16 \\
xm3600 & text-nl & 1 & 8.42 & 8.40 & 13 & 0.87 & 0.56 & 1.54 & 35.20 \\
xm3600 & text-no & 1 & 8.76 & 8.74 & 13 & 0.92 & 0.62 & 1.48 & 32.58 \\
xm3600 & text-pl & 1 & 10.27 & 10.25 & 14 & 0.90 & 0.66 & 1.36 & 26.63 \\
xm3600 & text-pt & 1 & 8.86 & 8.82 & 13 & 1.08 & 0.74 & 1.47 & 31.88 \\
xm3600 & text-ro & 1 & 9.09 & 9.06 & 14 & 1.65 & 1.07 & 1.54 & 35.07 \\
xm3600 & text-ru & 1 & 10.29 & 10.26 & 14 & 1.09 & 0.80 & 1.36 & 26.49 \\
xm3600 & text-sv & 1 & 8.72 & 8.68 & 13 & 0.82 & 0.55 & 1.49 & 32.95 \\
xm3600 & text-sw & 1 & 8.58 & 8.54 & 13 & 0.99 & 0.66 & 1.52 & 34.03 \\
xm3600 & text-te & 1 & 8.47 & 8.44 & 13 & 0.71 & 0.46 & 1.53 & 34.84 \\
xm3600 & text-th & 1 & 8.60 & 8.57 & 12 & 1.12 & 0.81 & 1.40 & 28.33 \\
xm3600 & text-tr & 1 & 10.03 & 10.00 & 14 & 0.98 & 0.70 & 1.40 & 28.37 \\
xm3600 & text-uk & 1 & 10.65 & 10.62 & 14 & 1.07 & 0.82 & 1.31 & 23.93 \\
xm3600 & text-vi & 1 & 8.68 & 8.65 & 12 & 1.61 & 1.17 & 1.38 & 27.65 \\
xm3600 & text-zh & 1 & 9.63 & 9.60 & 14 & 1.43 & 0.98 & 1.45 & 31.22 \\
spin & chameleon-512 & 1 & 11.24 & 11.20 & 12 & 6.00 & 5.62 & 1.07 & 6.37 \\
spin & compvis-vq-f8-64 & 1 & 9.75 & 9.71 & 10 & 5.00 & 4.87 & 1.03 & 2.54 \\
spin & compvis-vq-f8-256 & 1 & 9.65 & 9.62 & 10 & 5.00 & 4.83 & 1.04 & 3.49 \\
spin & compvis-vq-imagenet-f16-1024-256 & 1 & 8.77 & 8.74 & 9 & 4.50 & 4.39 & 1.03 & 2.55 \\
spin & llamagen-vq-ds16-c2i & 1 & 13.97 & 13.95 & 14 & 7.00 & 6.99 & 1.00 & 0.18 \\
cc12m & chameleon-512 & 1 & 11.28 & 11.25 & 12 & 6.00 & 5.64 & 1.06 & 6.03 \\
cc12m & compvis-vq-f8-64 & 1 & 9.76 & 9.73 & 10 & 5.00 & 4.88 & 1.02 & 2.40 \\
cc12m & compvis-vq-f8-256 & 1 & 9.64 & 9.60 & 10 & 5.00 & 4.82 & 1.04 & 3.62 \\
cc12m & compvis-vq-imagenet-f16-1024-256 & 1 & 8.73 & 8.70 & 9 & 4.50 & 4.36 & 1.03 & 3.01 \\
cc12m & llamagen-vq-ds16-c2i & 1 & 13.89 & 13.86 & 14 & 7.00 & 6.94 & 1.01 & 0.80 \\
ilsvrc & chameleon-512 & 1 & 11.27 & 11.24 & 12 & 6.00 & 5.64 & 1.06 & 6.08 \\
ilsvrc & compvis-vq-f8-64 & 1 & 9.78 & 9.74 & 10 & 5.00 & 4.89 & 1.02 & 2.21 \\
ilsvrc & compvis-vq-f8-256 & 1 & 9.69 & 9.66 & 10 & 5.00 & 4.85 & 1.03 & 3.09 \\
ilsvrc & compvis-vq-imagenet-f16-1024-256 & 1 & 8.78 & 8.75 & 9 & 4.50 & 4.39 & 1.02 & 2.40 \\
ilsvrc & llamagen-vq-ds16-c2i & 1 & 13.98 & 13.96 & 14 & 7.00 & 6.99 & 1.00 & 0.13 \\
coco & chameleon-512 & 2 & 18.80 & 18.79 & 19 & 9.50 & 9.40 & 1.01 & 1.03 \\
coco & compvis-vq-f8-64 & 2 & 18.29 & 18.28 & 19 & 9.50 & 9.14 & 1.04 & 3.74 \\
coco & compvis-vq-f8-256 & 2 & 18.12 & 18.11 & 19 & 9.50 & 9.06 & 1.05 & 4.65 \\
coco & compvis-vq-imagenet-f16-1024-256 & 2 & 17.08 & 17.05 & 18 & 9.00 & 8.54 & 1.05 & 5.13 \\
coco & llamagen-vq-ds16-c2i & 2 & 18.94 & 18.92 & 19 & 9.50 & 9.47 & 1.00 & 0.30 \\
coco & text-ar & 2 & 14.99 & 14.97 & 18 & 9.00 & 7.50 & 1.20 & 16.71 \\
coco & text-bn & 2 & 14.17 & 14.15 & 17 & 8.50 & 7.09 & 1.20 & 16.63 \\
coco & text-cs & 2 & 15.07 & 15.05 & 18 & 9.00 & 7.54 & 1.19 & 16.27 \\
coco & text-da & 2 & 12.97 & 12.95 & 17 & 8.50 & 6.49 & 1.31 & 23.70 \\
coco & text-de & 2 & 13.80 & 13.78 & 17 & 8.50 & 6.90 & 1.23 & 18.82 \\
coco & text-el & 2 & 13.46 & 13.43 & 17 & 8.50 & 6.73 & 1.26 & 20.85 \\
coco & text-es & 2 & 12.87 & 12.84 & 17 & 8.50 & 6.43 & 1.32 & 24.30 \\
coco & text-fa & 2 & 13.07 & 13.04 & 17 & 8.50 & 6.53 & 1.30 & 23.12 \\
coco & text-fi & 2 & 15.41 & 15.39 & 18 & 9.00 & 7.70 & 1.17 & 14.40 \\
coco & text-fil & 2 & 12.30 & 12.28 & 16 & 8.00 & 6.15 & 1.30 & 23.11 \\
coco & text-fr & 2 & 12.93 & 12.90 & 17 & 8.50 & 6.46 & 1.32 & 23.97 \\
coco & text-hi & 2 & 13.01 & 12.99 & 17 & 8.50 & 6.51 & 1.31 & 23.46 \\
coco & text-hr & 2 & 14.84 & 14.81 & 18 & 9.00 & 7.42 & 1.21 & 17.58 \\
coco & text-hu & 2 & 14.57 & 14.55 & 18 & 9.00 & 7.29 & 1.24 & 19.04 \\
coco & text-id & 2 & 13.28 & 13.25 & 17 & 8.50 & 6.64 & 1.28 & 21.89 \\
coco & text-it & 2 & 13.27 & 13.24 & 17 & 8.50 & 6.63 & 1.28 & 21.96 \\
coco & text-he & 2 & 15.24 & 15.22 & 18 & 9.00 & 7.62 & 1.18 & 15.31 \\
coco & text-ja & 2 & 12.08 & 12.06 & 16 & 8.00 & 6.04 & 1.32 & 24.50 \\
coco & text-ko & 2 & 13.48 & 13.46 & 17 & 8.50 & 6.74 & 1.26 & 20.69 \\
coco & text-mi & 2 & 10.90 & 10.88 & 16 & 8.00 & 5.45 & 1.47 & 31.89 \\
coco & text-nl & 2 & 13.16 & 13.13 & 17 & 8.50 & 6.58 & 1.29 & 22.59 \\
coco & text-no & 2 & 13.11 & 13.09 & 17 & 8.50 & 6.56 & 1.30 & 22.87 \\
coco & text-pl & 2 & 15.07 & 15.05 & 18 & 9.00 & 7.54 & 1.19 & 16.26 \\
coco & text-pt & 2 & 13.02 & 12.99 & 17 & 8.50 & 6.51 & 1.31 & 23.43 \\
coco & text-ro & 2 & 13.55 & 13.52 & 17 & 8.50 & 6.78 & 1.25 & 20.29 \\
coco & text-ru & 2 & 14.78 & 14.76 & 18 & 9.00 & 7.39 & 1.22 & 17.90 \\
coco & text-sv & 2 & 13.28 & 13.25 & 17 & 8.50 & 6.64 & 1.28 & 21.89 \\
coco & text-sw & 2 & 13.86 & 13.83 & 17 & 8.50 & 6.93 & 1.23 & 18.49 \\
coco & text-te & 2 & 15.02 & 15.00 & 18 & 9.00 & 7.51 & 1.20 & 16.53 \\
coco & text-th & 2 & 13.10 & 13.08 & 17 & 8.50 & 6.55 & 1.30 & 22.94 \\
coco & text-tr & 2 & 14.18 & 14.15 & 17 & 8.50 & 7.09 & 1.20 & 16.60 \\
coco & text-uk & 2 & 14.75 & 14.73 & 18 & 9.00 & 7.37 & 1.22 & 18.06 \\
coco & text-vi & 2 & 12.45 & 12.42 & 16 & 8.00 & 6.22 & 1.29 & 22.20 \\
coco & text-zh & 2 & 14.08 & 14.06 & 17 & 8.50 & 7.04 & 1.21 & 17.15 \\
xm3600 & chameleon-512 & 2 & 18.79 & 18.77 & 19 & 9.50 & 9.40 & 1.01 & 1.09 \\
xm3600 & compvis-vq-f8-64 & 2 & 16.68 & 16.63 & 17 & 1.95 & 1.91 & 1.02 & 1.87 \\
xm3600 & compvis-vq-f8-256 & 2 & 18.10 & 18.09 & 19 & 9.50 & 9.05 & 1.05 & 4.72 \\
xm3600 & compvis-vq-imagenet-f16-1024-256 & 2 & 17.04 & 17.01 & 18 & 9.00 & 8.52 & 1.06 & 5.35 \\
xm3600 & llamagen-vq-ds16-c2i & 2 & 18.94 & 18.92 & 19 & 9.50 & 9.47 & 1.00 & 0.32 \\
xm3600 & text-ar & 2 & 14.49 & 14.43 & 16 & 0.79 & 0.72 & 1.10 & 9.44 \\
xm3600 & text-bn & 2 & 11.30 & 11.27 & 14 & 0.52 & 0.42 & 1.24 & 19.31 \\
xm3600 & text-cs & 2 & 13.64 & 13.59 & 15 & 0.60 & 0.55 & 1.10 & 9.06 \\
xm3600 & text-da & 2 & 13.13 & 13.11 & 15 & 0.88 & 0.77 & 1.14 & 12.48 \\
xm3600 & text-de & 2 & 14.04 & 13.98 & 16 & 1.51 & 1.32 & 1.14 & 12.28 \\
xm3600 & text-el & 2 & 14.16 & 14.10 & 15 & 0.75 & 0.71 & 1.06 & 5.62 \\
xm3600 & text-es & 2 & 12.83 & 12.80 & 15 & 1.19 & 1.02 & 1.17 & 14.44 \\
xm3600 & text-fa & 2 & 13.66 & 13.61 & 16 & 1.37 & 1.17 & 1.17 & 14.62 \\
xm3600 & text-fi & 2 & 14.58 & 14.51 & 16 & 0.78 & 0.71 & 1.10 & 8.90 \\
xm3600 & text-fil & 2 & 12.35 & 12.32 & 15 & 1.21 & 1.00 & 1.21 & 17.64 \\
xm3600 & text-fr & 2 & 12.82 & 12.79 & 16 & 1.72 & 1.38 & 1.25 & 19.86 \\
xm3600 & text-hi & 2 & 11.30 & 11.27 & 15 & 1.60 & 1.21 & 1.33 & 24.68 \\
xm3600 & text-hr & 2 & 14.47 & 14.46 & 16 & 0.97 & 0.88 & 1.11 & 9.54 \\
xm3600 & text-hu & 2 & 14.41 & 14.39 & 16 & 0.98 & 0.88 & 1.11 & 9.96 \\
xm3600 & text-id & 2 & 12.84 & 12.80 & 15 & 1.43 & 1.22 & 1.17 & 14.40 \\
xm3600 & text-it & 2 & 13.70 & 13.65 & 16 & 1.55 & 1.33 & 1.17 & 14.35 \\
xm3600 & text-he & 2 & 14.58 & 14.52 & 16 & 1.40 & 1.28 & 1.10 & 8.88 \\
xm3600 & text-ja & 2 & 12.85 & 12.82 & 15 & 1.55 & 1.33 & 1.17 & 14.33 \\
xm3600 & text-ko & 2 & 14.00 & 13.98 & 16 & 1.09 & 0.96 & 1.14 & 12.52 \\
xm3600 & text-mi & 2 & 11.54 & 11.51 & 14 & 0.73 & 0.60 & 1.21 & 17.55 \\
xm3600 & text-nl & 2 & 12.50 & 12.48 & 15 & 0.88 & 0.73 & 1.20 & 16.67 \\
xm3600 & text-no & 2 & 12.92 & 12.90 & 15 & 0.95 & 0.82 & 1.16 & 13.87 \\
xm3600 & text-pl & 2 & 14.26 & 14.23 & 16 & 0.92 & 0.82 & 1.12 & 10.87 \\
xm3600 & text-pt & 2 & 13.40 & 13.37 & 15 & 1.14 & 1.02 & 1.12 & 10.68 \\
xm3600 & text-ro & 2 & 13.49 & 13.46 & 16 & 1.77 & 1.49 & 1.19 & 15.69 \\
xm3600 & text-ru & 2 & 14.30 & 14.28 & 16 & 1.13 & 1.01 & 1.12 & 10.63 \\
xm3600 & text-sv & 2 & 13.17 & 13.14 & 15 & 0.83 & 0.73 & 1.14 & 12.21 \\
xm3600 & text-sw & 2 & 12.98 & 12.96 & 15 & 1.04 & 0.90 & 1.16 & 13.45 \\
xm3600 & text-te & 2 & 12.11 & 12.06 & 15 & 0.71 & 0.57 & 1.24 & 19.29 \\
xm3600 & text-th & 2 & 12.63 & 12.58 & 15 & 1.30 & 1.09 & 1.19 & 15.82 \\
xm3600 & text-tr & 2 & 14.23 & 14.22 & 16 & 1.01 & 0.90 & 1.12 & 11.06 \\
xm3600 & text-uk & 2 & 14.49 & 14.48 & 16 & 1.11 & 1.01 & 1.10 & 9.41 \\
xm3600 & text-vi & 2 & 12.96 & 12.94 & 15 & 1.91 & 1.65 & 1.16 & 13.60 \\
xm3600 & text-zh & 2 & 14.31 & 14.25 & 16 & 1.52 & 1.36 & 1.12 & 10.57 \\
spin & chameleon-512 & 2 & 18.80 & 18.78 & 19 & 9.50 & 9.40 & 1.01 & 1.05 \\
spin & compvis-vq-f8-64 & 2 & 18.26 & 18.25 & 19 & 9.50 & 9.13 & 1.04 & 3.90 \\
spin & compvis-vq-f8-256 & 2 & 18.09 & 18.08 & 19 & 9.50 & 9.05 & 1.05 & 4.77 \\
spin & compvis-vq-imagenet-f16-1024-256 & 2 & 17.06 & 17.04 & 18 & 9.00 & 8.53 & 1.05 & 5.21 \\
spin & llamagen-vq-ds16-c2i & 2 & 18.94 & 18.92 & 19 & 9.50 & 9.47 & 1.00 & 0.30 \\
cc12m & chameleon-512 & 2 & 18.58 & 18.57 & 19 & 9.50 & 9.29 & 1.02 & 2.20 \\
cc12m & compvis-vq-f8-64 & 2 & 18.23 & 18.22 & 19 & 9.50 & 9.12 & 1.04 & 4.05 \\
cc12m & compvis-vq-f8-256 & 2 & 17.75 & 17.73 & 19 & 9.50 & 8.87 & 1.07 & 6.60 \\
cc12m & compvis-vq-imagenet-f16-1024-256 & 2 & 16.76 & 16.73 & 18 & 9.00 & 8.38 & 1.07 & 6.90 \\
cc12m & llamagen-vq-ds16-c2i & 2 & 18.85 & 18.83 & 19 & 9.50 & 9.42 & 1.01 & 0.81 \\
ilsvrc & chameleon-512 & 2 & 18.76 & 18.74 & 19 & 9.50 & 9.38 & 1.01 & 1.26 \\
ilsvrc & compvis-vq-f8-64 & 2 & 18.29 & 18.28 & 19 & 9.50 & 9.14 & 1.04 & 3.76 \\
ilsvrc & compvis-vq-f8-256 & 2 & 18.09 & 18.08 & 19 & 9.50 & 9.04 & 1.05 & 4.80 \\
ilsvrc & compvis-vq-imagenet-f16-1024-256 & 2 & 17.01 & 16.99 & 18 & 9.00 & 8.51 & 1.06 & 5.49 \\
ilsvrc & llamagen-vq-ds16-c2i & 2 & 18.93 & 18.91 & 19 & 9.50 & 9.47 & 1.00 & 0.36 \\
\end{longtable}
\end{tiny}

\section{Segmentation Granularity}
\label{app:purity}

In \autoref{sec:segmentation}, we explore at what level visual tokens/words correlate with parts/sub-parts/wholes of objects in images. To analyze the co-occurrence of wholes, parts, and sub-parts, we primarily leverage the SPIN dataset, discussed in \autoref{app:datasets}, which provides labeled annotations for each of these levels in the images. To compute co-occurrence statistics, we first extract a part-label-to-visual-token co-occurrence frequency matrix for each tokenizer and dataset. Each entry $(i,j)$ of the matrix represents the number of times that visual token $z_i$ co-occurs with part-label $y_j$ (which could represent a whole, part, or sub-part). From this co-occurrence matrix, we compute three metrics—Part Purity, Visual Token Purity, and Part-Normalized Mutual Information—as described in \citet{hsu2021hubert}.

\textbf{Part Purity:}  Part purity describes the average probability of the most likely part-label for each visual-token, representing how accurately parts are assigned to the corresponding visual tokens. It is computed as:
\begin{equation}
    \text{Part Purity (PP)} = \mathbb{E}_{z} \left[ p(y^*(z) \mid z) \right]
\end{equation}
where $z$ is a visual token cluster, $y^*(z)$ denotes the most likely part-label for a given visual-token $z$, $p(y^*(z) \mid z)$ is the conditional probability of the most likely part-label $y^*(z)$ given the visual-token $z$, and $\mathbb{E}_{z}$ is the expectation over all visual tokens. In practice, we draw these probabilities from the normalized empirical co-occurrence matrix.

\textbf{Visual Token Purity:} Visual token purity measures how well images containing the same part-label are consistently assigned to the same visual tokens. It is computed as:
\begin{equation}
    \text{Visual Token Purity (VTP)} = \mathbb{E}_{y} \left[ p(z^*(y) \mid y) \right]
\end{equation}
where $y$ is a part-label, $z^*(y)$ represents the most likely visual-token for a given part-label $y$, $p(z^*(y) \mid y)$ is the conditional probability of the most likely visual-token $z^*(y)$ given the part-label $y$, and $\mathbb{E}_{y}$ is the expectation over all part-labels. Similar to part-purity, these probabilities are derived from the normalized empirical co-occurrence matrix.

\paragraph{Part-Normalized Mutual Information:} Part-normalized mutual information (PNMI) is an information-theoretic metric that quantifies the percentage of uncertainty about a part-label eliminated after observing a visual-token. It is computed as:
\begin{equation}
    \text{PNMI} = \frac{I(y; z)}{H(y)} = \frac{H(y) - H(y \mid z)}{H(y)} = 1 - \frac{H(y \mid z)}{H(y)}
\end{equation}
where $I(y; z)$ is the mutual information between part-labels $y$ and visual tokens $z$, $H(y)$ is the entropy of the part-labels, and $H(y \mid z)$ is the conditional entropy of the part-labels given the visual tokens. The entropy values are computed from the empirical co-occurrence frequency matrix, where each entry represents the joint probability $p(y, z)$ of a part-label $y$ and a visual-token $z$ co-occurring. Specifically, $H(y)$ is computed as:
\begin{equation}
    H(y) = - \sum_{i} p(y_i) \log p(y_i)
\end{equation}
where $p(y_i)$ is the marginal probability of part-label $y_i$, derived by summing the joint probabilities $p(y_i, z_j)$ across all visual-tokens $z_j$. Similarly, the conditional entropy $H(y \mid z)$ is computed as:
\begin{equation}
    H(y \mid z) = - \sum_{j} p(z_j) \sum_{i} p(y_i \mid z_j) \log p(y_i \mid z_j)
\end{equation}
where $p(y_i \mid z_j)$ is the conditional probability of part-label $y_i$ given visual-token $z_j$, derived from the co-occurrence matrix by normalizing the joint probabilities $p(y_i, z_j)$ by $p(z_j)$, the marginal probability of the visual-token $z_j$. Higher PNMI values indicate that more information about the part-label is captured by the visual-token assignments.

\section{Topological Alignment of Vision and Language Tokens}
\label{app:glove}

\subsection{GloVe Embedding of Vision and Language Tokens}

In order to get continuous representations of the vision and language token spaces, we employ GloVe embeddings \cite{pennington2014glove}. GloVe (Global Vectors for Word Representation) is a word embedding technique that captures semantic relationships between words by training on global word co-occurrence statistics. Unlike local context methods like Word2Vec \citep{church2017word2vec}, GloVe constructs a matrix from word co-occurrence counts in a corpus and factorizes this matrix to generate dense vector representations. These embeddings reflect the relative meanings of words, allowing similar words to have similar vectors in the latent space. GloVe aims to learn word embeddings by factorizing a token co-occurrence matrix. The model minimizes a weighted least squares objective function:
\begin{equation}
J = \sum_{i,j=1}^{V} f(X_{ij}) \left( w_i^\top \tilde{w}_j + b_i + \tilde{b}_j - \log X_{ij} \right)^2
\end{equation}

where $X_{ij}$ is the co-occurrence count of token $i$ with token $j$, $w_i$ and $\tilde{w}_j$ are the token vectors for token $i$ and $j$, $b_i$ and $\tilde{b}_j$ are the bias terms, and $f(X_{ij})$ is a token co-occurrence based weighting function to discount frequent co-occurrences.

In all of the analysis methods below, before applying analysis we whiten the data before normalization to avoid significant scale effects:
\begin{equation}
X'_{ij} = \frac{X_{ij}}{\sigma_j}, \quad \sigma_j = \sqrt{\frac{1}{n} \sum_{i=1}^{n} \left( X_{ij} - \mu_j \right)^2}, \quad \mu_j = \frac{1}{n} \sum_{i=1}^{n} X_{ij}
\end{equation}
where $ X_{ij} $ is the original value of the i-th data point in the j-th feature, $ \sigma_j $ is the standard deviation of the j-th feature, and $ \mu_j $ is the mean of the j-th feature.

\subsection{Compound Probabilistic Context-free Grammars}\label{app:cpcfgs}

\subsubsection{Background}

Here we describe the basic background and formulation of Compound Probabilistic Context-free grammars (C-PCFGs) for convenience, much of this content is sourced from ~\citep{kim-etal-2019-compound}, which we point readers to for a more thorough treatment of the topic. 

C-PCFGs extend the PCFG formalism.
PCFGs are defined by a 5-tuple $\mathcal{G}=(S,\mathcal{N},\mathcal{P},\Sigma,\mathcal{R})$, consisting of a start symbol $S$, a set of non-terminals $\mathcal{N}$, a set of pre-terminals $\mathcal{P}$, a set of terminals $\Sigma$, and a set of derivation rules $\mathcal{R}$:
\begin{align*}
S\to A && A\in \mathcal{N} \\
A\to BC && A\in \mathcal{N}, B,C\in \mathcal{N}\cup \mathcal{P} \\
T\to w && T\in \mathcal{P}, w\in \Sigma 
\end{align*}
The derivation rules are probabilistic, with their distribution denoted as $\bm{\pi}=\{\pi_r\}_{r\in\mathcal{R}}$. 
Inference may be performed efficiently over them using the inside algorithm~\citep{baker1979trainable}. 
In neural variants of PCFGs, this distribution may be formulated as follows: 
\begin{align*}
    \pi _{S\to A}=\frac{\text{exp}(\bm{u}_A^{\top} f_1(\bm{w}_S))}{\Sigma _{A'\in\mathcal{N}}\text{exp}(\bm{u}_{A'}^\top f_1(\bm{w}_S))} \\
    \pi _{A\to BC}=\frac{\text{exp}(\bm{u}_{BC}^{\top} \bm{w}_A)}{\Sigma _{B'C'\in\mathcal{M}}\text{exp}(\bm{u}_{B'C'}^\top \bm{w}_A)} \\
    \pi _{T\to w}=\frac{\text{exp}(\bm{u}_w^{\top} f_2(\bm{w}_T))}{\Sigma _{w'\in\Sigma}\text{exp}(\bm{u}_{w'}^\top f_2(\bm{w}_T))} \\
\end{align*}
where $\bm{u}$ are transformation vectors for each production rule, $\bm{w}$ are learnable parameter vectors for each symbol, and $f_1$ and $f_2$ are neural networks.

Compound PCFGs~\citep{kim-etal-2019-compound} formulate rule probabilities as a compound probability distribution~\citep{robbins1956empirical}:
\begin{align*}
    \bm{z}\sim p_{\gamma}(\bm{z}) && \pi _{\bm{z}} = f_{\lambda}(\bm{z}, \bm{E}_{\mathcal{G}})
\end{align*}
Where $\bm{z}$ is a latent variable generated by a prior distribution (a spherical Gaussian) and $\bm{E}_{\mathcal{G}}=\{\bm{w}_N|N\in \{S\}\cup \mathcal{N} \cup \mathcal{P}\}$ denotes the set of symbol embeddings. 
Rule probabilities $\bm{\pi}_{\bm{z}}$ are conditioned on this latent: 
\begin{align*}
    \pi _{\bm{z}, S\to A}\propto \text{exp}(\bm{u}_{A}^\top f_1 ([\bm{w}_S;\bm{z}])), \\
    \pi _{\bm{z}, A\to BC}\propto \text{exp}(\bm{u}_{BC}^\top [\bm{w}_A;\bm{z}]), \\
    \pi _{\bm{z}, T\to w}\propto \text{exp}(\bm{u}_{w}^\top f_2 ([\bm{w}_T;\bm{z}]))
\end{align*}
The latent $\bm{z}$ allows global information to be shared across parsing decisions, while simultaneously respecting the context-free assumption when $\bm{z}$ is fixed, allowing for efficient inference as before.

C-PCFGs are optimized with variational methods~\citep{kingma2013auto}, since the introduction of $\bm{z}$ makes inference intractable.
At inference time, given a sentence $\bm{x}$, the variational inference network $q_\phi$ is used to produce the latent $\bm{z} = \bm{\mu}_\phi (g(\mathcal{E}(\bm{x})))$.
Here, $g$ is a sentence encoder used to generate a vector representation given token embeddings $\mathcal{E}(\bm{x})$.
For more details on C-PCFGs, we point readers to \citet{kim-etal-2019-compound}.

\subsubsection{Parse Trees}

\begin{figure}[ht]
    \centering
    \begin{subfigure}[b]{0.49\textwidth}
        \centering
        \includegraphics[width=\textwidth]{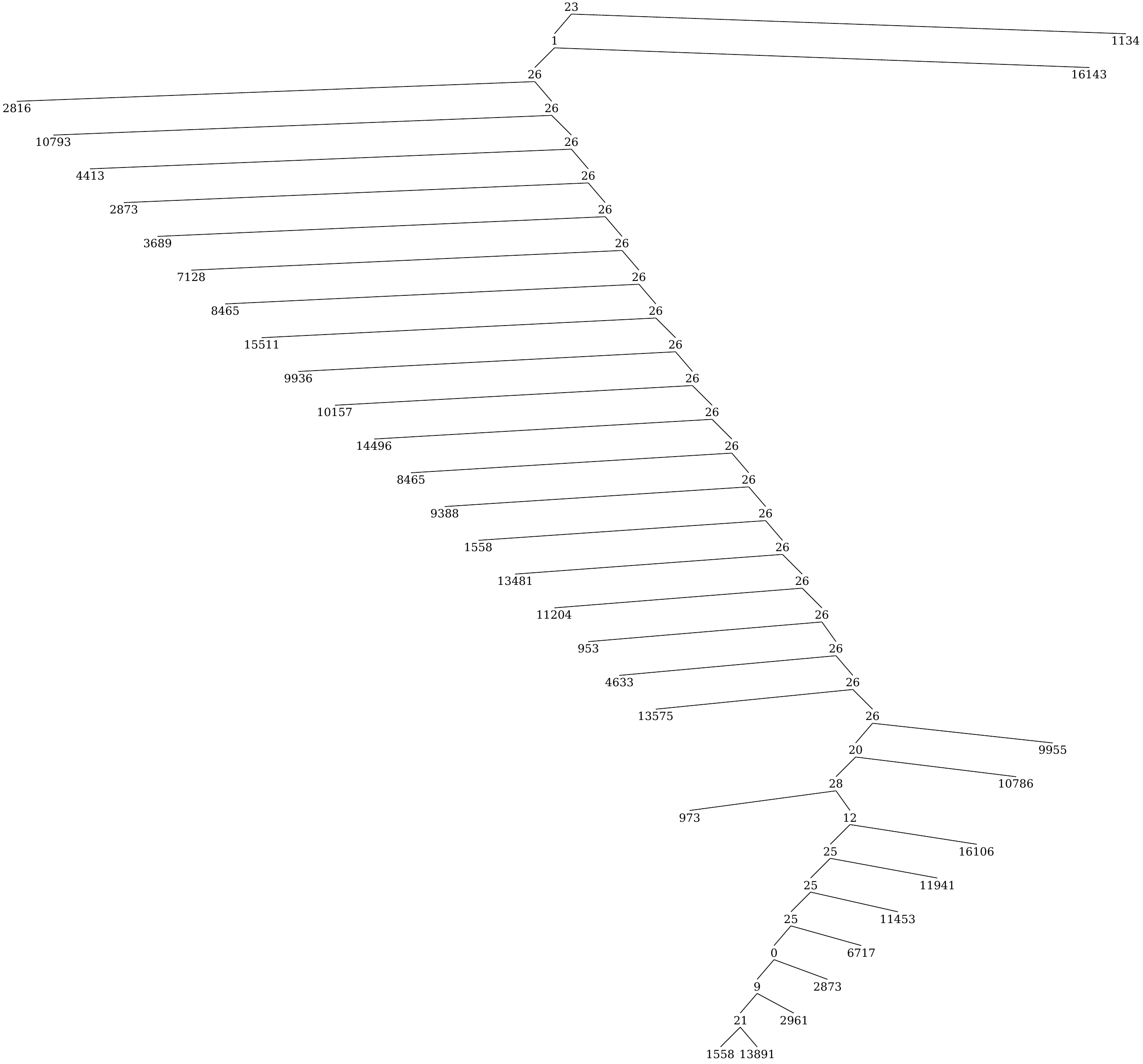}
        \caption{XM3600}
    \end{subfigure}
    \hfill
    \begin{subfigure}[b]{0.39\textwidth}
        \centering
        \includegraphics[width=\textwidth]{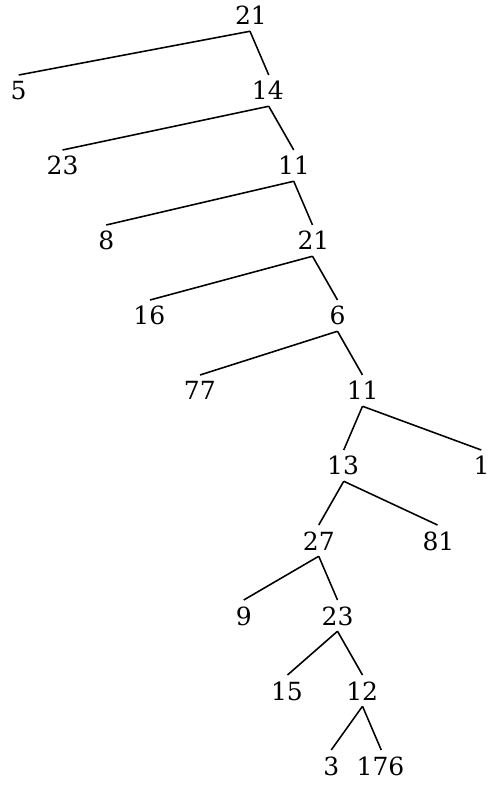}
        \caption{COCO-DE}
    \end{subfigure}

    \vskip\baselineskip

    \begin{subfigure}[b]{0.49\textwidth}
        \centering
        \includegraphics[width=\textwidth]{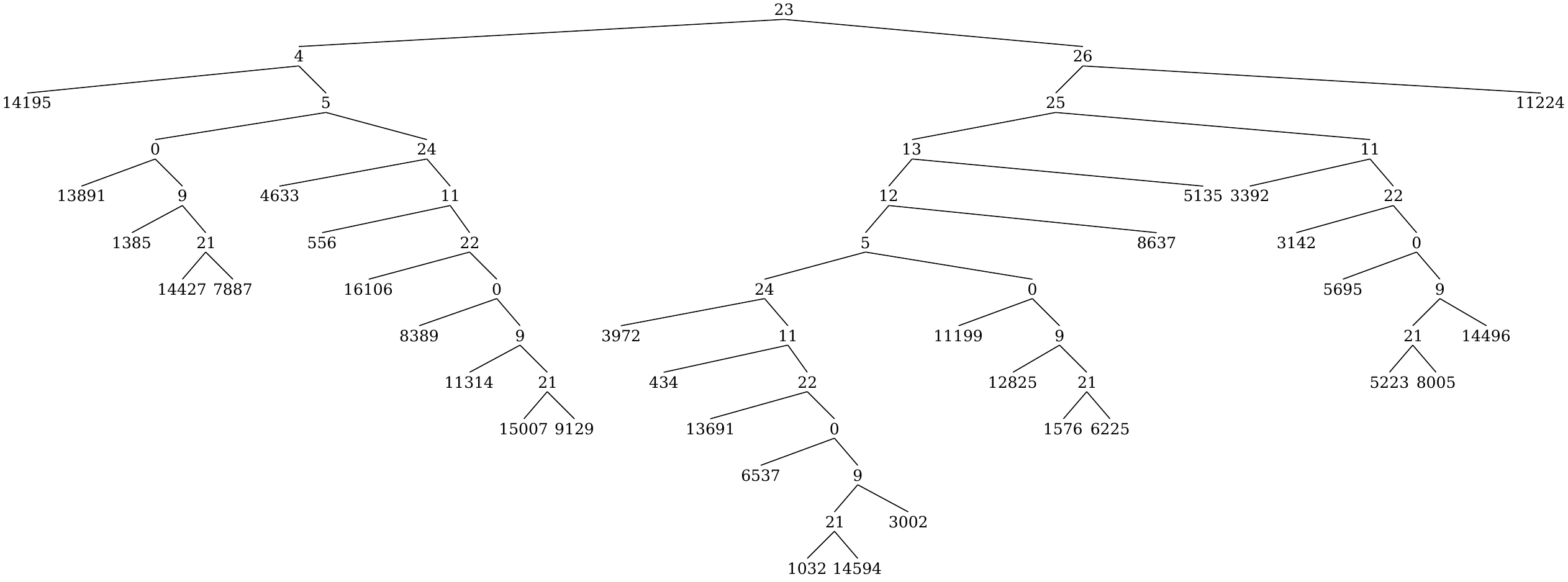}
        \caption{COCO-VQ}
    \end{subfigure}
    \hfill
    \begin{subfigure}[b]{0.49\textwidth}
        \centering
        \includegraphics[width=\textwidth]{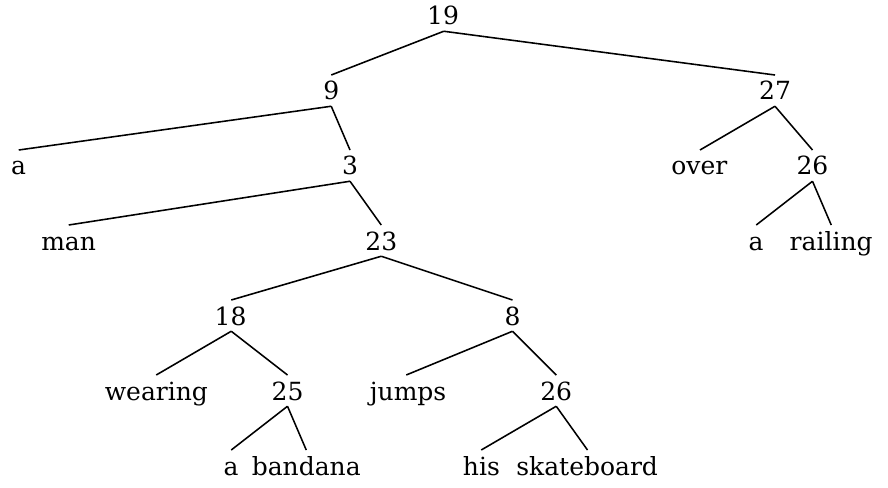}
        \caption{COCO-EN}
    \end{subfigure}
    \hfill
    \begin{subfigure}[b]{0.49\textwidth}
        \centering
        \includegraphics[width=\textwidth]{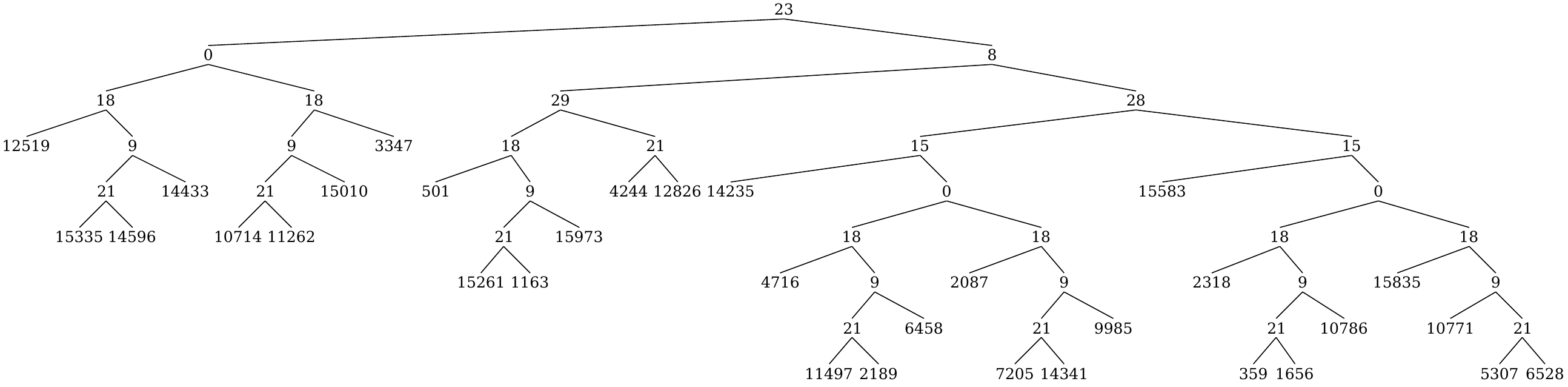}
        \caption{CC12M}
    \end{subfigure}
    \hfill
    \begin{subfigure}[b]{0.49\textwidth}
        \centering
        \includegraphics[width=\textwidth]{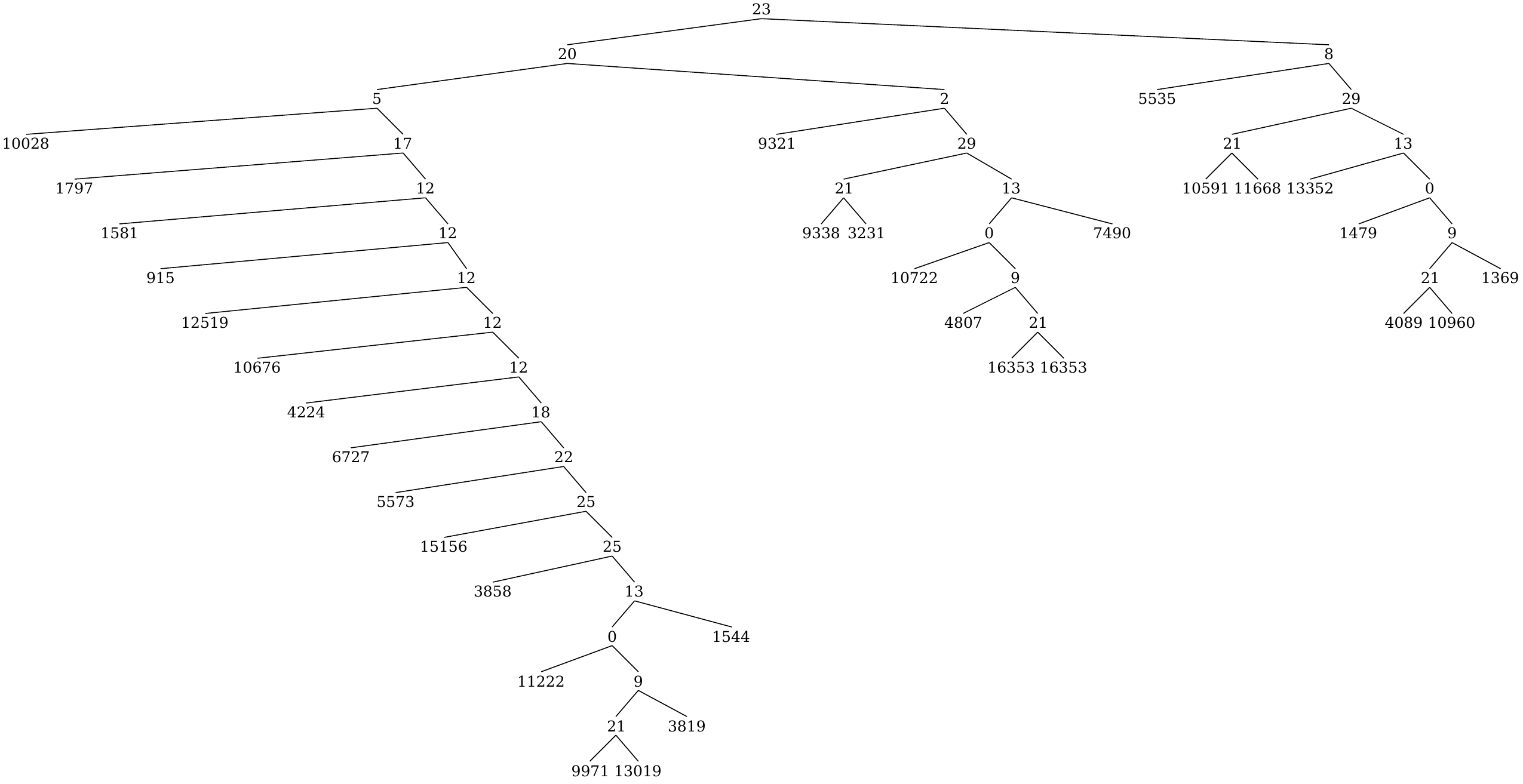}
        \caption{SPIN}
    \end{subfigure}
    
    \vskip\baselineskip

    \begin{subfigure}[b]{0.49\textwidth}
        \centering
        \includegraphics[width=\textwidth]{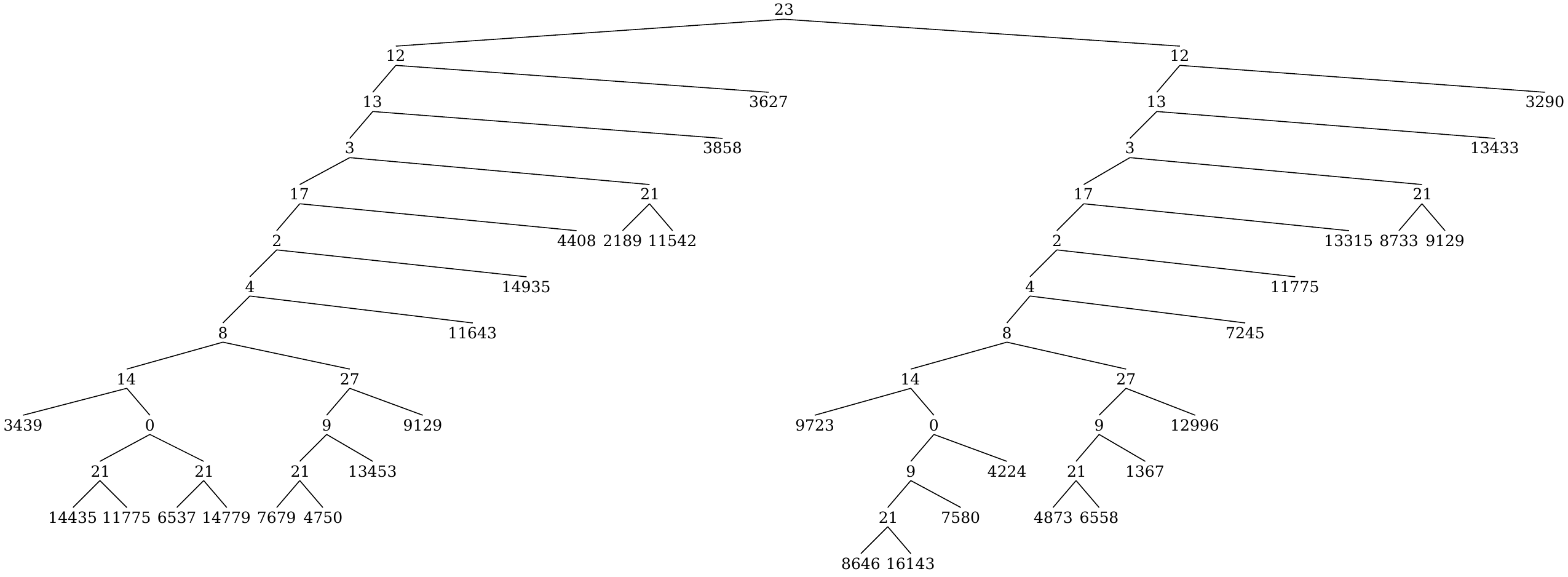}
        \caption{ILSVRC}
    \end{subfigure}

    \caption{Example parse trees for different datasets.}
    \label{fig:parse_trees}
\end{figure}

In Figure \ref{fig:parse_trees} we show an example parse tree generated with a learned grammar for each dataset. 

\subsection{Procrustes Analysis}

Procrustes Analysis is a statistical method used to compare the shapes or structures of two datasets by finding an optimal transformation (including translation, scaling, and rotation) that minimizes the distance between corresponding points in the datasets. The resulting transformation provides insight into how closely the datasets align in their geometry. Procrustes Analysis minimizes the distance between two matrices $X$ and $Y$ by finding the optimal translation, scaling, and rotation. The goal is to solve:

\begin{equation}
\min_{R, b, c} \| bXR + c - Y \|_F
\end{equation}

where: $X$ and $Y$ are the two point sets (matrices) being compared, $R$ is the optimal rotation matrix, $b$ is a scaling factor, $c$ is the translation vector,
and $|\cdot\|_F$ is the Frobenius norm.

For Procrustes analysis, it is required that the two matrices to be aligned have identical shape. Because the number of tokens is different in the vision and language cases, in our experiments we use K-means to quantize the different token embedding spaces to 256 centers, which we then compare topologically. This has the downside of reducing the topological comparisons to more global structure comparisons, however means that we can run experiments on point-to-point coherence. 

\begin{figure}[t]
    \centering
    \begin{subfigure}[b]{0.49\textwidth}
        \centering
        \includegraphics[width=\textwidth]{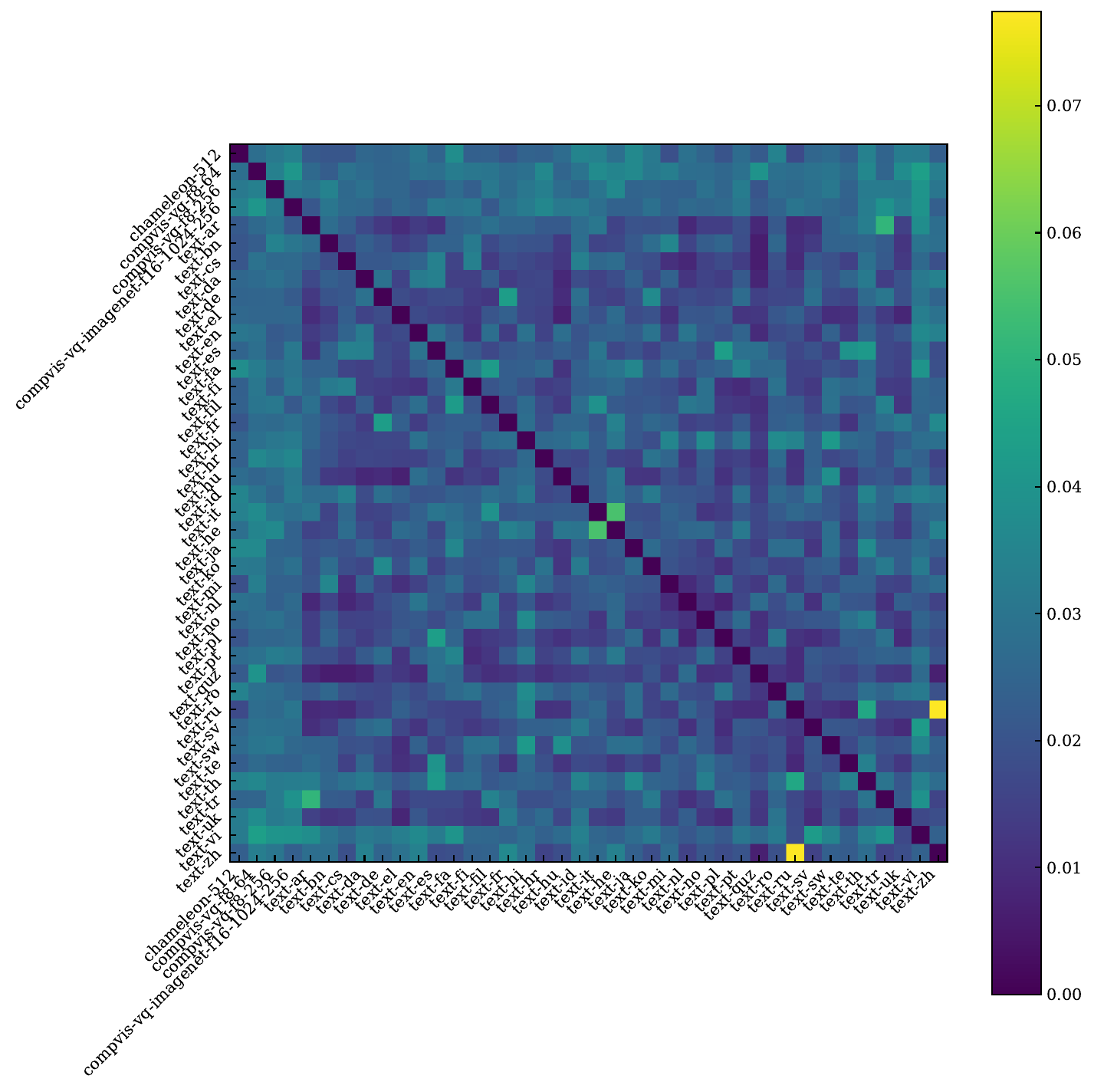}
        \caption{XM-3600}
        \label{fig:procrustes_xm3600}
    \end{subfigure}
    \hfill
    \begin{subfigure}[b]{0.49\textwidth}
        \centering
        \includegraphics[width=\textwidth]{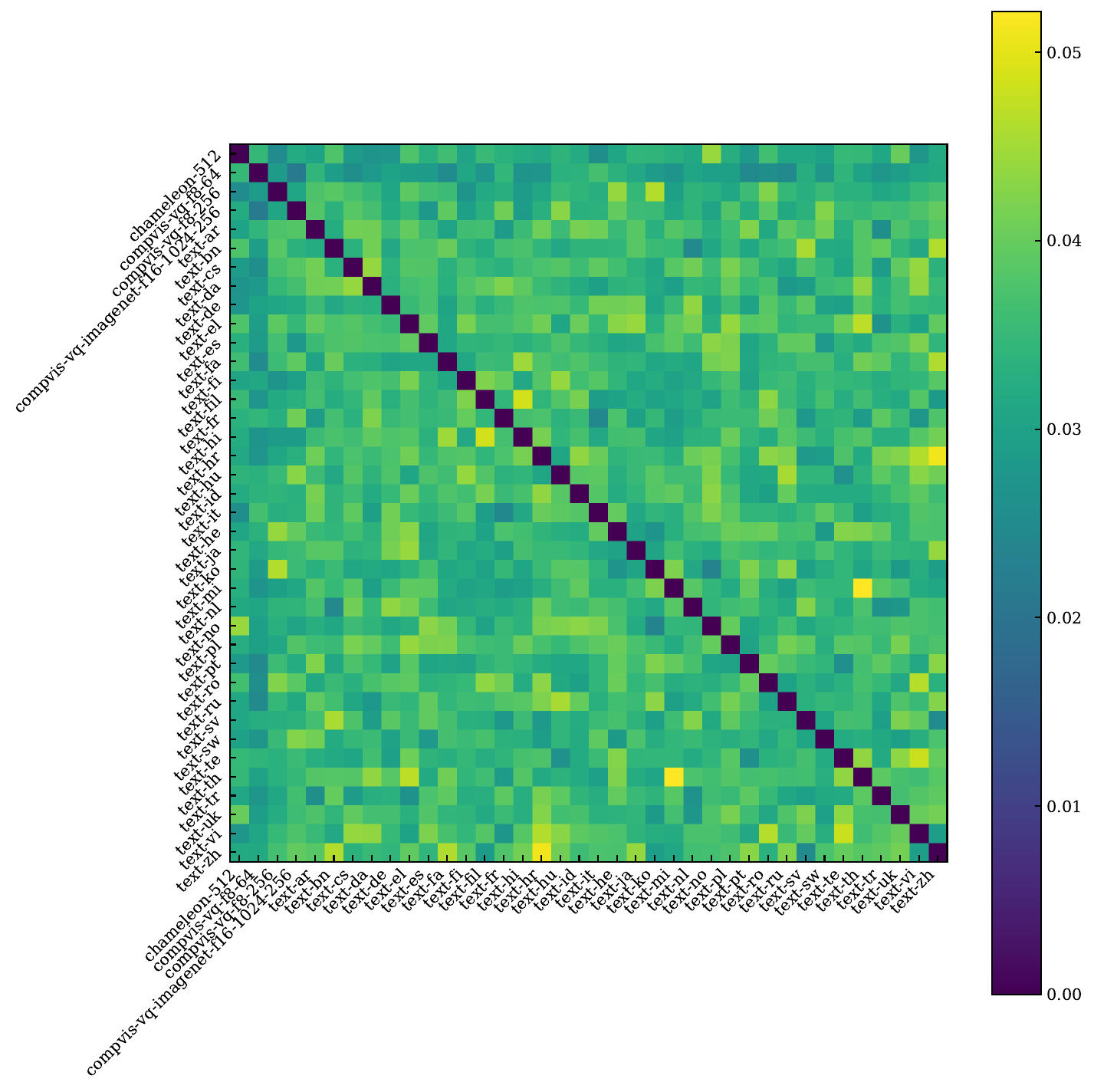}
        \caption{MS-COCO}
        \label{fig:procrustes_coco}
    \end{subfigure}
    \caption{Procrustes similarity matrices for XM-3600 and MS-COCO across all models and languages.}
    \label{fig:procrustes}
\end{figure}

Full results for our Procrustes analysis are given in \autoref{fig:procrustes}. For simplicity and clarity, in \autoref{fig:procrustes}, we replace the distance with $1 - \text{distance}$ to get a similarity measure, and set the diagonal to 0 (even though the diagonal similarity is naturally 1), in order to avoid contrast issues on the off-diagonals. 

\subsection{Directed Hausdorff Distance}

Directed Hausdorff Distance is a measure used to quantify the degree of mismatch between two point sets by calculating the greatest distance from a point in one set to the nearest point in the other set. It considers only the largest such deviation in one direction, making it useful for determining the extent to which one set is contained within or approximates another. The directed Hausdorff distance from set $A$ to set $B$ is defined as:

\begin{equation}
d_H(A, B) = \max_{a \in A} \min_{b \in B} \| a - b \|
\end{equation}

where $A$ and $B$ are two point sets, and $ \| a - b \| $ is the Euclidean distance between point $a \in A$ and point $b \in B$.

\begin{figure}[t]
    \centering
    \begin{subfigure}[b]{0.49\textwidth}
        \centering
        \includegraphics[width=\textwidth]{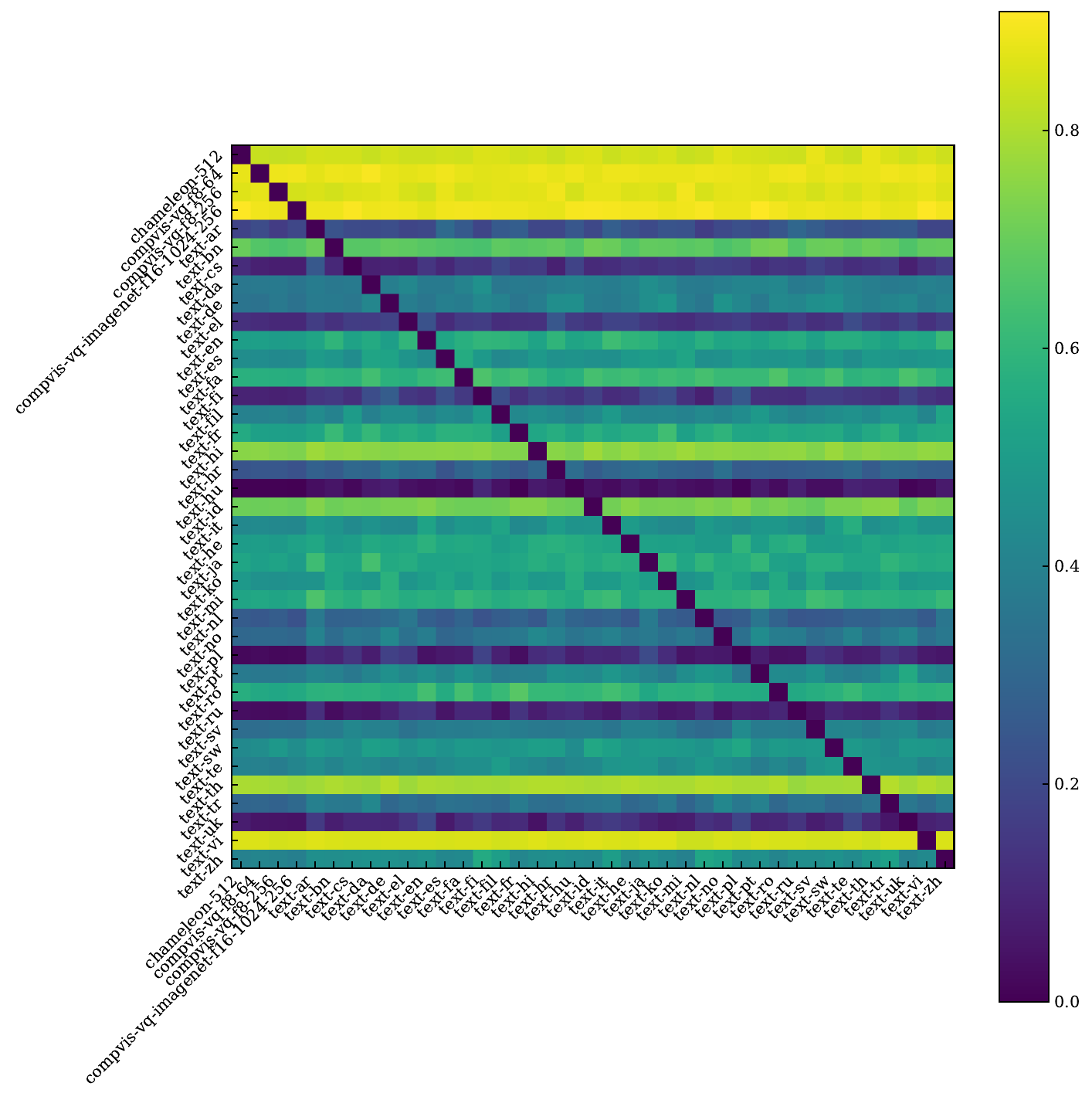}
        \caption{XM-3600}
        \label{fig:haussdorf_xm3600}
    \end{subfigure}
    \hfill
    \begin{subfigure}[b]{0.49\textwidth}
        \centering
        \includegraphics[width=\textwidth]{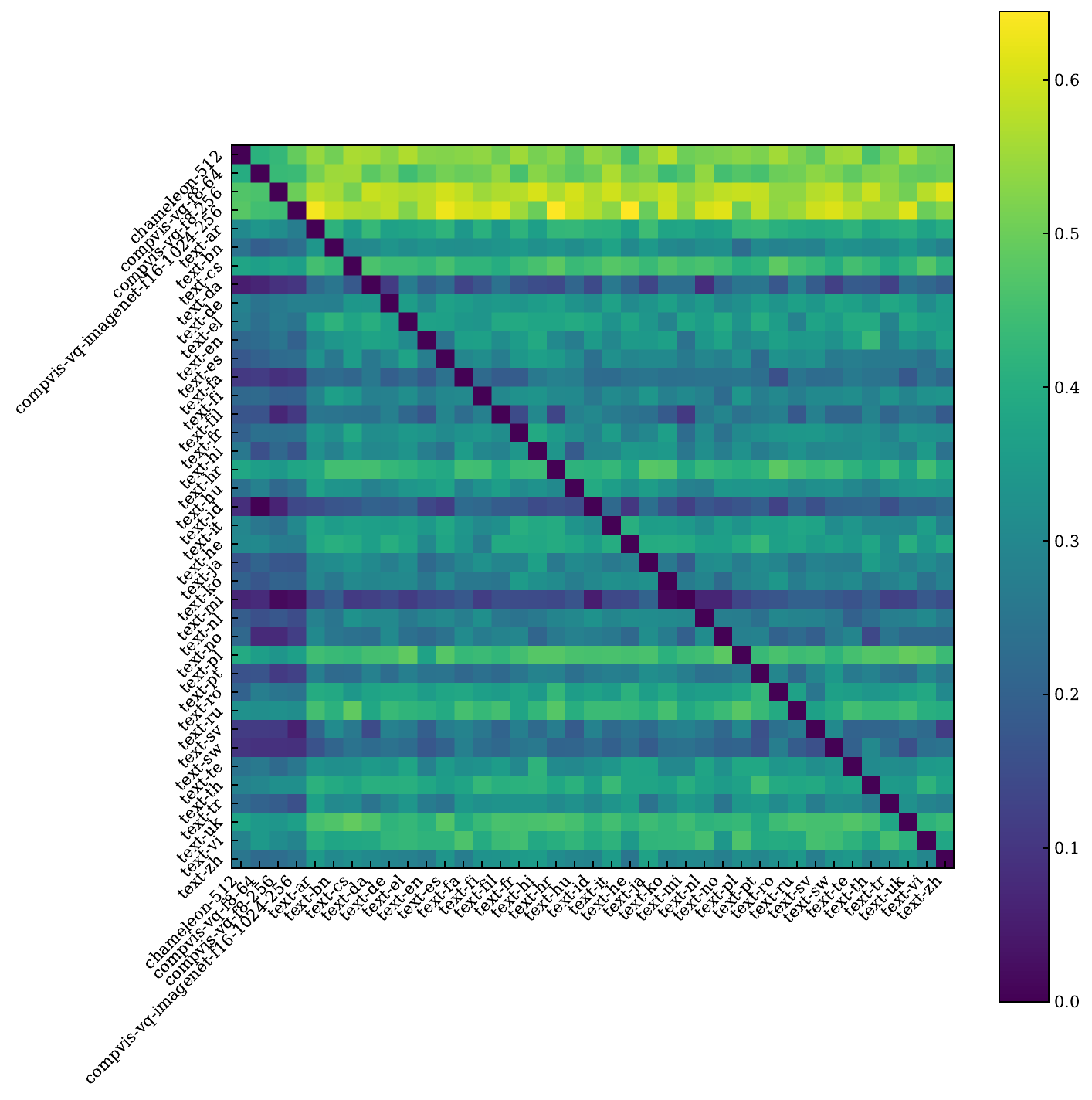}
        \caption{MS-COCO}
        \label{fig:haussdorf_coco}
    \end{subfigure}
    \caption{Directed Haussdorf \textit{similarity} matrices for XM-3600 and MS-COCO across all models and languages.}
    \label{fig:haussdorf}
\end{figure}

Similar to \autoref{fig:procrustes}, for simplicity and clarity, in \autoref{fig:haussdorf} we replace the distance with $1 - \text{distance}$ to get a similarity measure, and set the diagonal to 0 (even though the diagonal similarity is naturally 1), in order to avoid contrast issues on the off-diagonals. 

Full results for the directed Haussdorf distance are given in \autoref{fig:haussdorf}.

\end{document}